# Domain-Adapted Molecular Language Models for Efficient Search of Make-on-Demand Libraries

Henrik Wille,[a] Luis-Finley Schütz,[a] and Felix Strieth-Kalthoff *,[a,b]

a University of Wuppertal, School of Mathematics and Natural Sciences, Gaußstr. 20, 42119 Wuppertal, Germany

b University of Wuppertal, Interdisciplinary Center of Machine Learning and Data Analytics, Gaußstr. 20, 42119 Wuppertal, Germany

* Corresponding author: strieth-kalthoff@uni-wuppertal.de

Pretrained molecular language models are increasingly used as molecular encoders for learning structure–property relationships. However, their practical suitability for molecular discovery within and beyond their pretraining domain remains unclear. Herein, we systematically benchmark four molecular language models across six virtual molecular libraries spanning drug discovery, organic materials, and catalysis. Native molecular language model embeddings show substantial variation in discovery performance across libraries, whereas molecular fingerprints provide a consistently strong and robust baseline. Consistent with a potential domain–representation mismatch, we show that explicit domain adaptation substantially improves representation performance. Fine-tuning molecular language model encoders on structures from the target virtual library consistently improves sample efficiency, with several adapted encoders emerging as the top-performing representations across the benchmark tasks. These results show that molecular representation quality depends strongly on the target domain and that explicit adaptation can improve the practical utility of molecular foundation models. More broadly, our findings establish domain-adapted molecular representations as a promising strategy for sample-efficient adaptive decision making in virtual screening and self-driving laboratories.

## Introduction

The rapid discovery of functional molecules is central to many areas of chemistry, including medicinal chemistry, agricultural chemistry, and materials science.[1–3] In recent years, data-driven decision-making has become increasingly important for closing the loop between molecular property measurements and the selection of subsequent candidate molecules, particularly via the use of machine learning methods.[4–9] In many practical settings, these candidates are selected from a predefined, make-on-demand virtual library, whose members are enumerated from available molecular building blocks and validated synthetic reactions. Such libraries constrain the search to candidate molecules that are typically accessible at a feasible experimental cost.[10,11]

Despite recent advances in generative molecular design,[12,13] virtual library search remains widely used in real-world discovery scenarios.[14] Two recent developments have reinforced the importance of this approach. First, commercially available virtual libraries, most prominently the Enamine REAL space, have grown to the billion-molecule scale, covering an increasingly broad range of molecular structures.[15] Although exhaustive virtual screening of such libraries is infeasible, data-driven decision-making can identify promising candidates by computationally evaluating only a small, adaptively selected fraction of the library.[16,17] Second, automated laboratories are well suited for accessing molecules from combinatorial libraries, if candidate synthesis relies on a small number of well-defined synthetic procedures. When coupled with automated property measurements, data-driven decision-making enables closed design–make–test–analyze loops (Fig. 1a).[18,19] Although the specific costs and evaluation budgets differ between virtual screening and experimental discovery settings, both scenarios require models that effectively learn molecular structure–property relationships from limited property evaluations.

Molecular representation is central to learning and generalizing such structure–property relationships within a virtual library. Over the last decade, traditional cheminformatic representations,[20] such as physicochemical descriptors and structural fingerprints,[21] have been complemented by learned embeddings from deep neural networks, including graph neural networks[22,23] and molecular language models (Fig. 1b).[24,25] Such models, often discussed in the context of molecular foundation models,[26] learn representations through self-supervised pretraining on large collections of molecular structures. Particularly transformer models, that operate on SMILES notation and are trained with established objectives from the field of language modeling, have gained widespread attention as transferable representations for molecular property prediction and adaptive decision making.[27–32]

In molecular discovery scenarios, however, pretrained molecular encoders may be poorly matched to the target virtual library (Fig. 1c). Common pretraining datasets typically originate from widely used molecular databases and catalogs, including PubChem, ChEMBL and ZINC, which are dominated by drug-like molecules. As a result, the learned embedding spaces are primarily shaped by this region of chemical space. A focused drug-discovery library may occupy only a small region of this space, whereas libraries designed for materials chemistry or catalysis may occupy different regions altogether. As a result, a virtual library for a given molecular discovery problem can therefore be shifted relative to the pretraining distribution, or compressed within the model's embedding space. Such observations motivate a representation–domain mismatch hypothesis: candidate molecules may be insufficiently differentiated by the representation, which can limit the learning of structure–property relationships, especially in the presence of property cliffs.[17,33] This raises the broader question of how well pretrained molecular representations are suited for iterative discovery within and beyond their pre-training domain. Importantly, however, a distribution shift or embedding contraction alone does not establish that a representation is inadequate. For molecular discovery, representation quality ultimately depends on how effectively the representation supports the downstream surrogate model and decision-making procedure.

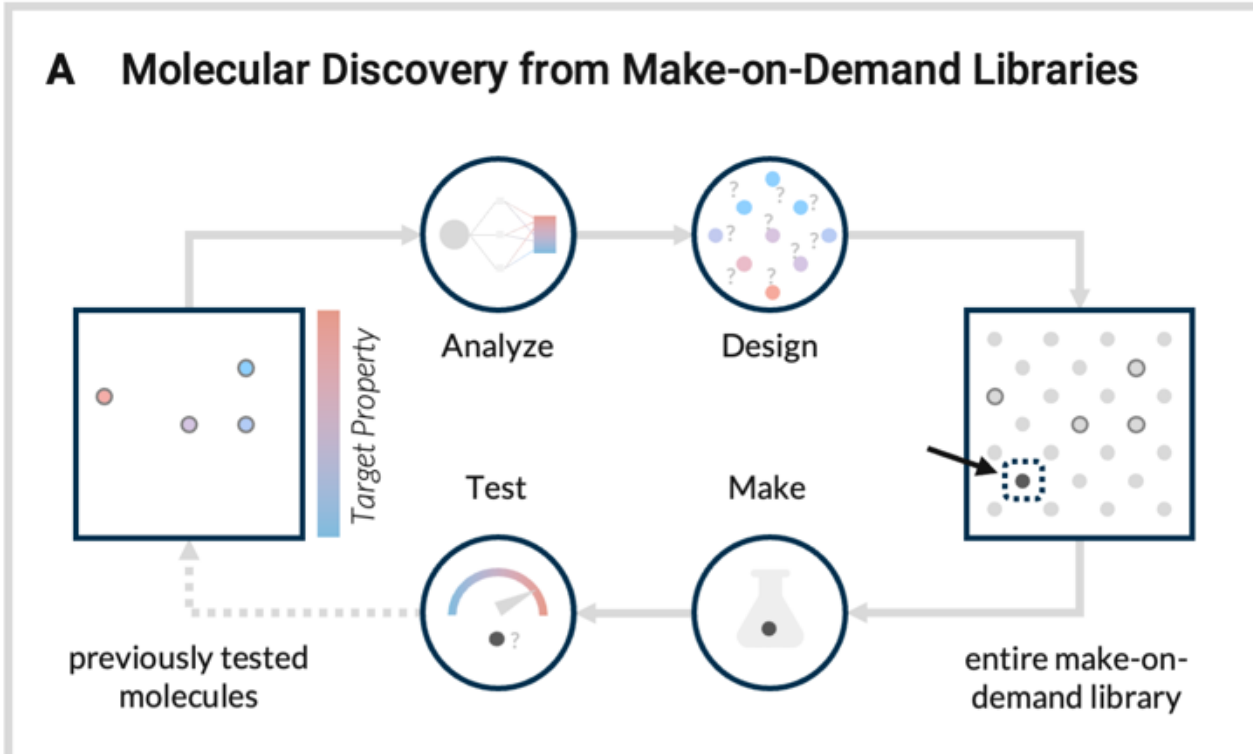

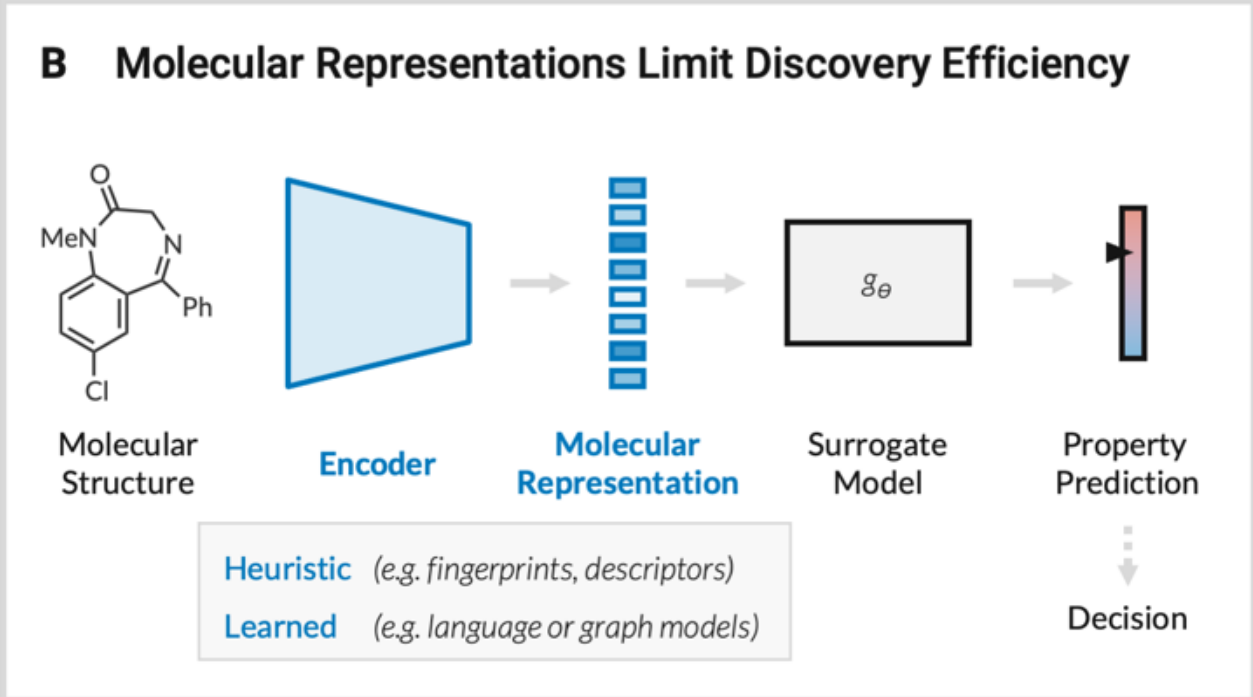

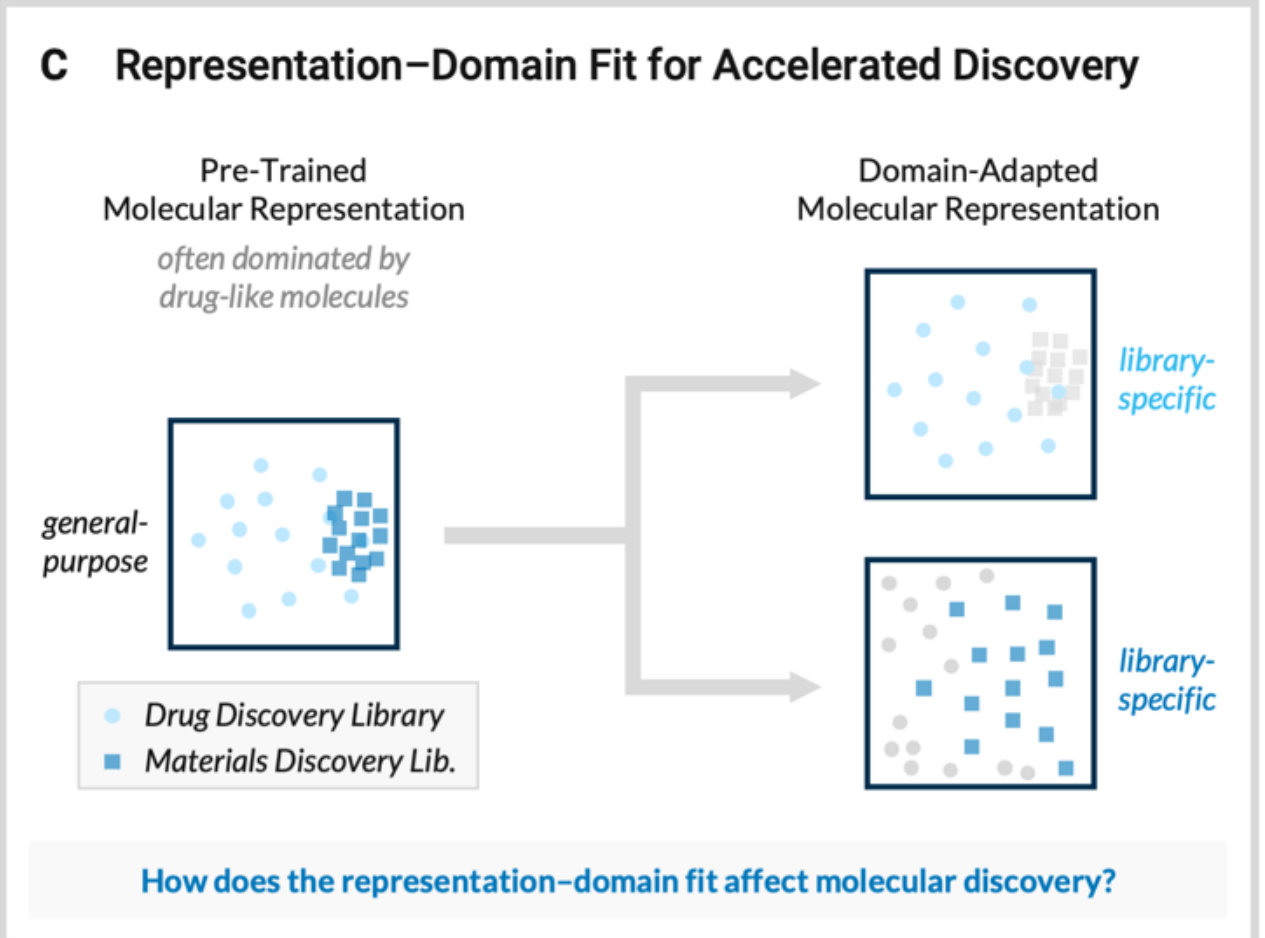

**Figure 1**: a) Design–Make–Test–Analyze cycles for discovering functional molecules from virtual make-on-demand libraries. b) Molecular representations as central factor governing sample-efficient discovery. c) Schematic two-dimensional projections of a pretrained molecular representation for a drug discovery library (circles, light blue) and a virtual library for materials discovery (squares, dark blue). Domain adaptation may improve the ability to differentiate molecules within a virtual library.

Herein, we examine the association between representation–domain fit and the efficiency of iterative molecular discovery. Inspired by recent demonstrations of task- or domain-specific fine-tuning for molecular property prediction,[34–38] we investigate whether potential mismatches can be mitigated by domain adaptation to the distribution of a given virtual library (Fig. 1c). Importantly, for practical use in iterative discovery, such adaptation should rely only on access to molecular structures from the candidate library and should not require costly property labels. To this end, we perform systematic benchmarks of molecular discovery efficiency under both virtual screening and experimental discovery budgets across six virtual libraries spanning drug discovery, materials chemistry, and catalysis. Overall, we aim to identify principles that govern efficient molecular discovery and provide practical guidance for selecting and adapting molecular representations in new discovery settings. We foresee that such insights could enable more efficient use of computational and experimental resources in real-world molecular design campaigns, including virtual screening for drug discovery, or self-driving laboratories for the discovery of functional molecular materials.

## Results and Discussion

Benchmark experiments were performed on six virtual libraries covering drug discovery, materials chemistry, and catalysis. The drug discovery tasks comprised a library of ~1.9 million molecules from ZINC15 with docking scores against an AmpC β-lactamase (*Zinc-AmpC*);[39] a focused subset of ~1.5 million ZINC15 molecules satisfying the rule-of-four criteria, with docking scores against 8-oxoguanine DNA glycosylase (*RO4-OGG1*);[40] and ~2.1 million molecules from the Enamine high-throughput screening collection, with docking scores against a thymidylate kinase (*Enamine-TMPK*).[17] The remaining three libraries represent applications outside drug discovery: a set of ~2.3 million molecules from the Harvard Clean Energy Project,[41] labeled with density functional theory-derived power conversion efficiencies (*HCEP-PCE*);[42] a combinatorial library of ~180,000 potential organic laser emitters with computed fluorescence oscillator strengths (*Laser-OSC*);[43] and ~300,000 phosphine ligands from the Kraken database[44] with predicted enantioselectivities in a Pd-catalyzed cross-coupling (*Kraken-ES*).[45] Further information on library construction, together with detailed descriptions of the molecular and target property distributions, is provided in the Supplementary Information (SI).

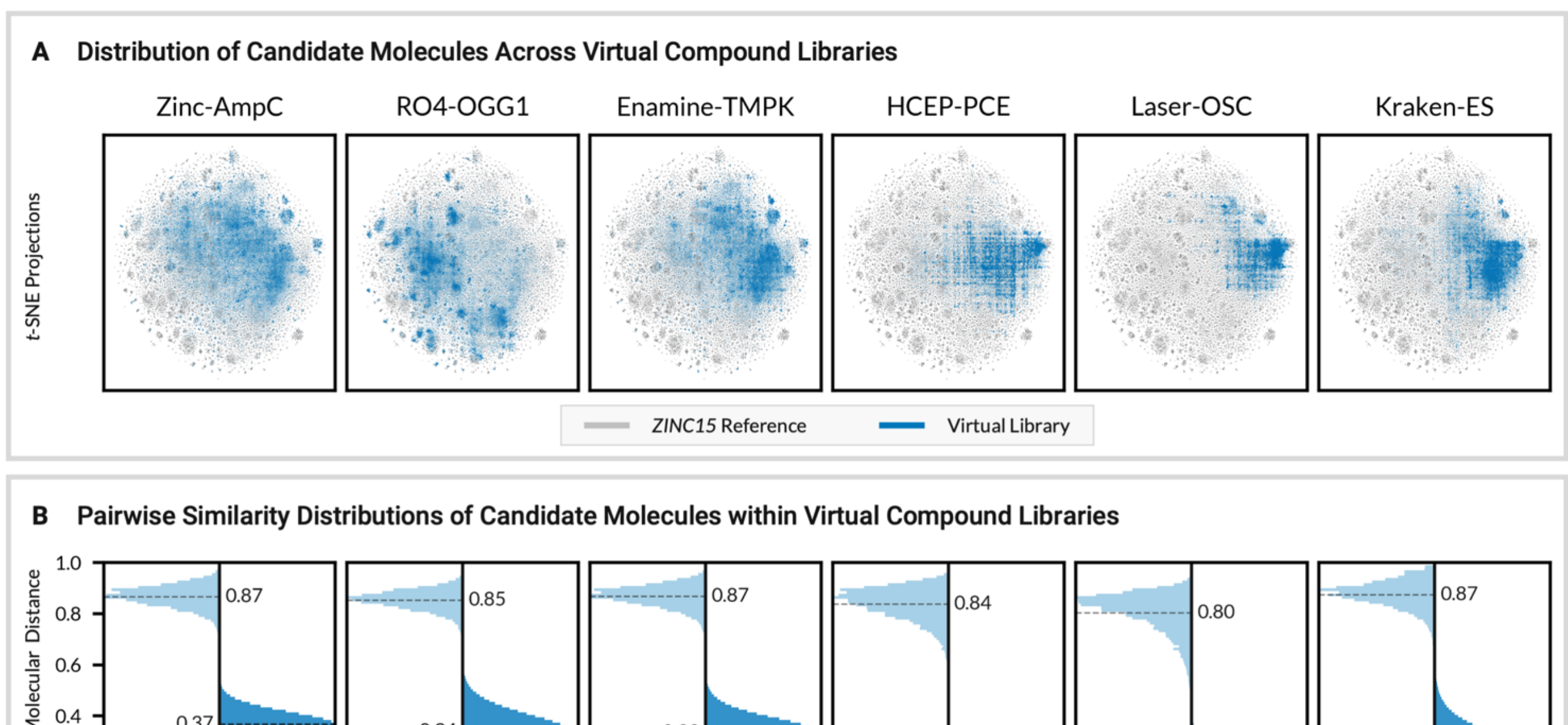

**Figure 2:** Distribution of candidate molecules in the virtual libraries relative to a reference library of 25,000 randomly selected compounds from the ZINC15 library. Top: A *t*-distributed stochastic neighbor embedding (*t*-SNE) model was fitted to Morgan fingerprints of all compounds in the reference library, and molecules from the target libraries were projected into the same *t*-SNE embedding space. Bottom: Distributions of pairwise molecular distances for 5 million molecule pairs sampled from within each library, computed using Tanimoto similarity on molecular fingerprints (light blue, left) and cosine distance on *MolFormer* embeddings (dark blue, right).

**Virtual molecular libraries occupy distinct regions of chemical and representation space.**

Chemical space analysis confirms that these libraries show systematic differences in molecular distribution (Fig. 2). To investigate distribution shifts and possible representation–domain mismatches, we collected a reference set of 25,000 randomly selected ZINC15 molecules as a proxy for the drug-like chemical space represented in common pretraining datasets.[46] Qualitative projection of the target relative to this reference set (Fig. 2a) suggests a substantial overlap for the three drug-discovery libraries, whereas the materials and catalysis-focused libraries show the expected domain shift. Pairwise molecular distance distributions between the reference and each target library (see SI) support this conclusion, indicating that the libraries outside drug discovery could be well suited for investigating potential representation–domain mismatches.

We therefore investigated the distribution of these libraries within the embedding spaces of several pretrained molecular language models with different transformer architectures:

- *ChemBERTa*, a *RoBERTa*-type encoder (12M parameters), trained through masked-token prediction on 100k molecules from ZINC15 and ChEMBL;[47,48]
- *MolFormer*, an encoder-only transformer with linear attention and rotary positional embeddings (45M parameters), pretrained on approximately 1.1B molecules from ZINC and PubChem;[49]
- *T5Chem*, a T5 encoder–decoder model (110M encoder parameters) pretrained on 33.5M training samples;[50] and
- *SmiTed*, a transformer-based encoder–decoder model with a learned, autoencoder-based pooling and reconstruction framework (55M encoder parameters), pretrained on ~91M curated SMILES strings from PubChem.[51]

Exact checkpoints, embedding dimensions, and pooling procedures are provided in the SI.

Analysis of pairwise distance distributions in embedding space shows a substantial contraction for the non-drug libraries (Fig. 2b), indicating that the molecules occupy a more restricted region of embedding space. Similar dispersion was observed across all molecular language models, with *SmiTed* producing particularly narrow embedding distributions (see SI for further details). While these findings may provide an initial indication of a potential representation–domain mismatch, it should be noted that a library occupying a narrow region in embedding space does not necessarily imply the loss of information for learning structure–property relationships. We therefore treat this embedding dispersion as a candidate diagnostic, and evaluate its relationship to discovery efficiency in the following sections.

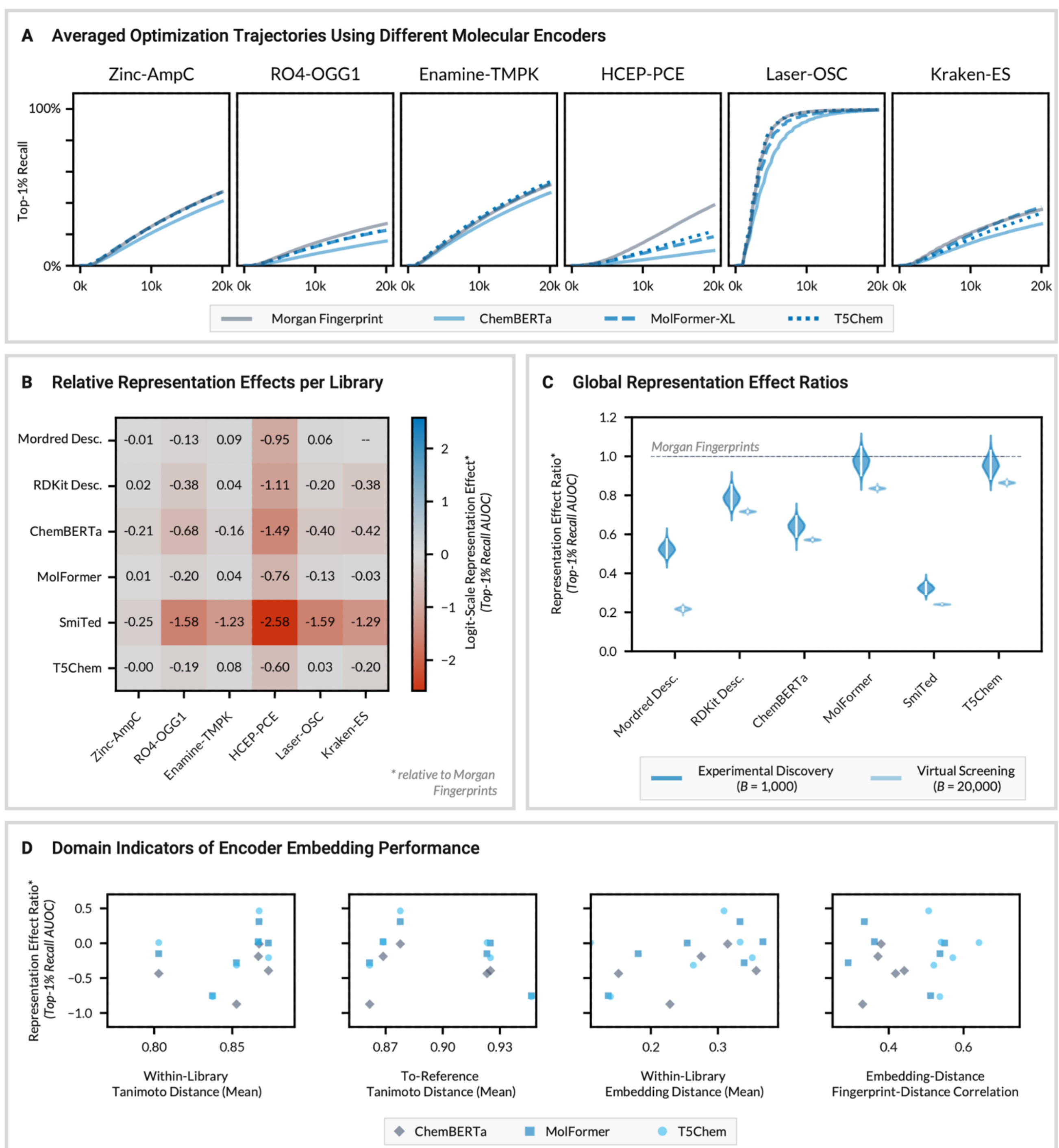

| | Zinc-AmpC | RO4-OGG1 | Enamine-TMPK | HCEP-PCE | Laser-OSC | Kraken-ES |
|---|---|---|---|---|---|---|
| Mordred Desc. | -0.01 | -0.13 | 0.09 | -0.95 | 0.06 | -- |
| RDKit Desc. | 0.02 | -0.38 | 0.04 | -1.11 | -0.20 | -0.38 |
| ChemBERTa | -0.21 | -0.68 | -0.16 | -1.49 | -0.40 | -0.42 |
| MolFormer | 0.01 | -0.20 | 0.04 | -0.76 | -0.13 | -0.03 |
| SmiTed | -0.25 | -1.58 | -1.23 | -2.58 | -1.59 | -1.29 |
| T5Chem | -0.00 | -0.19 | 0.08 | -0.60 | 0.03 | -0.20 |

**Figure 3:** Evaluation of molecular representations for iterative discovery in virtual molecular libraries. a) Top-1% recall trajectories averaged over 20 independent runs. All experiments used a Laplace-approximated neural network surrogate and an upper confidence bound candidate ranking with independent top-*k* batch selection. b) Library-specific representation effects, and c) global representation effects from the hierarchical Bayesian analysis of run-level AUOC values, reported relative to Morgan fingerprints. d) Exploratory correlations between library distribution descriptors and benchmark optimization performance. Further experimental details are provided in the Supplementary Information. AUOC: area under the optimization curve.

**Native molecular language model representations show variable performance across tasks.**

Motivated by these observations, we evaluated the molecular representations in benchmark tasks of iterative, model-based discovery across two complementary acquisition-budget regimes. For the virtual screening scenario, we used a total budget of 20,000 property evaluations in batches of 1,000, representing feasible budgets for computational campaigns in which e.g. molecular docking is used for property evaluation. The experimental discovery regime was limited to batches of 50 evaluations up to a total budget of 1,000 evaluations, approximating the budget constraints of a self-driving laboratory scenario in which each evaluation requires molecular synthesis and experimental property measurement. After each acquisition round, a neural network surrogate model with a Laplace approximation for uncertainty quantification was trained on all previously evaluated candidates. An upper confidence bound acquisition function then was used to rank unevaluated candidates, and select the top-scoring candidates as a batch. Comparisons with alternative surrogate models and acquisition strategies are provided in the SI.

For each experiment, we recorded optimization trajectories for the recall of the top 1% library candidates as a function of the cumulative evaluation budget. All experiments were repeated 20 times using independently sampled seed populations. From each trajectory, we computed the area under the optimization curve (AUOC) and fit a hierarchical Bayesian model to the run-level AUOC values, in order to separate global representation effects from library-specific contributions and variation between seed populations.

Morgan fingerprints were consistently among the strongest native representations across the six discovery tasks. A quantitative comparison of traditional cheminformatic representations and molecular language model embeddings shows that, while significant library-by-library differences in top-1% candidate recall trajectories are observed (Fig. 3a), Morgan fingerprints consistently rank in the best-performing representations. Statistical analysis reveals that all other tested representations, including descriptors and language model embeddings, showed significant negative global posterior effects relative to this baseline. In line with recent literature precedents, these findings confirm the utility of Morgan fingerprints as a robust baseline for molecular discovery.[51] Descriptor-based representations show greater variation among libraries. For certain libraries, especially those from drug discovery, the posterior effects relative to fingerprints are close to zero, whereas performance is lower for other libraries (see Fig. 3b and SI).

Native molecular-language-model embeddings varied substantially in discovery performance. *MolFormer* and *T5Chem* were the strongest native molecular language model representations, and approached the performance of Morgan fingerprints in several libraries (Fig. 3b). However, neither model consistently outperforms fingerprints across the full set of benchmark tasks, as evidenced by negative global posterior effects (Fig. 3c). *ChemBERTa* embeddings show a significantly more negative effect, which may reflect the smaller model size and the potentially lower expressivity of the resulting embeddings. Native *SmiTed* embeddings show the lowest global effect across all molecular representations tested. This behavior is consistent with the extremely narrow within-library embedding distances observed during the geometric analysis (Fig. S1–S24), indicating that the frozen *SmiTed* embeddings may not be suitable for downstream property prediction without simultaneous adaptation of the encoder or the learned pooling mechanism.

The observed ordering of representation was largely consistent across the virtual screening and experimental discovery regimes (Fig. 3c), although differences were less pronounced in the experimental discovery scenario. Under the lower experimental budget, Morgan fingerprints and the embeddings from *MolFormer* and *T5Chem* show overlapping global posterior effects. One possible explanation for this observation is a stronger contribution from initial-sample variability and exploration. Especially during the first iterations of an experimental discovery campaign, the surrogate model has very limited information about the target property landscape, and candidate selection may be strongly influenced by the initial population and predictive uncertainty. In contrast, once the surrogate is provided with sufficient molecule–property pairs to learn advanced structure–activity relationships, expressive representations may become more important for refining these relationships. While a formal investigation of this hypothesis is beyond the scope of this study, our findings confirm that the general trends across representations remain consistent under both budget regimes investigated.

Overall, these findings identify only a weak relation between representation–domain fit and experimental optimization efficiency. On the one hand, it was notable that both physicochemical descriptors and molecular language model embeddings consistently underperformed molecular fingerprints on the *HCEP-PCE* library, which also showed the largest domain shift from the ZINC15 reference (mean Tanimoto distance of 0.95). This pattern would be compatible with the representation–domain mismatch hypothesis. On the other hand, language model embeddings supported relatively efficient discovery in the *Laser-OSC* (*T5Chem*) and *Kraken-ES* (*MolFormer*) libraries, despite significant domain shifts (mean Tanimoto distances of 0.92 and 0.93, respectively). Indeed, exploratory performance correlations identify only a weak negative correlation between pairwise Tanimoto distances to the reference library and the relative posterior effects for different molecular language model embeddings (Fig. 3d). Likewise, within-library embedding dispersion, which we identified above as a diagnostic of potential representation–domain mismatches, shows only a weak positive correlation with relative representation effects. Further correlations between optimization behavior with readily accessible descriptors of the molecular distribution did not reveal stable, interpretable trends.

**Domain adaptation improves discovery efficiency.**

Given these weak indications of possible representation–domain mismatches, we evaluated whether the observed underperformance of molecular language models could be mitigated by explicit domain adaptation. We envisioned that, by fine-tuning on molecular structures from the target library, the representation could become more sensitive to structural variation within the virtual library, thereby improving learned structure–property relationships and, in turn, increasing discovery efficiency. For this purpose, we evaluated three conceptually distinct strategies: (i) continued language-model-style pretraining with the model's original masked-token prediction or span corruption objective; (ii) contrastive representation adaption, using triplets of similar and dissimilar molecules from the library in combination with a temperature-scaled multiple-negatives ranking loss;[52,53] and (iii) fingerprint-supervised auxiliary-label domain adaptation,[38] in which a prediction head was trained to reconstruct Morgan fingerprint bits from the encoder representation. All encoder updates were performed using parameter-efficient fine-tuning through low-rank adaptation (LoRA).[54] Full data sampling procedures, losses, and hyperparameters are described in the SI.

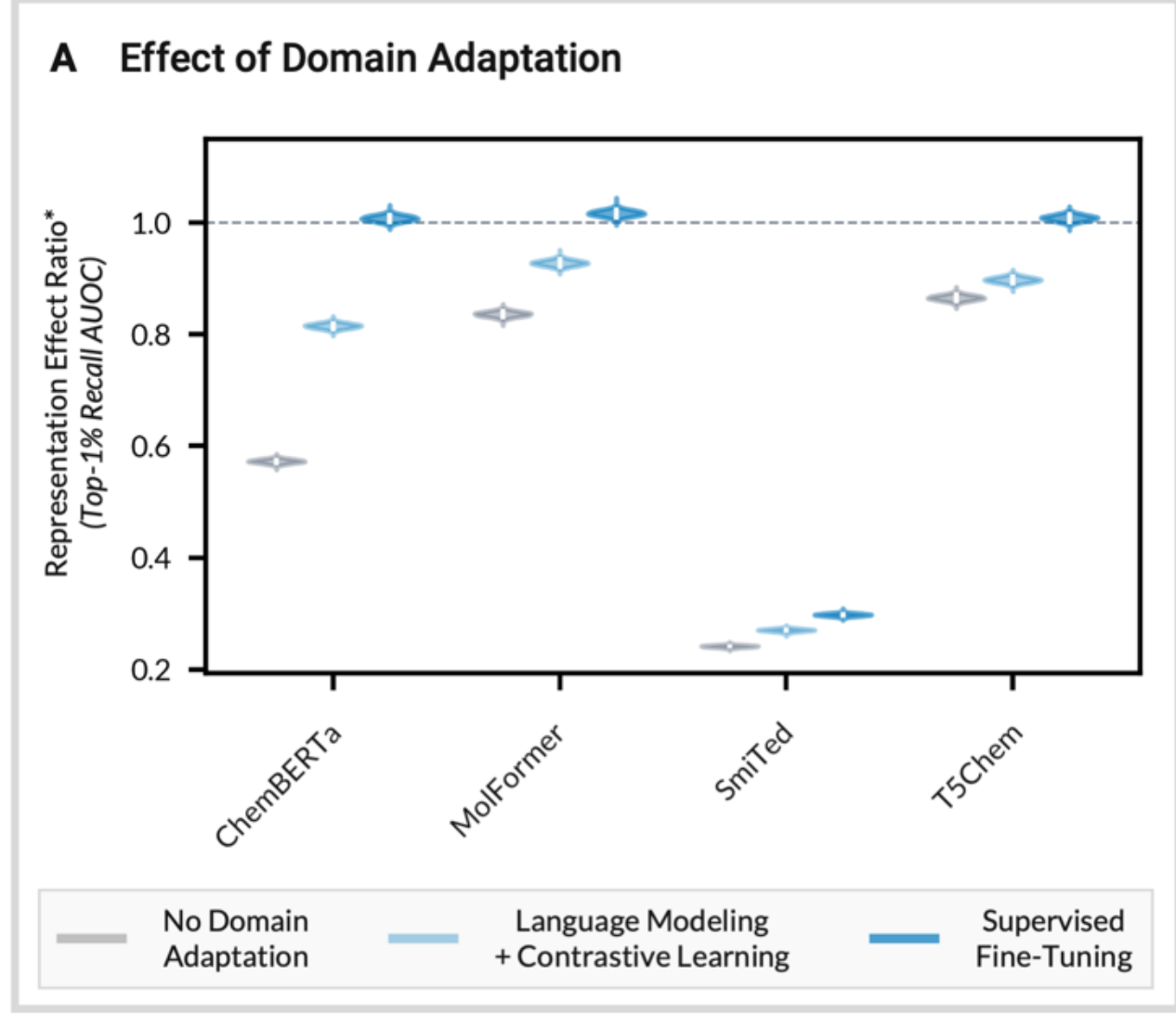

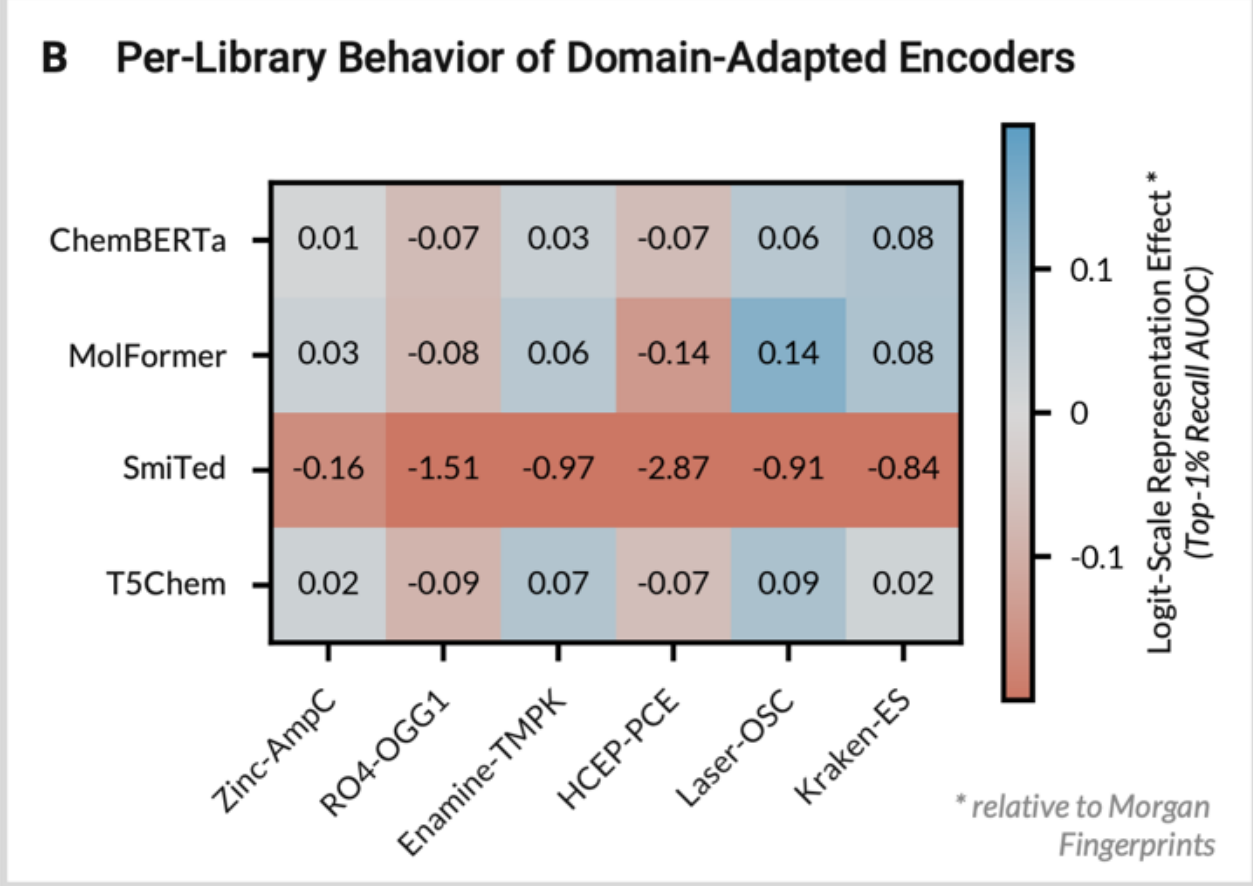

**Figure 4:** Evaluation of domain-adapted transformer embeddings for iterative discovery in virtual molecular libraries. a) Global representation effects, as obtained from a hierarchical Bayesian analysis of raw trajectories. b) Relative effects of supervised domain adaptation, as obtained from a hierarchical Bayesian analysis of raw trajectories.

Our findings show that optimization efficiency increased significantly when domain-adapted molecular language model embeddings are used. Across all tested libraries and molecular language models, auxiliary-label adaptation produced the largest average gains (Fig. 4a). Notably, the adapted *ChemBERTa*, *MolFormer*, and *T5Chem* embeddings show positive global effects relative to Morgan fingerprints, with broadly similar trends across several libraries and acquisition-budget regimes (see SI for further details).

*ChemBERTa* benefited most strongly from domain adaptation. Although native embeddings showed significantly negative average effects relative to Morgan fingerprints, the adapted substantially reduced or reversed these effects. This result shows that even a comparatively compact encoder can yield a practical embedding space for efficient optimization, even when the native embedding space is not well-suited to the discovery task. Similarly, we find that *MolFormer* and *T5Chem* improve upon their comparatively strong native performance. *SmiTed* embeddings follow the same directional trend, but the adapted model retained a clear negative posterior effect relative to Morgan fingerprints.

These findings indicate that domain adaptation combines complementary information from pretraining and finetuning,[36] rather than merely reproducing Morgan fingerprints. If the auxiliary-label fine-tuning strategy transformed the encoders into approximate fingerprint calculators, their behavior in molecular discovery scenarios would be expected to approach, but not systematically exceed, that of fingerprints. We hypothesize that the use of fingerprints as auxiliary objectives introduces a bias towards library-specific molecular substructures and local chemical environments, while the pretrained encoder retains more global structural similarity concepts from its initial pretraining corpus. The auxiliary-label strategy therefore provides a simple route for incorporating general chemical expertise and domain-specific information into the molecular language model representation.

Contrastive domain adaptation also improves discovery performance relative to the native embeddings, albeit with smaller gains than those obtained using auxiliary-label finetuning. Nonetheless, the use of a contrastive training objective was beneficial. Continued pretraining with only the original language model objective provided significantly smaller changes (see SI for further details), confirming that the contrastive component contributes directly to the change in representation performance. The observed differences between these three adaptation strategies highlight the value of chemically informed training objectives, and indicate possible directions for further improvement. We foresee that combining complementary domain adaptation objectives, e.g. by combining contrastive learning with auxiliary-label strategies based on fingerprints and physicochemical descriptors, will lead to further improvements in sample-efficient optimization. Similarly, a systematic investigation of fine-tuning hyperparameters is expected to be beneficial, but was beyond the scope of this study.

## Conclusions

Overall, this study shows that molecular discovery from virtual libraries can involve substantial discrepancies between the geometry of pretrained molecular representations and the structure of the target library. Molecular language models are commonly pretrained on corpora dominated by drug-like molecules, whereas virtual libraries for materials discovery and catalysis can occupy distinct regions of chemical space. These differences are reflected in shifted and comparatively narrow distributions within the corresponding embedding spaces, and may influence discovery efficiency. However, these differences explained only little of the observed performance differences in benchmark experiments of molecular discovery. The performance of molecular language model embeddings varies considerably across virtual libraries, and this variation can only partly be attributed to the representation–domain fit. By contrast, molecular fingerprints perform robustly across chemical domains, target properties, and acquisition budgets and outperform the native molecular language model encoders on average. These findings reinforce molecular fingerprints as a strong default representation for iterative molecular discovery.

The main practical result is that shortcomings of the native molecular language model representations can be effectively mitigated through library-specific domain adaptation. Fine-tuning the molecular language model encoders using molecular fingerprints as auxiliary labels provides consistent performance gains and allows adapted *ChemBERTa*, *MolFormer*, and *T5Chem* embeddings to exceed the Morgan fingerprint baseline. Because this adaptation requires only molecular structures sampled from the target virtual library, it can be performed at the outset of a discovery campaign, and could be applied to billion-scale virtual libraries.

The scope of these conclusions is naturally limited by the breadth of the virtual libraries included in our benchmarks, and confirmation across a broader range of molecular distributions and target properties would therefore be valuable. However, such validation is particularly difficult outside drug discovery, where few sufficiently large and consistently labeled molecular libraries are available for systematic benchmarking. Representation performance may also depend on the surrogate model, acquisition function, and batch-selection strategy, although the additional benchmarks indicate that the main trends are robust across several methodological choices. Updating the molecular encoder directly during surrogate model training may further improve its expressivity, but this strategy substantially increases the computational cost of each iteration and may become prohibitive for large virtual libraries.

Overall, our findings provide a basis for developing more effective domain adaptation objectives that incorporate additional forms of chemical information. Together, these developments could extend the utility of molecular foundation models beyond their use as general-purpose feature extractors and enable representations tailored to the molecular domain and decision problem of a specific discovery campaign. In both computational virtual screening and self-driving laboratories, we foresee that domain-adapted representations can improve sample efficiency and accelerate discovery.

## Author contributions

Conceptualization: F. S.-K.; Data curation: H. W., L.-F. S., F. S.-K.; Formal analysis: H.W., L.-F. S., F. S.-K.; Funding acquisition: F .S.-K.; Investigation: H. W., L.-F. S., F. S.-K.; Methodology: F. S.-K.; Project administration: F. S.-K.; Resources: F. S.-K.; Software: F. S.-K.; Supervision: F. S.-K.; Validation: H. W., L.-F. S., F. S.-K.; Visualization: F. S.-K.; Writing – original draft: F. S.-K.; Writing – review & editing: H. W.; L.-F. S.; F. S.-K.

## Conflicts of interest

There are no conflicts to declare.

## Data availability

All experiments reported in this paper were performed using the BAYLEYS (Bayesian Library Exploration and Virtual Screening) Python package, which is available on *GitHub* at https://github.com/fsk-lab/bayleys. Python scripts for reproducing the described experiments and performing the associated data analyses are provided in the same repository. A snapshot of this repository, together with the complete molecular libraries, experiment configurations, and experimental results is archived on *Zenodo* (https://doi.org/10.5281/zenodo.21907928).

## Acknowledgements

Computations were performed on the *Pleiades* HPC center at the University of Wuppertal. Financial support from the Deutsche Forschungsgemeinschaft (SPP2363, "Molecular Machine Learning“, Grant No. 497260357) is gratefully acknowledged. The authors thank Mohammad Haddadnia (Harvard University) for helpful discussions.

*Supplementary Information*

# Domain-Adapted Molecular Language Models for Efficient Search of Make-on-Demand Libraries

Henrik Wille,[a,†] Luis-Finley Schütz,[a,†] and Felix Strieth-Kalthoff [a,b,*]

a University of Wuppertal, School of Mathematics and Natural Sciences, Gaußstr. 20, 42119 Wuppertal, Germany

b University of Wuppertal, Interdisciplinary Center of Machine Learning and Data Analytics, Gaußstr. 20, 42119 Wuppertal, Germany

# 1 Experimental Setup

All experiments were performed using our *Python* library BAYLEYS (**Bay**esian **L**ibrary **E**xploration and Virtual **S**creening). The code base for reproducing all experiments in this paper are provided on *GitHub* (https://github.com/fsk-lab/bayleys). The molecular libraries, as well as specific implementation details and experimental procedures, are described below.

## 1.1 General Procedure: Iterative Optimization

To benchmark the efficiency of iterative optimization algorithms, we used virtual molecular libraries for which computed target properties were available for all molecules in the library. The libraries used in our benchmark experiments are described in detail in Section 2. In the context of our benchmark experiments, "observing" the target property corresponds to a lookup in the full library. Before optimization, molecular representation vectors (see Sections 1.3 and 1.4) for all molecules in the library were pre-computed and cached.

Iterative optimization experiments were performed in a batch-wise manner. For initialization, a batch of $b$ molecules were drawn randomly from the full library, and the corresponding target properties were "observed". These molecules formed the seed dataset of observations. The surrogate model (see Section 1.5) was then trained on this dataset using the precomputed representation vectors as inputs. From the trained model, the predictive mean and uncertainty for all remaining molecules in the virtual library were inferred. These predictions were used to compute an acquisition function value for each unevaluated molecule. The next batch of $b$ molecules was selected for evaluation according to a batch acquisition policy (see Section 1.6). The target properties of these molecules were then "observed", and the corresponding data points were added to the dataset of observations. This procedure was repeated until a given budget $B$ of observations was exhausted. Each experiment was repeated 20 times with different seed datasets of observations. To enable paired comparisons, the same initial datasets were used across different methods for each library.

All experiments were performed in two settings:

- Virtual Screening Setting: $b = 1{,}000$  $B = 20{,}000$
- Experimental Discovery Setting: $b = 50$  $B = 1{,}000$

For quantitative evaluation, we extract optimization trajectories that record the recall of top-$k$ candidates (top-1000 or top-1%) as a function of the number of experimental observations. Averaged optimization trajectories over the 20 repeats are shown for visual comparison of optimizer performance.

For a quantitative statistical comparison between optimizers, we compute the normalized area under the optimization curve (AUOC) through trapezoidal integration, and then separate library, model, and seed effects using a hierarchical Bayesian regression approach.

For that purpose, we analyzed the normalized AUOC values for each library (indexed *i*), optimizer (indexed *j*) and seed (index *k*) using a hierarchical beta regression model

$$y_{ijk} \sim \mathrm{Beta}\left(\mu_{ijk} \cdot \phi, \left(1 - \mu_{ijk}\right) \cdot \phi\right)$$

where $\mu_{ijk}$ is the expected normalized AUC, and $\phi$ is the concentration parameter of the beta distribution.

The expected AUC was modeled on the logit scale, representing the paired design described above.

$$\mathrm{logit}\left(\mu_{ijk}\right) = \alpha + a_i + s_{ik} + \beta_j + \gamma_{ij}$$

The individual terms reflect the sources of variability in our experimental design.

- $\alpha$ is the global intercept.
- $a_i$ is the library-specific baseline effect of library *i*.
- $\beta_j$ is the average effect of optimizer *j* relative to a chosen reference optimizer.
- $s_{ik}$ is the library-specific effect of seed *k* in library *i*, accounting for the paired experimental design. Seed populations from different libraries were treated as unrelated.
- $\gamma_{ij}$ is the library-specific deviation of optimizer *j* from its average effect.

We model $\beta_j$, $s_{ik}$, and $\gamma_{ij}$ as centered varying effects from a normal distribution prior, controlled by shared scale parameters from a half-normal distribution. $\alpha$, $a_i$ and $\beta_j$ were assigned normal priors. For $\phi$, we used a log-normal prior.

The model was built using Monte-Carlo integration using the *pymc* library.[1] Posterior sampling was performed using a No-U-Turn-Sampler with four independent chains. Each chain comprised 2000 tuning iterations and 2000 retained posterior draws. Sampling used diagonal mass-matrix adaptation and a target acceptance probability of 0.9.

From the obtained posterior samples, we calculate relative optimizer effects on the logit scale. For optimizer *j*, the relative effect compared to the reference optimizer is

$$\theta_j = \beta_j \ .$$

In addition, we compute library-specific effects as

$$\theta_{ij} = \beta_j + \gamma_{ij}$$

Both values are independent of library and seed effects, and we report statistics of $\theta_j$ and $\theta_{ij}$ over 8,000 independent posterior samples.

From these logit-scale effects, effect ratios were calculated as

$$\Theta = e^{\theta} \ .$$

An effect ration of 1 indicates no difference from the reference optimizer, values above indicate larger expected AUC values compared to the optimizer, i.e., improved performance, whereas values below 1 indicate lower expected AUC values.

### 1.2 Heuristic Molecular Representations

The following three heuristic molecular representations were evaluated in our benchmark experiments.

***Morgan Fingerprints***[2] were used as bit vectors with a radius of 2 and a vector size of 2048. Fingerprints were generated using the *RDKit* (version 2026.3.1).[3]

***Mordred Descriptors*** were generated with the *Mordred* library.[4] Only descriptors derived from the two-dimensional structure were included. Descriptors with zero variance across the full virtual library were removed.

***RDKit Descriptors*** were generated by calculating all descriptors implemented in *RDKit.Chem.Descriptors* (version 2026.3.1).[3]

### 1.3 Molecular Language Models

We evaluated four molecular language models pre-trained on large corpora of SMILES strings. For each model, we used the indicated Hugging Face checkpoint, and obtained molecular embeddings from the model encoder.

***ChemBERTa***[5] (checkpoint: *seyonec/ChemBERTa-zinc-base-v1*) is a RoBERTa-style encoder-only transformer model, using 5 layers and 12 attention heads. ChemBERTa was pre-trained with masked language modeling (15% masked input tokens) pre-trained on 100,000 SMILES strings from the ZINC library. Molecular embeddings were obtained from canonicalized SMILES by pooling encoder outputs.

***MolFormer***[6] (checkpoint: *ibm/MoLFormer-XL-both-10pct*) is an encoder-only transformer that uses linear attention and rotary positional embeddings. MolFormer was pre-trained with masked language modeling on about 1.1B unlabeled molecules from the PubChem and ZINC databases. Molecular embeddings were obtained from canonicalized SMILES by pooling encoder outputs.

***T5Chem***[7] (checkpoint: *GT4SD/multitask-text-and-chemistry-t5-base-augm*) is a T5-base encoder–decoder transformer trained as a multi-task model for chemistry and natural language. It was trained with text-to-text prompting on five tasks: forward reaction prediction, retrosynthesis, molecular captioning, text-conditional molecule generation, and paragraph-to-action extraction. The model was trained on Pistachio-derived reaction pairs, paragraph-to-action procedure data, and ChEBI-20 molecule description pairs (33.5M training samples across tasks). Molecular embeddings were obtained from canonicalized SMILES by pooling encoder outputs.

***SmiTed***[8] (checkpoint: bisectgroup/materials-smi-ted-fork) is an encoder-decoder model with a BERT-like token encoder and an autoencoder model for SMILES reconstruction. Pre-training combined masked token prediction for the encoder with a reconstruction loss for SMILES reconstruction from the autoencoder. The model was pre-trained on 91M curated SMILES from the PubChem library. Molecular embeddings were obtained from canonicalized SMILES, using the latent space of the autoencoder.

For domain adaptation of these models to the individual virtual libraries, we added a projection head, implemented as a fully connected neural network (one hidden layer of 1024 neurons), on the molecular embeddings. The model was then fine-tuned by training this projection head and a low-rank adaptation (LoRA) module applied to the weights of the original language model. LoRA was used as implemented in the *peft-finetuning* library.[9] Unless otherwise noted, finetuning was performed using the *Adam* optimizer with a learning rate of $5.0 \cdot 10^{-4}$ for 25 epochs. LoRA was applied to the transformer *key* and *query* modules, with LoRA parameters of $r = 8$ and $\alpha = 16$. Fine-tuned embeddings were obtained by pooling embeddings from the fine-tuned encoder.

Here, we considered two different fine-tuning tasks:

- ***Language model finetuning*** was tested on the *MolFormer* encoder, by pre-training with a BERT-style masked-token prediction loss (15%).
- ***Contrastive finetuning*** used a weighted sum of a language modeling loss and a contrastive loss. The language modeling task followed the objective used during the original model training, using the original language modeling head. For contrastive learning, positive and negative pairs were generated through a fingerprint-similarity-based strategy. For each example, a positive pair was sampled from the library as a molecule with a Tanimoto similarity of > 0.7. In case no positive example could be sampled, a randomized version of the respective SMILES string was sampled as a fallback. Negative pairs were generated by sampling a molecule with a Tanimoto similarity of < 0.1. We then computed a batch-based triplet contrastive loss, as implemented by Henderson *et al*.[10] Here, we computed the loss as a weighted sum between the original language modeling loss (0.99) and the contrastive learning loss (0.01).
- ***Supervised finetuning*** was performed by reconstructing molecular fingerprints from the projection head outputs. Morgan fingerprints with radius 2 and length 1024 were used as targets. The model was trained using a mean squared error (MSE) loss for quantifying reproduction quality.

### 1.4 Surrogate Models

Surrogate models were trained on target values that were standardized to zero mean and unit variance.

***Gaussian Process*** models were implemented using *GPyTorch*.[11] All Gaussian Process models used a Matern 5/2 Kernel with automatic relevance detection, except for models trained on molecular fingerprints, which used a Tanimoto Kernel from the *gauche* library.[12] Following the recommendations by Hvarfner *et al.*,[13] Kernel lengthscale was initialized as $\sqrt{d}$, where $d$ is the dimensionality of the molecular representation. Models were trained by minimizing the negative marginal log-likelihood using the *Adam* optimizer with a learning rate of 0.05.

***Feedforward Neural Network Models with Laplace Approximation***: Feedforward neural networks were implemented in *Pytorch*. The networks were trained for up to 1,500 epochs using the *Adam* optimizer with a batch size of 128, a learning rate of $1.0 \cdot 10^{-3}$, a weight decay of $1.0 \cdot 10^{-4}$, and a dropout rate of 0.1. Early stopping on a held-out validation dataset was used with a patience of 50 epochs. Predictive posterior distributions were obtained by fitting a last-layer Laplace approximation after model training, using the *Laplace-Torch* library.[14]

***Random Forest*** models were implemented using *scikit-learn*[15] with 200 estimators and otherwise default hyperparameters. Predictive variances were estimated from the variance of predictions across trees.

### 1.5 Acquisition Functions and Batch Acquisition Policies

After surrogate model training, predictive posteriors were obtained by computing the predictive mean $\mu(x)$ and variance $\sigma^2(x)$ for each unevaluated molecule in the virtual library. These values were used to calculate acquisition function values for each molecule.

Optimization was treated as a maximization problem for all virtual libraries. In this context, we used the following two acquisition functions.

The upper confidence bound (UCB) acquisition function is defined as

$$UCB(x) = \mu(x) + \beta \cdot \sigma(x)$$

where the parameter $\beta$ controls the exploration–exploitation trade-off.

The log expected improvement (LogEI) acquisition function is defined as

$$\log EI(x) = \log\big(z \cdot \Phi(z) + \sigma(x) \cdot \varphi(z)\big)$$

where $z = \frac{\mu(x) - f(x^*) - \beta}{\sigma}$. $\varphi$ and $\Phi$ are the probability density function and the cumulative density function of the normal distribution, respectively. Again, the parameter $\beta$ controls the exploration–exploitation trade-off.

With these acquisition function values for each molecule, the following policies for batch acquisition were used.

***Top-K Acquisition*** selects a batch of candidates by sorting all candidates by their acquisition function value, and selecting the *K* candidates with the largest values.[16]

***Ensemble Acquisition*** generates an ensemble of *K* acquisition functions using *K* different values of the exploration parameter $\beta$, equally spaced between 0 and $\beta_{\mathrm{max}}$. For each value of $\beta$, the candidate that maximizes the corresponding acquisition function is selected for the next batch.[17] After acquisition of a candidate molecule, the selected molecule was removed from the pool of unevaluated molecules before proceeding to the next acquisition function.

## 2 Molecular Library Distributions

In this work, we evaluated a total of six virtual libraries from the literature that contained ground-truth target property labels for all candidates. For all molecular libraries, we performed a systematic analysis of the underlying molecule distribution using Morgan Fingerprints (radius 2, length 2048, details see above) and all transformer encoders described in section 1.3. Specifically, we quantify the dispersion of within-library similarities based on a sample of 5M random pairs, using Tanimoto similarities for molecular fingerprints and cosine similarities for transformer embeddings.

To investigate the relation to established libraries of drug-like molecules, we sampled a reference set of 25,000 molecules from the ZINC15 library. Similarity of the target virtual library to this reference library was then investigated in three complementary ways:

- A *Uniform Manifold Approximation* (UMAP) model[18] was fitted on the molecular fingerprints (Morgan Fingerprints, radius 2, length 2048, details see above) for all molecules in the reference library. The molecular fingerprints for all molecules in the library of interest were then projected into the same UMAP embedding space. All UMAP models were built using the *umap* package.[19]
- A *T-Distributed Stochastic Neighbor Embedding* (*t*-SNE) model[20] was fitted on the molecular fingerprints (Morgan fingerprints, radius 2, length 2048, details see above) for all molecules in the reference library. The molecular fingerprints for all molecules in the library of interest were then projected into the same *t*-SNE embedding space. All *t*-SNE models were built using the *openTSNE* library.[21]
- We sampled 5M random pairs of molecules from the target library and the reference library, respectively, and quantified molecular similarity as described above.

For all libraries, the histograms of within-library and to-reference similarity, the dimensionality-reduced projections, and the histogram of target properties, are shown in Figures S1–S24. Characteristics of the respective distributions are given in Tables S1–S6.

## 2.1 1.9M Molecules from the Zinc Library Docked against AmpC (*Zinc-AmpC*)

From the original dataset by Shoichet and co-workers, who docked a sample from the Zinc15 virtual library (~96M molecules) against AmpC β-lactamase,[22] we selected a random subsample of 1,914,566 molecules, which corresponds to approx. 2% of the full library.

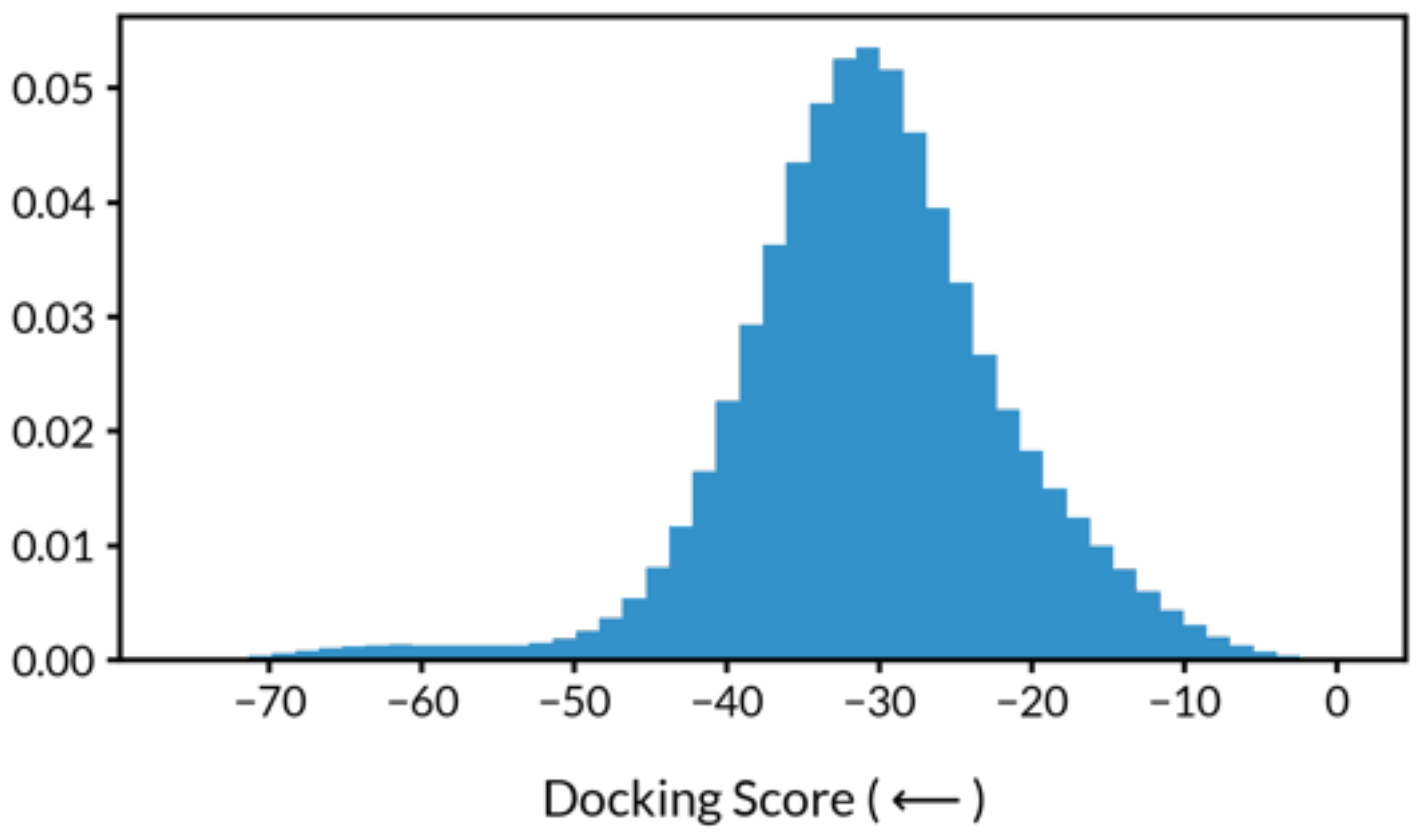

**Figure S 1**: Histogram of target properties (docking scores) for the *Zinc-AmpC* library.

**Table S 1**: Descriptive statistics of the similarity distributions of the *Zinc-AmpC* library for different molecular embeddings

| | ***Within Library Similarities*** | | | | ***Similarities To Reference*** | | | |
|---|---|---|---|---|---|---|---|---|
| | Mean | SD | Median | IQR | Mean | SD | Median | IQR |
| Morgan Fingerprint | 0.867 | 0.042 | 0.870 | 0.054 | 0.869 | 0.040 | 0.871 | 0.052 |
| ChemBERTa | 0.275 | 0.098 | 0.263 | 0.130 | 0.274 | 0.100 | 0.260 | 0.133 |
| MolFormer | 0.368 | 0.051 | 0.367 | 0.068 | 0.393 | 0.056 | 0.392 | 0.074 |
| SmiTed | 0.006 | 0.002 | 0.006 | 0.002 | 0.042 | 0.007 | 0.041 | 0.009 |
| T5Chem | 0.333 | 0.095 | 0.325 | 0.129 | 0.338 | 0.099 | 0.328 | 0.135 |

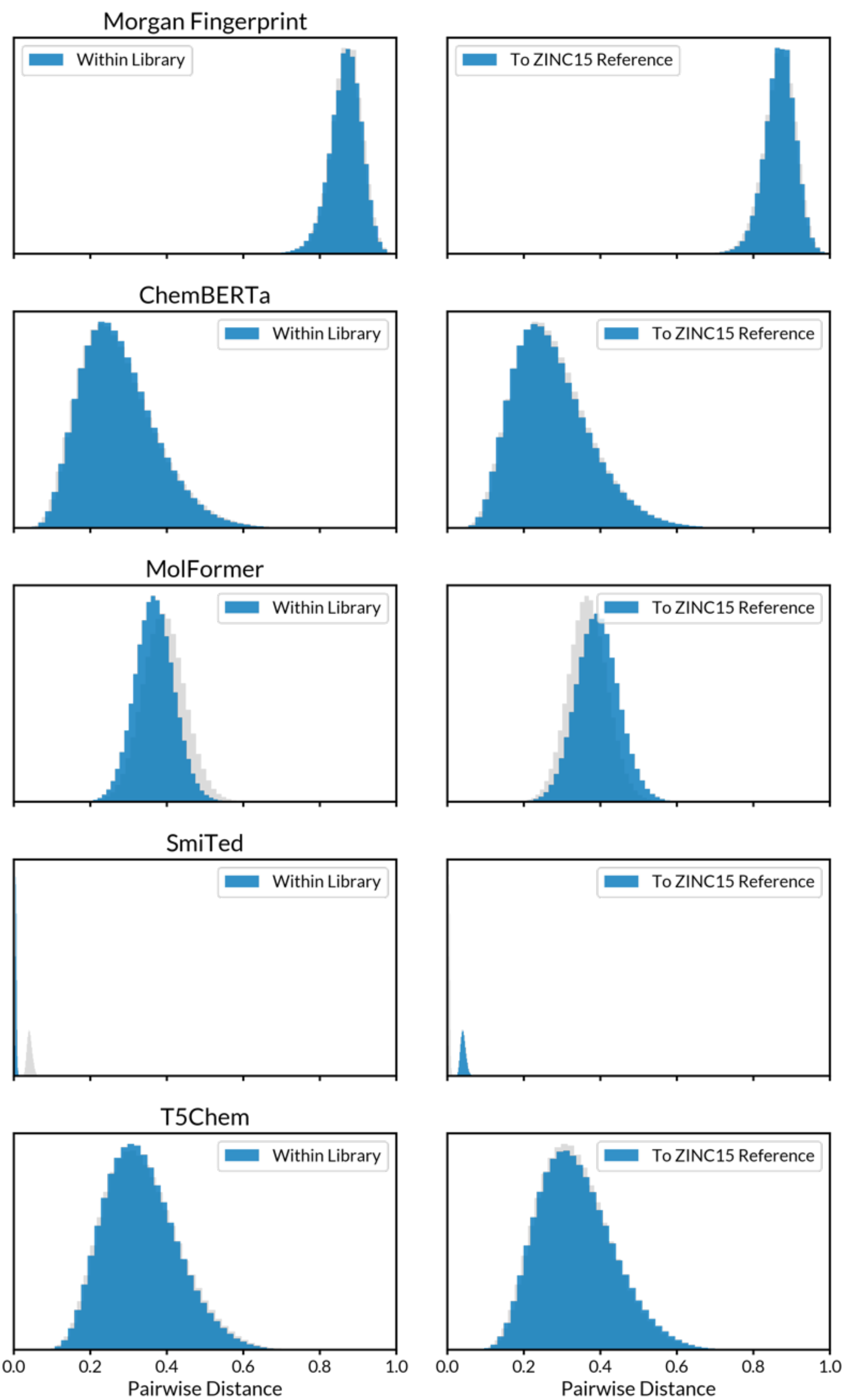

**Figure S 2**: Histograms of embedding similarities for the *Zinc-AmpC* library. Left column: within-library similarities. Right column: to-reference similarities.

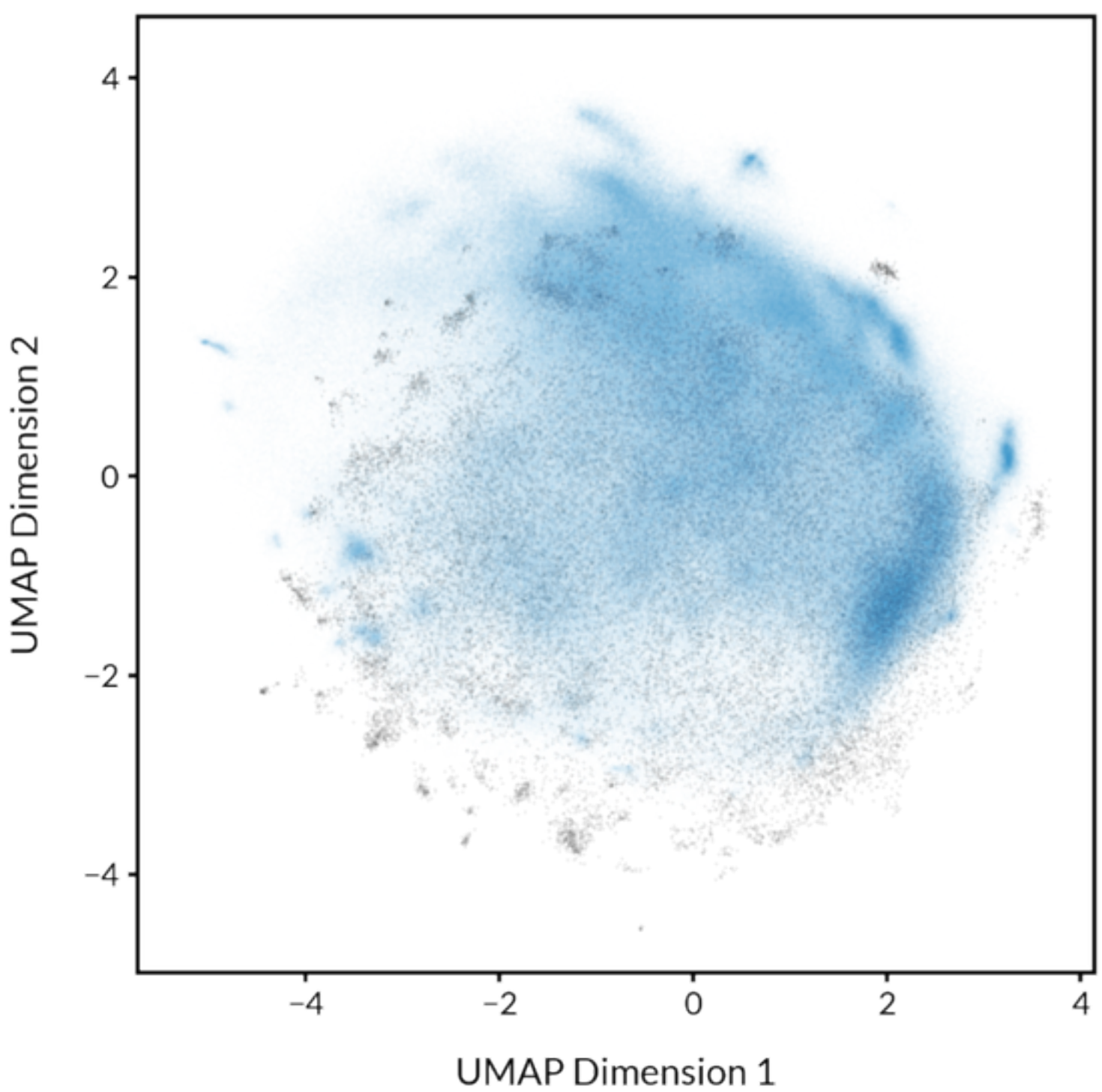

**Figure S 3**: Projection of molecular fingerprints of all candidate molecules from the *Zinc-AmpC* library into the UMAP coordinates trained on the fingerprints from the reference sample of 25k molecules from the ZINC15 library.

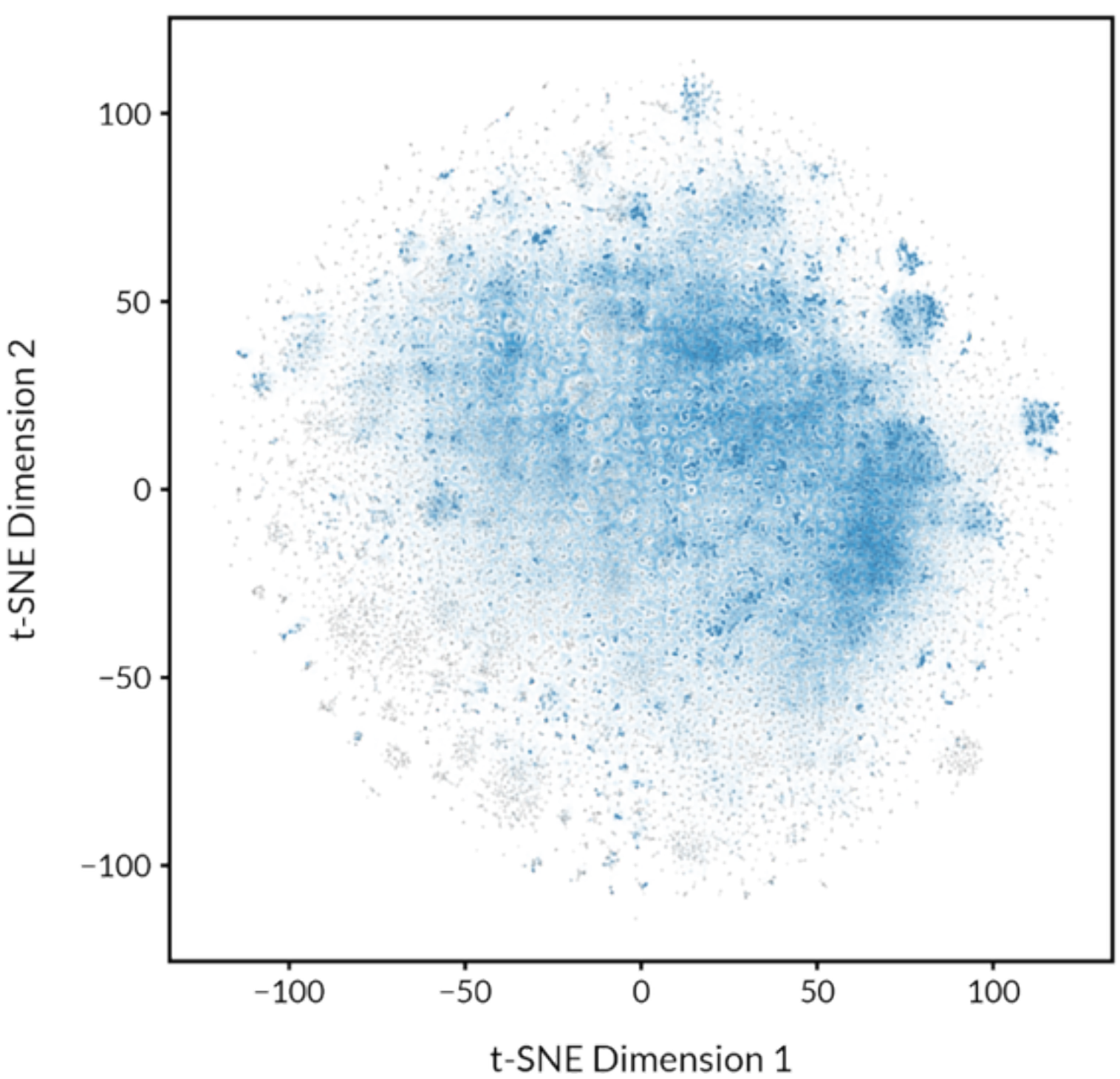

**Figure S 4**: Projection of molecular fingerprints of all candidate molecules from the *Zinc-AmpC* library into the *t*-SNE coordinates trained on the fingerprints from the reference sample of 25k molecules from the ZINC15 library.

### 2.2 1.5M Molecules from the RO4 Library Docked against OGG1 (*RO4-OGG1*)

From the original dataset by Luttens *et al.*,[23] who docked a sample from the Zinc15 virtual library that was filtered by "Rule-of-Four" criteria (~14,7M molecules) against oxoguanine glycosylase (OGG1), we selected a random subset of 1,470,000 molecules, which corresponds to approx. 10% of the full library.

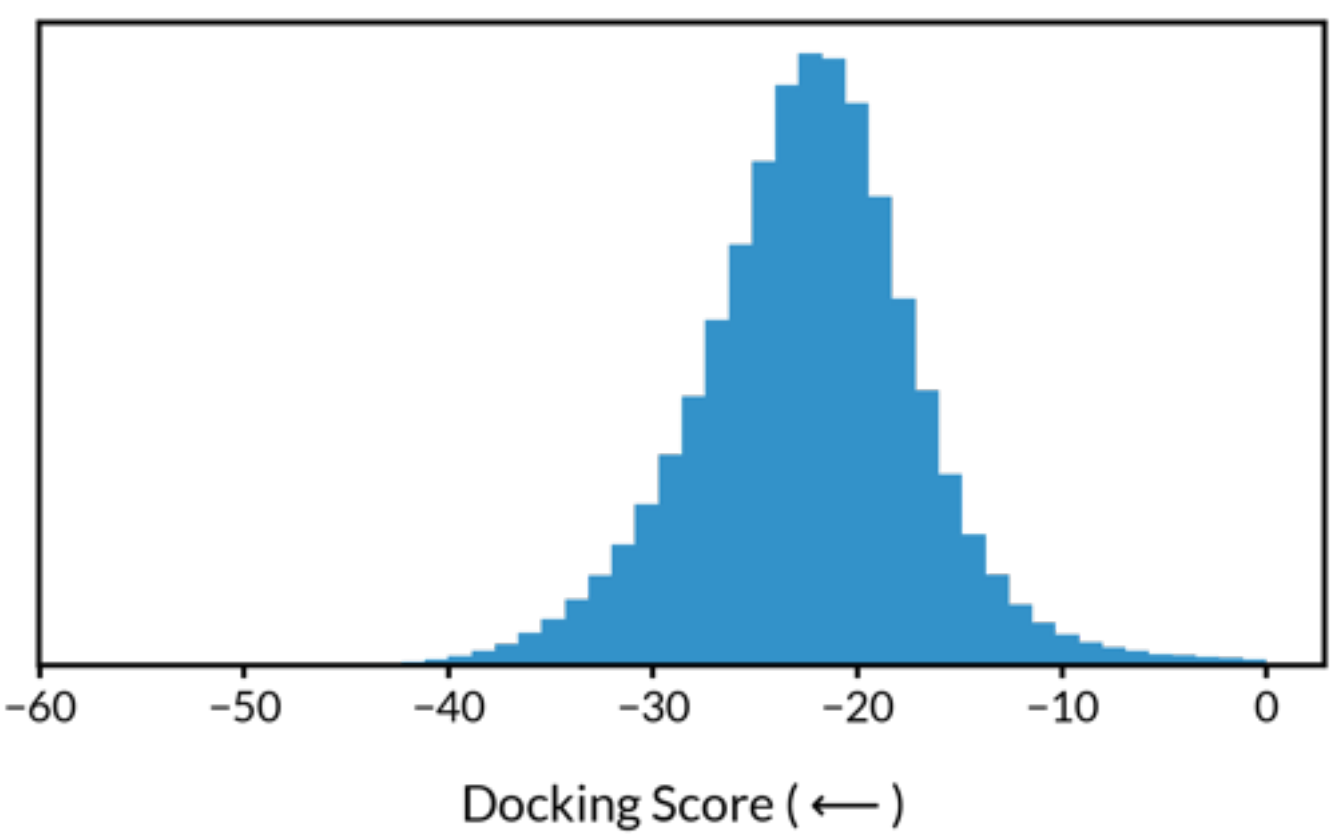

**Figure S 5**: Histogram of target properties (docking scores) for the *RO4-OGG1* library.

**Table S 2**: Descriptive statistics of the similarity distributions of the *RO4-OGG1* library for different molecular embeddings

| | ***Within Library Similarities*** | | | | ***Similarities To Reference*** | | | |
|---|---|---|---|---|---|---|---|---|
| | Mean | SD | Median | IQR | Mean | SD | Median | IQR |
| Morgan Fingerprint | 0.853 | 0.039 | 0.856 | 0.049 | 0.862 | 0.039 | 0.864 | 0.049 |
| ChemBERTa | 0.228 | 0.078 | 0.217 | 0.100 | 0.270 | 0.090 | 0.258 | 0.117 |
| MolFormer | 0.339 | 0.071 | 0.335 | 0.095 | 0.402 | 0.065 | 0.401 | 0.085 |
| SmiTed | 0.005 | 0.001 | 0.005 | 0.002 | 0.042 | 0.007 | 0.042 | 0.009 |
| T5Chem | 0.263 | 0.082 | 0.253 | 0.109 | 0.324 | 0.094 | 0.315 | 0.127 |

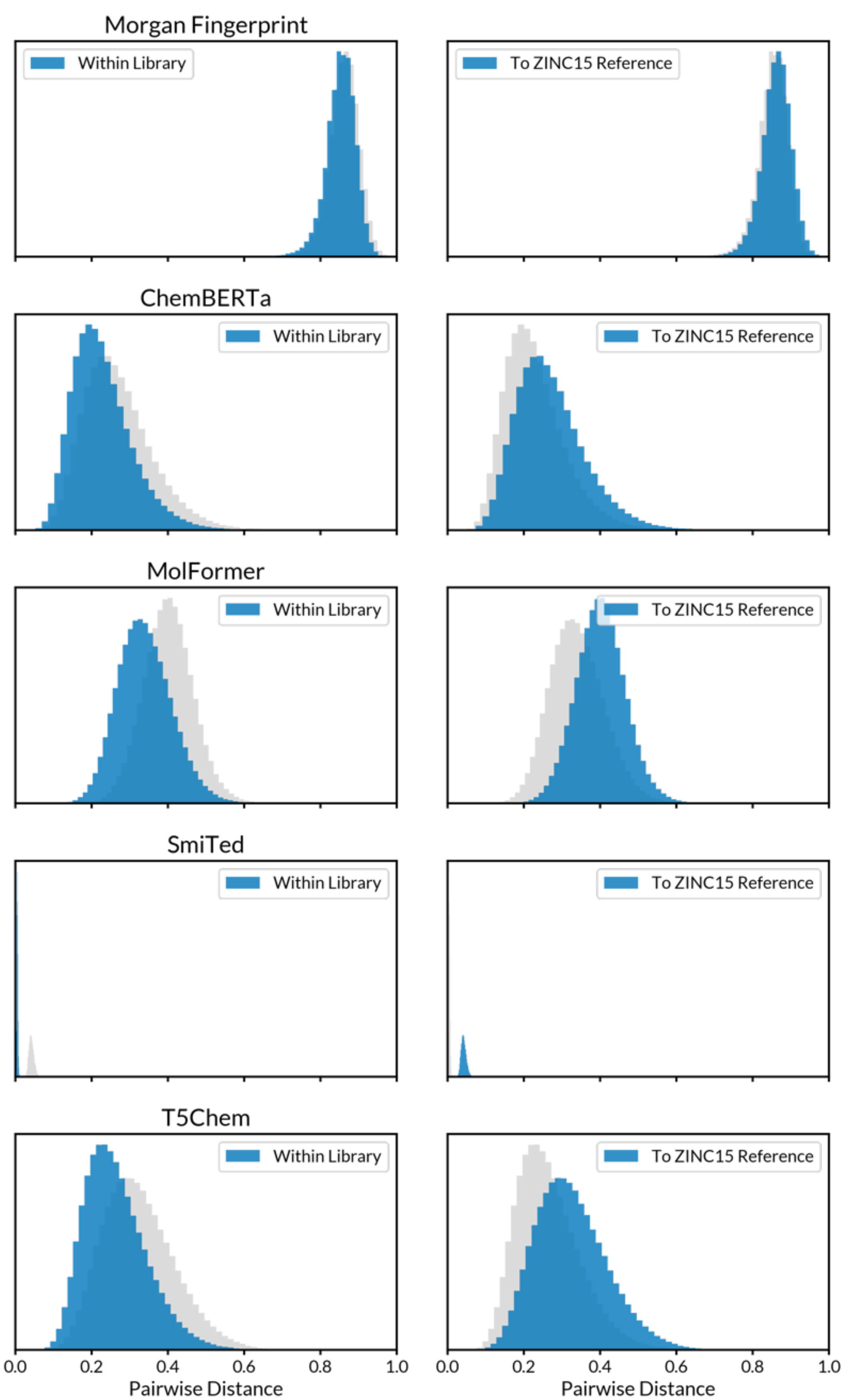

**Figure S 6**: Histograms of embedding similarities for the *RO4-OGG1* library. Left column: within-library similarities. Right column: to-reference similarities.

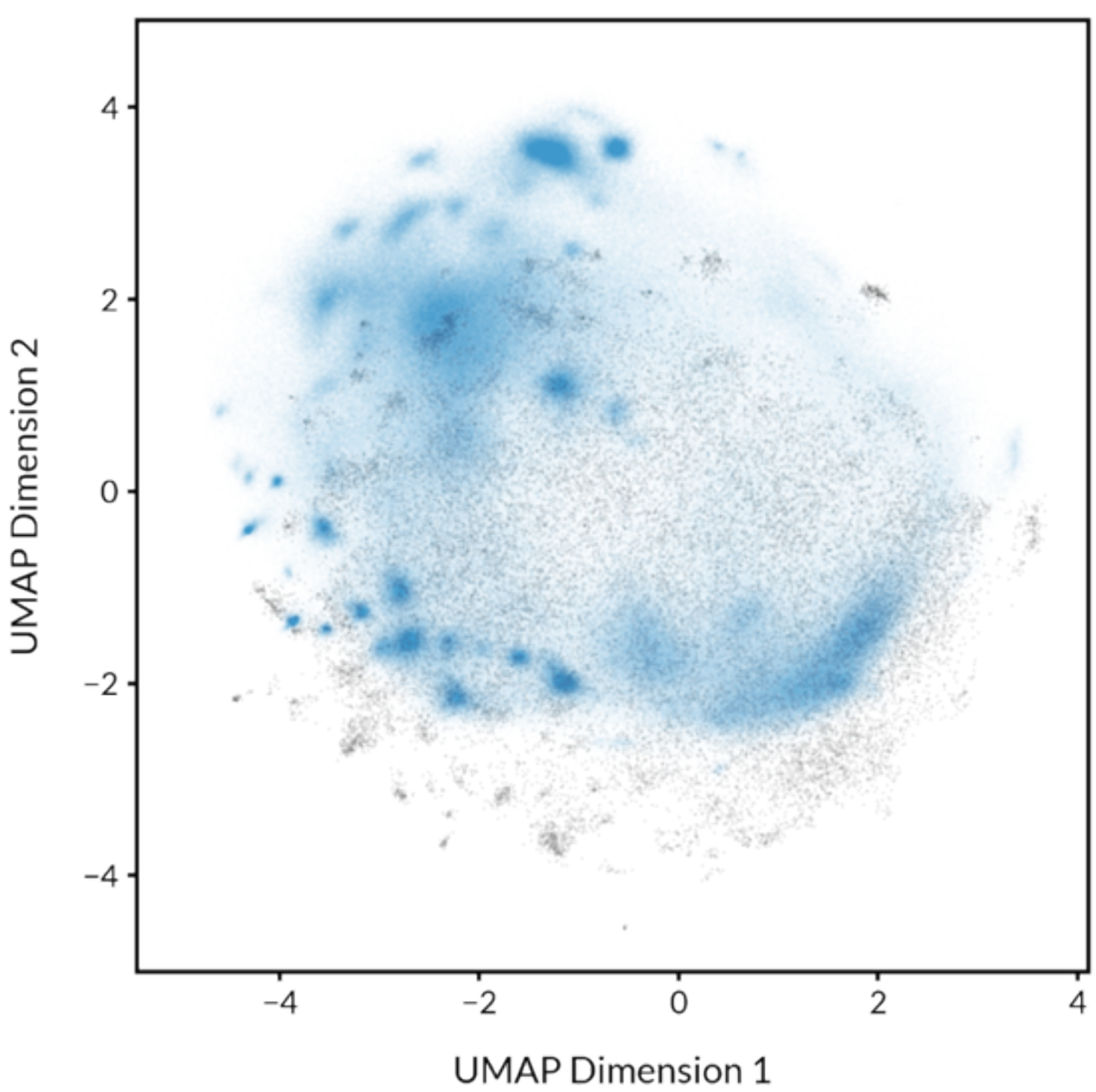

**Figure S 7**: Projection of molecular fingerprints of all candidate molecules from the *RO4-OGG1* library into the UMAP coordinates trained on the fingerprints from the reference sample of 25k molecules from the ZINC15 library.

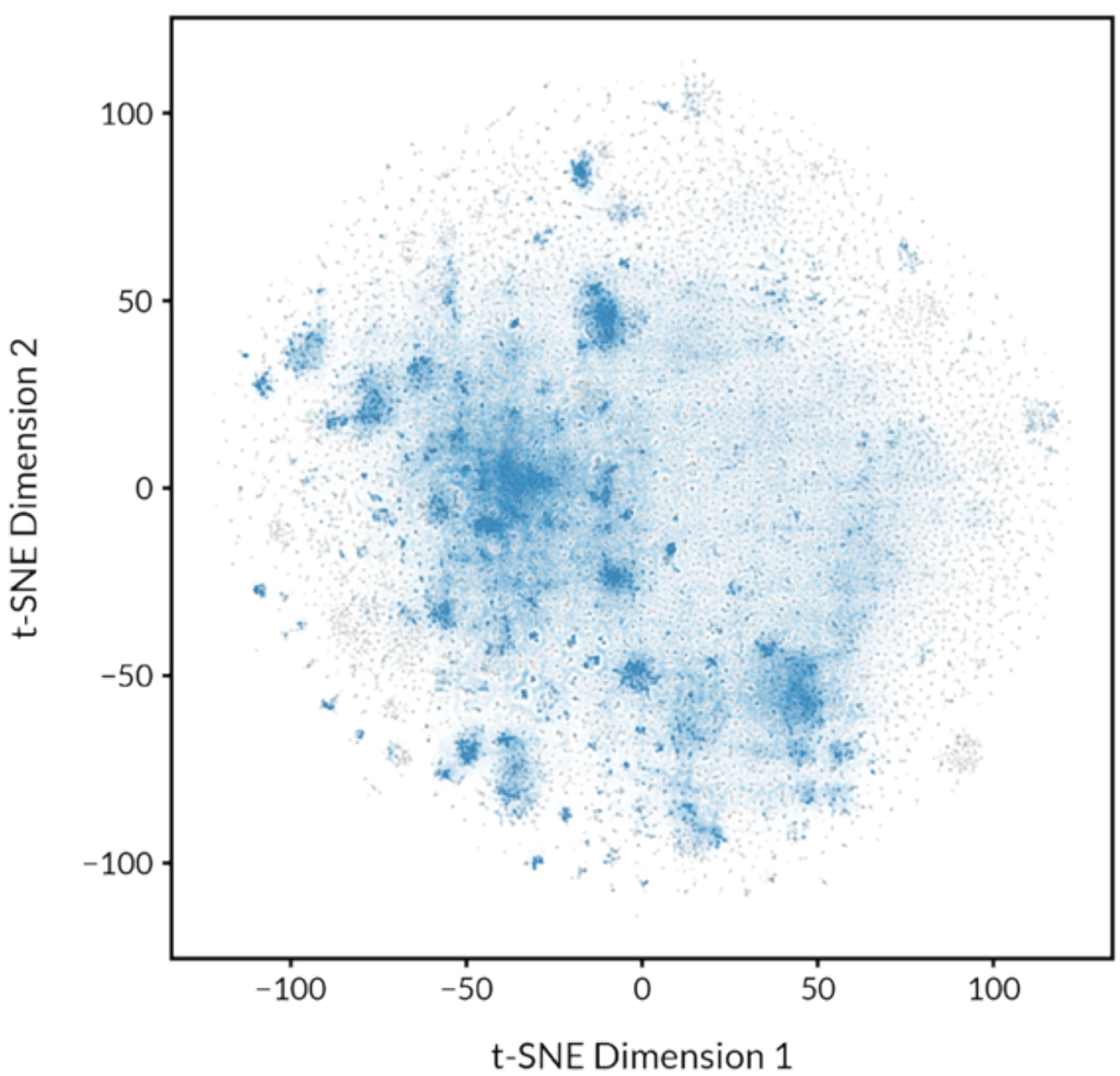

**Figure S 8**: Projection of molecular fingerprints of all candidate molecules from the *RO4-OGG1* library into the *t*-SNE coordinates trained on the fingerprints from the reference sample of 25k molecules from the ZINC15 library.

### 2.3 2.1M Molecules from the Enamine-HTS Collection Docked against TMPK (*Enamine-TMPK*)

The Enamine-HTS library, containing a total of 2,104,318 molecules, docked against a thymidylate kinase (TMPK) by Coley and co-workers.[16]

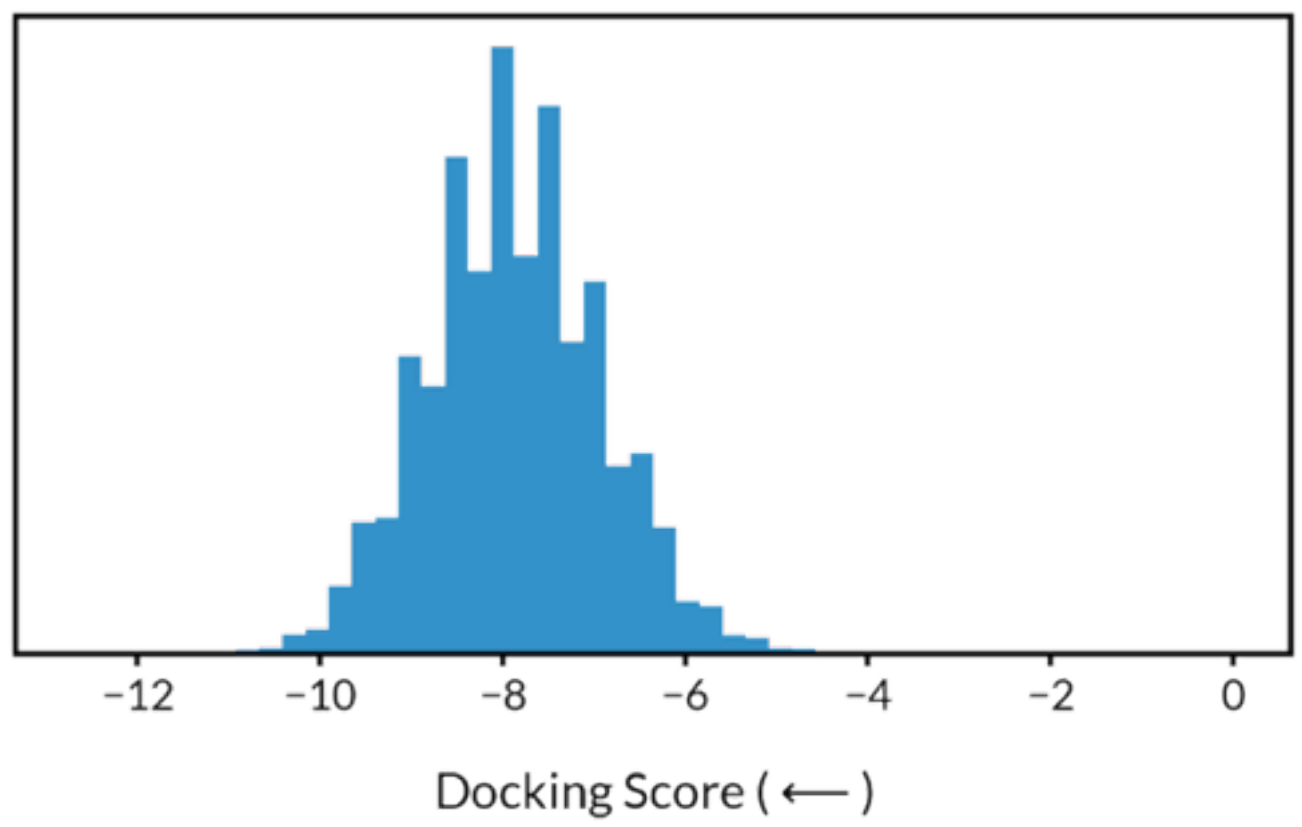

**Figure S 9**: Histogram of target properties (docking scores) for the *Enamine-TMPK* library.

**Table S 3**: Descriptive statistics of the similarity distributions of the *Enamine-TMPK* library for different molecular embeddings

| | ***Within Library Similarities*** | | | | ***Similarities To Reference*** | | | |
|---|---|---|---|---|---|---|---|---|
| | Mean | SD | Median | IQR | Mean | SD | Median | IQR |
| Morgan Fingerprint | 0.867 | 0.043 | 0.871 | 0.053 | 0.878 | 0.039 | 0.880 | 0.051 |
| ChemBERTa | 0.315 | 0.106 | 0.301 | 0.138 | 0.334 | 0.100 | 0.323 | 0.132 |
| MolFormer | 0.333 | 0.063 | 0.332 | 0.084 | 0.418 | 0.070 | 0.415 | 0.092 |
| SmiTed | 0.006 | 0.002 | 0.006 | 0.002 | 0.042 | 0.008 | 0.041 | 0.010 |
| T5Chem | 0.309 | 0.103 | 0.296 | 0.132 | 0.383 | 0.106 | 0.373 | 0.144 |

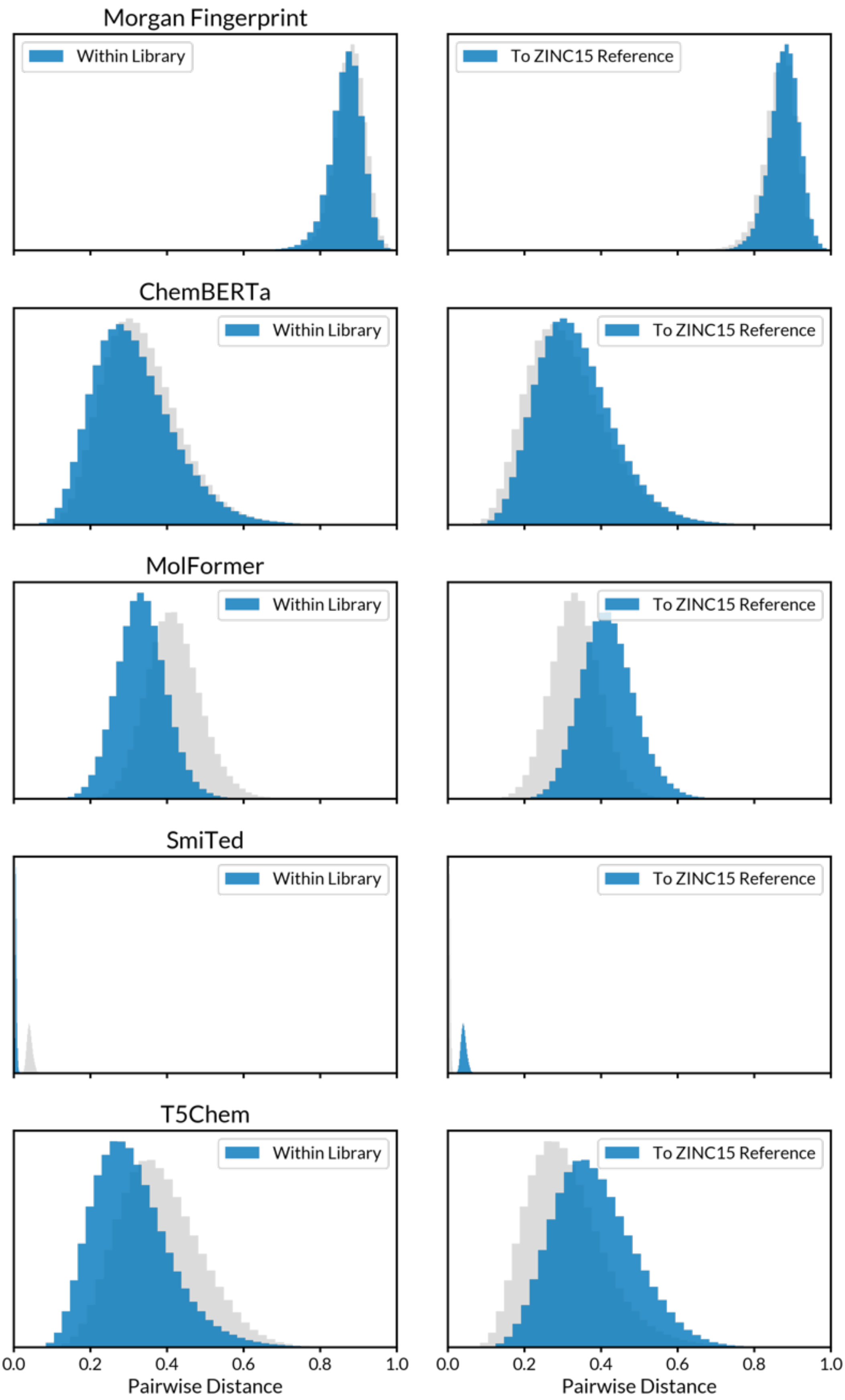

**Figure S 10**: Histograms of embedding similarities for the *Enamine-TMPK* library. Left column: within-library similarities,. Right column: to-reference similarities.

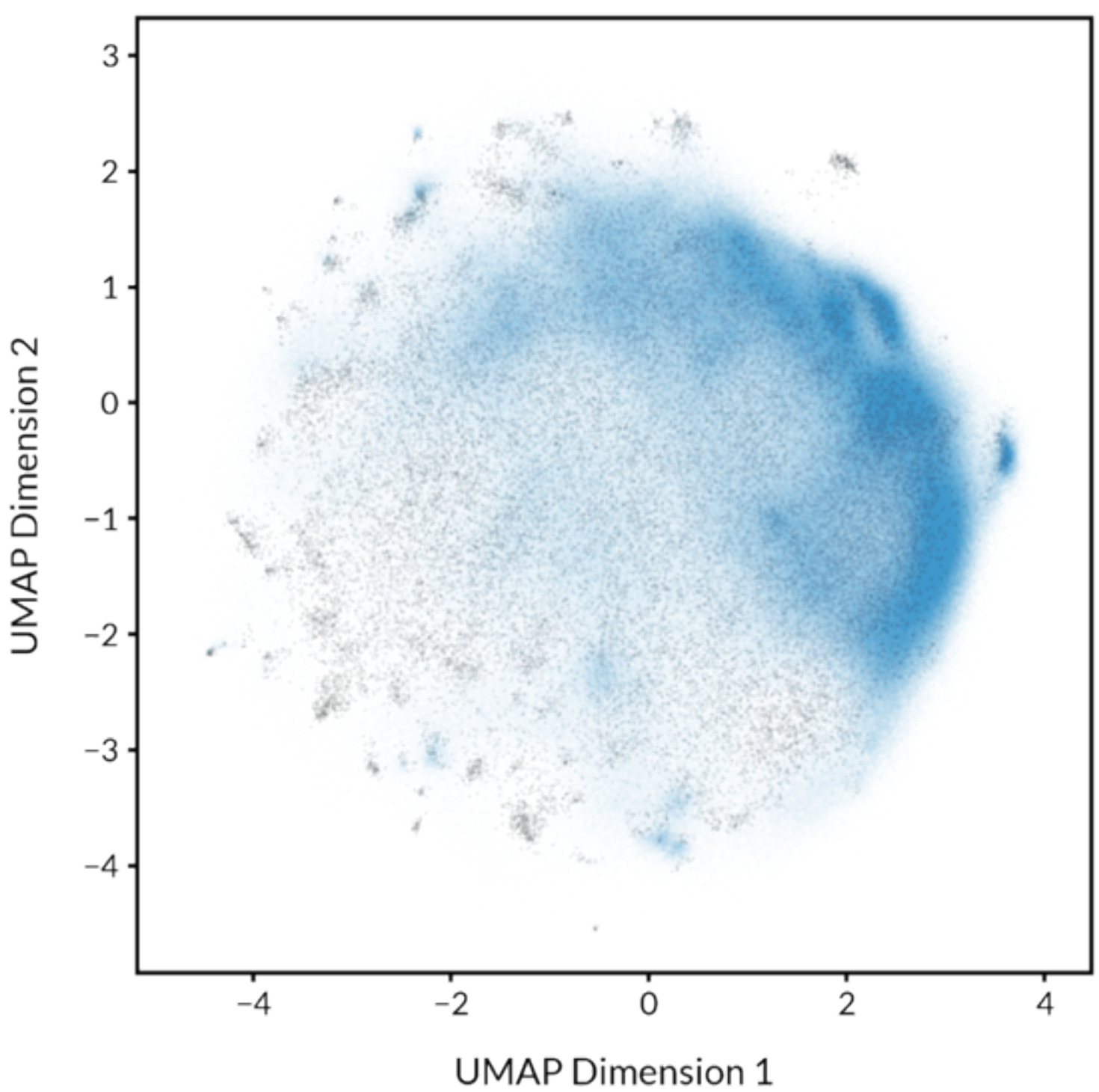

**Figure S 11**: Projection of molecular fingerprints of all candidate molecules from the *Enamine-TMPK* library into the UMAP coordinates trained on the fingerprints from the reference sample of 25k molecules from the ZINC15 library.

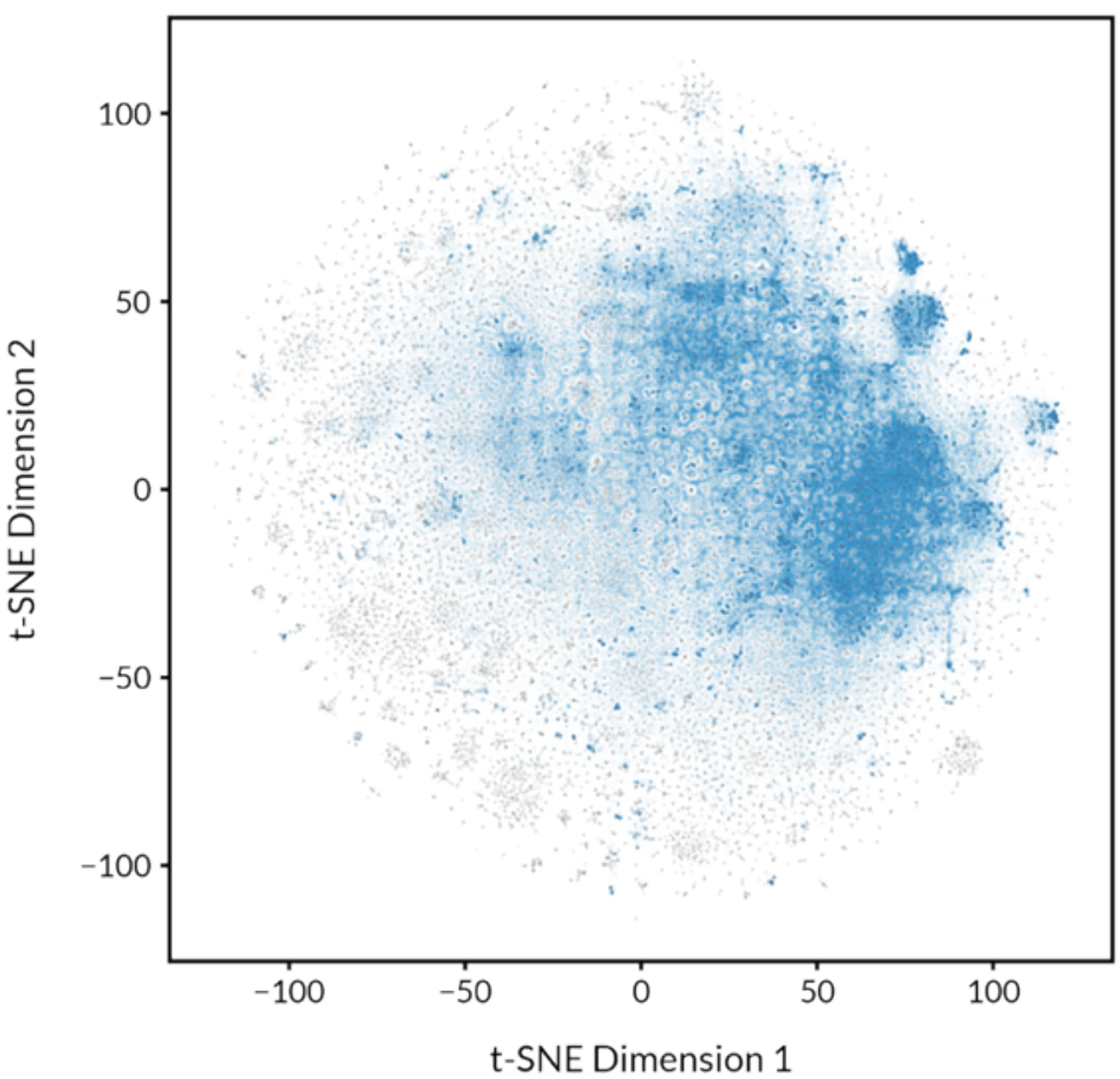

**Figure S 12**: Projection of molecular fingerprints of all candidate molecules from the *Enamine-TMPK* library into the *t*-SNE coordinates trained on the fingerprints from the reference sample of 25k molecules from the ZINC15 library.

## 2.4 2.3M Molecules from the Harvard Clean Energy Project (*HCEP-PCE*)

The dataset of 2,320,648 molecules from the Harvard Clean Energy Project,[24] along with DFT-computed power conversion efficiencies (PCE) for use in organic solar cells.

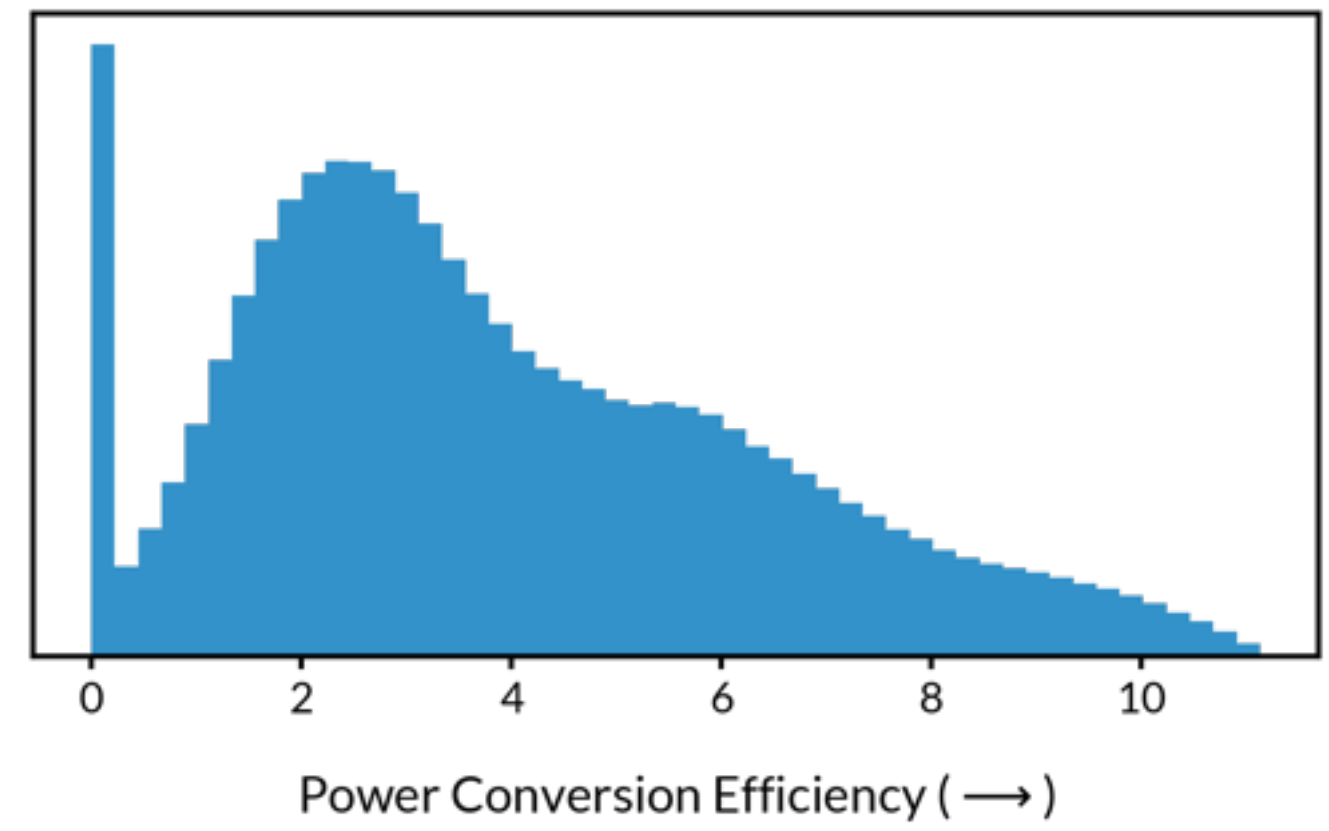

**Figure S 13**: Histogram of target properties (docking scores) for the *HCEP-PCE* library.

**Table S 4**: Descriptive statistics of the similarity distributions of *HCEP-PCE* library for different molecular embeddings

| | ***Within Library Similarities*** | | | | ***Similarities To Reference*** | | | |
|---|---|---|---|---|---|---|---|---|
| | Mean | SD | Median | IQR | Mean | SD | Median | IQR |
| Morgan Fingerprint | 0.837 | 0.064 | 0.848 | 0.081 | 0.946 | 0.027 | 0.947 | 0.036 |
| ChemBERTa | 0.232 | 0.096 | 0.220 | 0.131 | 0.566 | 0.109 | 0.570 | 0.150 |
| MolFormer | 0.136 | 0.041 | 0.132 | 0.055 | 0.509 | 0.090 | 0.517 | 0.130 |
| SmiTed | 0.008 | 0.003 | 0.008 | 0.004 | 0.036 | 0.006 | 0.035 | 0.008 |
| T5Chem | 0.139 | 0.060 | 0.129 | 0.073 | 0.589 | 0.115 | 0.590 | 0.157 |

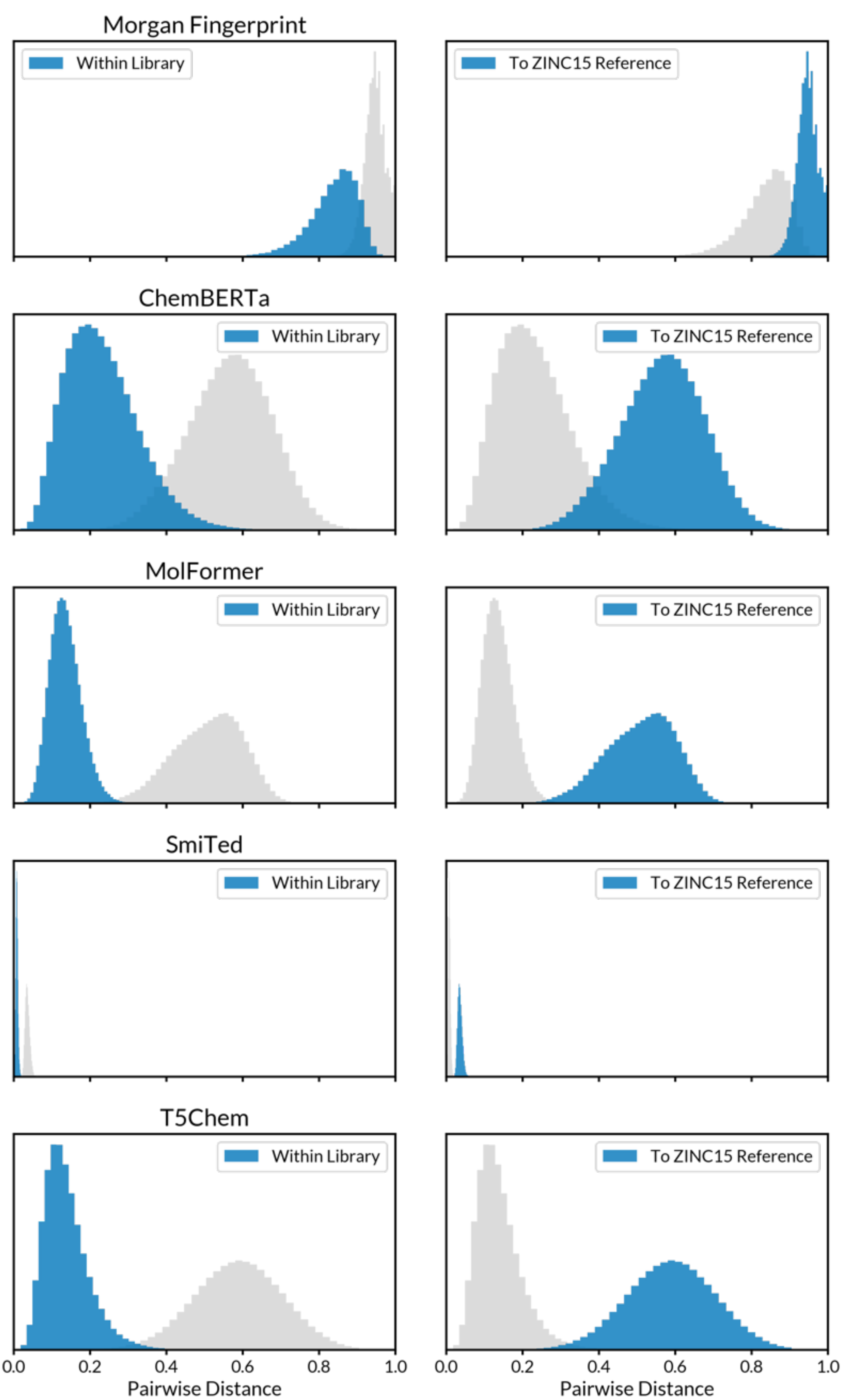

**Figure S 14**: Histograms of embedding similarities for the *HCEP-PCE* library. Left column: within-library similarities. Right column: to-reference similarities.

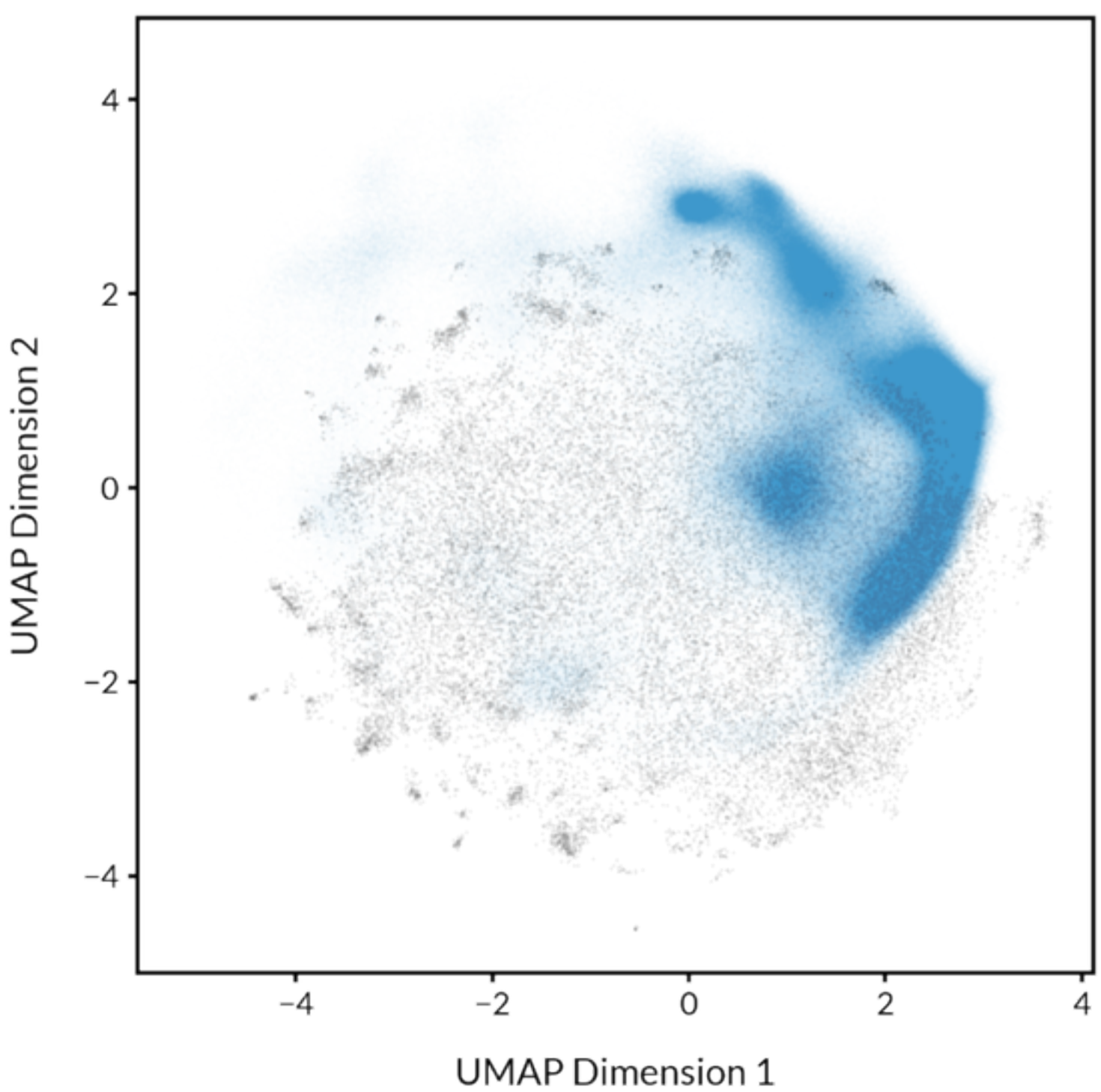

**Figure S 15**: Projection of molecular fingerprints of all candidate molecules from the *HCEP-PCE* library into the UMAP coordinates trained on the fingerprints from the reference sample of 25k molecules from the ZINC15 library.

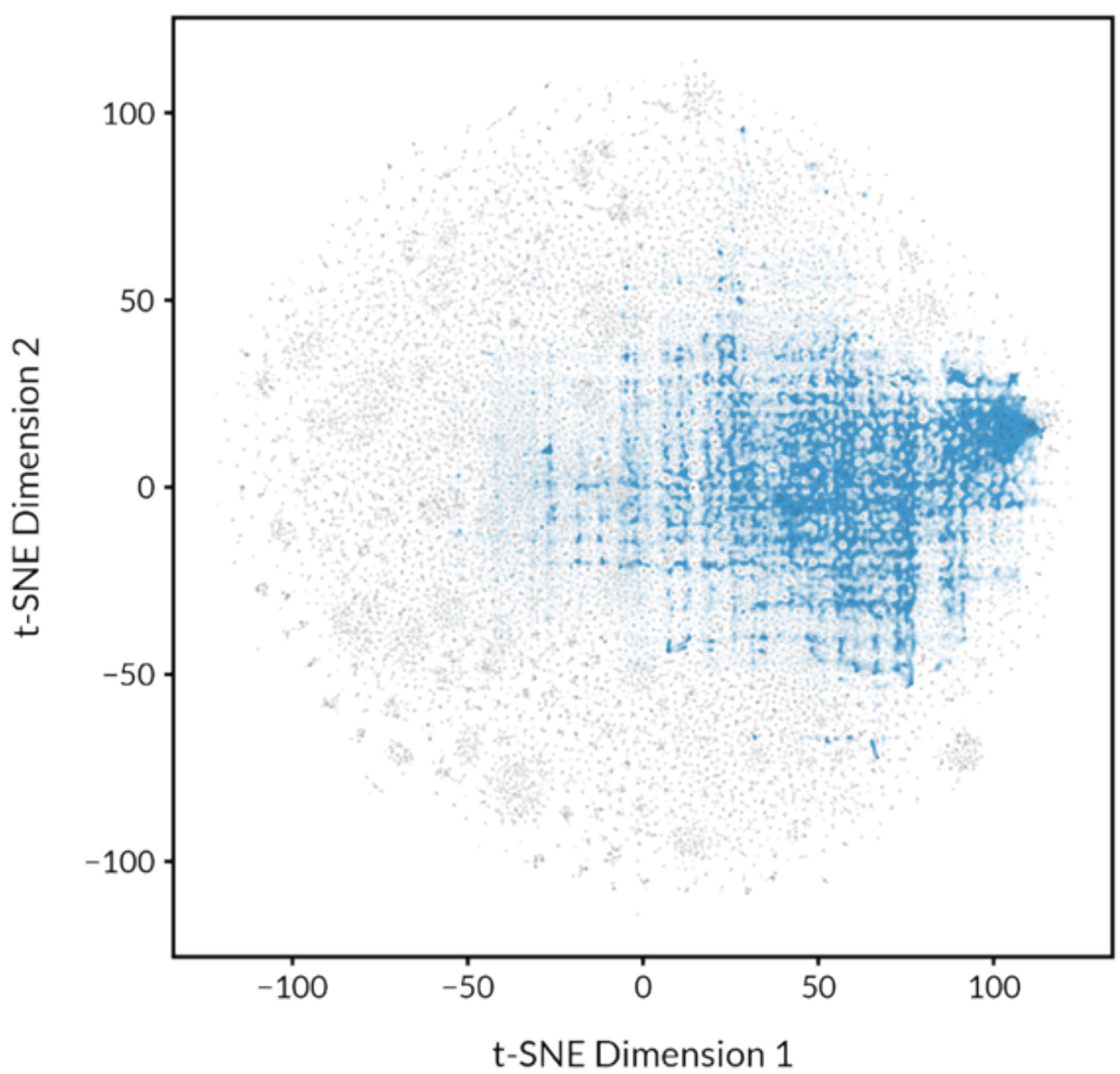

**Figure S 16**: Projection of molecular fingerprints of all candidate molecules from the *HCEP-PCE* library into the *t*-SNE coordinates trained on the fingerprints from the reference sample of 25k molecules from the ZINC15 library.

## 2.5 180k Organic Laser Candidates with Computed Oscillator Strengths (*Laser-OSC*)

The library of 182,888 conjugated organic molecules for use as thin-film organic laser emitters,[25] along with DFT-computed emission oscillator strengths (OSC).

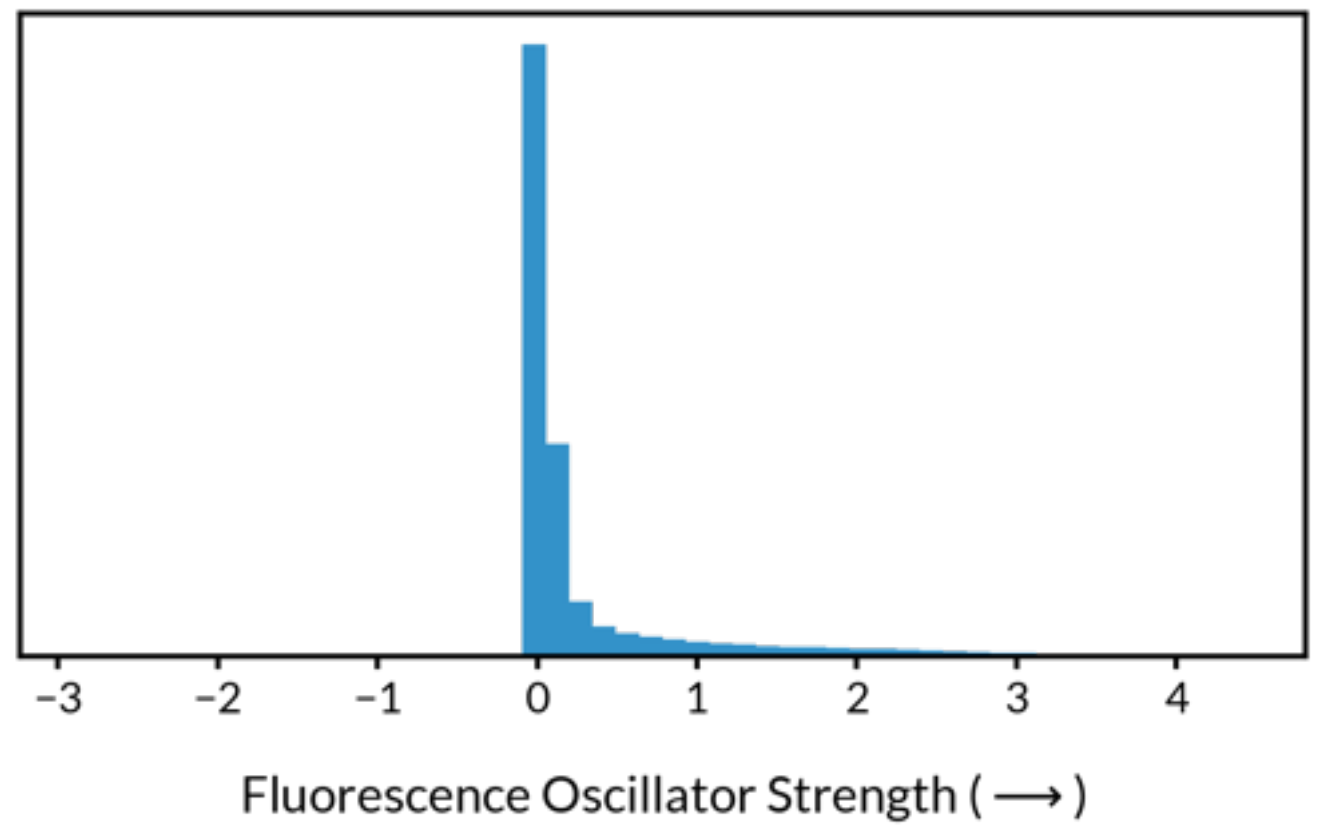

**Figure S 17**: Histogram of target properties (docking scores) for the *Laser-OSC* library.

**Table S 5**: Descriptive statistics of the similarity distributions of the *Laser-OSC* library for different molecular embeddings

| | ***Within Library Similarities*** | | | | ***Similarities To Reference*** | | | |
|---|---|---|---|---|---|---|---|---|
| | Mean | SD | Median | IQR | Mean | SD | Median | IQR |
| Morgan Fingerprint | 0.803 | 0.083 | 0.825 | 0.093 | 0.923 | 0.034 | 0.926 | 0.046 |
| ChemBERTa | 0.151 | 0.067 | 0.140 | 0.084 | 0.493 | 0.095 | 0.492 | 0.129 |
| MolFormer | 0.181 | 0.057 | 0.176 | 0.074 | 0.482 | 0.066 | 0.486 | 0.089 |
| SmiTed | 0.036 | 0.053 | 0.021 | 0.021 | 0.041 | 0.038 | 0.033 | 0.018 |
| T5Chem | 0.109 | 0.047 | 0.104 | 0.063 | 0.511 | 0.104 | 0.508 | 0.144 |

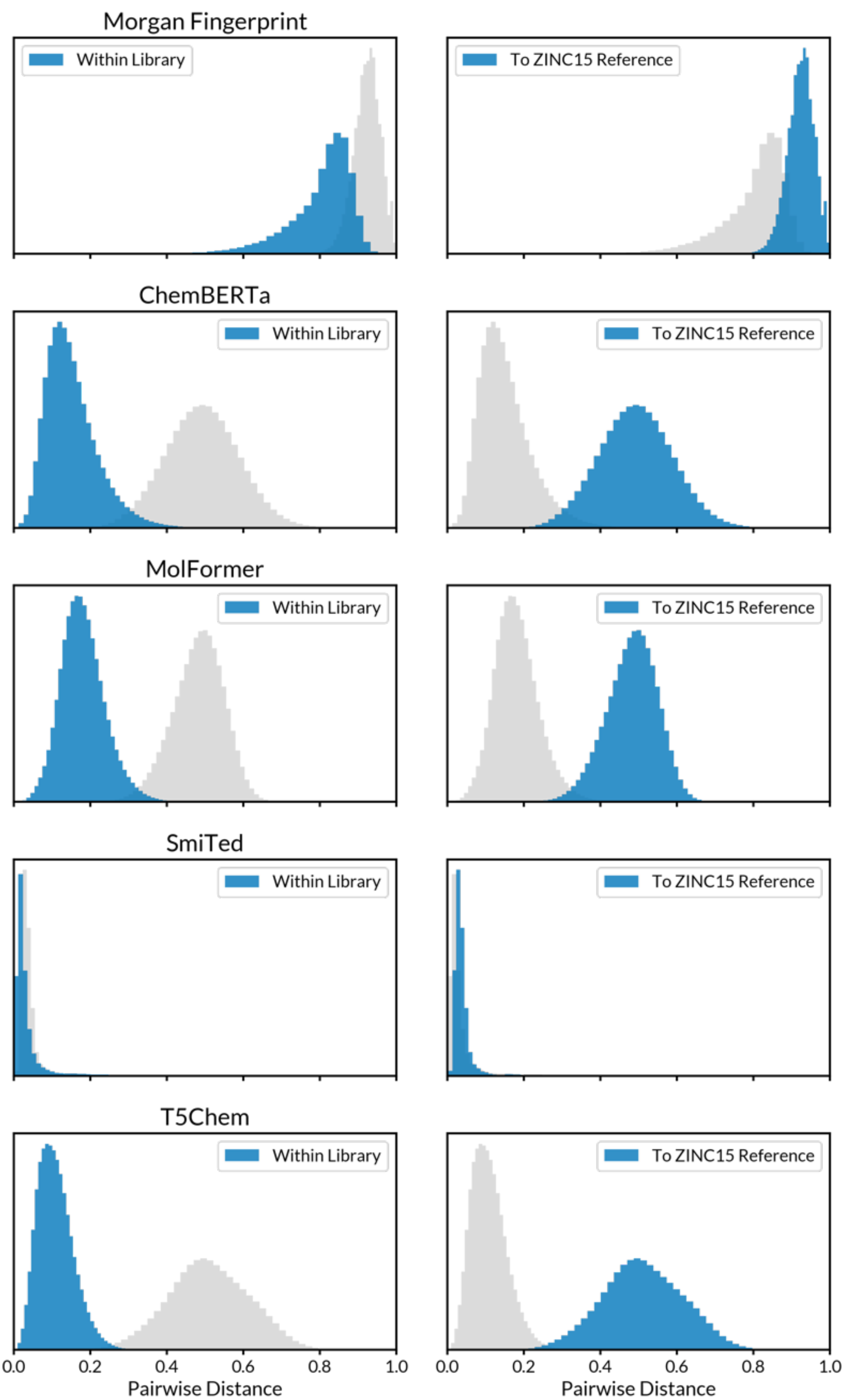

**Figure S 18**: Histograms of embedding similarities for the *Laser-OSC* library. Left column: within-library similarities. Right column: to-reference similarities.

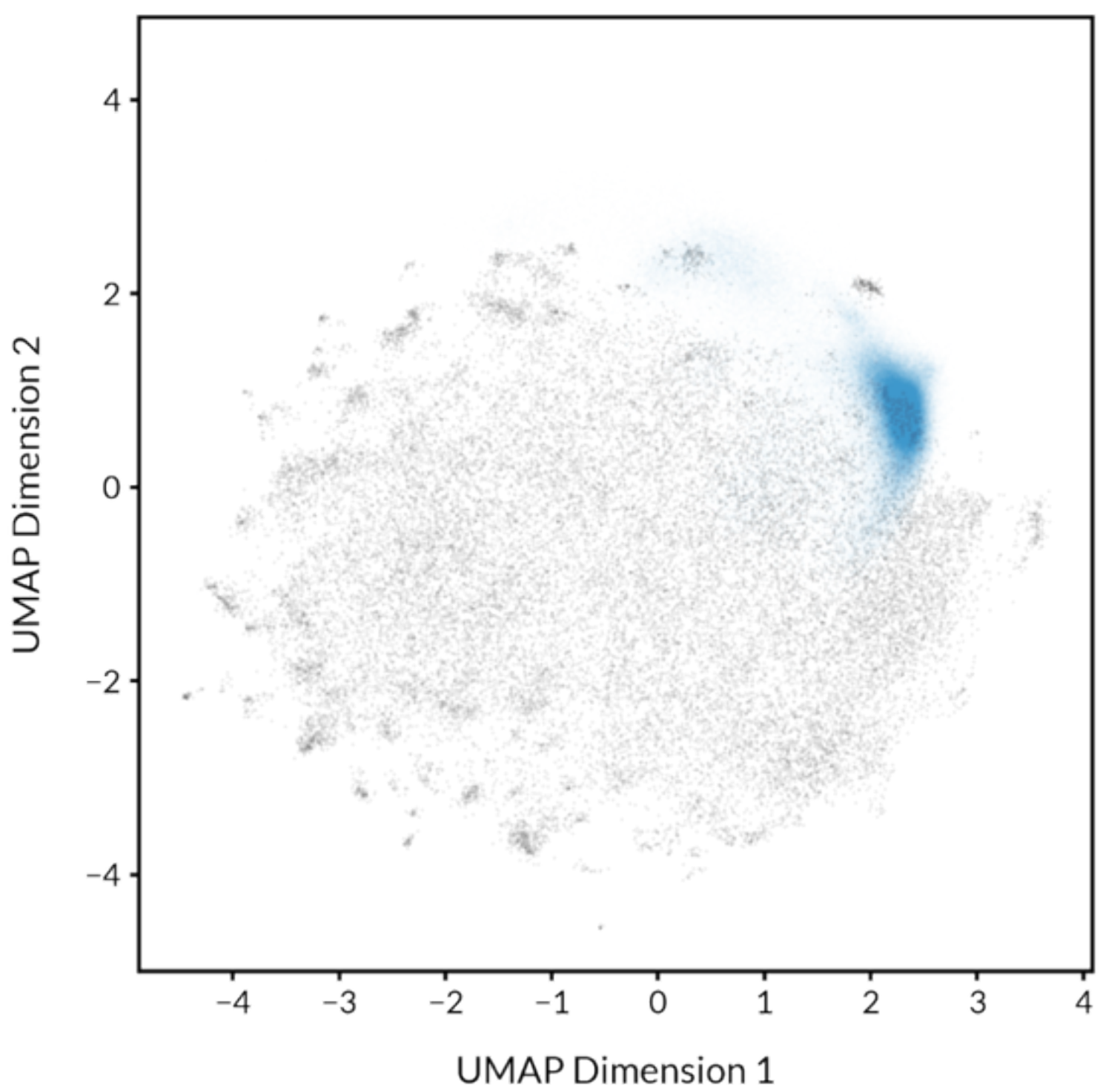

**Figure S 19**: Projection of molecular fingerprints of all candidate molecules from the *Laser-OSC* library into the UMAP coordinates trained on the fingerprints from the reference sample of 25k molecules from the ZINC15 library.

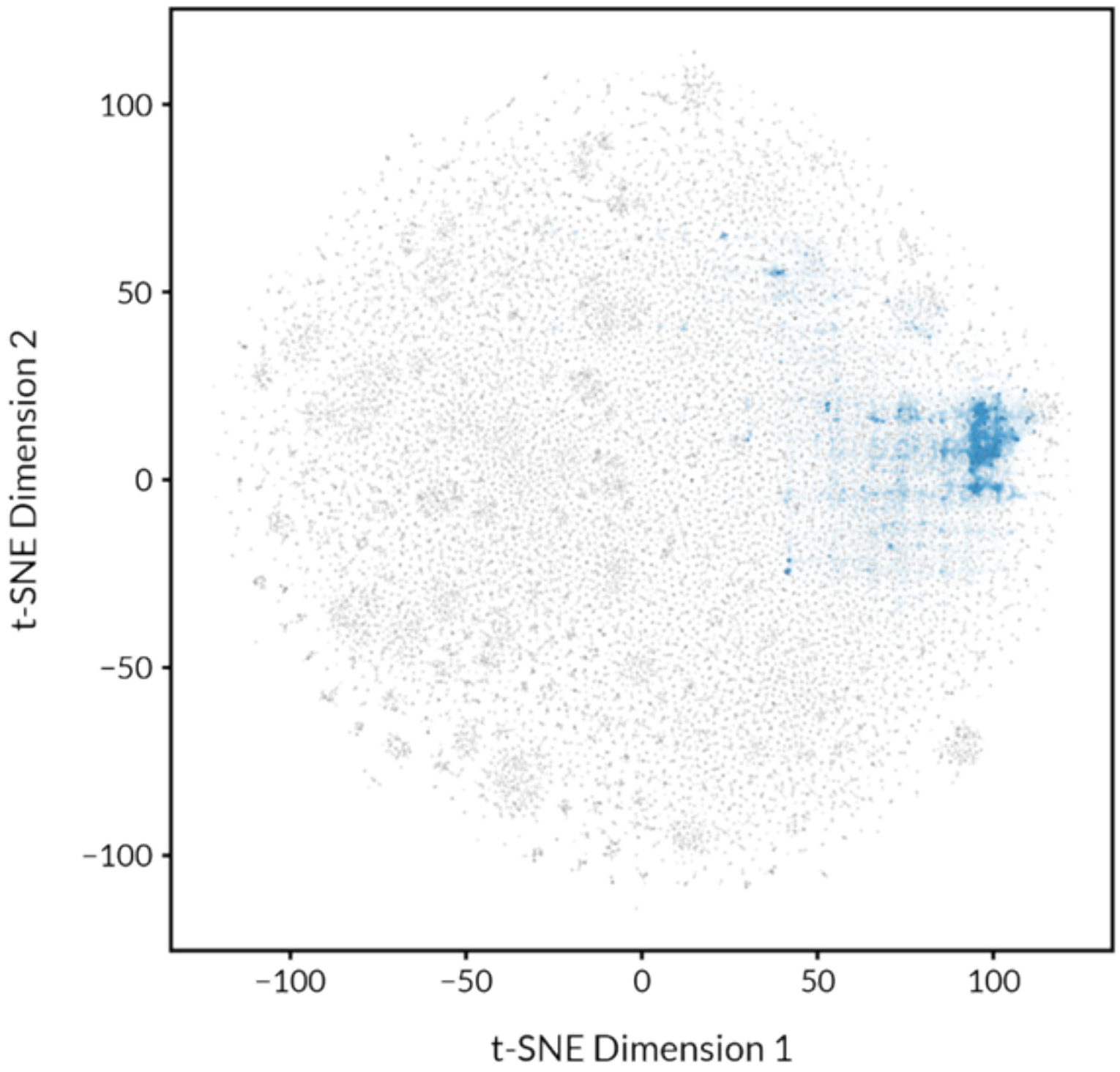

**Figure S 20**: Projection of molecular fingerprints of all candidate molecules from the *Laser-OSC* library into the *t*-SNE coordinates trained on the fingerprints from the reference sample of 25k molecules from the ZINC15 library.

## 2.6 300k Phosphine Ligands with Predicted Enantioselectivities (*Kraken-ES*)

The database of 300,000 phosphine ligands[26] with predicted enantioselectivity values ($\Delta\Delta G^{\ddagger}$) for an enantioselective C(sp3)–C(sp2) cross coupling reactions. The $\Delta\Delta G^{\ddagger}$ values were obtained by fitting a random forest model on the experimental data reported by Biscoe and co-workers,[27] using the ML-extrapolated phosphine descriptors from the Kraken database. The model was then applied to all posphines in the Kraken library to obtain $\Delta\Delta G^{\ddagger}$ values which were used as optimization targets.

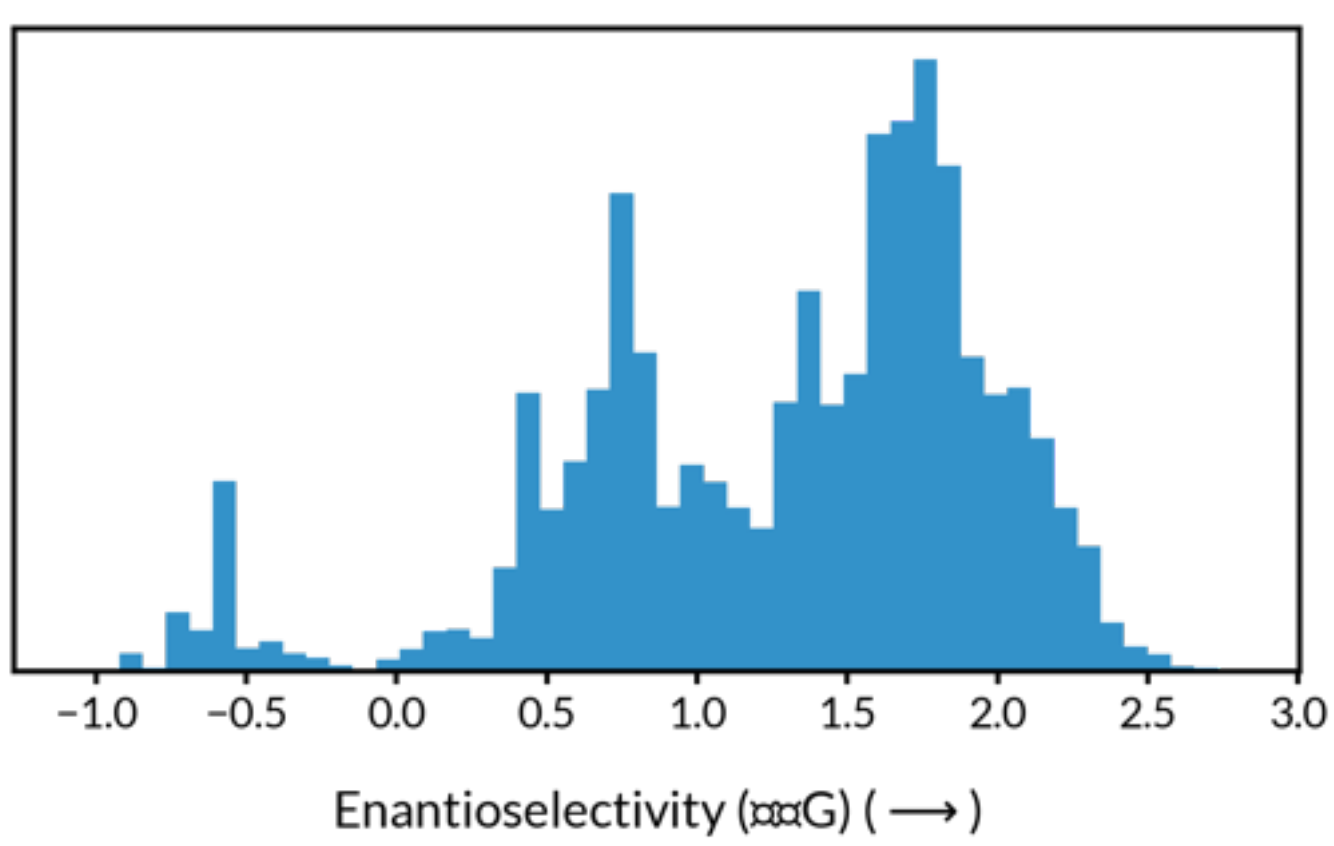

**Figure S 21**: Histogram of target properties (docking scores) for the *Kraken-ES* library.

**Table S 6**: Descriptive statistics of the similarity distributions of the *Kraken-ES* library for different molecular embeddings

| | ***Within Library Similarities*** | | | | ***Similarities To Reference*** | | | |
|---|---|---|---|---|---|---|---|---|
| | Mean | SD | Median | IQR | Mean | SD | Median | IQR |
| Morgan Fingerprint | 0.873 | 0.058 | 0.879 | 0.064 | 0.925 | 0.035 | 0.928 | 0.046 |
| ChemBERTa | 0.358 | 0.138 | 0.345 | 0.187 | 0.424 | 0.114 | 0.417 | 0.156 |
| MolFormer | 0.254 | 0.084 | 0.246 | 0.112 | 0.494 | 0.084 | 0.500 | 0.116 |
| SmiTed | 0.019 | 0.010 | 0.018 | 0.012 | 0.020 | 0.008 | 0.018 | 0.009 |
| T5Chem | 0.352 | 0.149 | 0.334 | 0.211 | 0.592 | 0.111 | 0.593 | 0.155 |

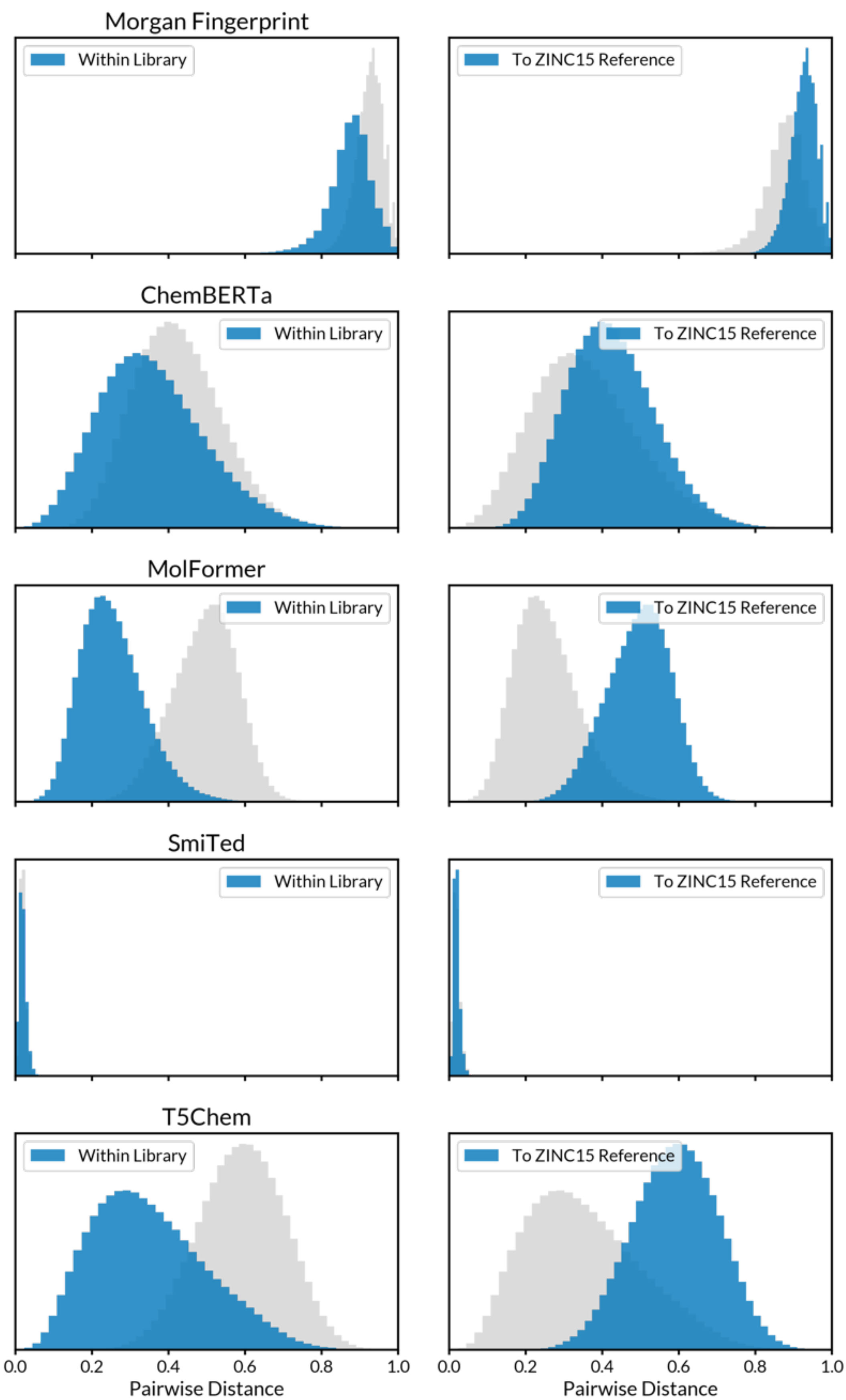

**Figure S 22**: Histograms of embedding similarities for the *Kraken-ES* library. Left column: within-library similarities. Right column: to-reference similarities.

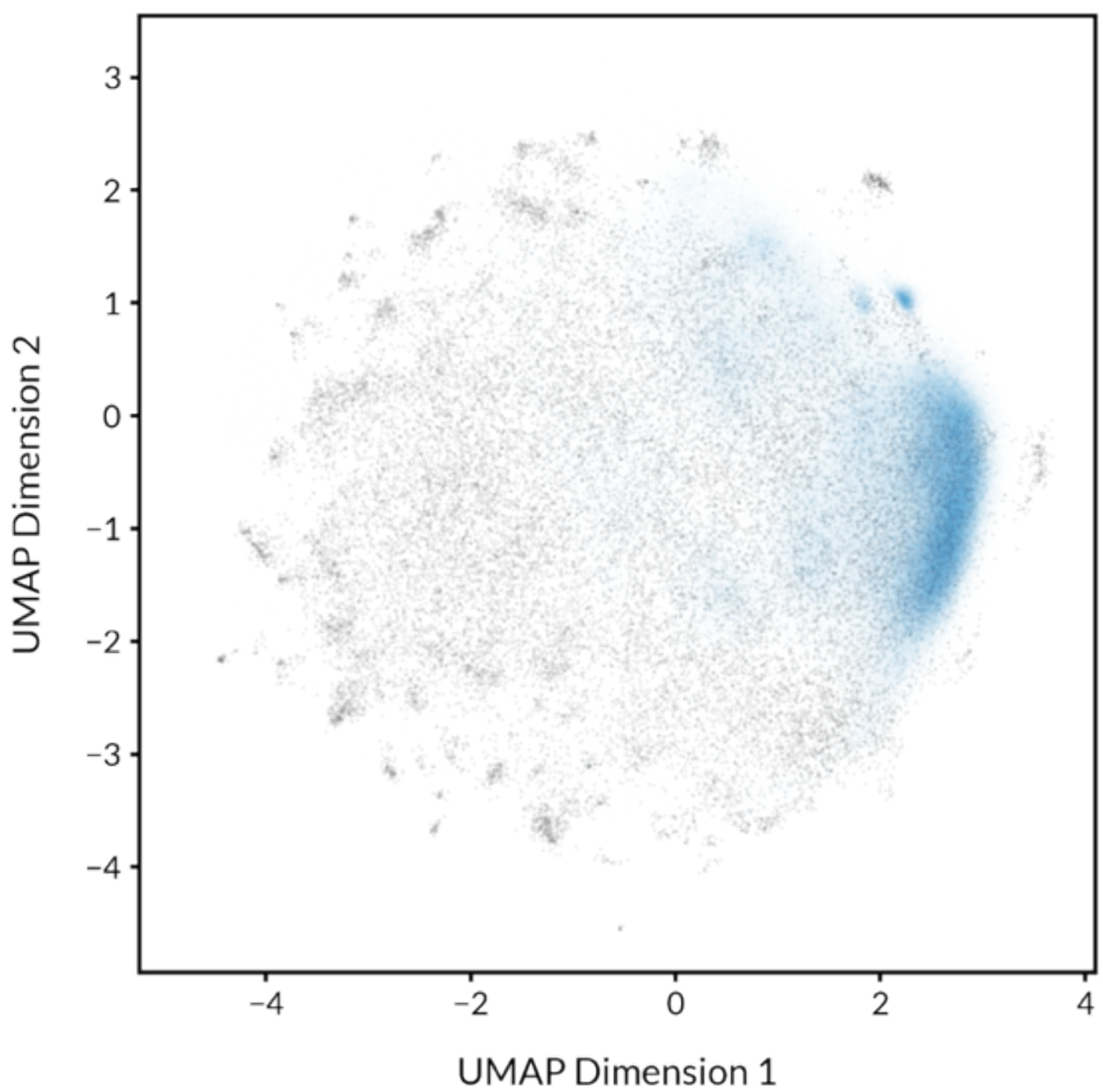

**Figure S 23**: Projection of molecular fingerprints of all candidate molecules from the *Kraken-ES* library into the UMAP coordinates trained on the fingerprints from the reference sample of 25k molecules from the ZINC15 library.

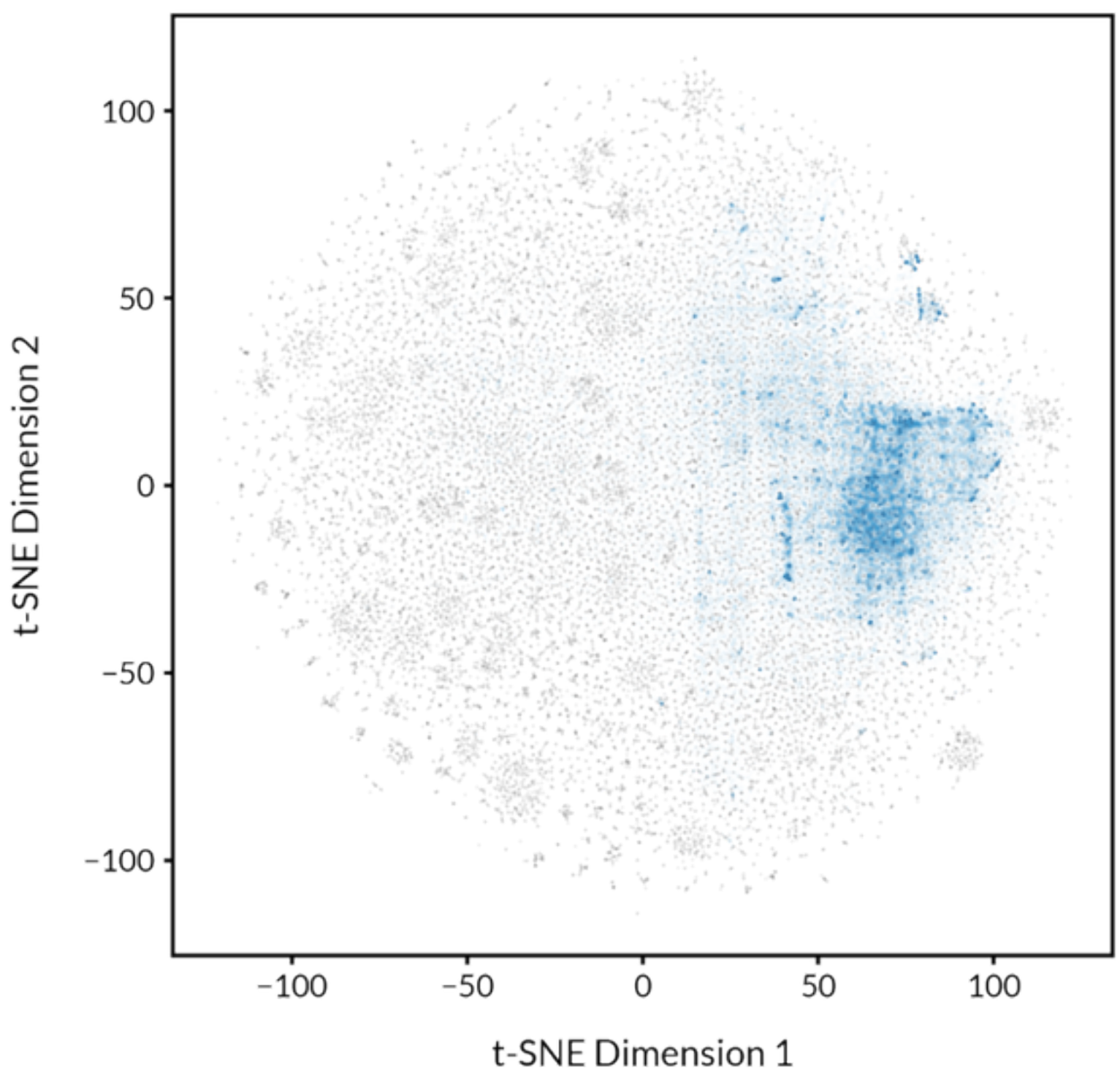

**Figure S 24**: Projection of molecular fingerprints of all candidate molecules from the *Kraken-ES* library into the *t*-SNE coordinates trained on the fingerprints from the reference sample of 25k molecules from the ZINC15 library.

# 3 Benchmarking Surrogate Models and Acquisition Policies

## 3.1 Surrogate Model Benchmarks

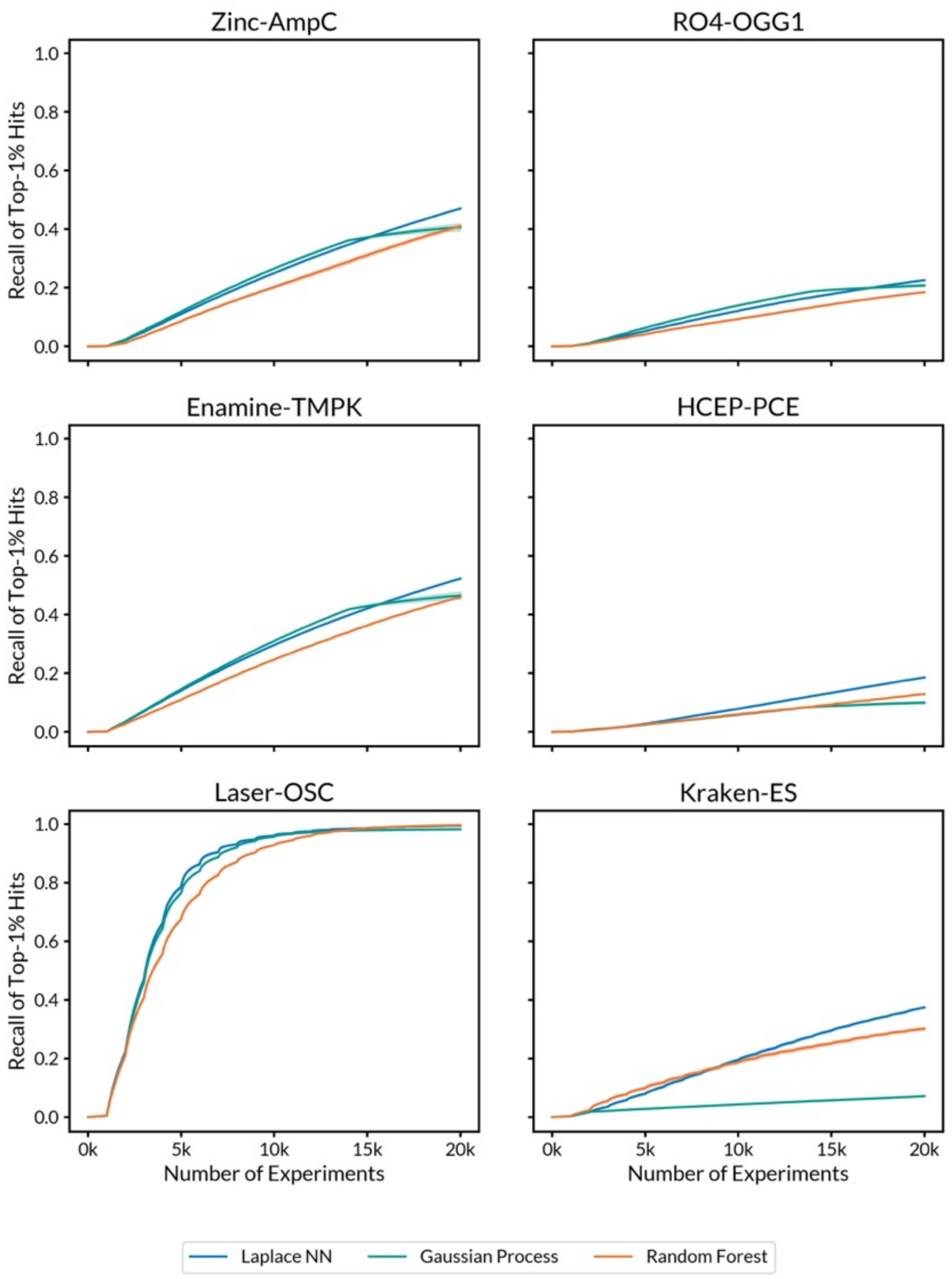

**Figure S 25**: Optimization trajectories for different surrogate models on MolFormer embeddings, evaluated over a budget of 20,000 experiments (batch size of 1,000 experiments). Trajectories are shown as the mean over 20 independent campaigns from different random seed populations. The shaded areas indicate the standard error of the mean.

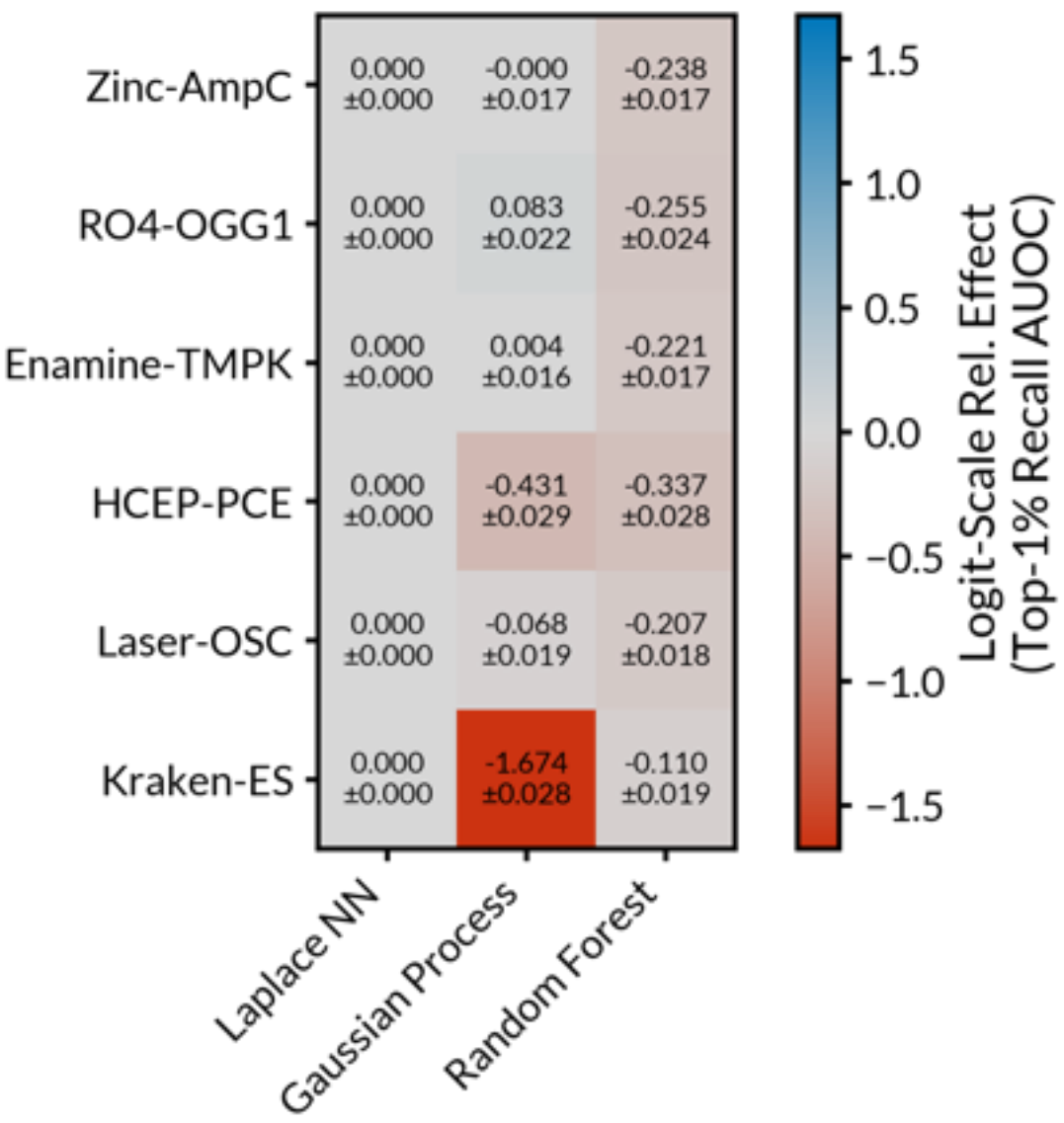

**Figure S 26**: Logit-scale effect of the surrogate model relative to a Laplace Neural network model, differentiated by molecular libraries. Experiments were performed using MolFormer embeddings in the virtual screening setting (20,000 experiments, batch size 1,000). Means and standard deviations were obtained from 8,000 posterior samples.

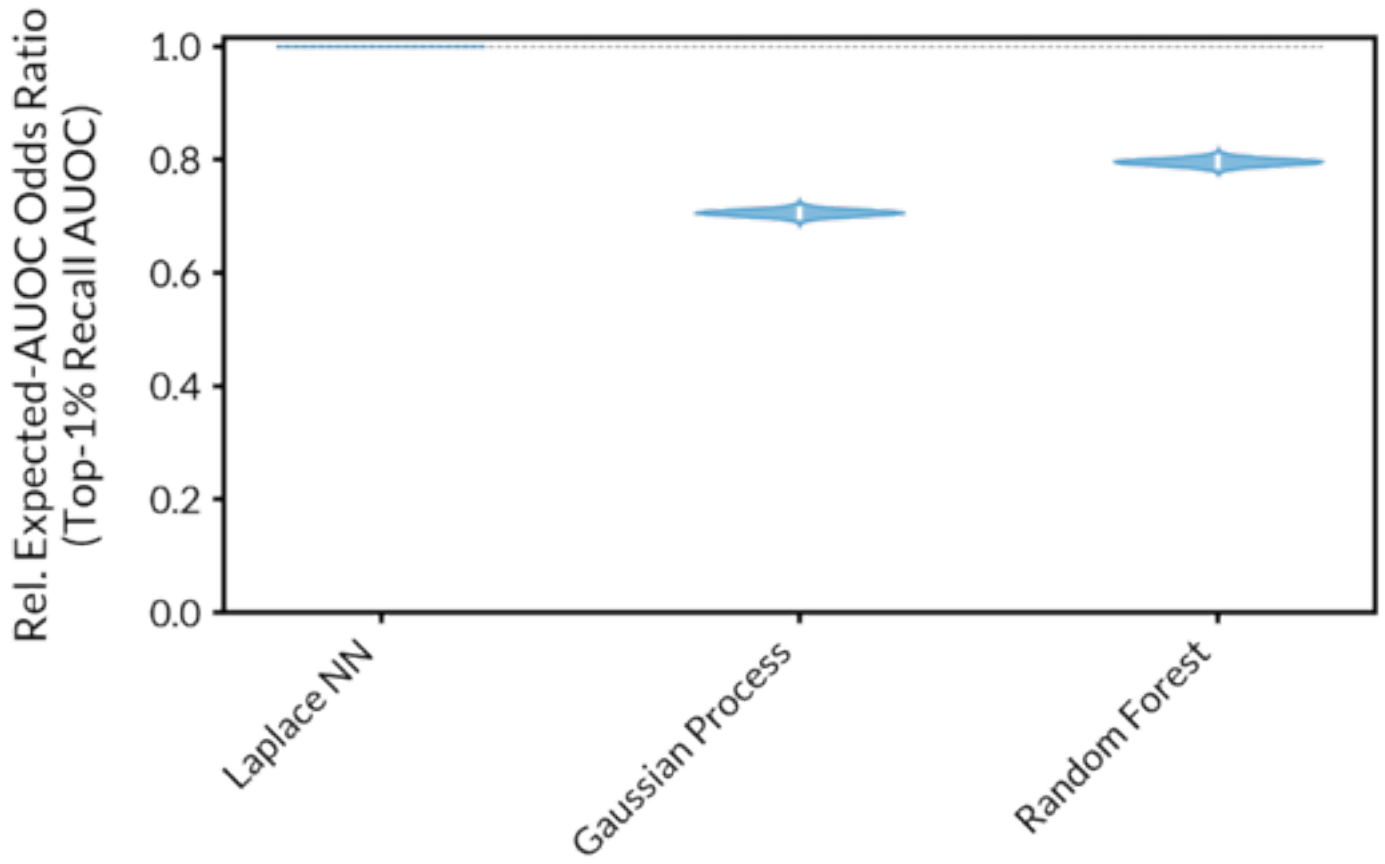

**Figure S 27**: Effect ratios of the surrogate model relative to a Laplace Neural network model, averaged over all libraries. Experiments were performed using MolFormer embeddings in the virtual screening setting (20,000 experiments, batch size 1,000). Bars within the violins indicate the 95% highest-density intervals.

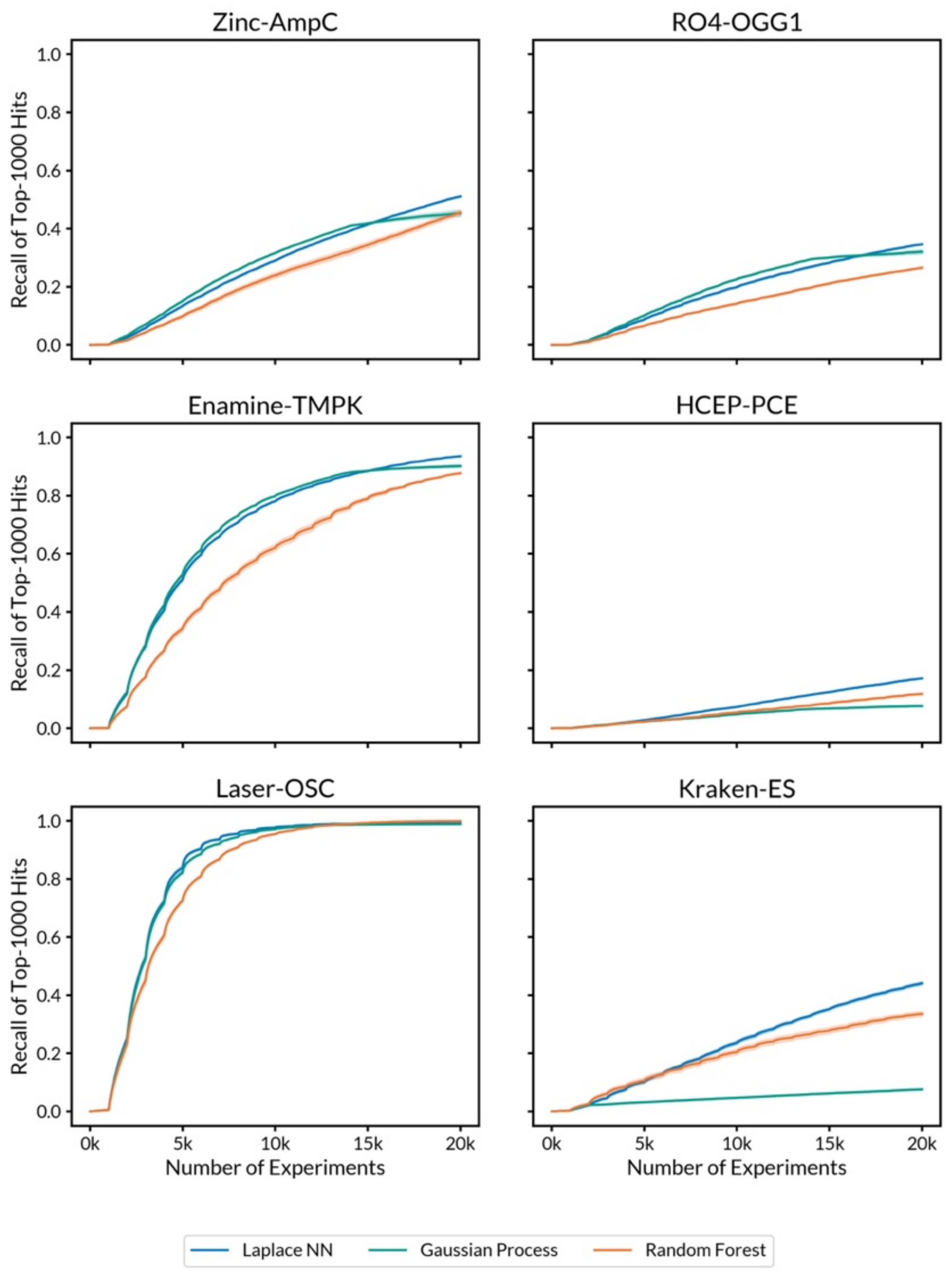

**Figure S 28**: Optimization trajectories for different surrogate models on MolFormer embeddings, evaluated over a budget of 20,000 experiments (batch size of 1,000 experiments). Trajectories are shown as the mean over 20 independent campaigns from different random seed populations. The shaded areas indicate the standard error of the mean.

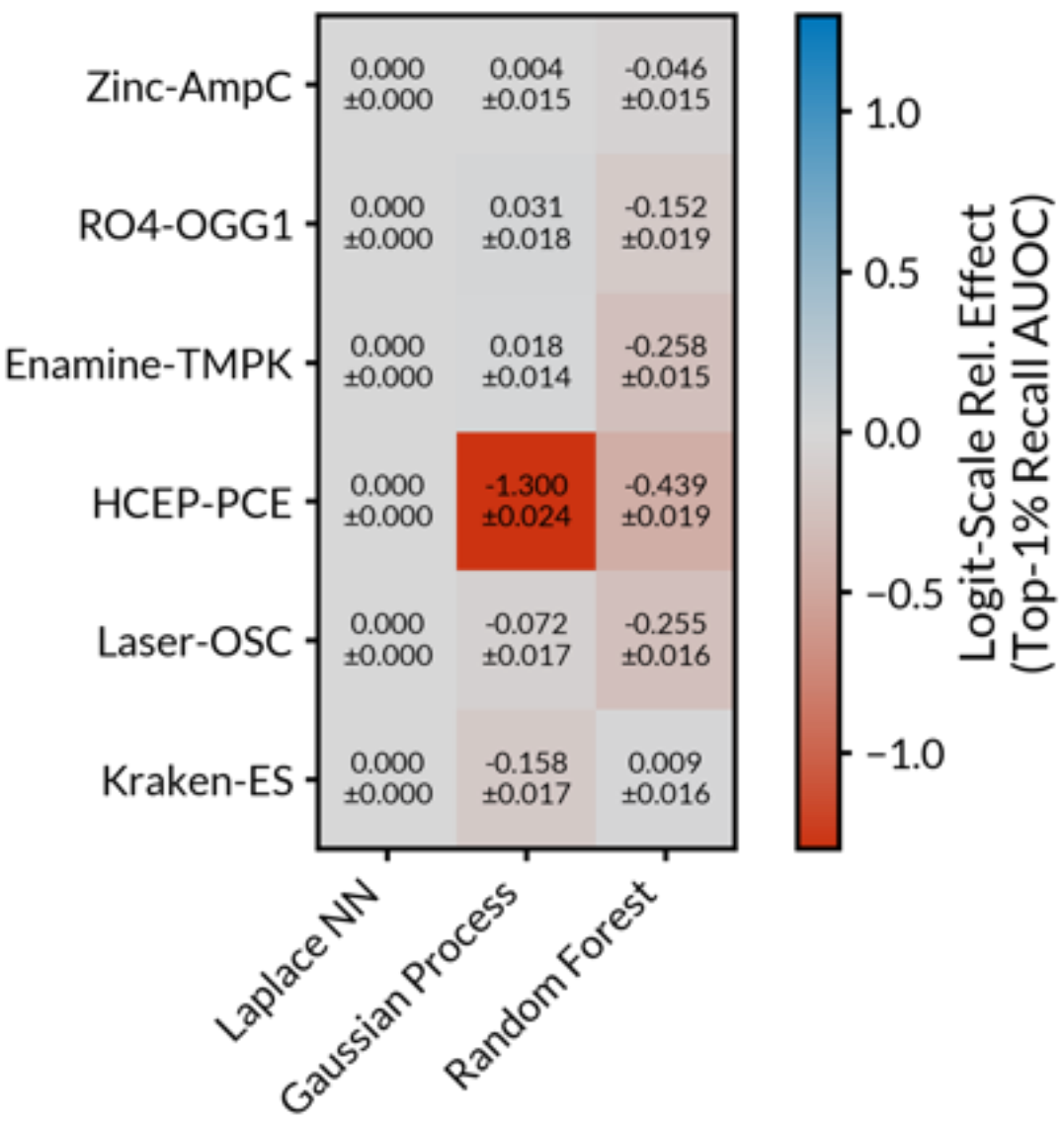

**Figure S 29**: Logit-scale effect of the surrogate model relative to a Laplace Neural network model, differentiated by molecular libraries. Experiments were performed using MolFormer embeddings in the virtual screening setting (20,000 experiments, batch size 1,000). Means and standard deviations were obtained from 8,000 posterior samples.

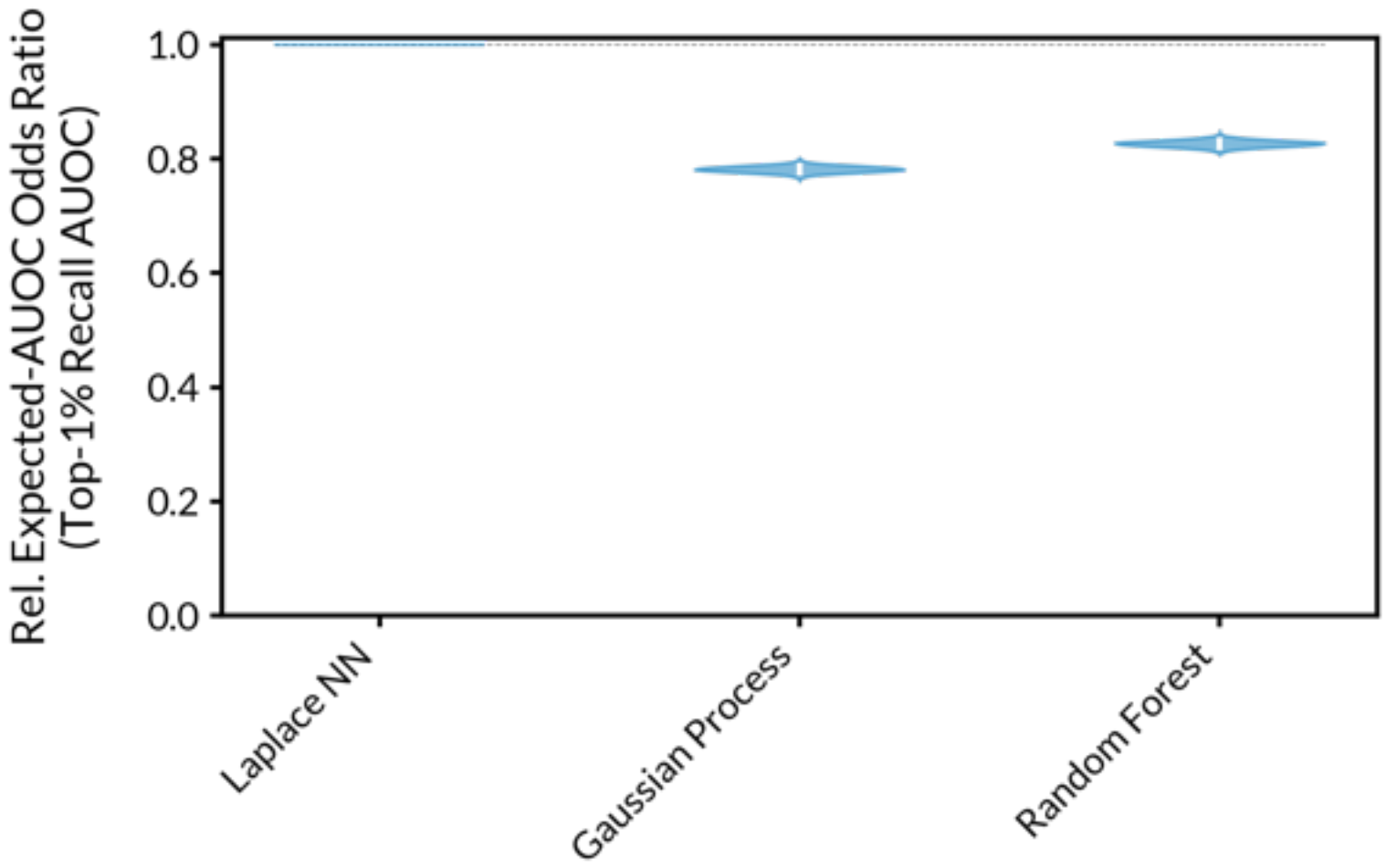

**Figure S 30**: Effect ratios of the surrogate model relative to a Laplace Neural network model, averaged over all libraries. Experiments were performed using MolFormer embeddings in the virtual screening setting (20,000 experiments, batch size 1,000). Bars within the violins indicate the 95% highest-density intervals.

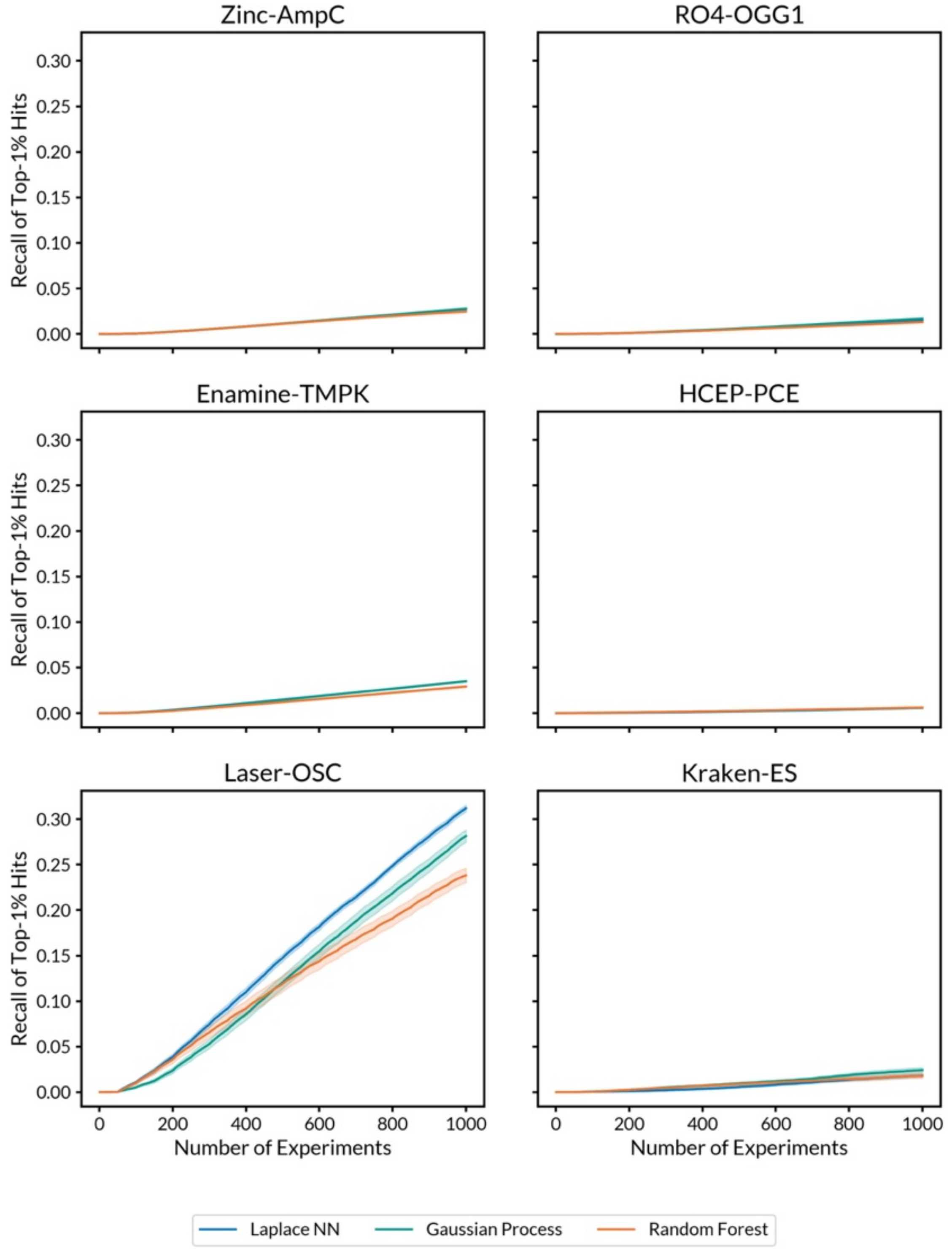

**Figure S 31**: Optimization trajectories for different surrogate models on MolFormer embeddings, evaluated over a budget of 1,000 experiments (batch size of 50 experiments). Trajectories are shown as the mean over 20 independent campaigns from different random seed populations. The shaded areas indicate the standard error of the mean.

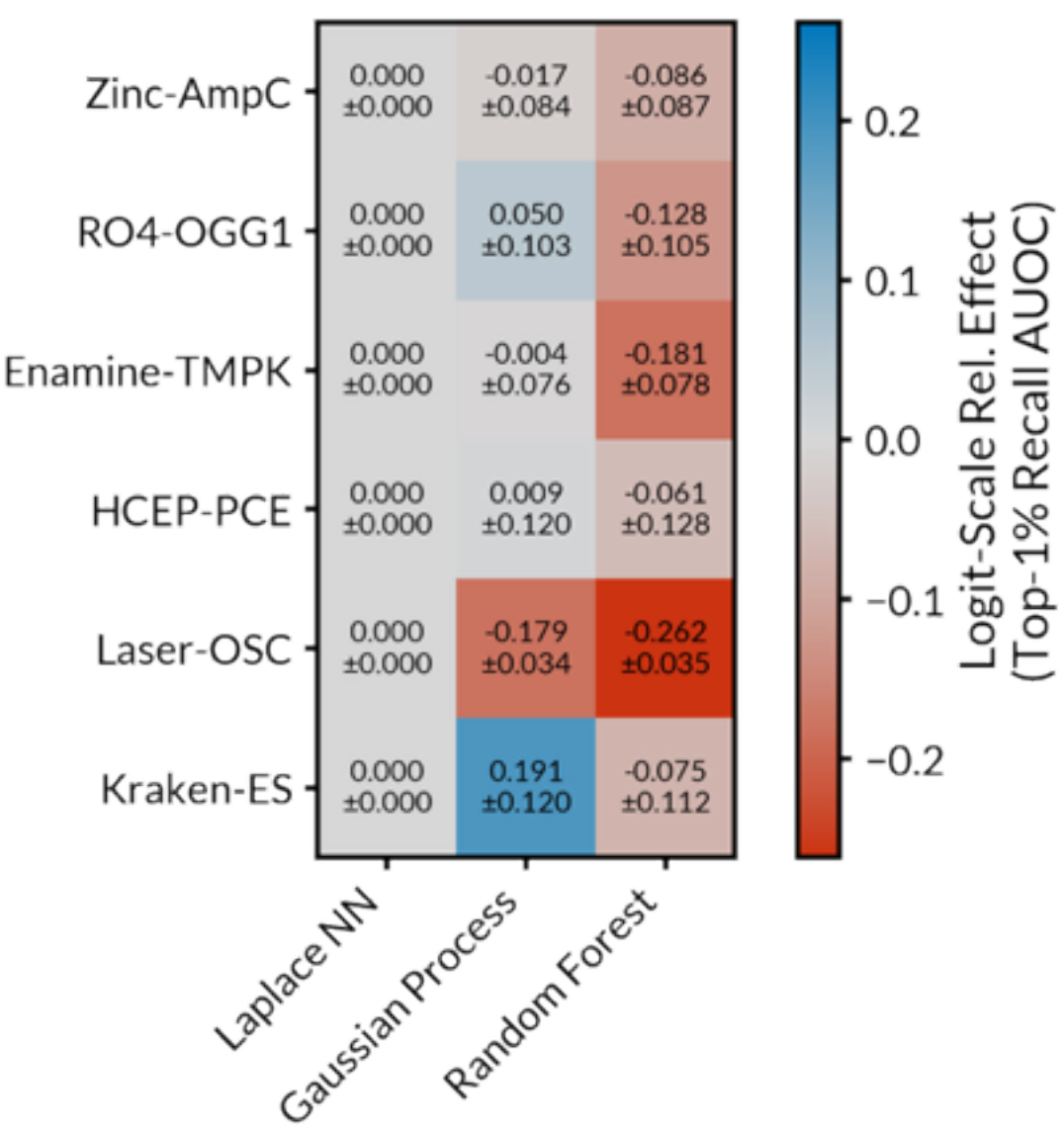

**Figure S 32**: Logit-scale effect of the surrogate model relative to a Laplace Neural network model, differentiated by molecular libraries. Experiments were performed using MolFormer embeddings in the experimental discovery setting (1,000 experiments, batch size 50). Means and standard deviations were obtained from 8,000 posterior samples.

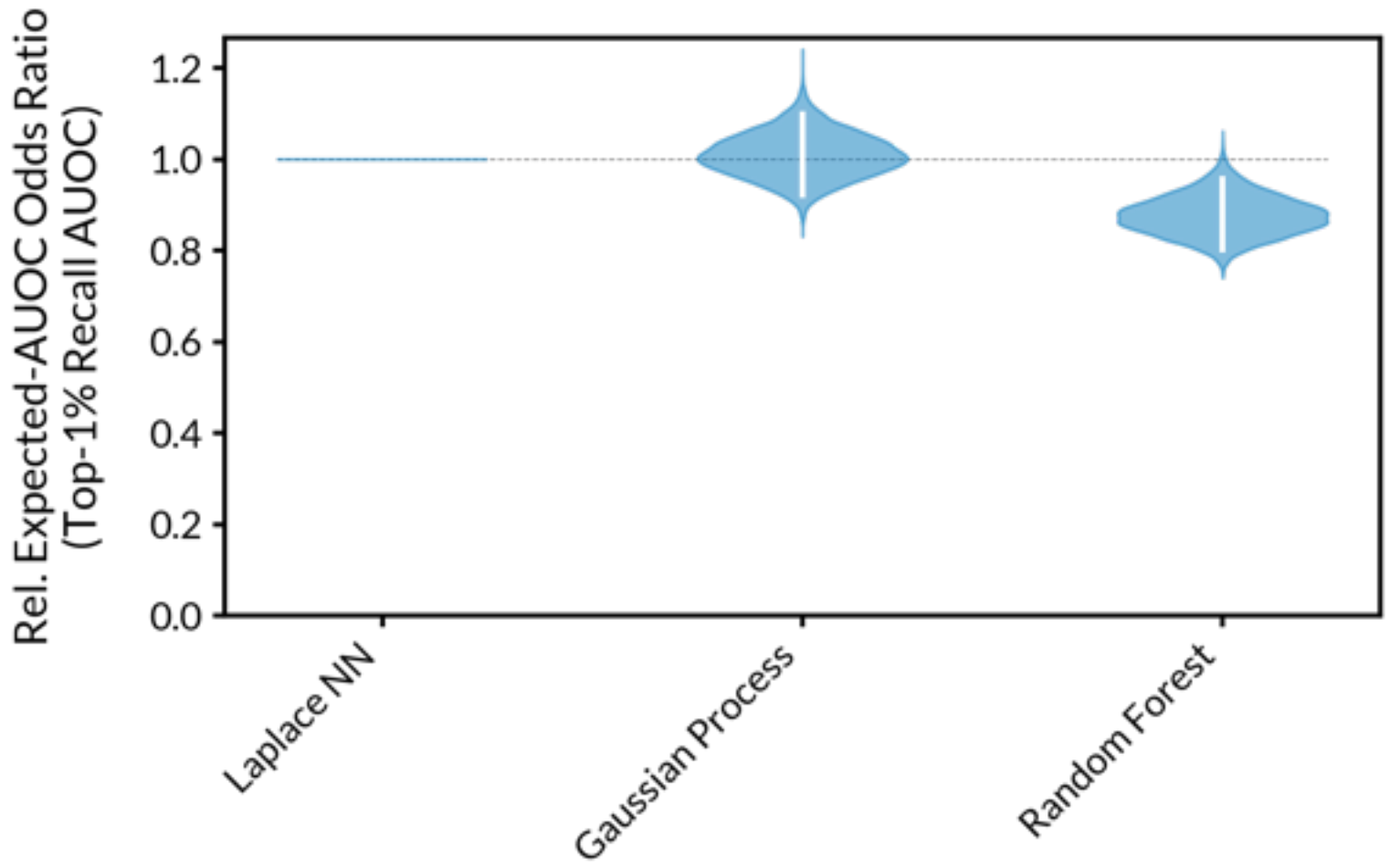

**Figure S 33**: Effect ratios of the surrogate model relative to a Laplace Neural network model, averaged over all libraries. Experiments were performed using MolFormer embeddings in the in the experimental discovery setting (1,000 experiments, batch size 50). Bars within the violins indicate the 95% highest-density intervals..

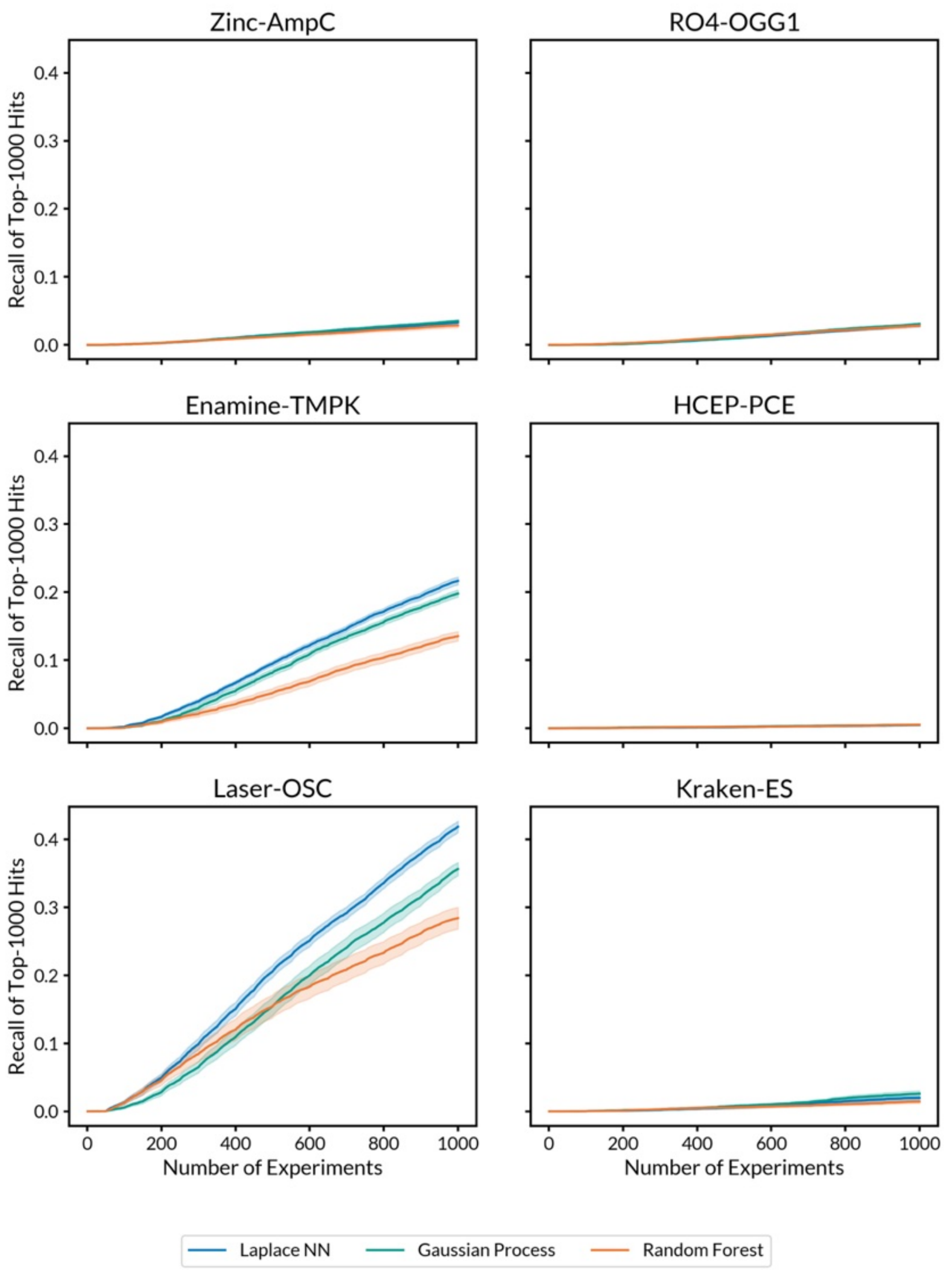

**Figure S 34**: Optimization trajectories for different surrogate models on MolFormer embeddings, evaluated over a budget of 1,000 experiments (batch size of 50 experiments). Trajectories are shown as the mean over 20 independent campaigns from different random seed populations. The shaded areas indicate the standard error of the mean.

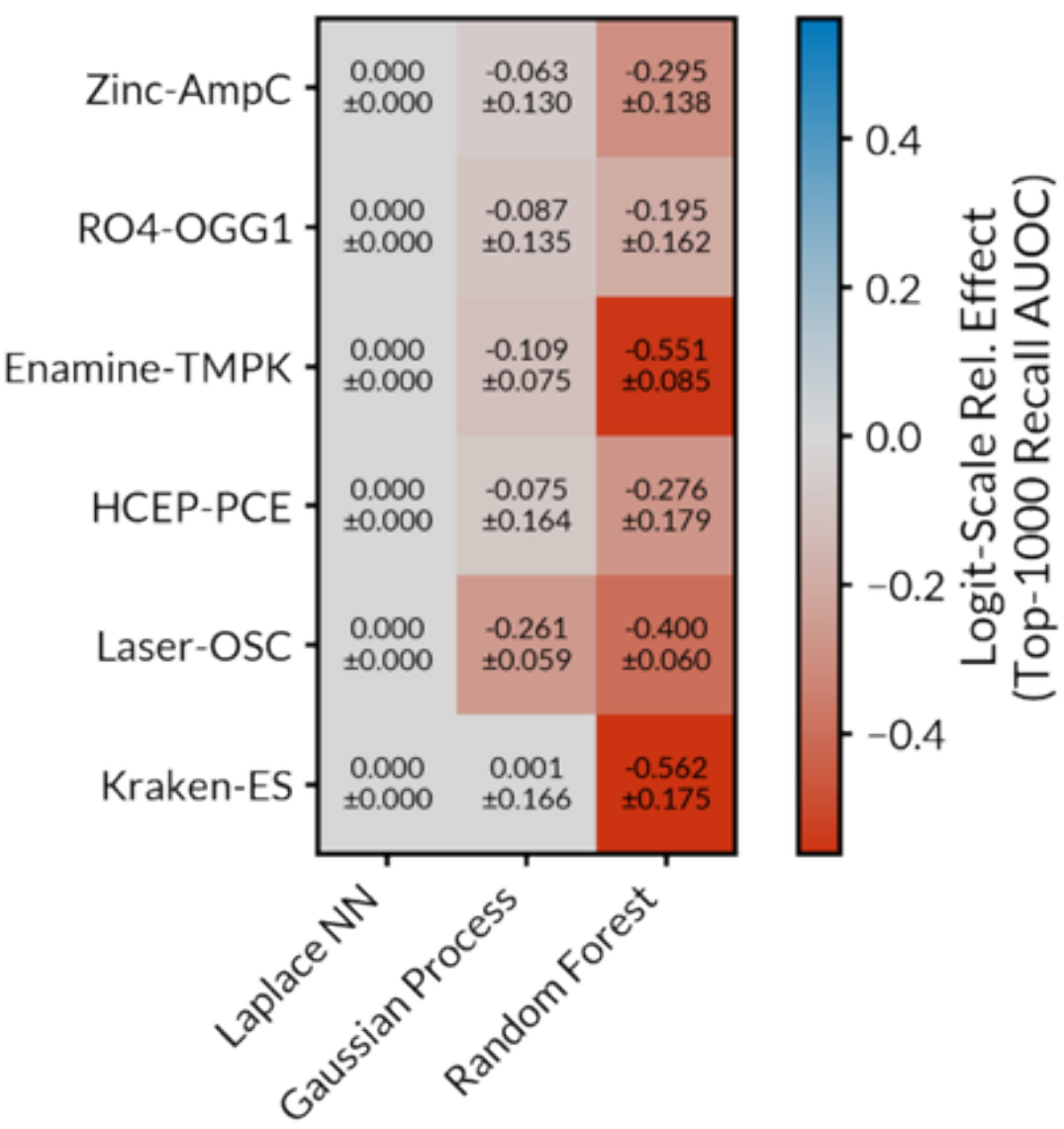

**Figure S 35**: Logit-scale effect of the surrogate model relative to a Laplace Neural network model, differentiated by molecular libraries. Experiments were performed using MolFormer embeddings in the experimental discovery setting (1,000 experiments, batch size 50). Means and standard deviations were obtained from 8,000 posterior samples.

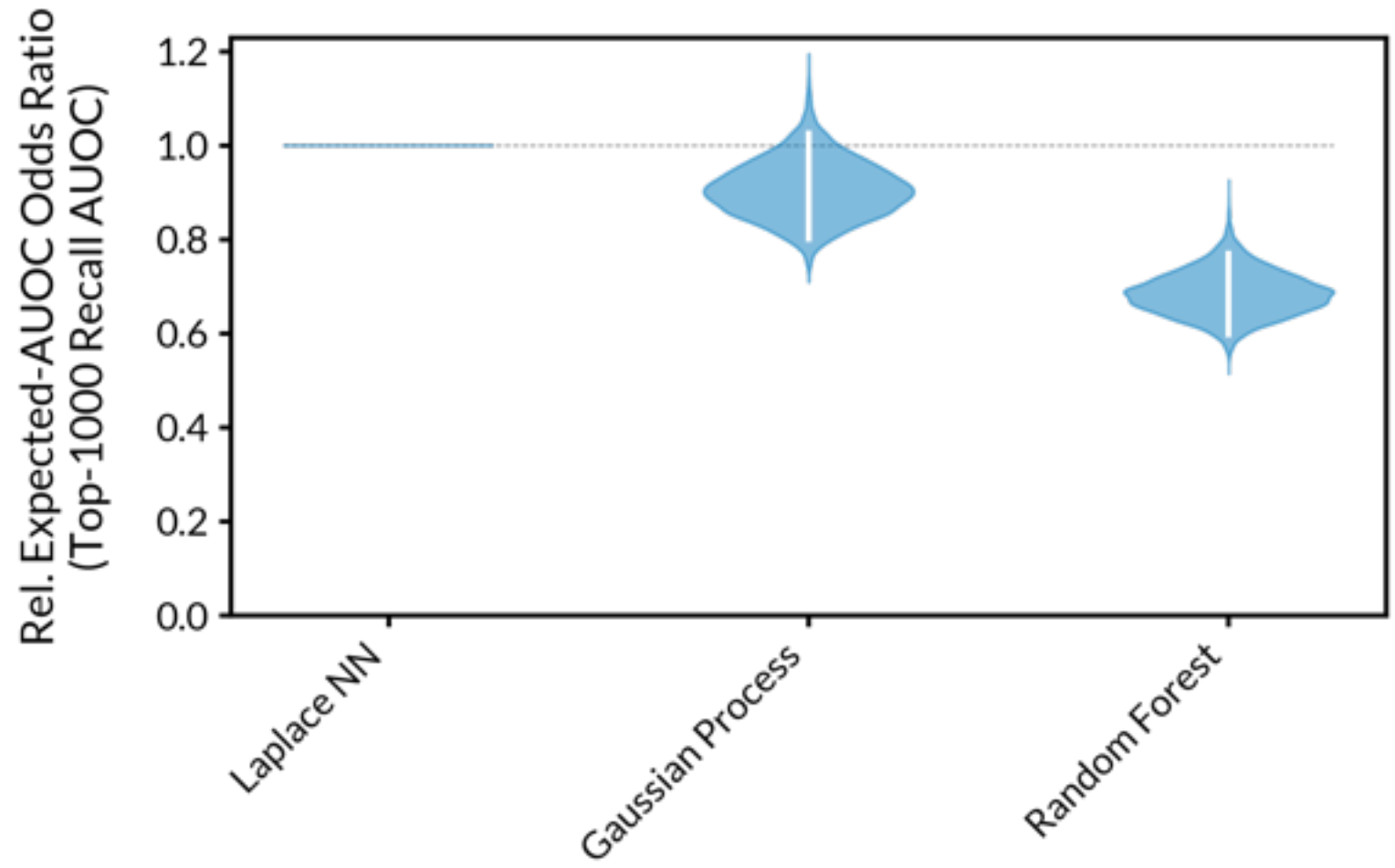

**Figure S 36**: Effect ratios of the surrogate model relative to a Laplace Neural network model, averaged over all libraries. Experiments were performed using MolFormer embeddings in the experimental discovery setting (1,000 experiments, batch size 50). Bars within the violins indicate the 95% highest-density intervals.

## 3.2 Acquisition Strategy Benchmarks

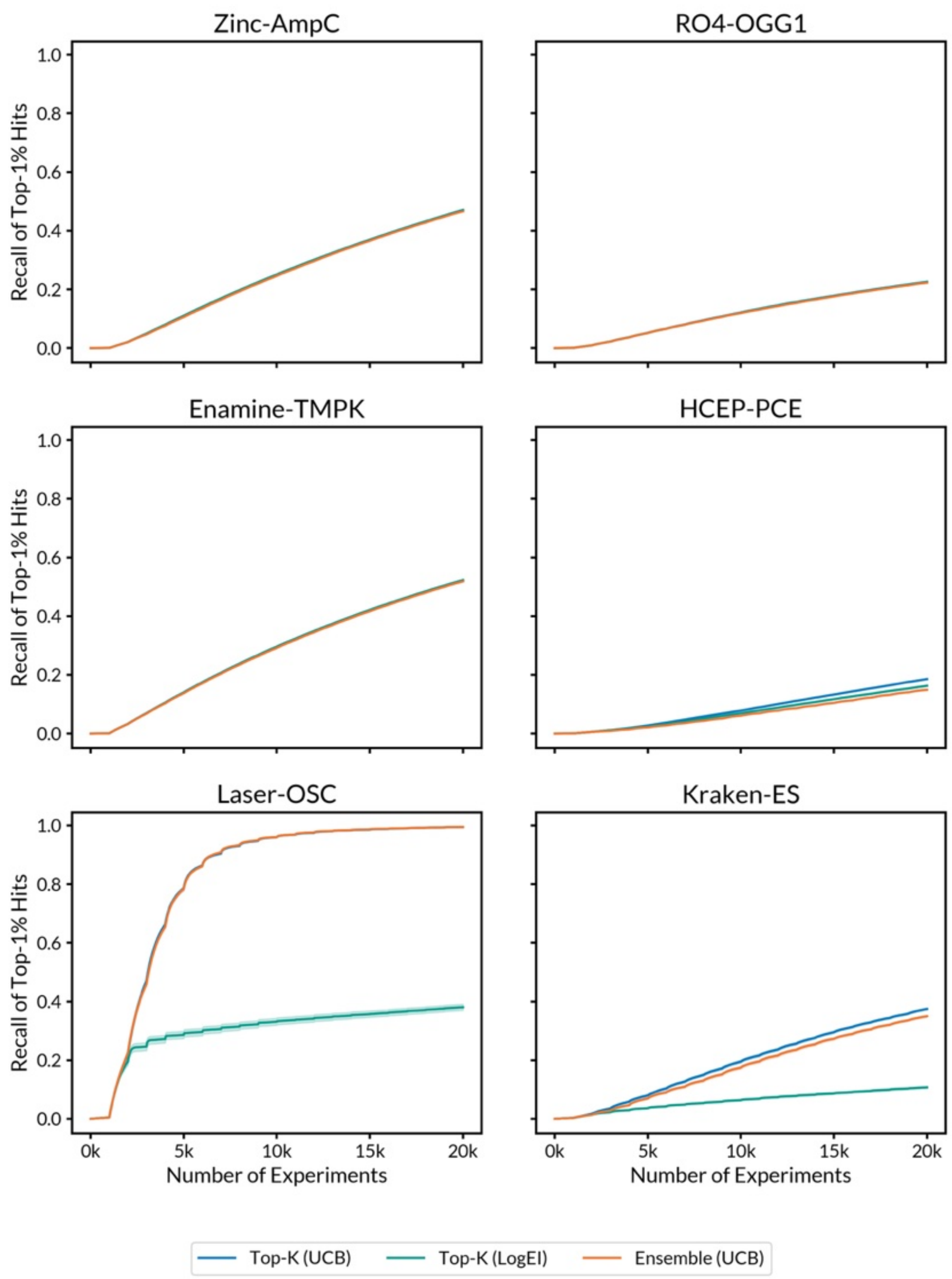

**Figure S 37**: Optimization trajectories for different acquisition strategies, using a Laplace Neural Network surrogate model on MolFormer embeddings, evaluated over a budget of 20,000 experiments (batch size of 1,000 experiments). Trajectories are shown as the mean over 20 independent campaigns from different random seed populations. The shaded areas indicate the standard error of the mean.

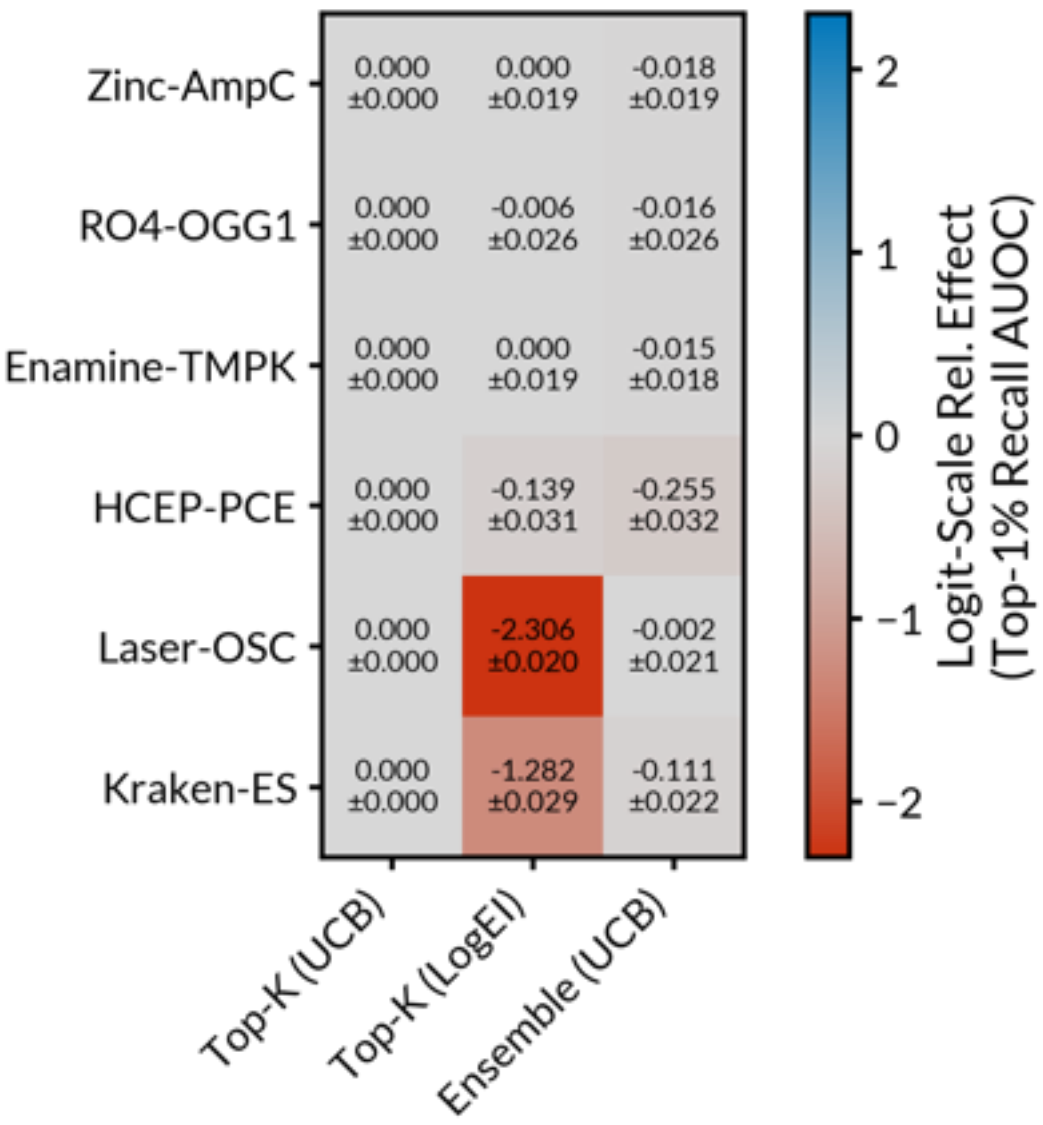

**Figure S 38**: Logit-scale effect of the acquisition strategy relative to Top-*K* acquisition with a UCB acquisition function, differentiated by molecular libraries. Experiments were performed using a Laplace Neural network model on MolFormer embeddings in the virtual screening setting (20,000 experiments, batch size 1,000). Means and standard deviations were obtained from 8,000 posterior samples.

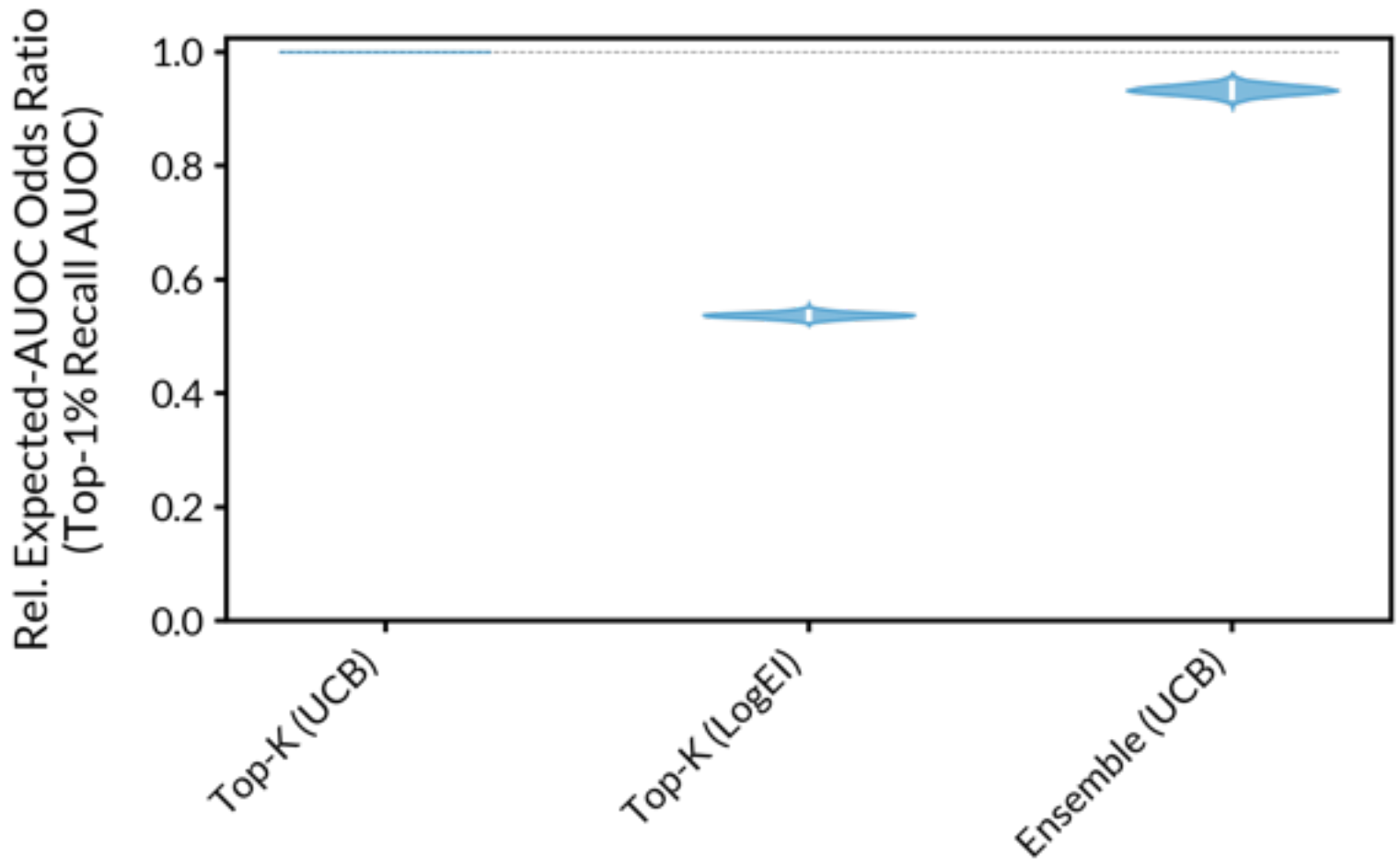

**Figure S 39**: Effect ratios of the acquisition strategy relative to Top-*K* acquisition with a UCB acquisition function, averaged over all libraries. Experiments were performed using a Laplace Neural network model on MolFormer embeddings in the virtual screening setting (20,000 experiments, batch size 1,000). Bars within the violins indicate the 95% highest-density intervals.

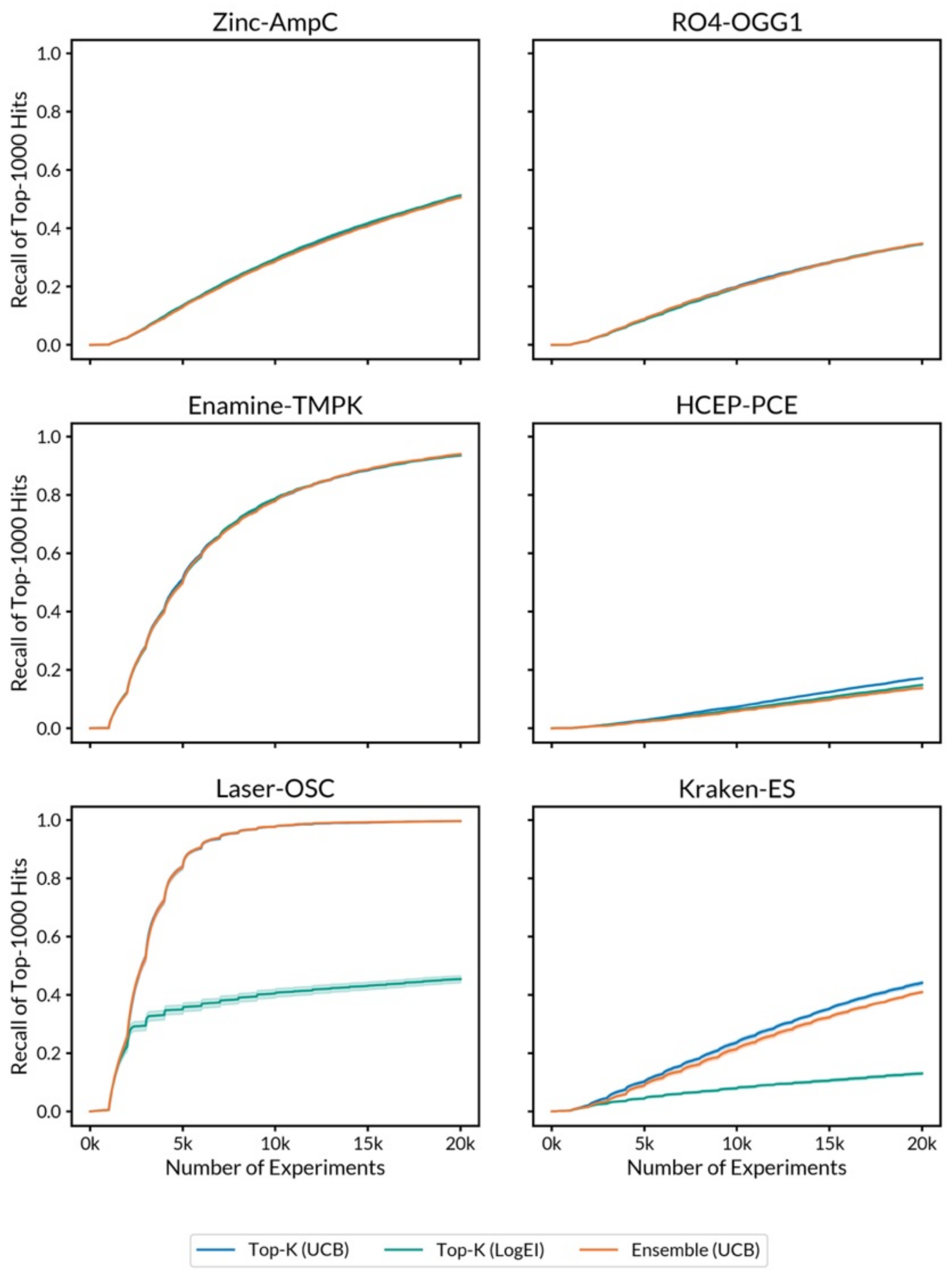

**Figure S 40**: Optimization trajectories for different acquisition strategies, using a Laplace Neural Network surrogate model on MolFormer embeddings, evaluated over a budget of 20,000 experiments (batch size of 1,000 experiments). Trajectories are shown as the mean over 20 independent campaigns from different random seed populations. The shaded areas indicate the standard error of the mean.

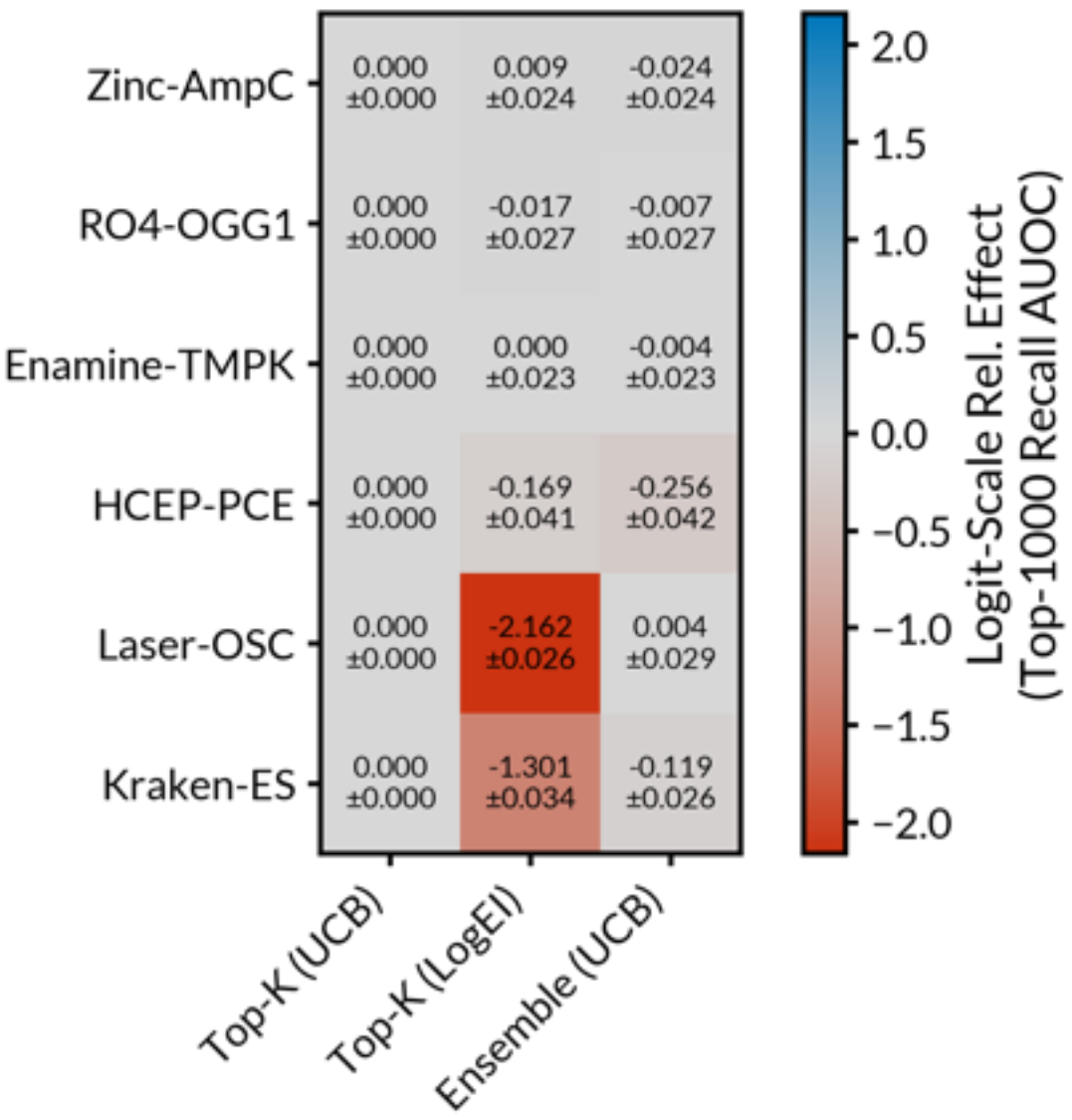

**Figure S 41**: Logit-scale effect of the acquisition strategy relative to Top-*K* acquisition with a UCB acquisition function, differentiated by molecular libraries. Experiments were performed using a Laplace Neural network model on MolFormer embeddings in the virtual screening setting (20,000 experiments, batch size 1,000). Means and standard deviations were obtained from 8,000 posterior samples.

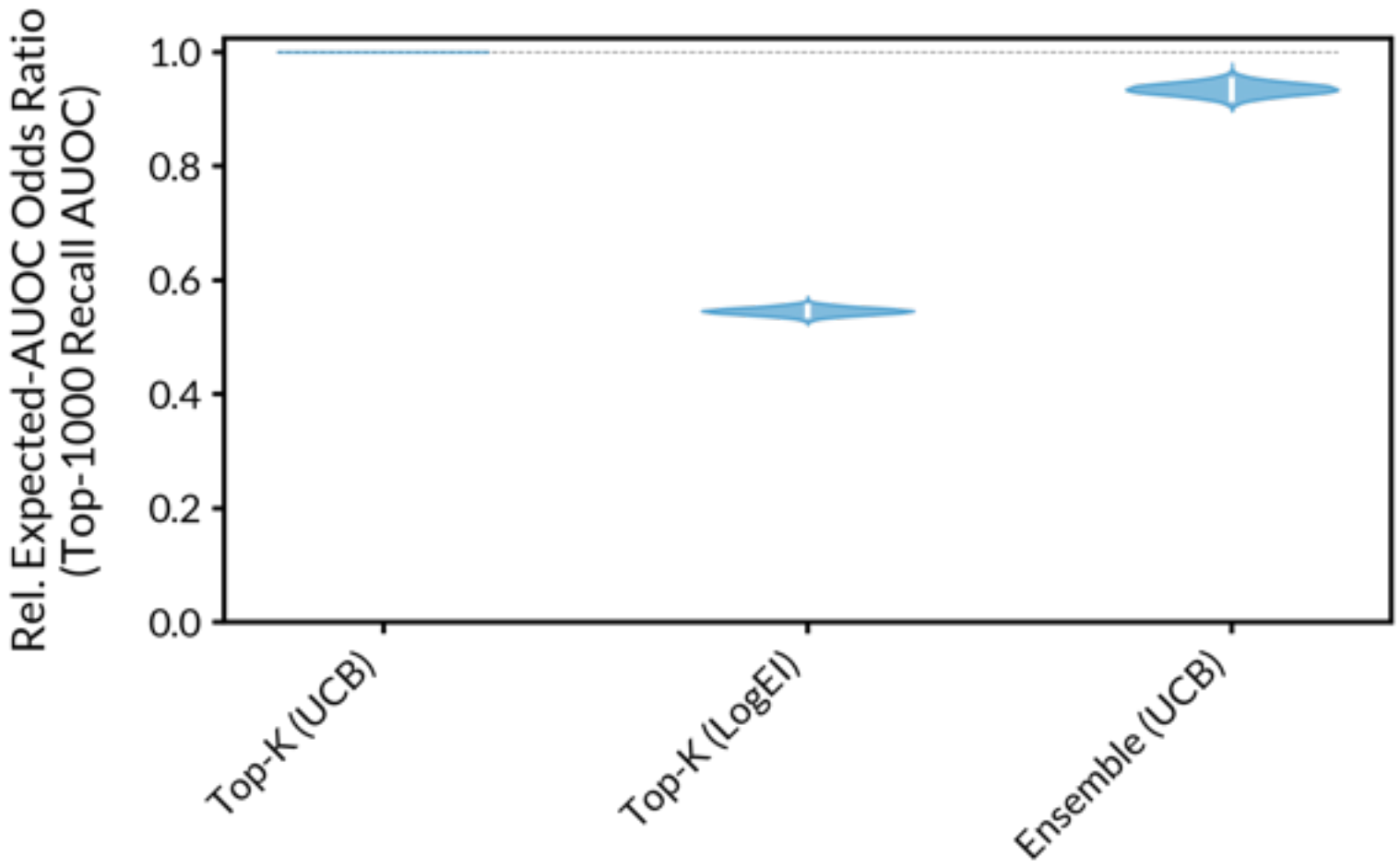

**Figure S 42**: Effect ratios of the acquisition strategy relative to Top-*K* acquisition with a UCB acquisition function, averaged over all libraries. Experiments were performed using a Laplace Neural network model on MolFormer embeddings in the virtual screening setting (20,000 experiments, batch size 1,000). Bars within the violins indicate the 95% highest-density intervals.

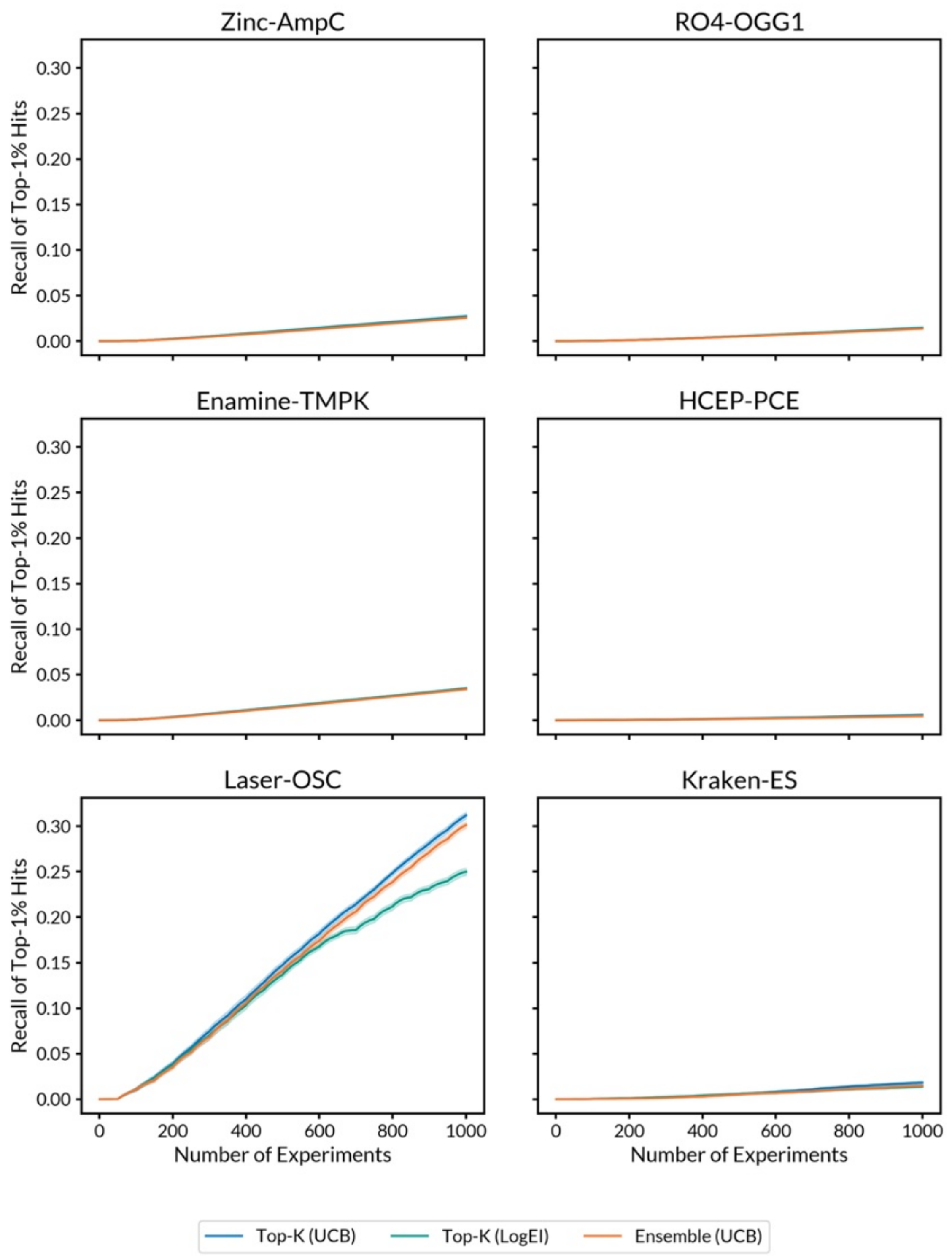

**Figure S 43**: Optimization trajectories for different acquisition strategies, using a Laplace Neural Network surrogate model on MolFormer embeddings, evaluated over a budget of 1,000 experiments (batch size of 50 experiments). Trajectories are shown as the mean over 20 independent campaigns from different random seed populations. The shaded areas indicate the standard error of the mean.

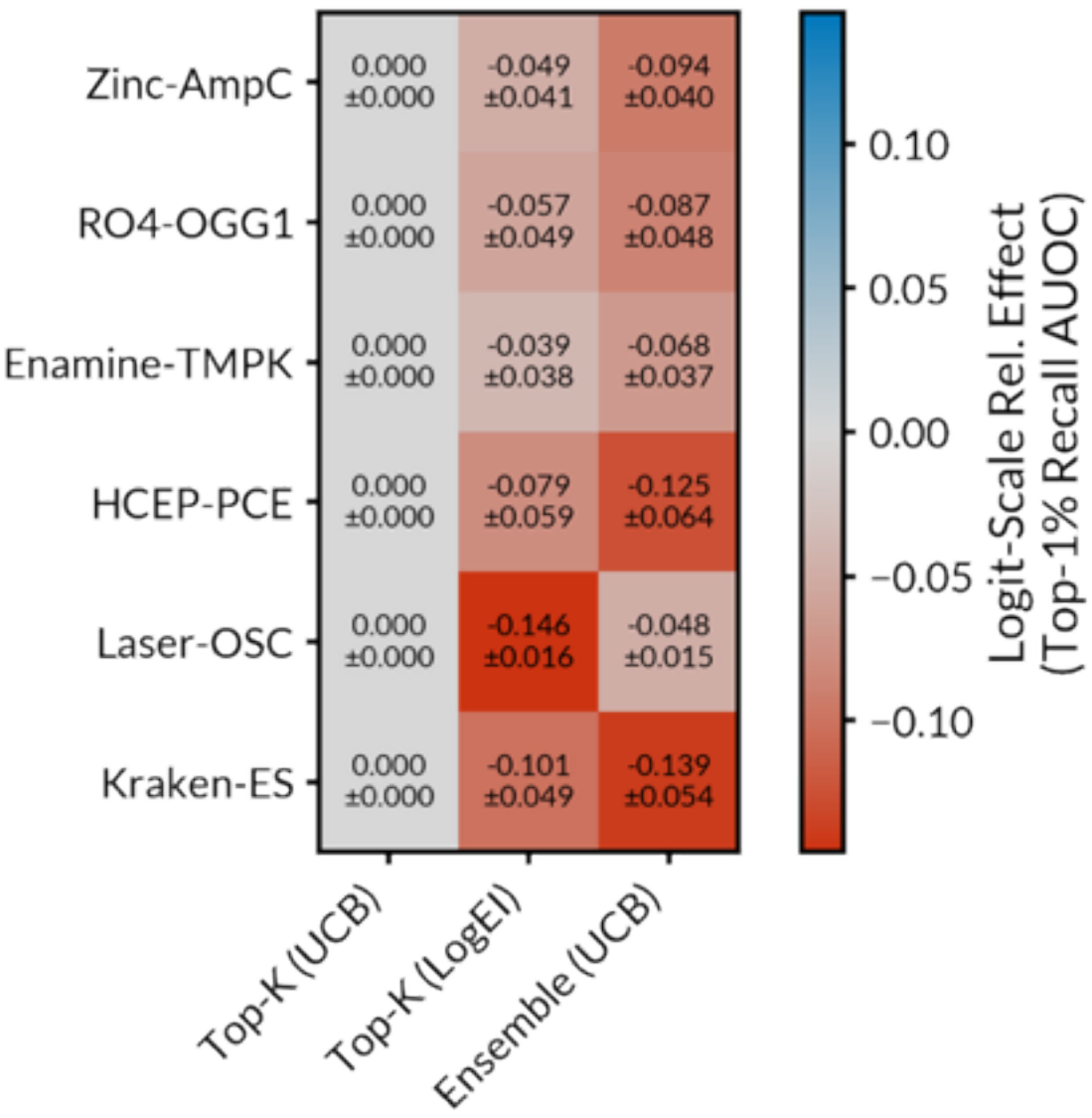

**Figure S 44**: Logit-scale effect of the acquisition strategy relative to Top-*K* acquisition with a UCB acquisition function, differentiated by molecular libraries. Experiments were performed using a Laplace Neural network model on MolFormer embeddings in the experimental discovery setting (1,000 experiments, batch size 50). Means and standard deviations were obtained from 8,000 posterior samples.

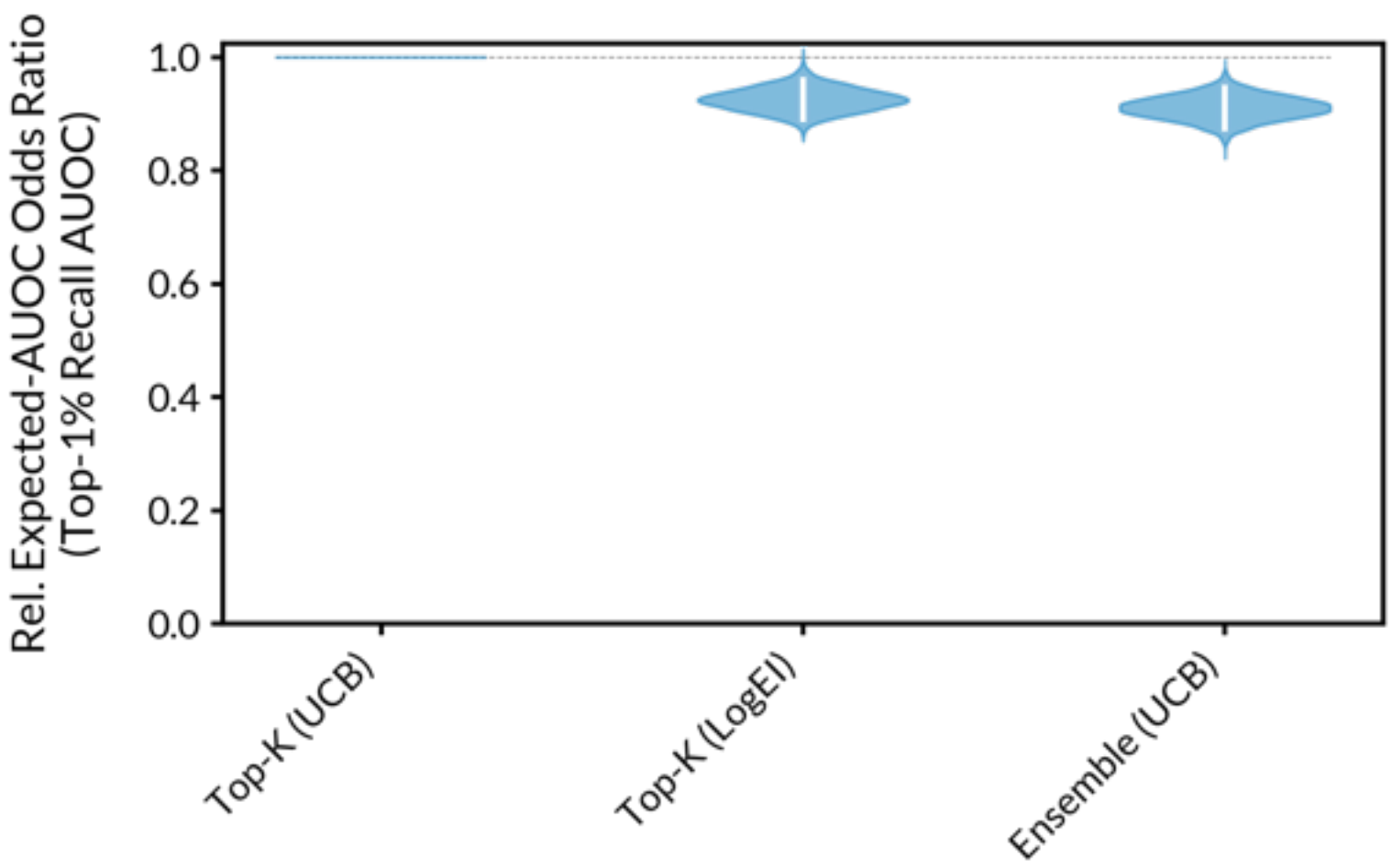

**Figure S 45**: Effect ratios of the acquisition strategy relative to Top-*K* acquisition with a UCB acquisition function, averaged over all libraries. Experiments were performed using a Laplace Neural network model on MolFormer embeddings in the experimental discovery setting (1,000 experiments, batch size 50). Bars within the violins indicate the 95% highest-density intervals.

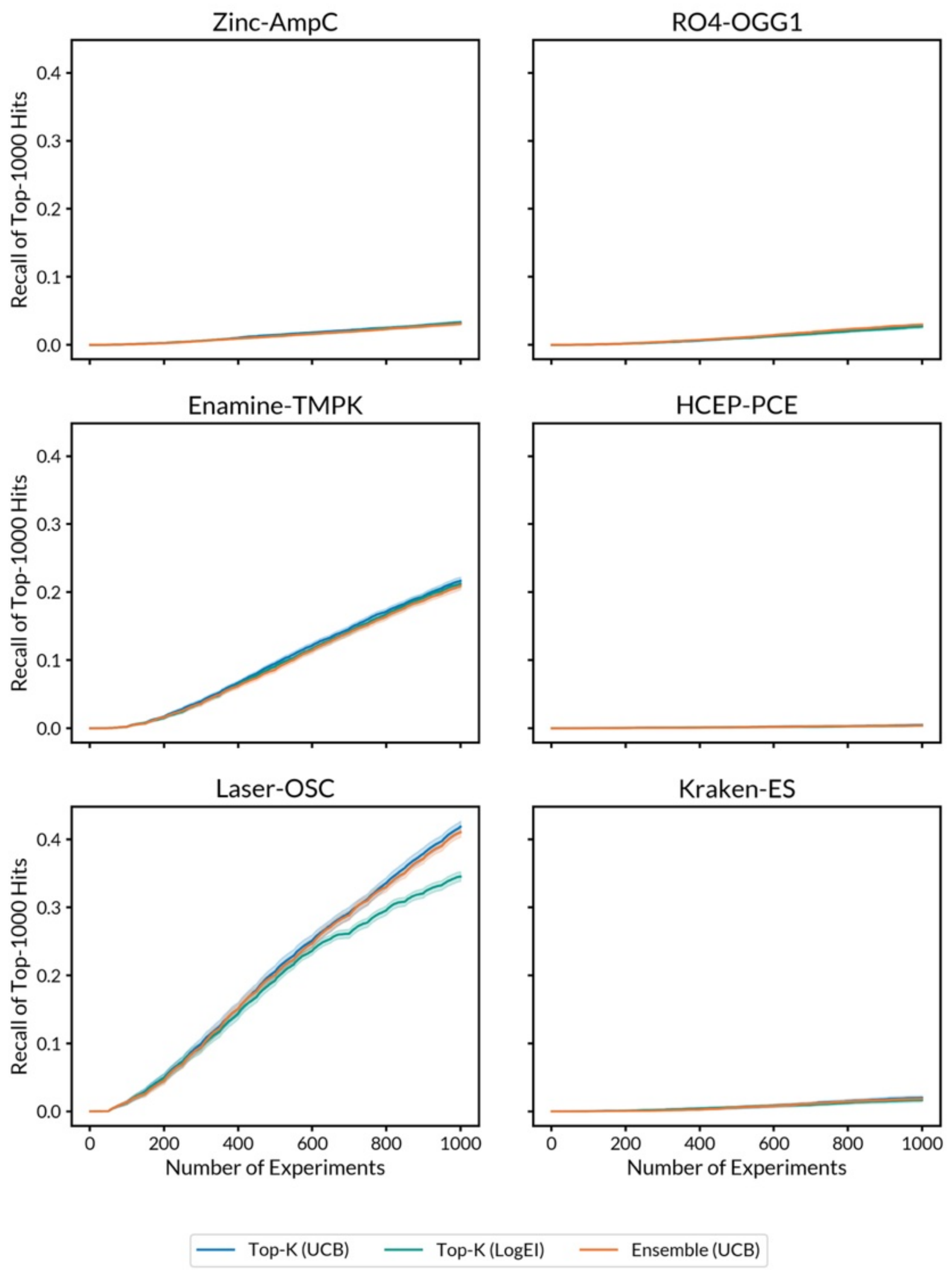

**Figure S 46**: Optimization trajectories for different acquisition strategies, using a Laplace Neural Network surrogate model on MolFormer embeddings, evaluated over a budget of 1,000 experiments (batch size of 50 experiments). Trajectories are shown as the mean over 20 independent campaigns from different random seed populations. The shaded areas indicate the standard error of the mean.

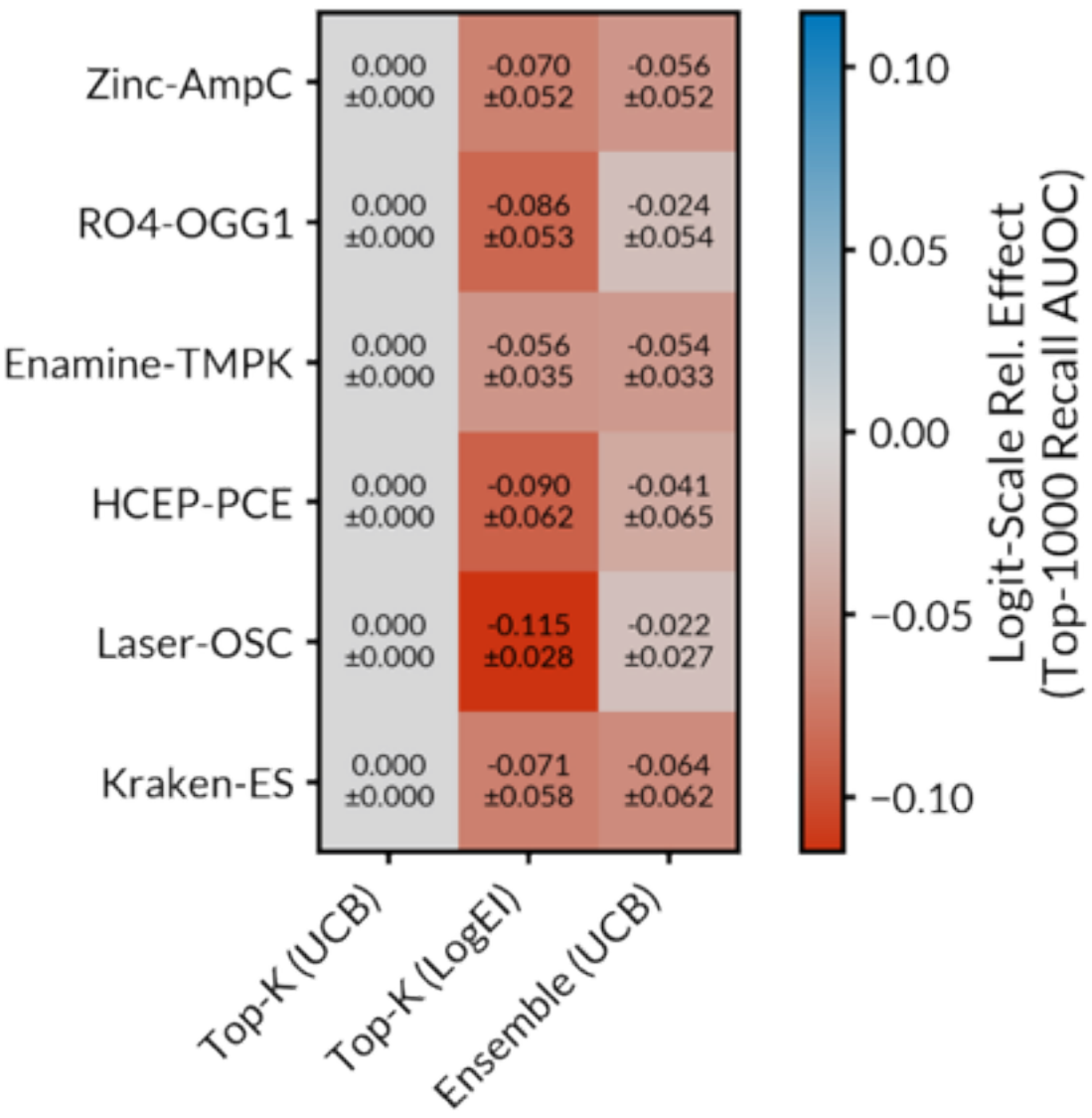

**Figure S 47**: Logit-scale effect of the acquisition strategy relative to Top-*K* acquisition with a UCB acquisition function, differentiated by molecular libraries. Experiments were performed using a Laplace Neural network model on MolFormer embeddings in the experimental discovery setting (1,000 experiments, batch size 50). Means and standard deviations were obtained from 8,000 posterior samples.

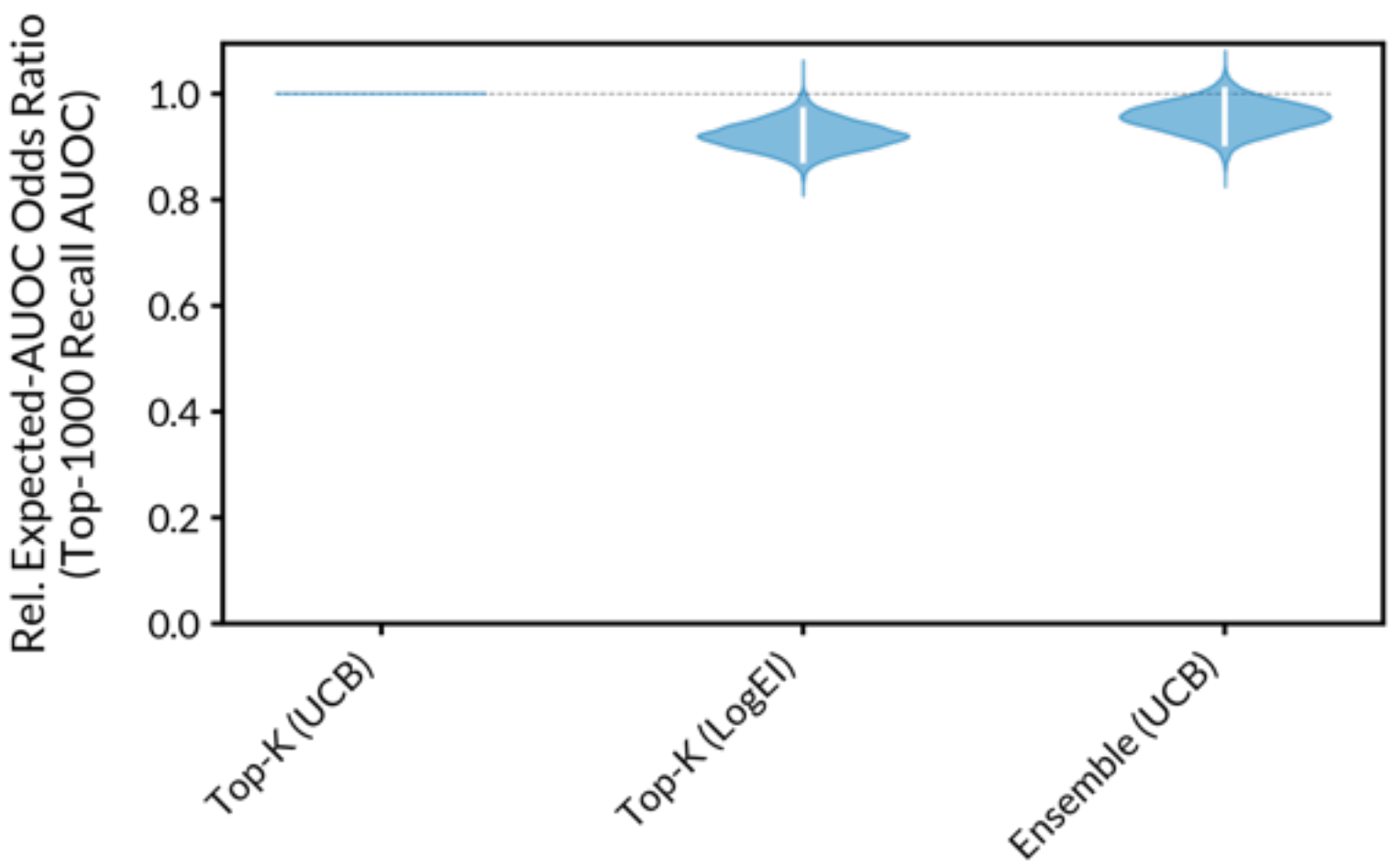

**Figure S 48**: Effect ratios of the acquisition strategy relative to Top-*K* acquisition with a UCB acquisition function, averaged over all libraries. Experiments were performed using a Laplace Neural network model on MolFormer embeddings in the experimental discovery setting (1,000 experiments, batch size 50). Bars within the violins indicate the 95% highest-density intervals.

# 4 Evaluating Molecular Representations

## 4.1 Heuristic Representations and Pre-Trained Transformer Embeddings

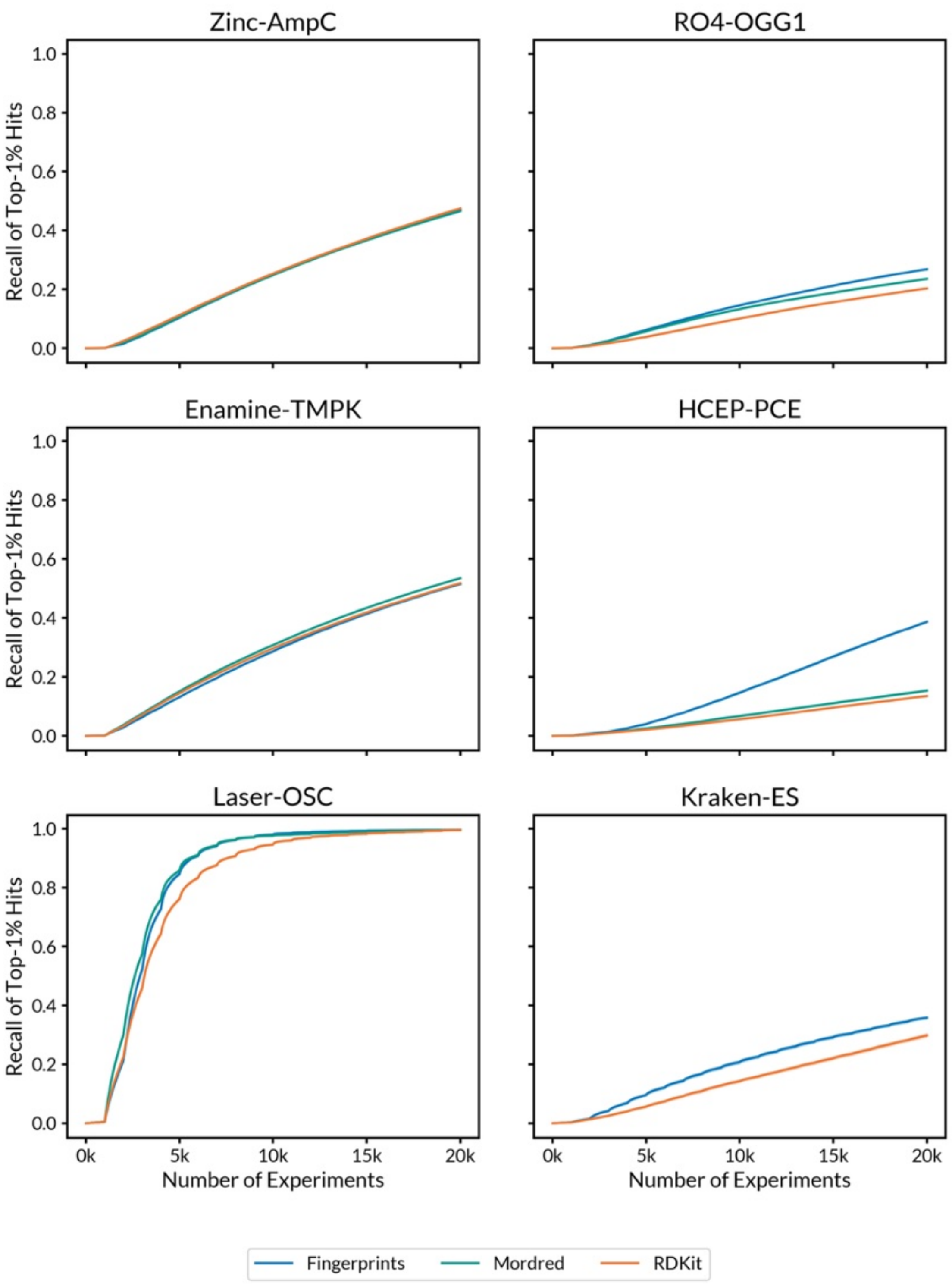

**Figure S 49**: Optimization trajectories for different domain-specific encoders using a Laplace Neural Network surrogate model, evaluated over a budget of 20,000 experiments (batch size of 1,000 experiments). Trajectories are shown as the mean over 20 independent campaigns from different random seed populations. The shaded areas indicate the standard error of the mean.

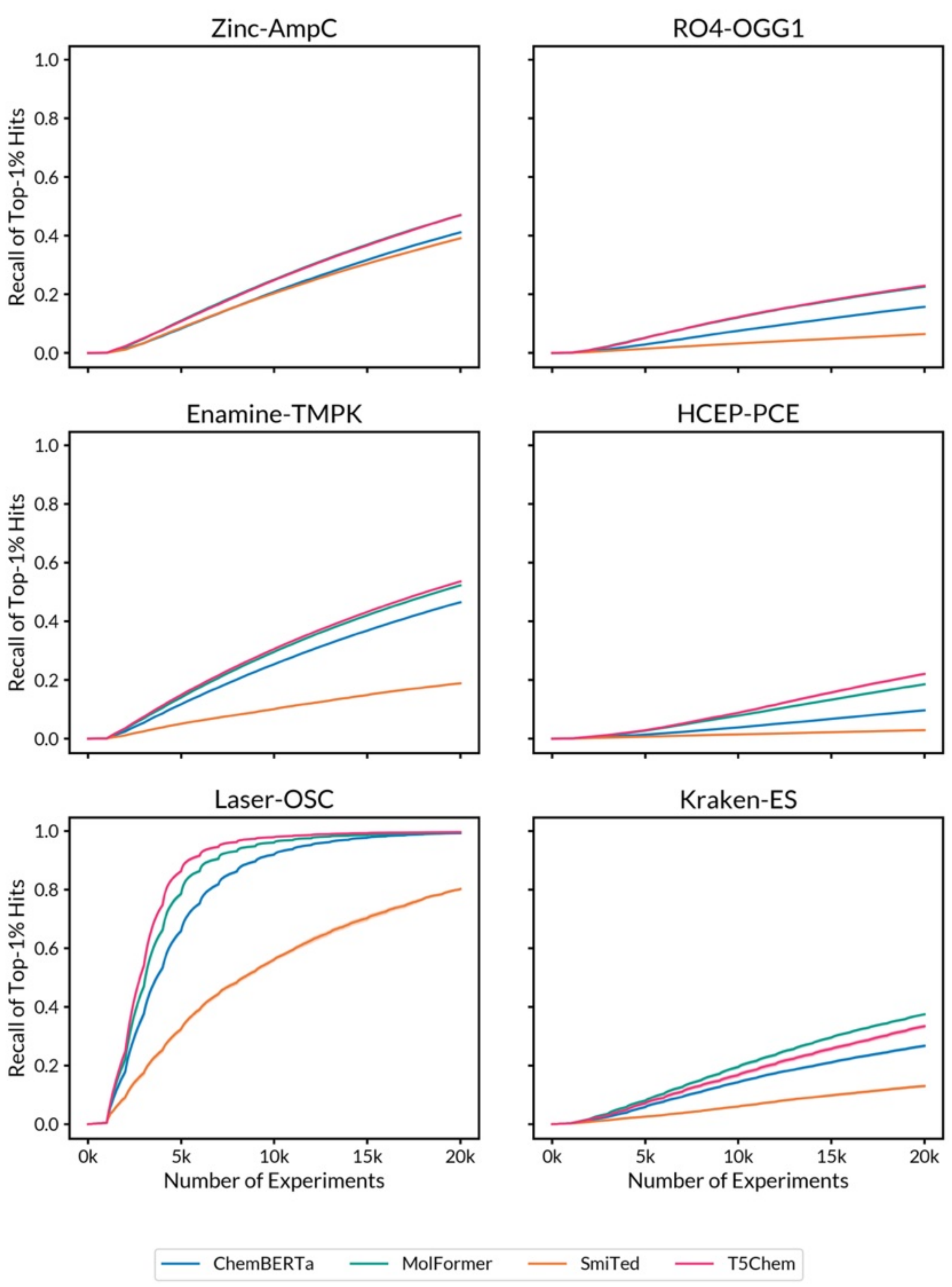

**Figure S 50**: Optimization trajectories for different molecular language model encoders using a Laplace Neural Network surrogate model, evaluated over a budget of 20,000 experiments (batch size of 1,000 experiments). Trajectories are shown as the mean over 20 independent campaigns from different random seed populations. The shaded areas indicate the standard error of the mean.

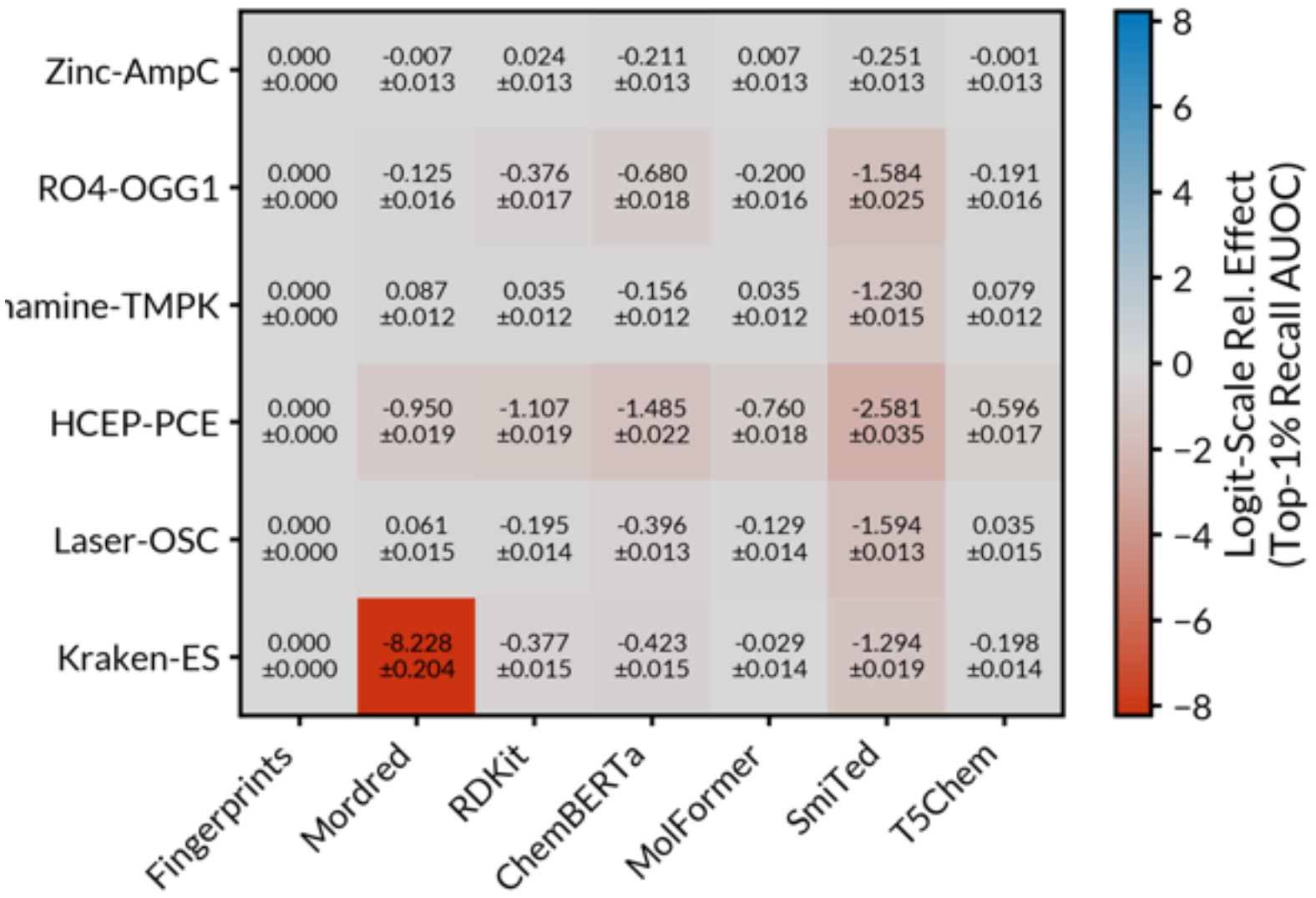

**Figure S 51**: Logit-scale effect of the molecular encoder relative to Morgan fingerprints, differentiated by molecular libraries. Experiments were performed using a Laplace Neural network model in the virtual screening setting (20,000 experiments, batch size 1,000). Means and standard deviations were obtained from 8,000 posterior samples.

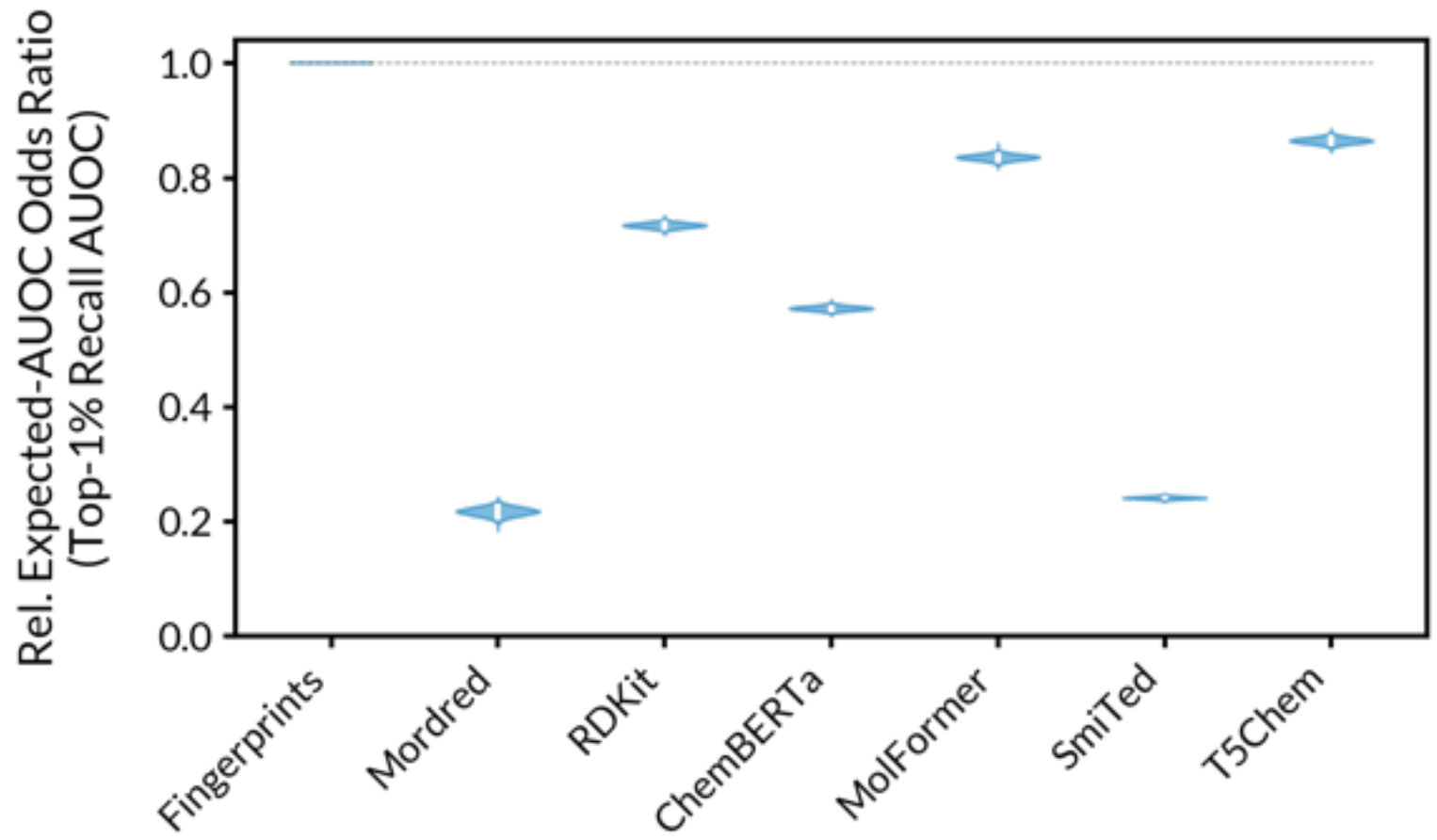

**Figure S 52**: Effect ratios of the molecular encoder relative to Morgan fingerprints, averaged over all libraries. Experiments were performed using a Laplace Neural network model in the virtual screening setting (20,000 experiments, batch size 1,000). Bars within the violins indicate the 95% highest-density intervals.

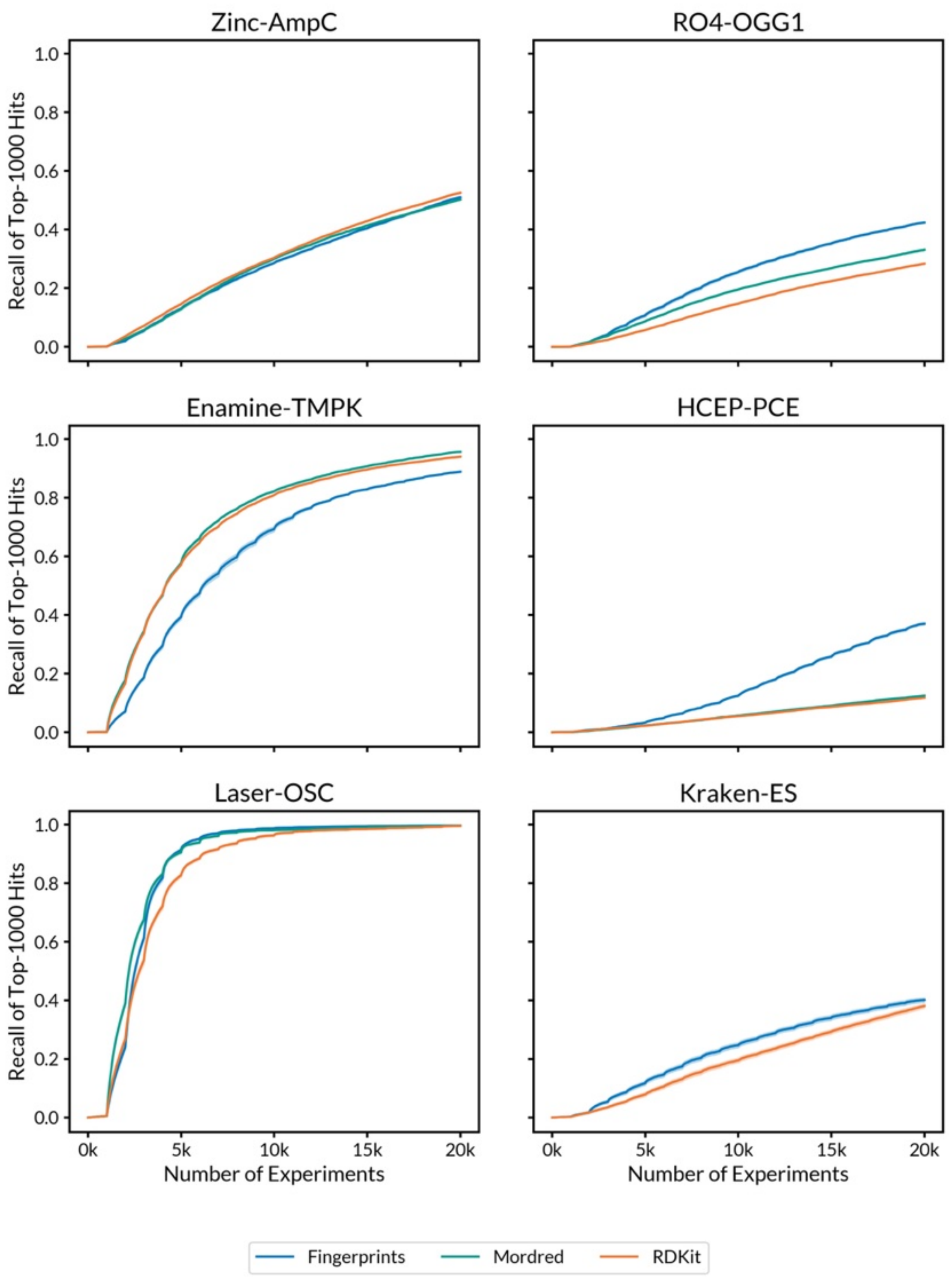

**Figure S 53**: Optimization trajectories for different domain-specific encoders using a Laplace Neural Network surrogate model, evaluated over a budget of 20,000 experiments (batch size of 1,000 experiments). Trajectories are shown as the mean over 20 independent campaigns from different random seed populations. The shaded areas indicate the standard error of the mean.

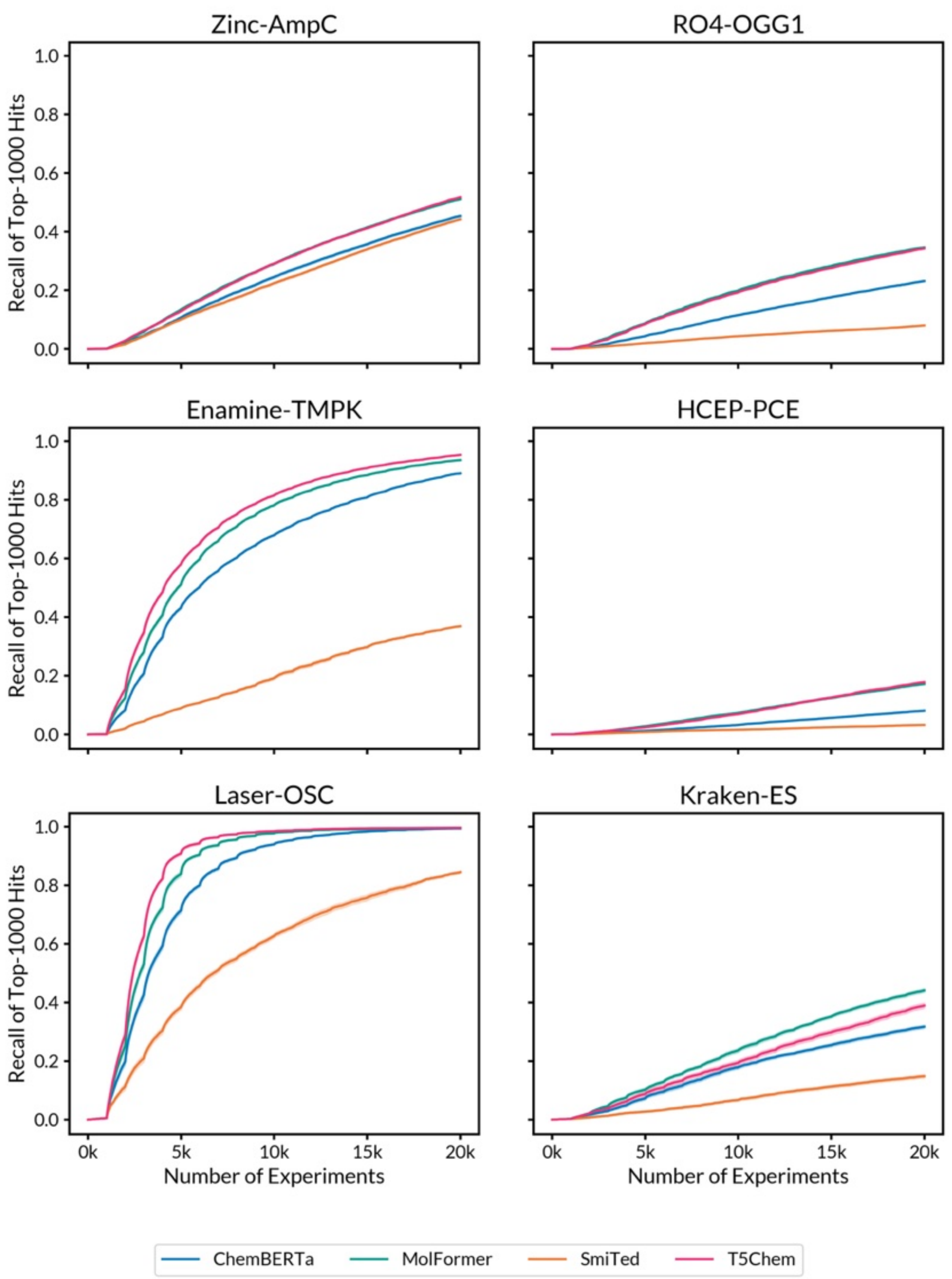

**Figure S 54**: Optimization trajectories for different molecular language model encoders using a Laplace Neural Network surrogate model, evaluated over a budget of 20,000 experiments (batch size of 1,000 experiments). Trajectories are shown as the mean over 20 independent campaigns from different random seed populations. The shaded areas indicate the standard error of the mean.

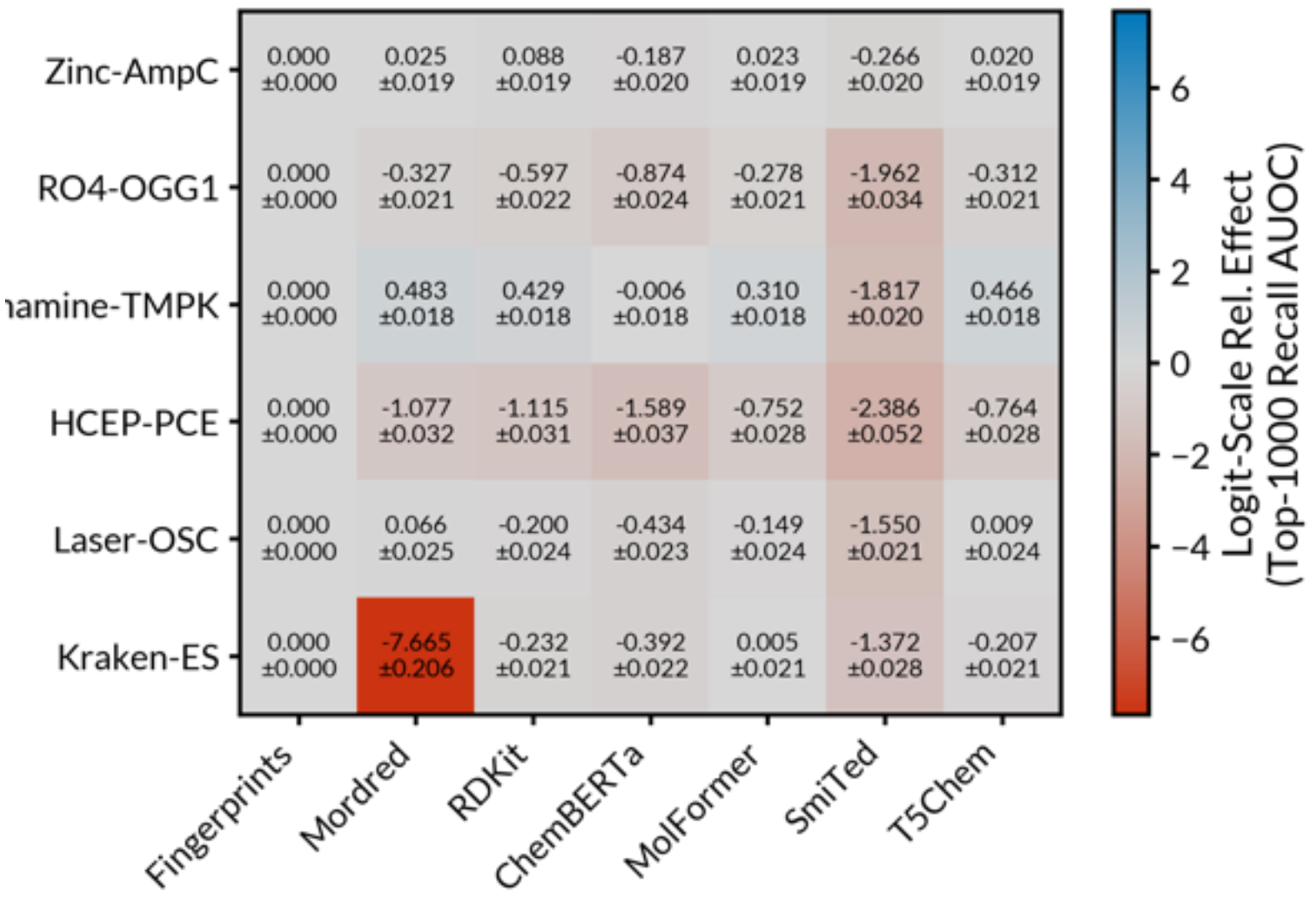

**Figure S 55**: Logit-scale effect of the molecular encoder relative to Morgan fingerprints, differentiated by molecular libraries. Experiments were performed using a Laplace Neural network model in the virtual screening setting (20,000 experiments, batch size 1,000). Means and standard deviations were obtained from 8,000 posterior samples.

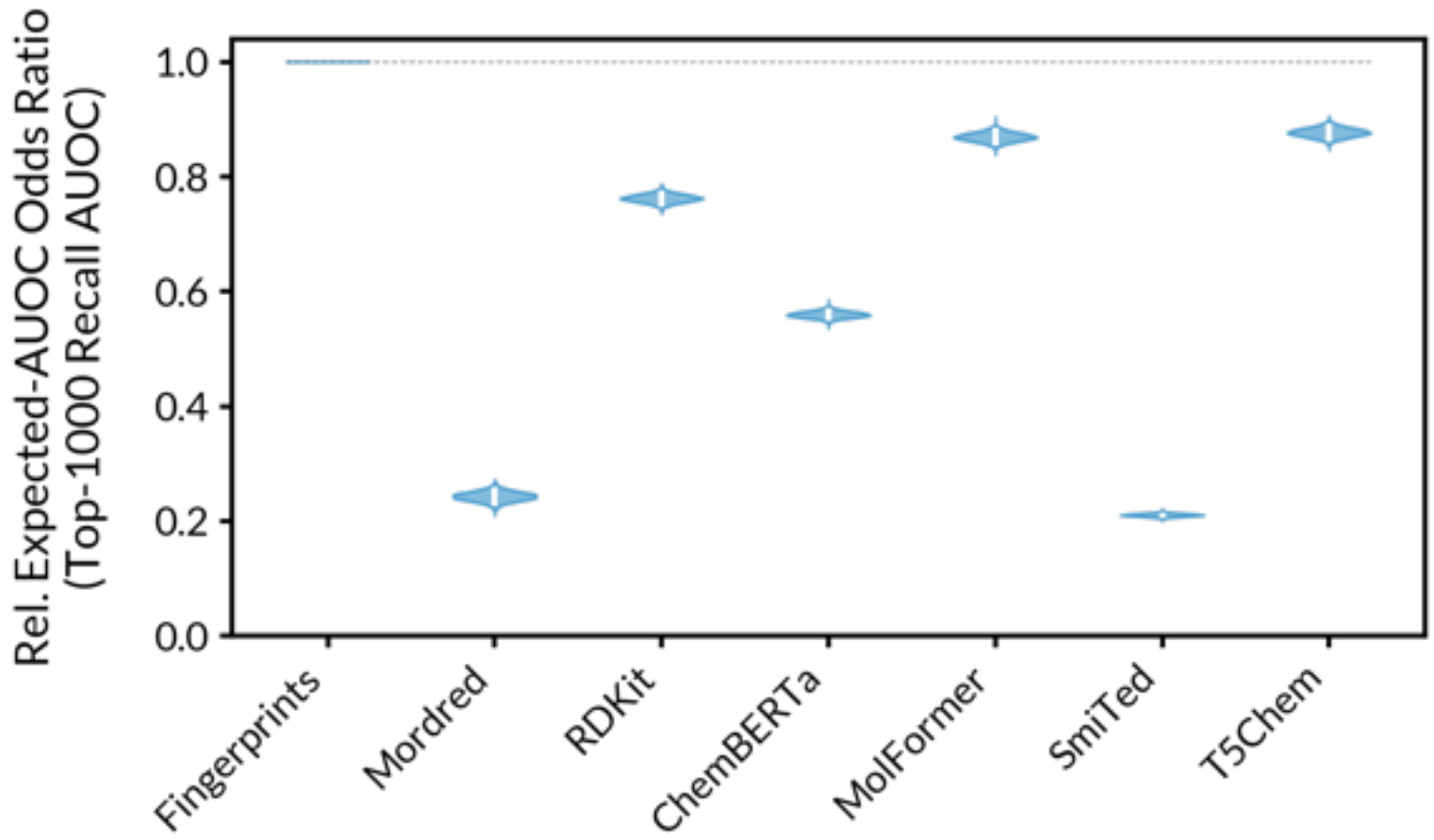

**Figure S 56**: Effect ratios of the molecular encoder relative to Morgan fingerprints, averaged over all libraries. Experiments were performed using a Laplace Neural network model in the virtual screening setting (20,000 experiments, batch size 1,000). Bars within the violins indicate the 95% highest-density intervals.

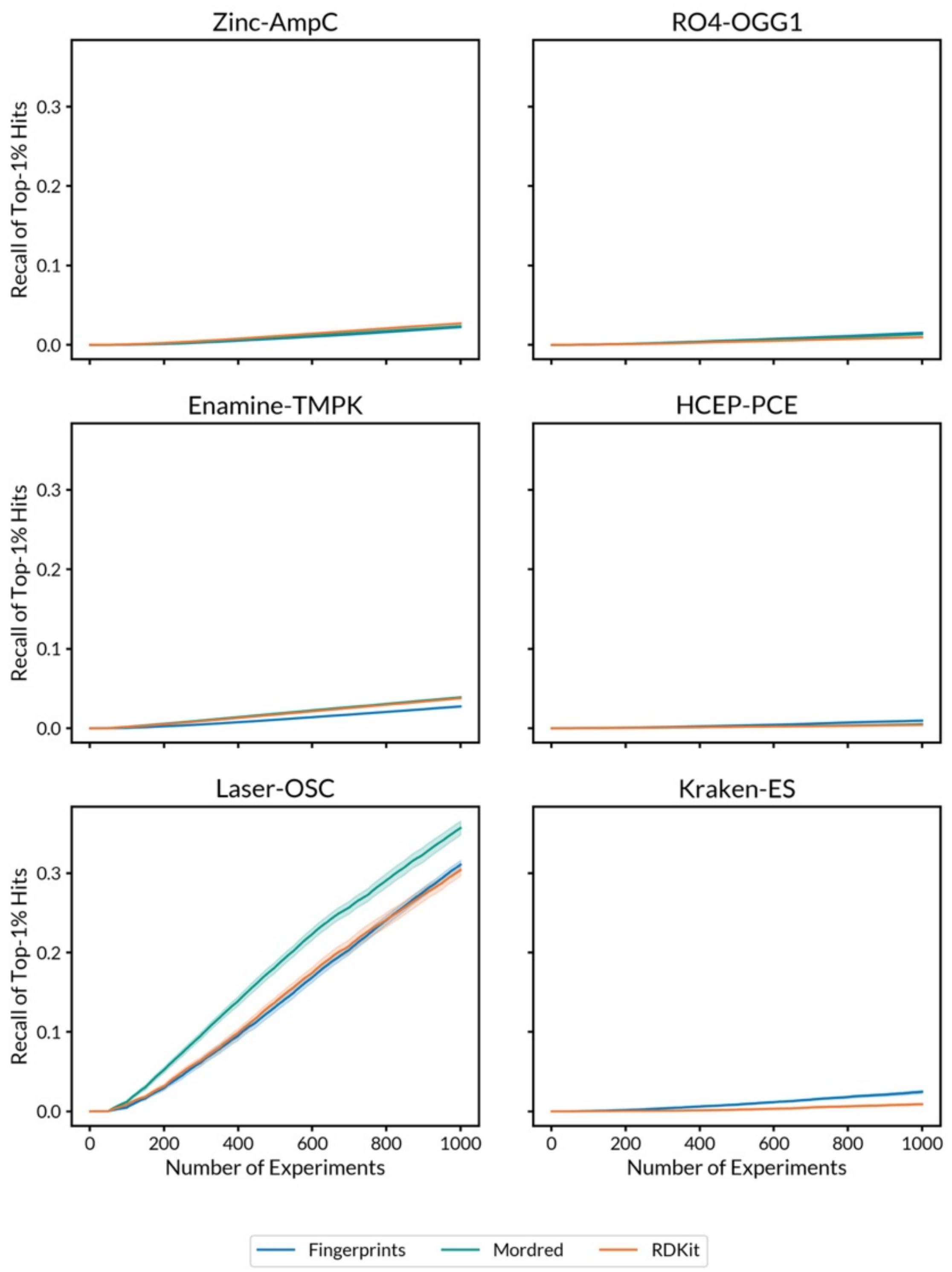

**Figure S 57**: Optimization trajectories for different domain-specific encoders using a Laplace Neural Network surrogate model, evaluated over a budget of 1,000 experiments (batch size of 50 experiments). Trajectories are shown as the mean over 20 independent campaigns from different random seed populations. The shaded areas indicate the standard error of the mean.

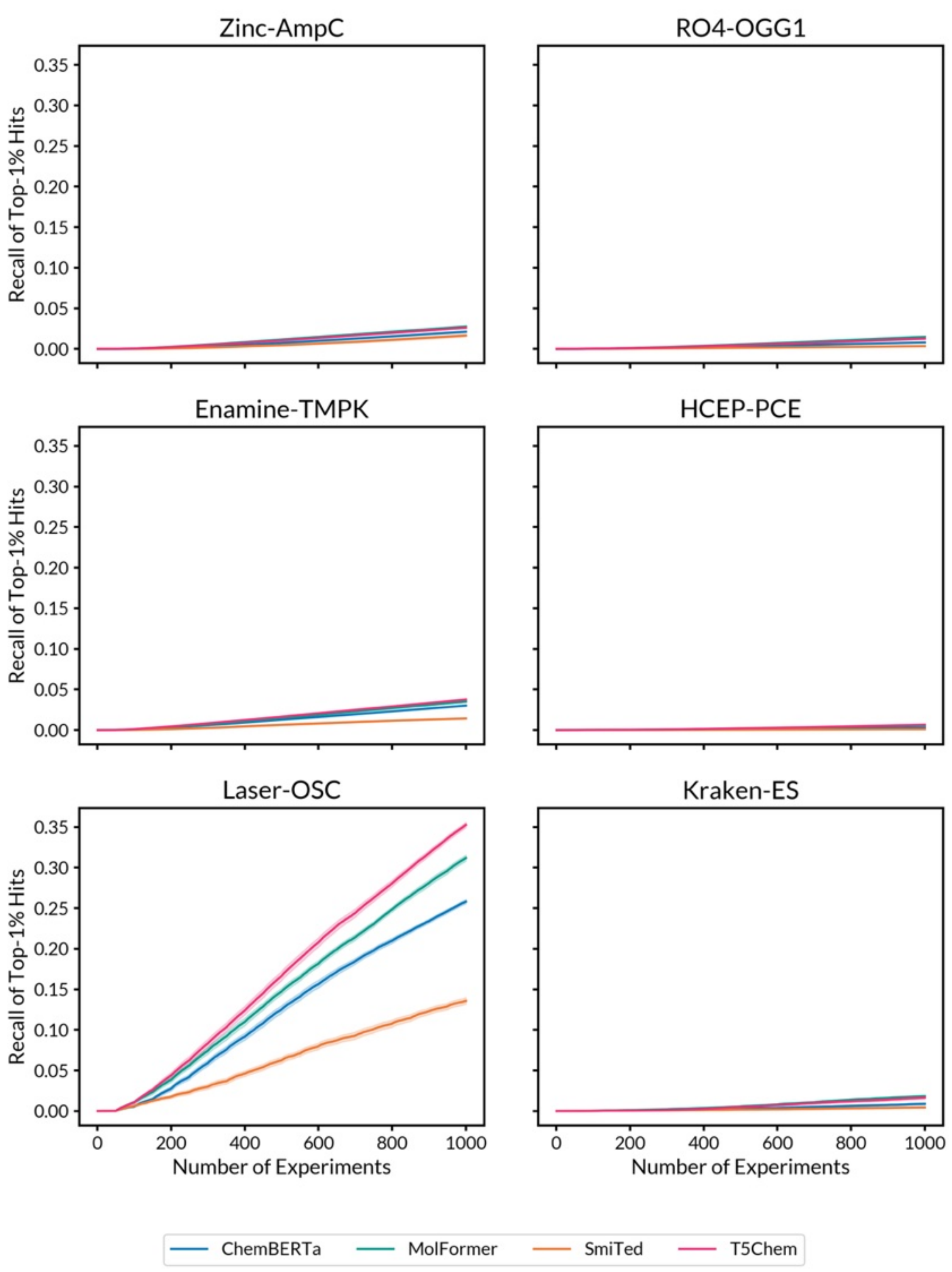

**Figure S 58**: Optimization trajectories for different molecular language model encoders using a Laplace Neural Network surrogate model, evaluated over a budget of 1,000 experiments (batch size of 50 experiments). Trajectories are shown as the mean over 20 independent campaigns from different random seed populations. The shaded areas indicate the standard error of the mean.

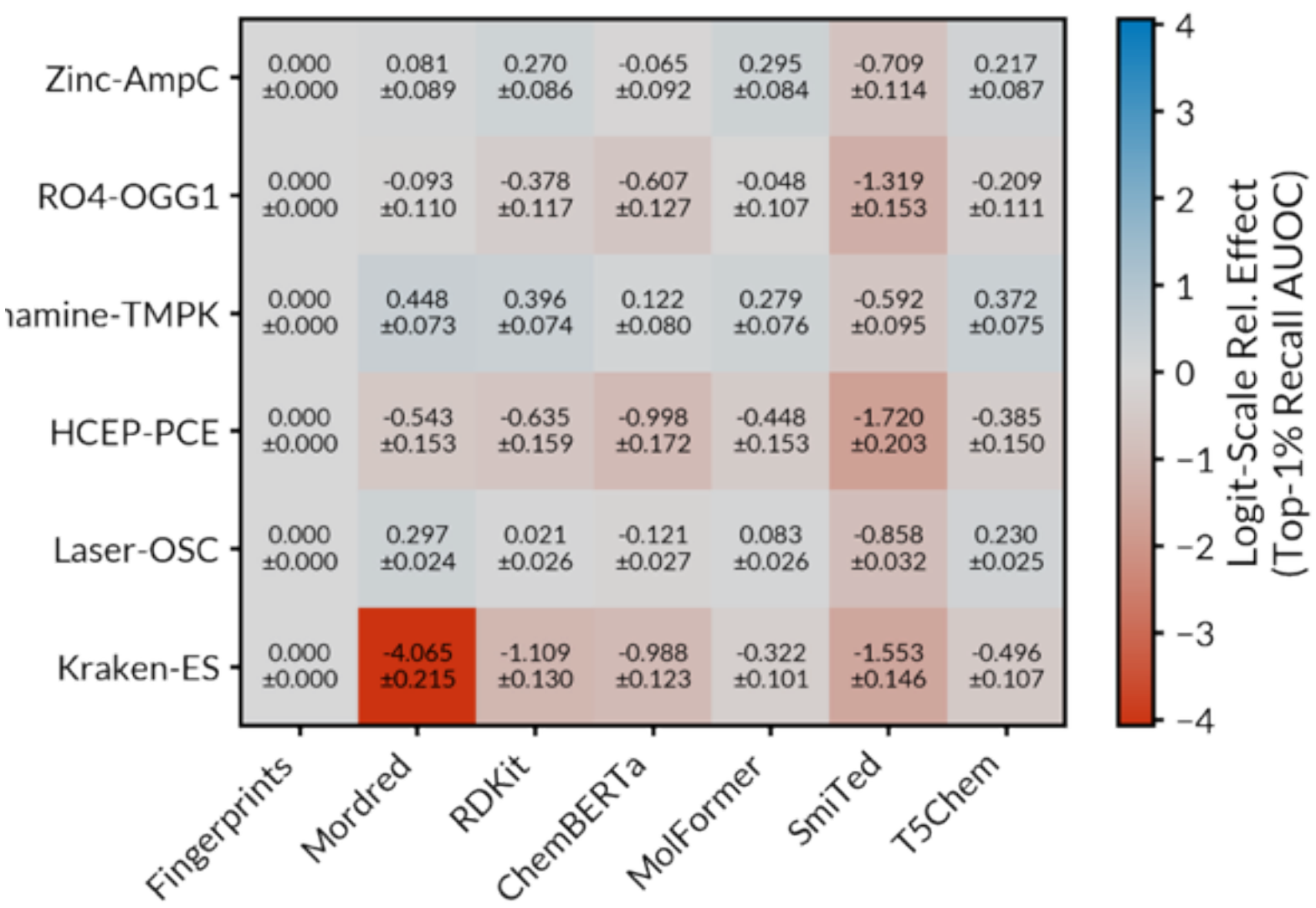

**Figure S 59**: Logit-scale effect of the molecular encoder relative to Morgan fingerprints, differentiated by molecular libraries. Experiments were performed using a Laplace Neural network model in the experimental discovery setting (1,000 experiments, batch size 50). Means and standard deviations were obtained from 8,000 posterior samples.

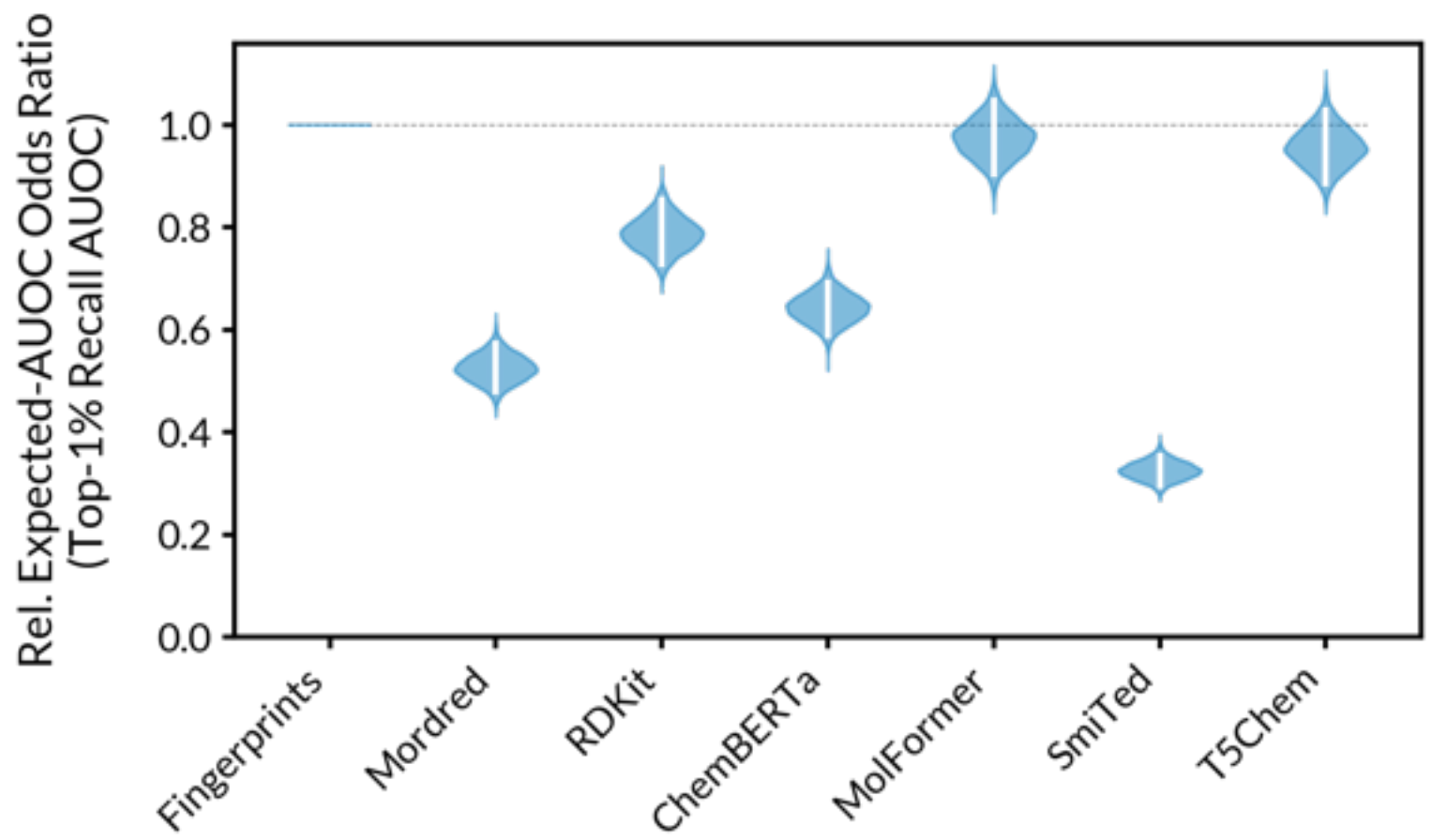

**Figure S 60**: Effect ratios of the molecular encoder relative to Morgan fingerprints, averaged over all libraries. Experiments were performed using a Laplace Neural network model in experimental discovery setting (1,000 experiments, batch size 1,000). Bars within the violins indicate the 95% highest-density intervals.

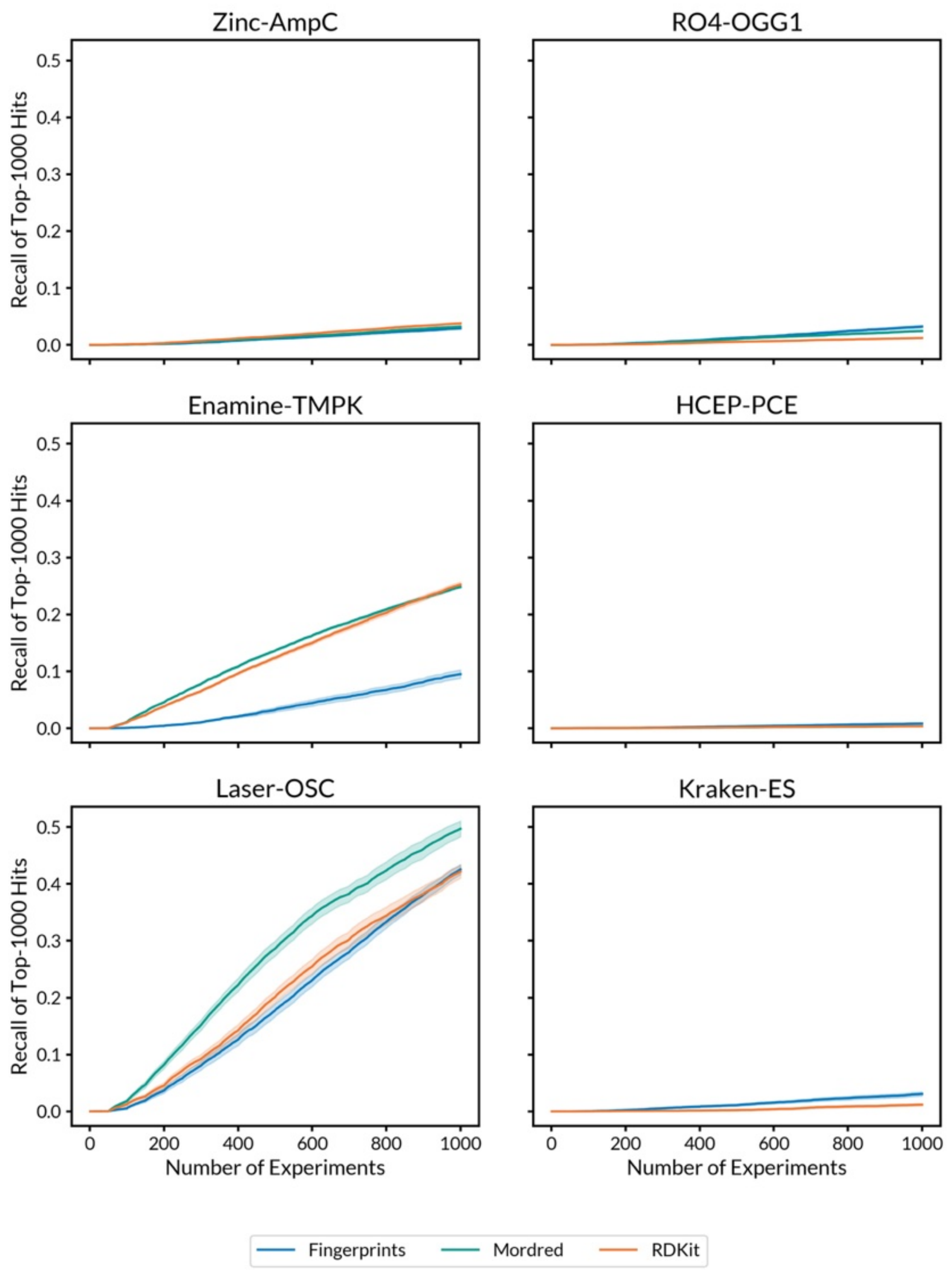

**Figure S 61**: Optimization trajectories for different domain-specific encoders using a Laplace Neural Network surrogate model, evaluated over a budget of 1,000 experiments (batch size of 50 experiments). Trajectories are shown as the mean over 20 independent campaigns from different random seed populations. The shaded areas indicate the standard error of the mean.

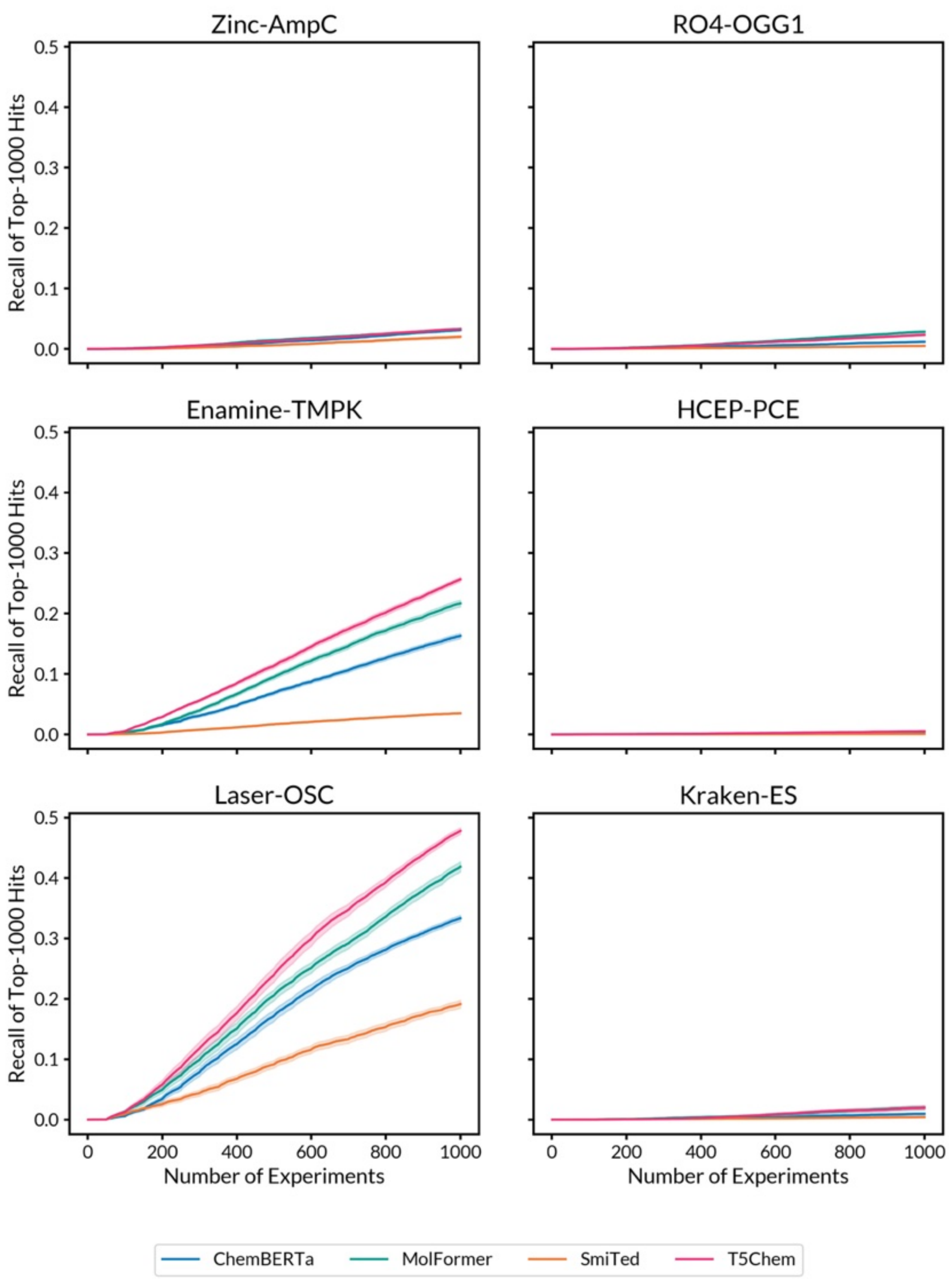

**Figure S 62**: Optimization trajectories for different molecular language model encoders using a Laplace Neural Network surrogate model, evaluated over a budget of 1,000 experiments (batch size of 50 experiments). Trajectories are shown as the mean over 20 independent campaigns from different random seed populations. The shaded areas indicate the standard error of the mean.

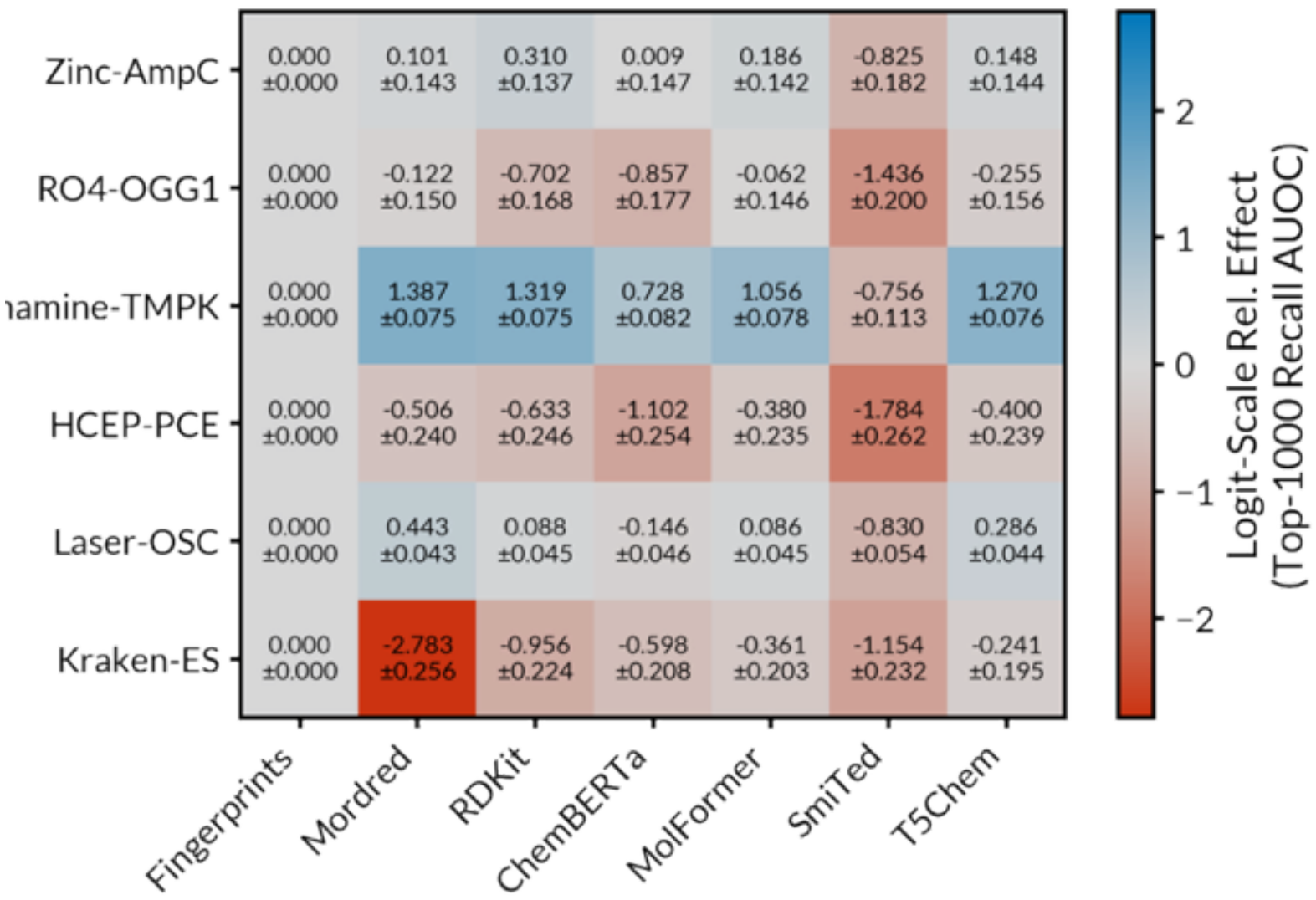

**Figure S 63**: Logit-scale effect of the molecular encoder relative to Morgan fingerprints, differentiated by molecular libraries. Experiments were performed using a Laplace Neural network model in the experimental discovery setting (1,000 experiments, batch size 50). Means and standard deviations were obtained from 8,000 posterior samples.

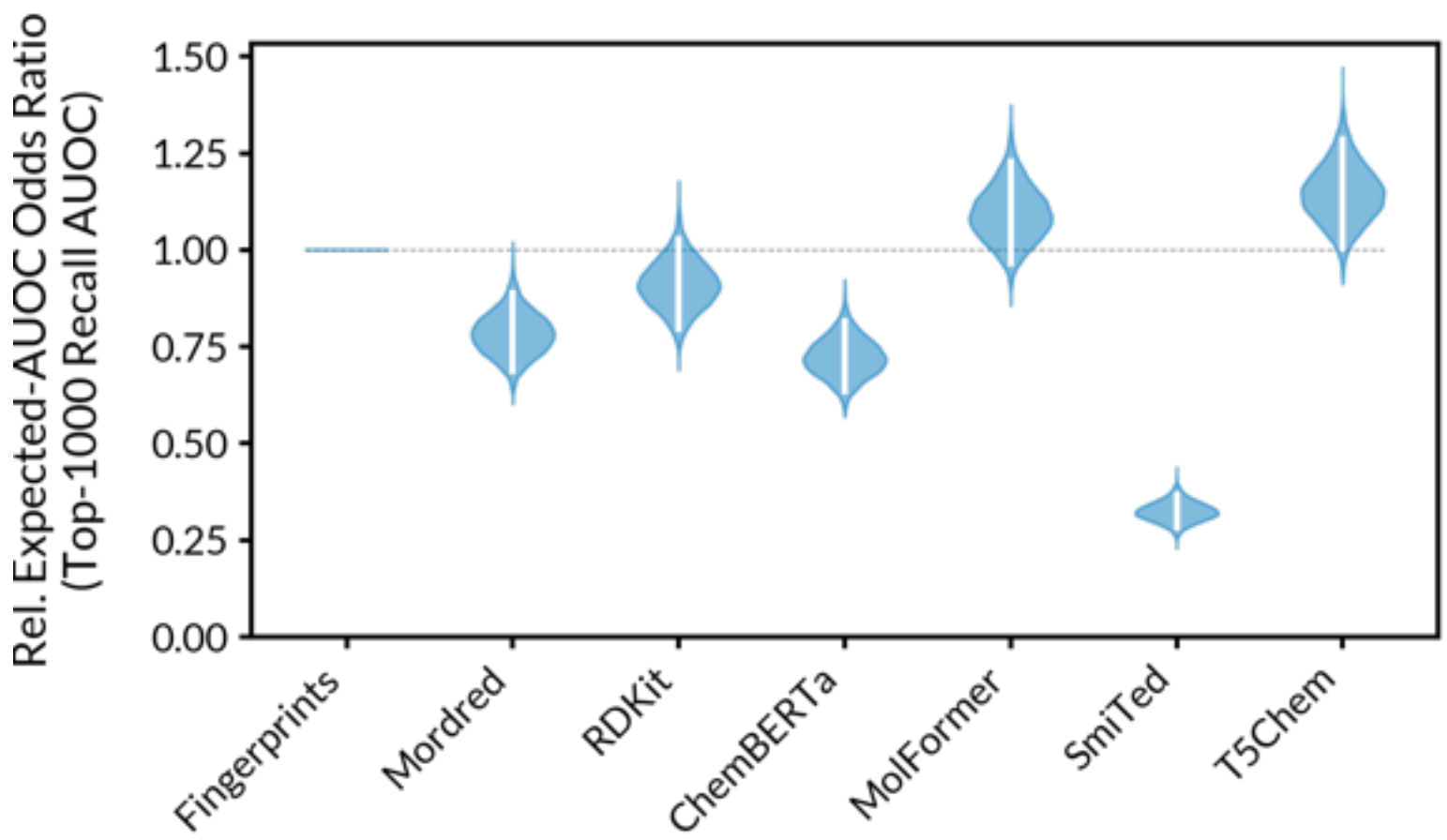

**Figure S 64**: Effect ratios of the molecular encoder relative to Morgan fingerprints, averaged over all libraries. Experiments were performed using a Laplace Neural network model in experimental discovery setting (1,000 experiments, batch size 1,000). Bars within the violins indicate the 95% highest-density intervals.

## 4.2 Transformer Pooling Strategies

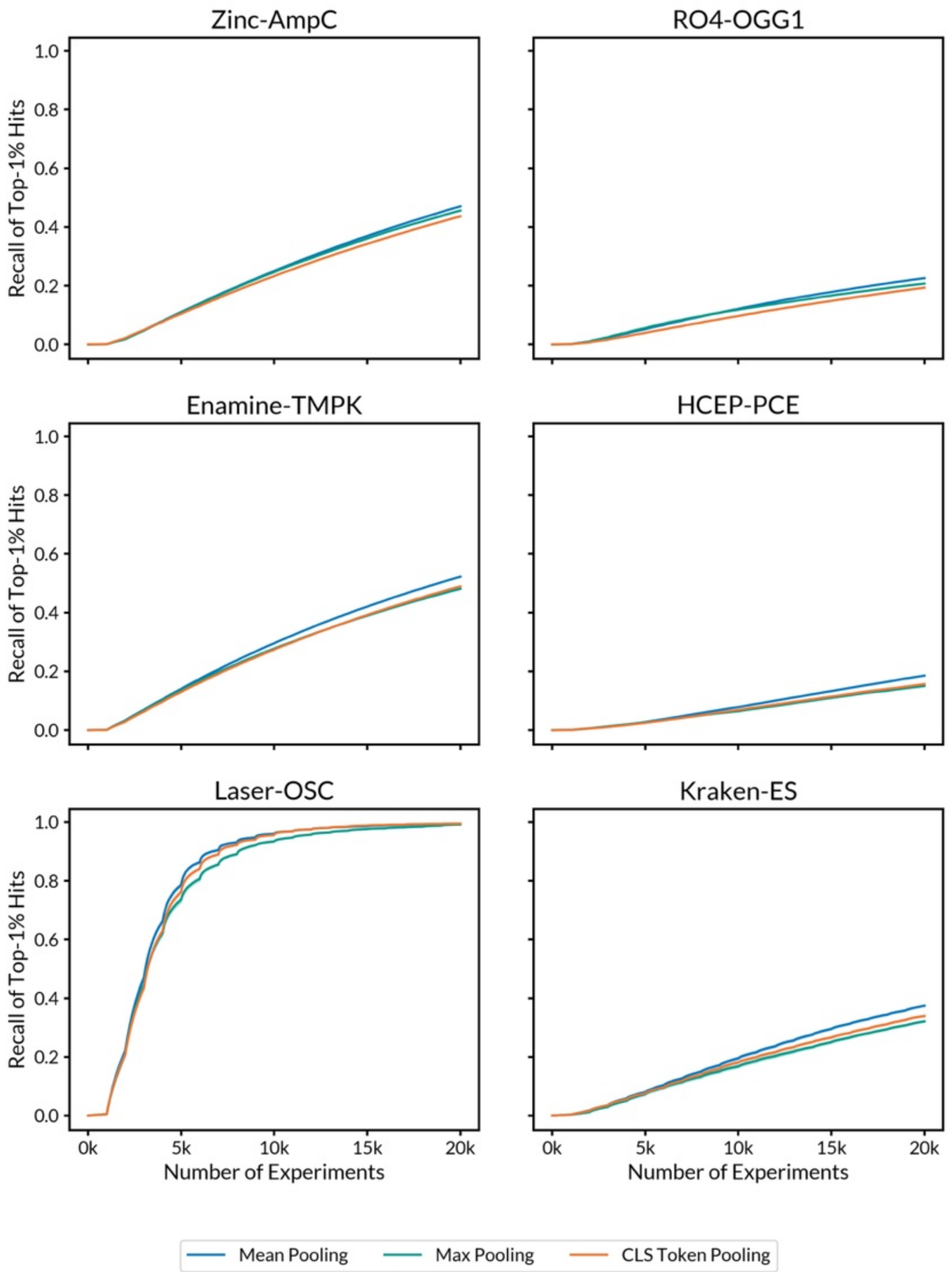

**Figure S 65**: Optimization trajectories for different pooling strategies on a MolFormer encoder using a Laplace Neural Network surrogate model, evaluated over a budget of 20,000 experiments (batch size of 1,000 experiments). Trajectories are shown as the mean over 20 independent campaigns from different random seed populations. The shaded areas indicate the standard error of the mean.

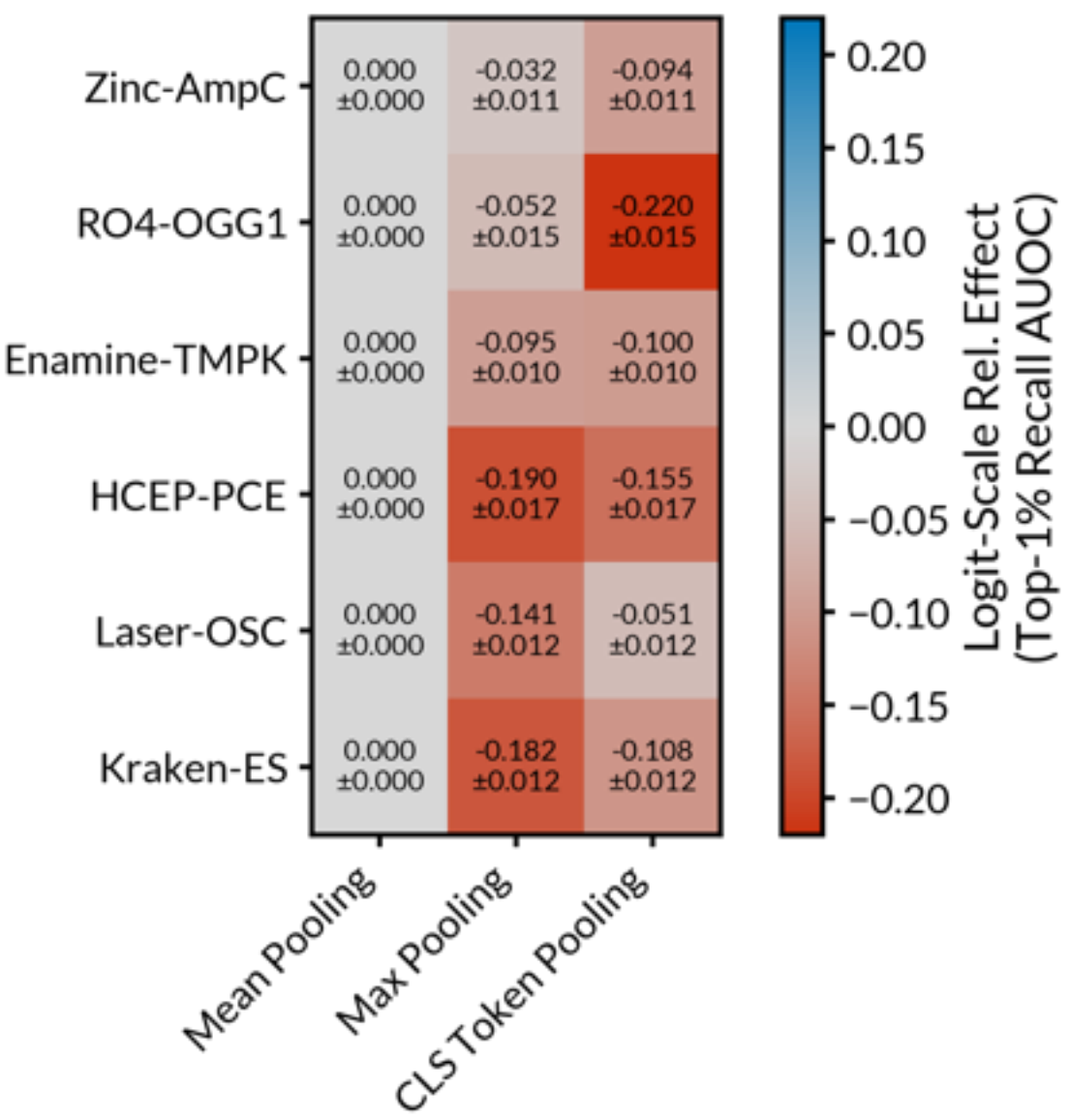

**Figure S 66**: Logit-scale effect of the transformer pooling strategy relative to mean pooling, differentiated by molecular libraries. Experiments were performed using a Laplace Neural network model on MolFormer embeddings in the virtual screening setting (20,000 experiments, batch size 1,000). Means and standard deviations were obtained from 8,000 posterior samples.

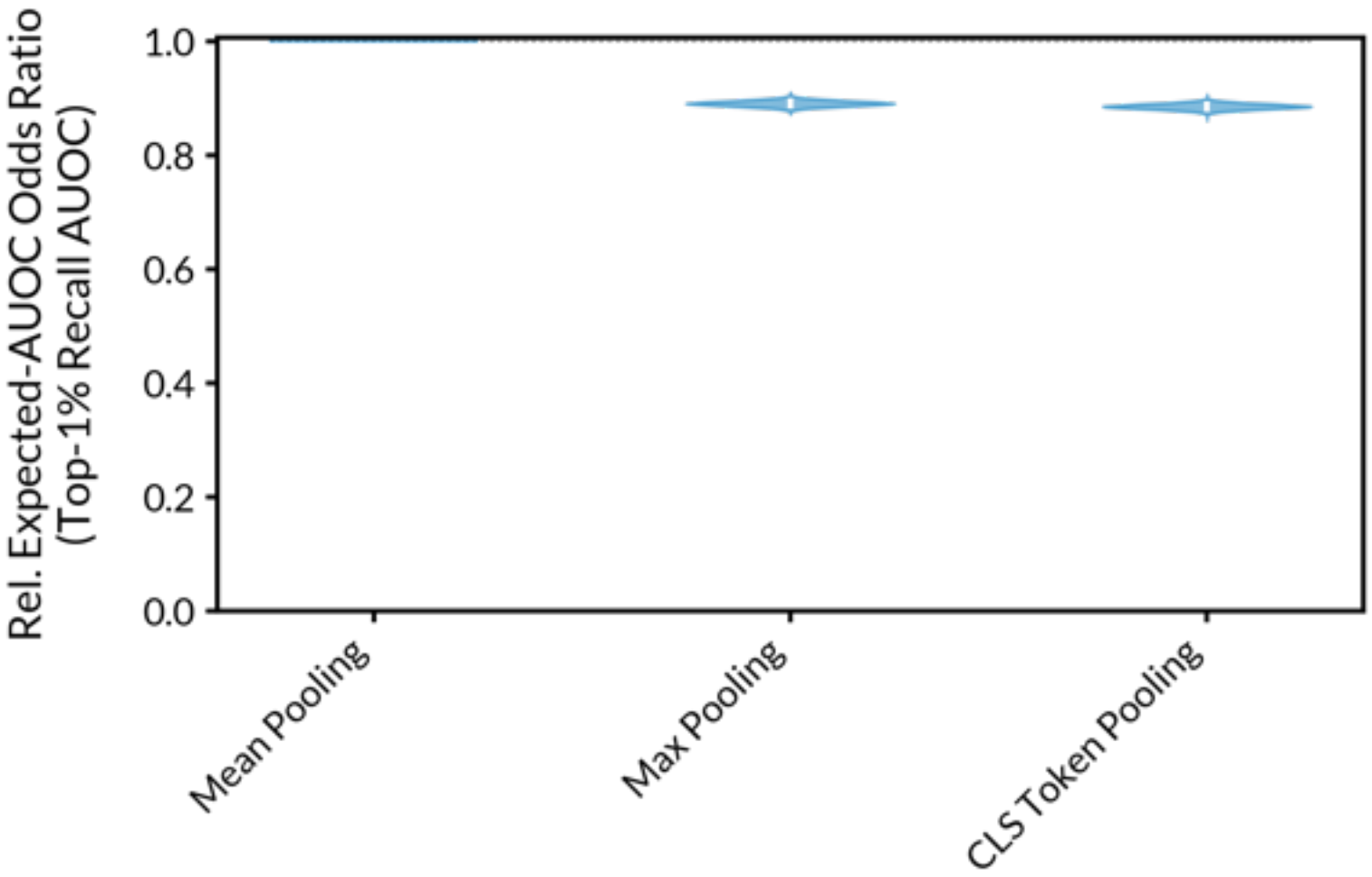

**Figure S 67**: Effect ratios of the transformer pooling strategy relative to mean pooling, averaged over all libraries. Experiments were performed using a Laplace Neural network model on MolFormer embeddings in the virtual screening setting (20,000 experiments, batch size 1,000). Bars within the violins indicate the 95% highest-density intervals.

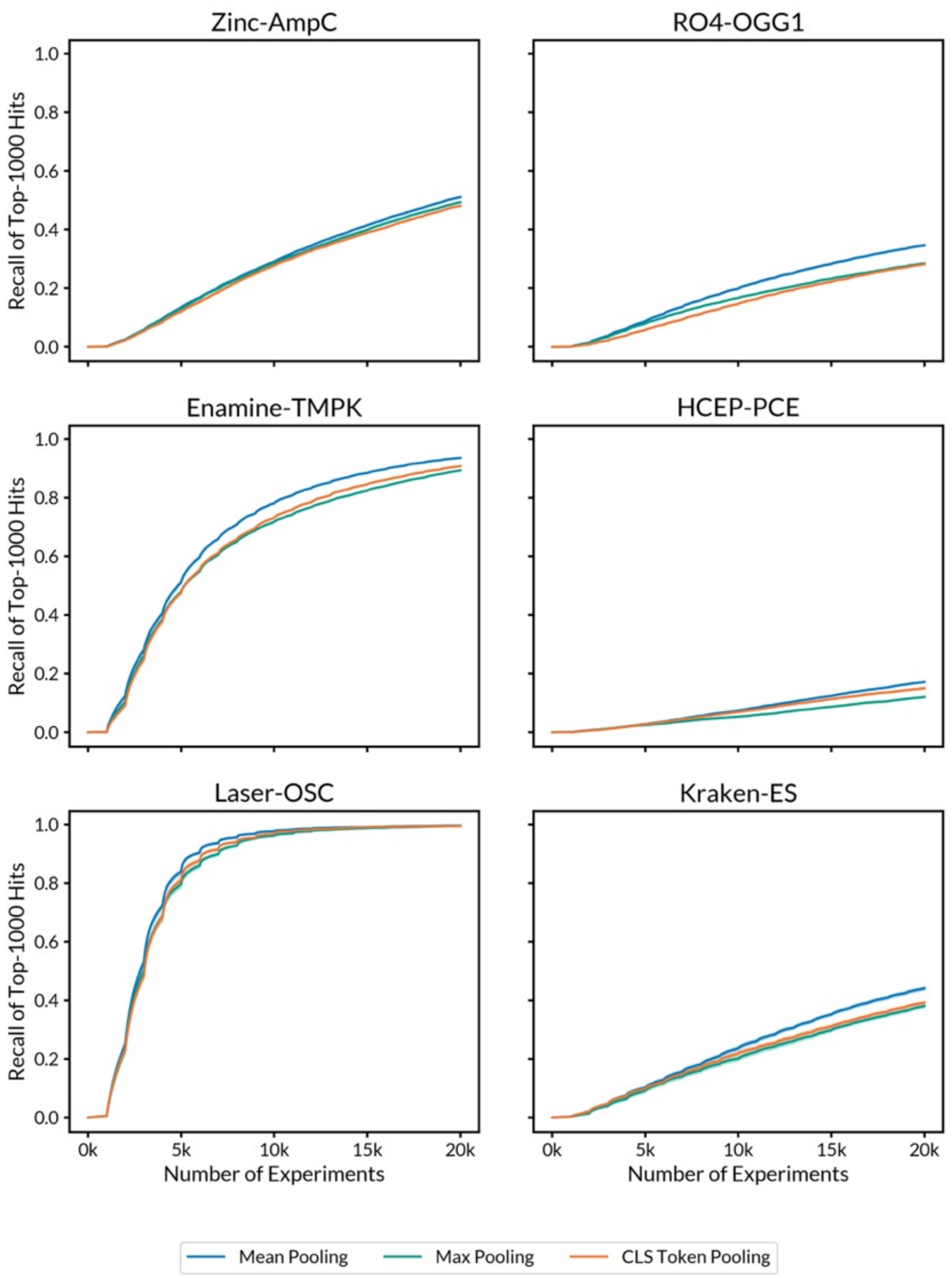

**Figure S 68**: Optimization trajectories for different pooling strategies on a MolFormer encoder using a Laplace Neural Network surrogate model, evaluated over a budget of 20,000 experiments (batch size of 1,000 experiments). Trajectories are shown as the mean over 20 independent campaigns from different random seed populations. The shaded areas indicate the standard error of the mean.

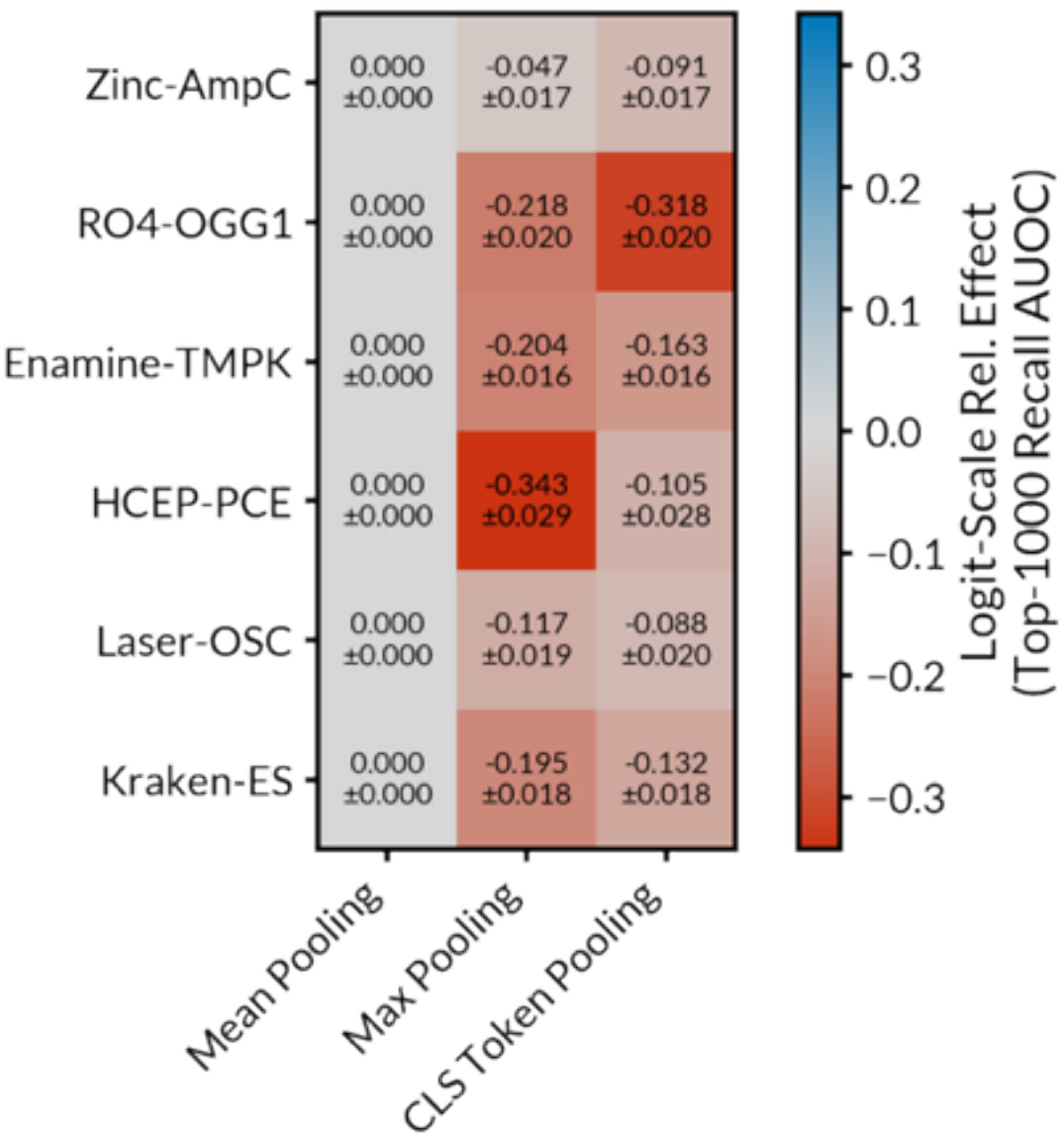

**Figure S 69**: Logit-scale effect of the transformer pooling strategy relative to mean pooling, differentiated by molecular libraries. Experiments were performed using a Laplace Neural network model on MolFormer embeddings in the virtual screening setting (20,000 experiments, batch size 1,000). Means and standard deviations were obtained from 8,000 posterior samples.

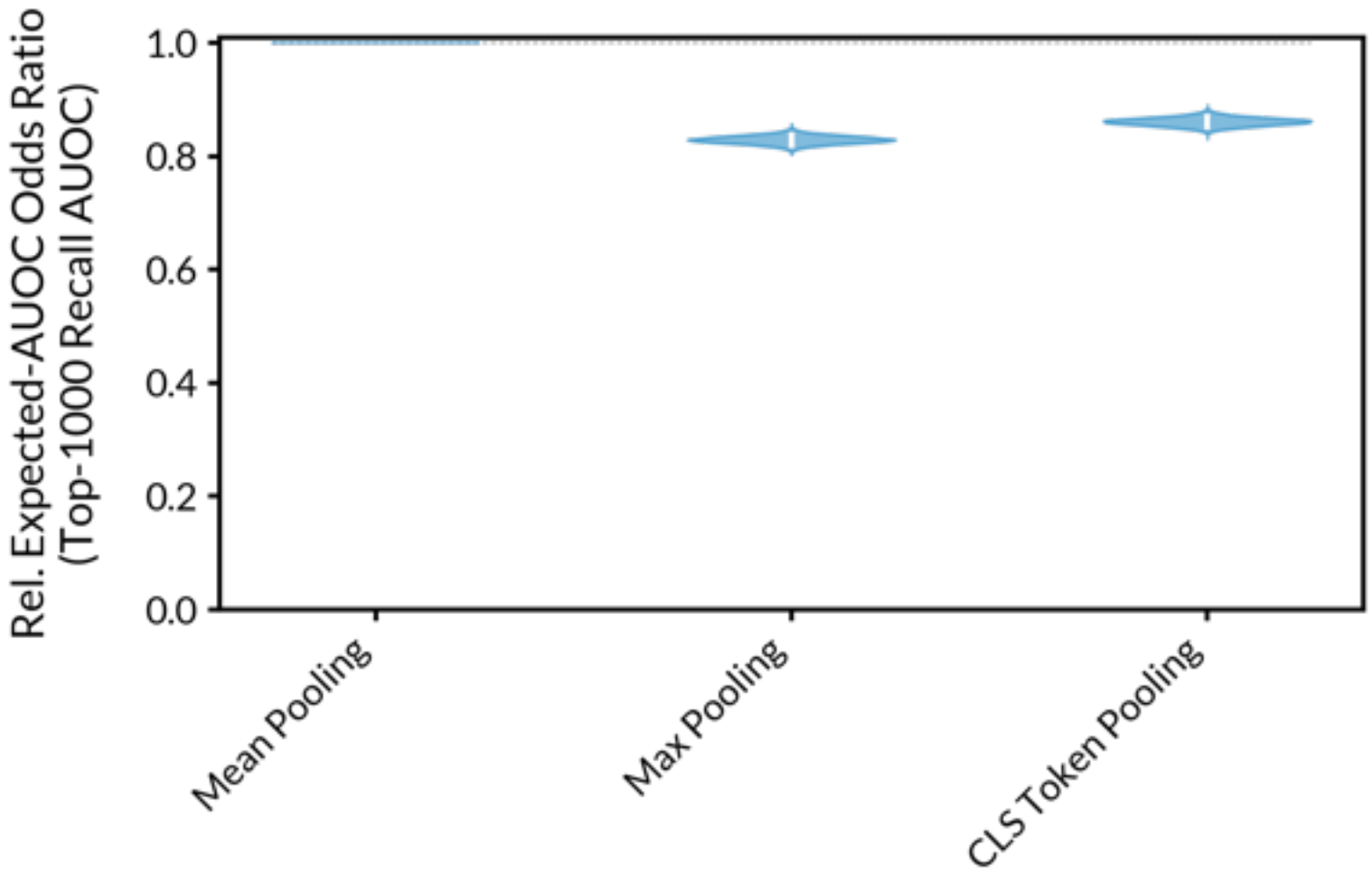

**Figure S 70**: Effect ratios of the transformer pooling strategy relative to mean pooling, averaged over all libraries. Experiments were performed using a Laplace Neural network model on MolFormer embeddings in the virtual screening setting (20,000 experiments, batch size 1,000). Bars within the violins indicate the 95% highest-density intervals.

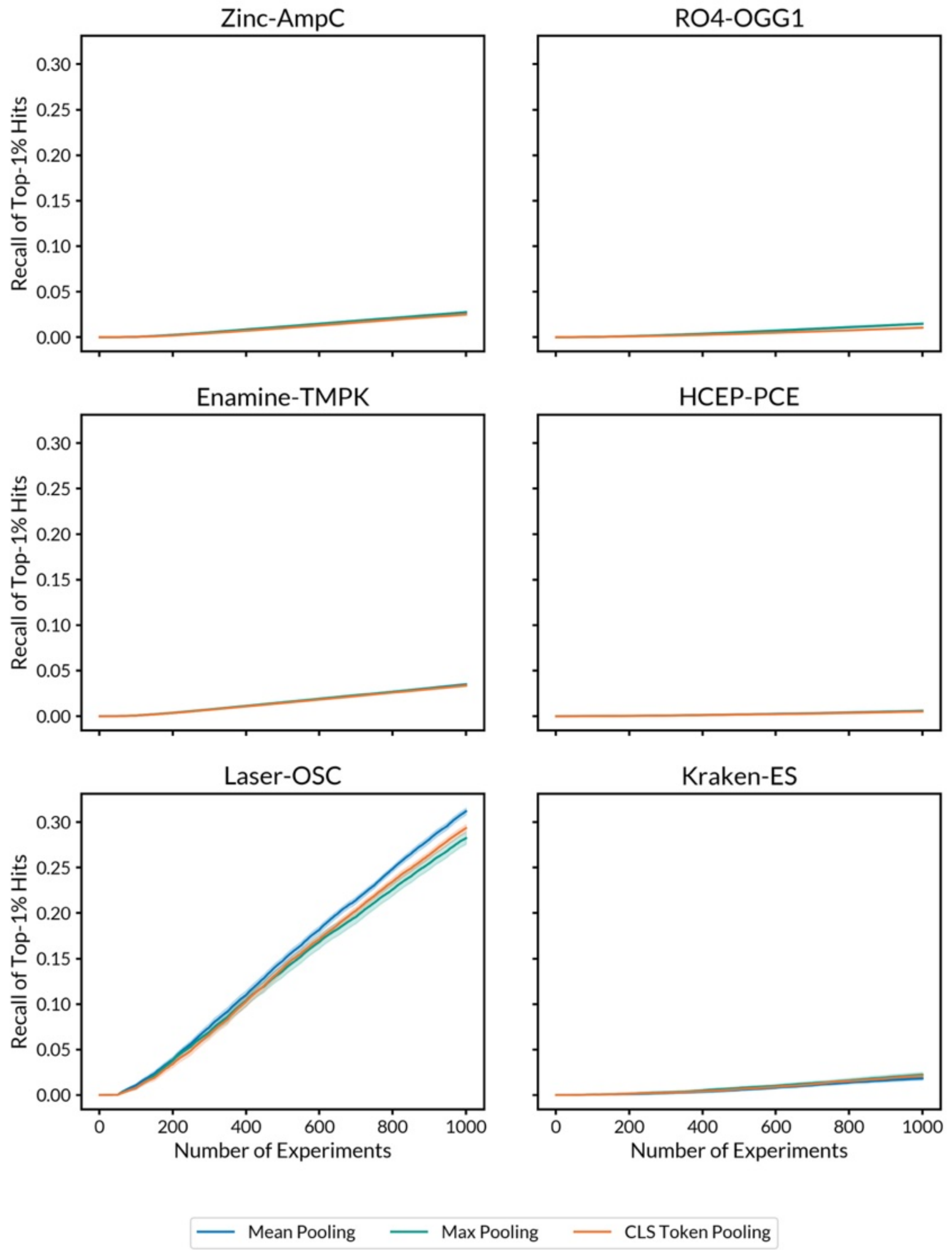

**Figure S 71**: Optimization trajectories for different pooling strategies on a MolFormer encoder using a Laplace Neural Network surrogate model, evaluated over a budget of 1,000 experiments (batch size of 50 experiments). Trajectories are shown as the mean over 20 independent campaigns from different random seed populations. The shaded areas indicate the standard error of the mean.

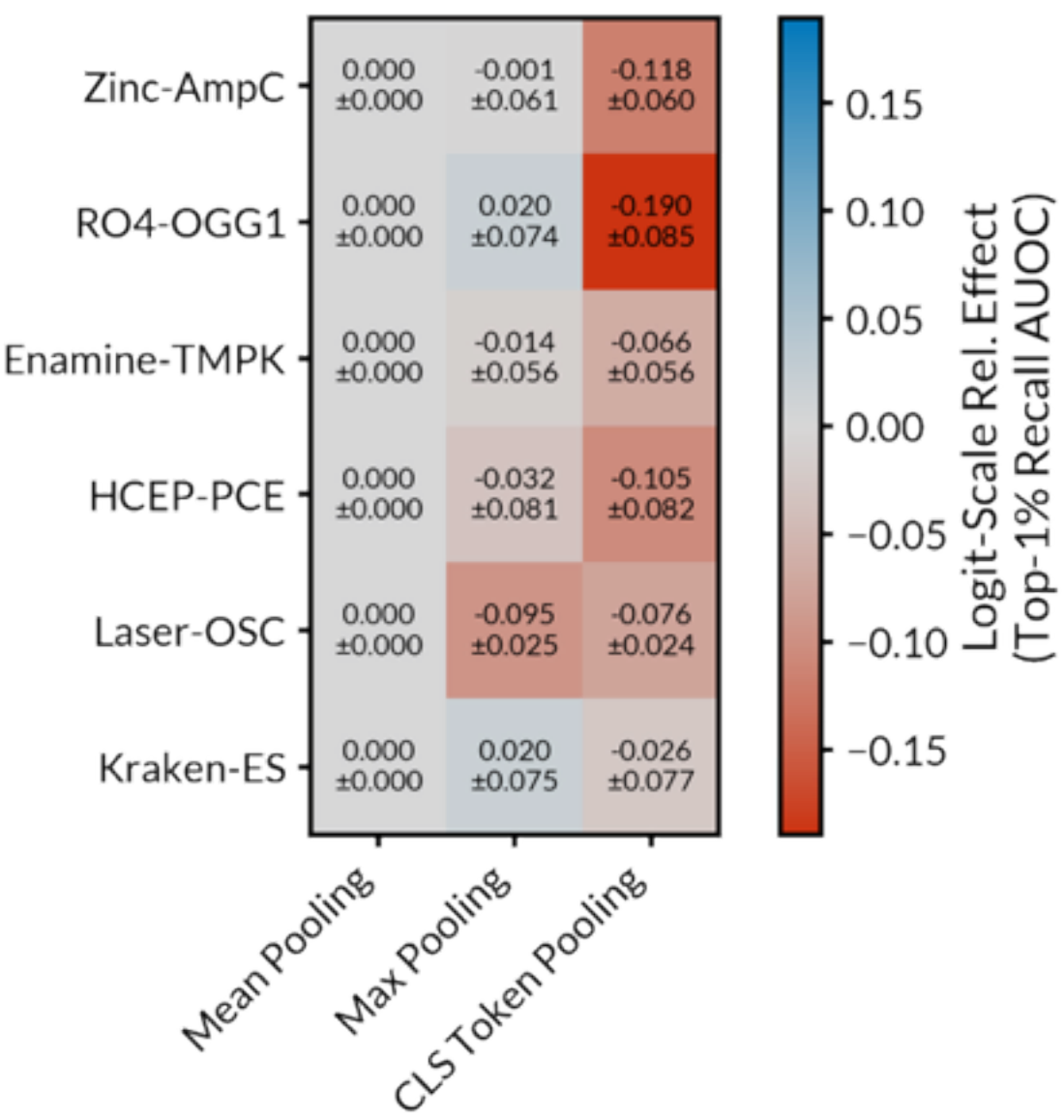

**Figure S 72**: Logit-scale effect of the transformer pooling strategy relative to mean pooling, differentiated by molecular libraries. Experiments were performed using a Laplace Neural network model on MolFormer embeddings in the experimental discovery setting (1,000 experiments, batch size 50). Means and standard deviations were obtained from 8,000 posterior samples.

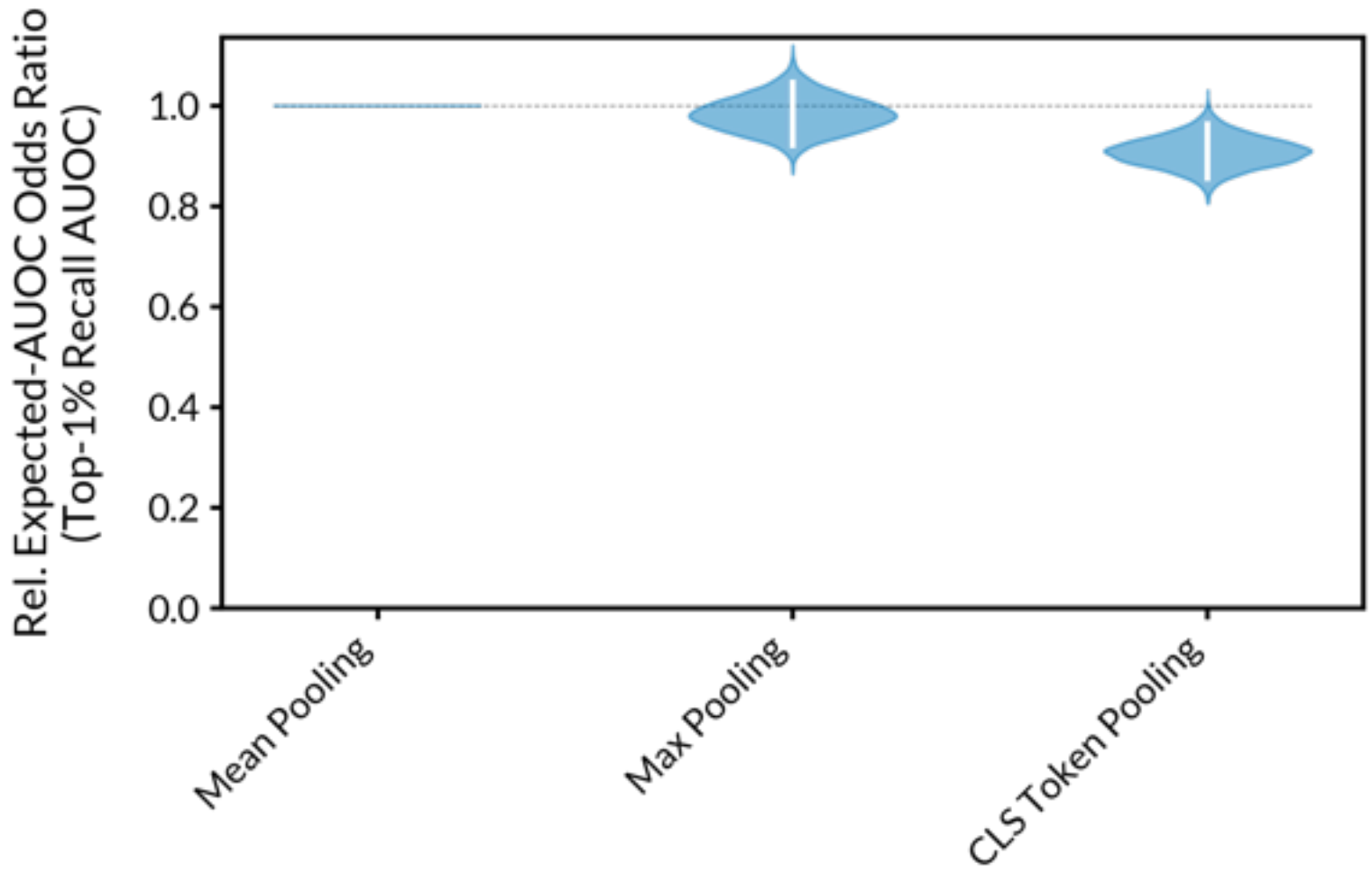

**Figure S 73**: Effect ratios of the transformer pooling strategy relative to mean pooling, averaged over all libraries. Experiments were performed using a Laplace Neural network model on MolFormer embeddings in experimental discovery setting (1,000 experiments, batch size 50). Bars within the violins indicate the 95% highest-density intervals.

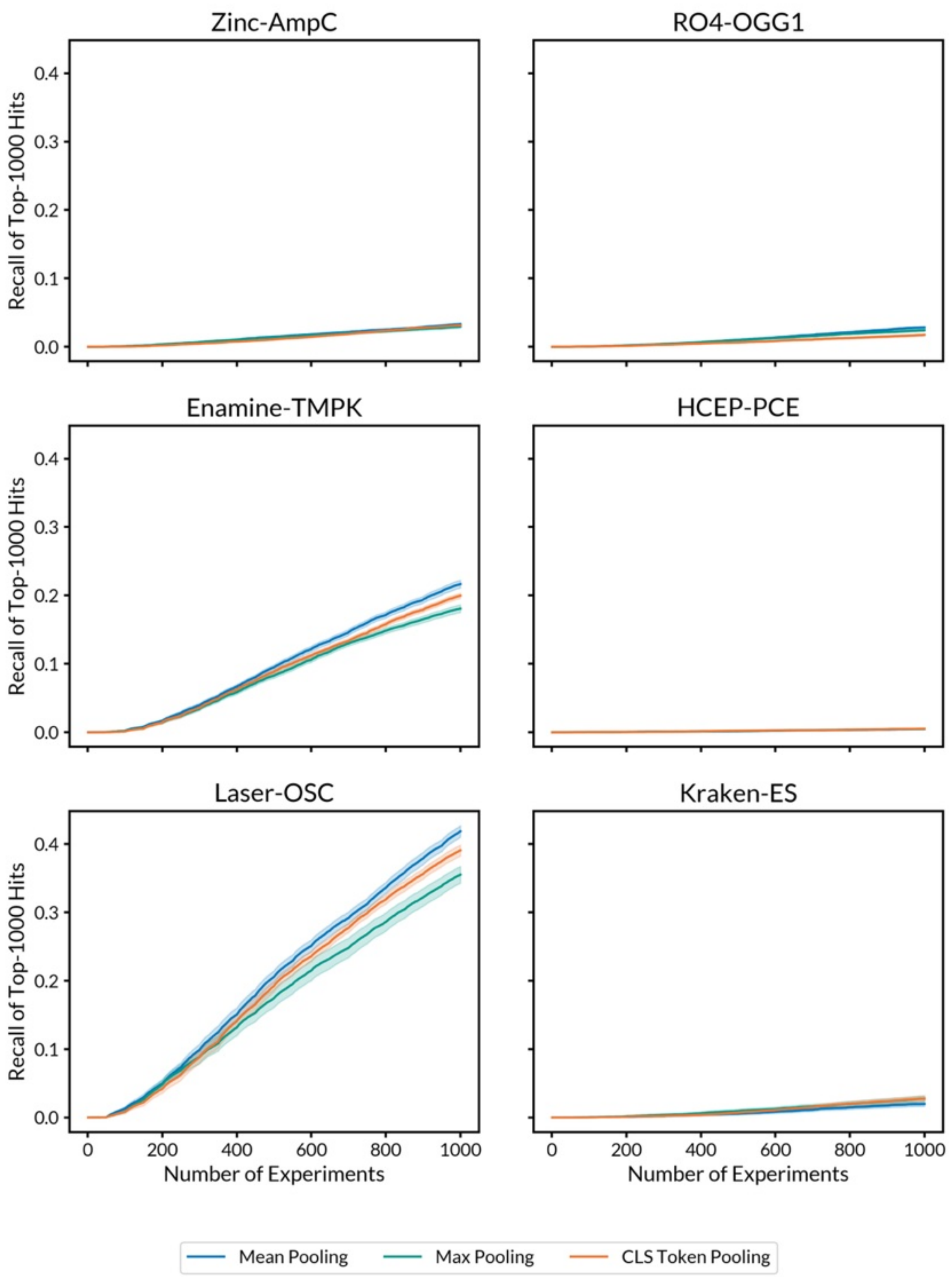

**Figure S 74**: Optimization trajectories for different pooling strategies on a MolFormer encoder using a Laplace Neural Network surrogate model, evaluated over a budget of 1,000 experiments (batch size of 50 experiments). Trajectories are shown as the mean over 20 independent campaigns from different random seed populations. The shaded areas indicate the standard error of the mean.

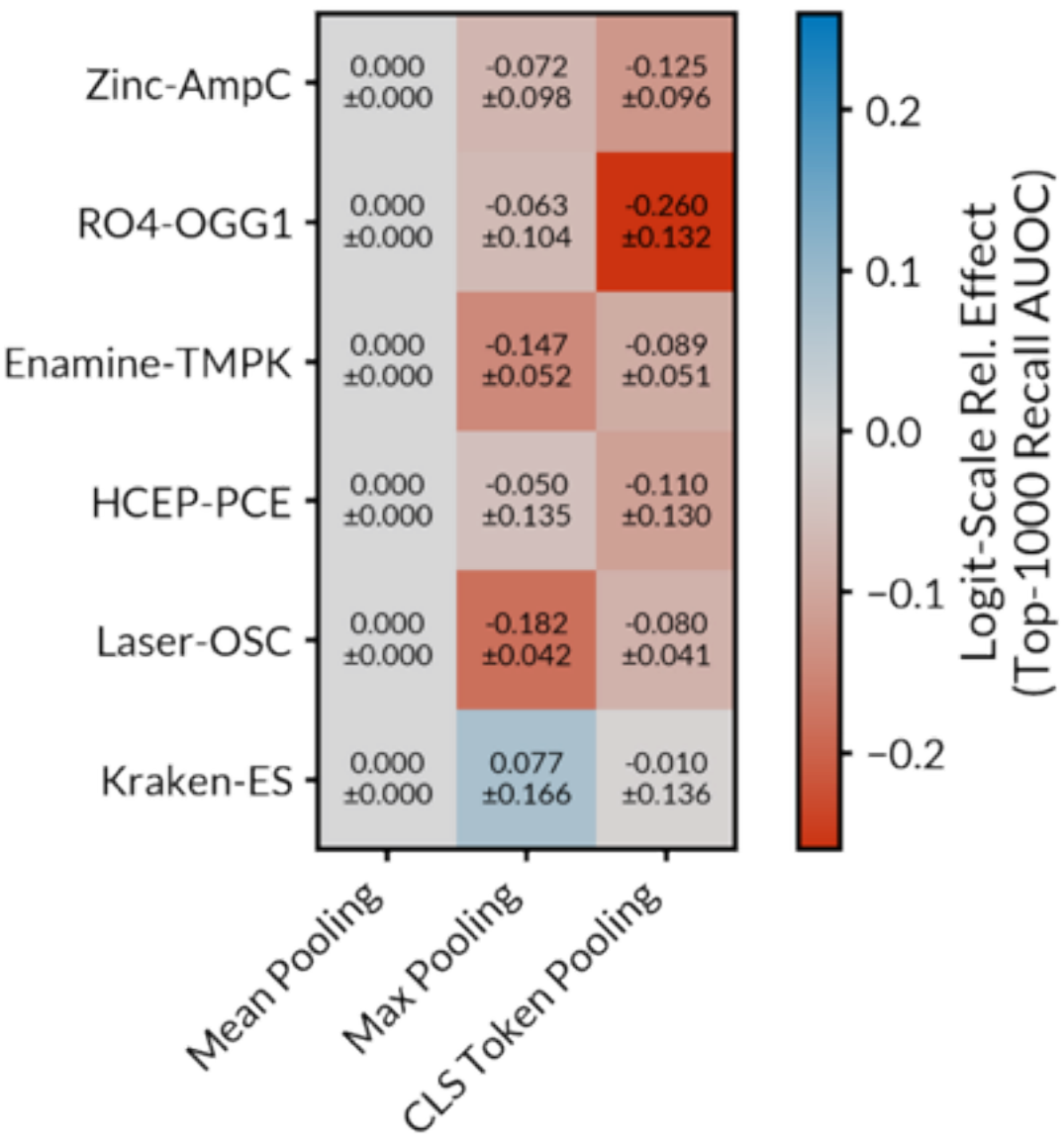

**Figure S 75**: Logit-scale effect of the transformer pooling strategy relative to mean pooling, differentiated by molecular libraries. Experiments were performed using a Laplace Neural network model on MolFormer embeddings in the experimental discovery setting (1,000 experiments, batch size 50). Means and standard deviations were obtained from 8,000 posterior samples.

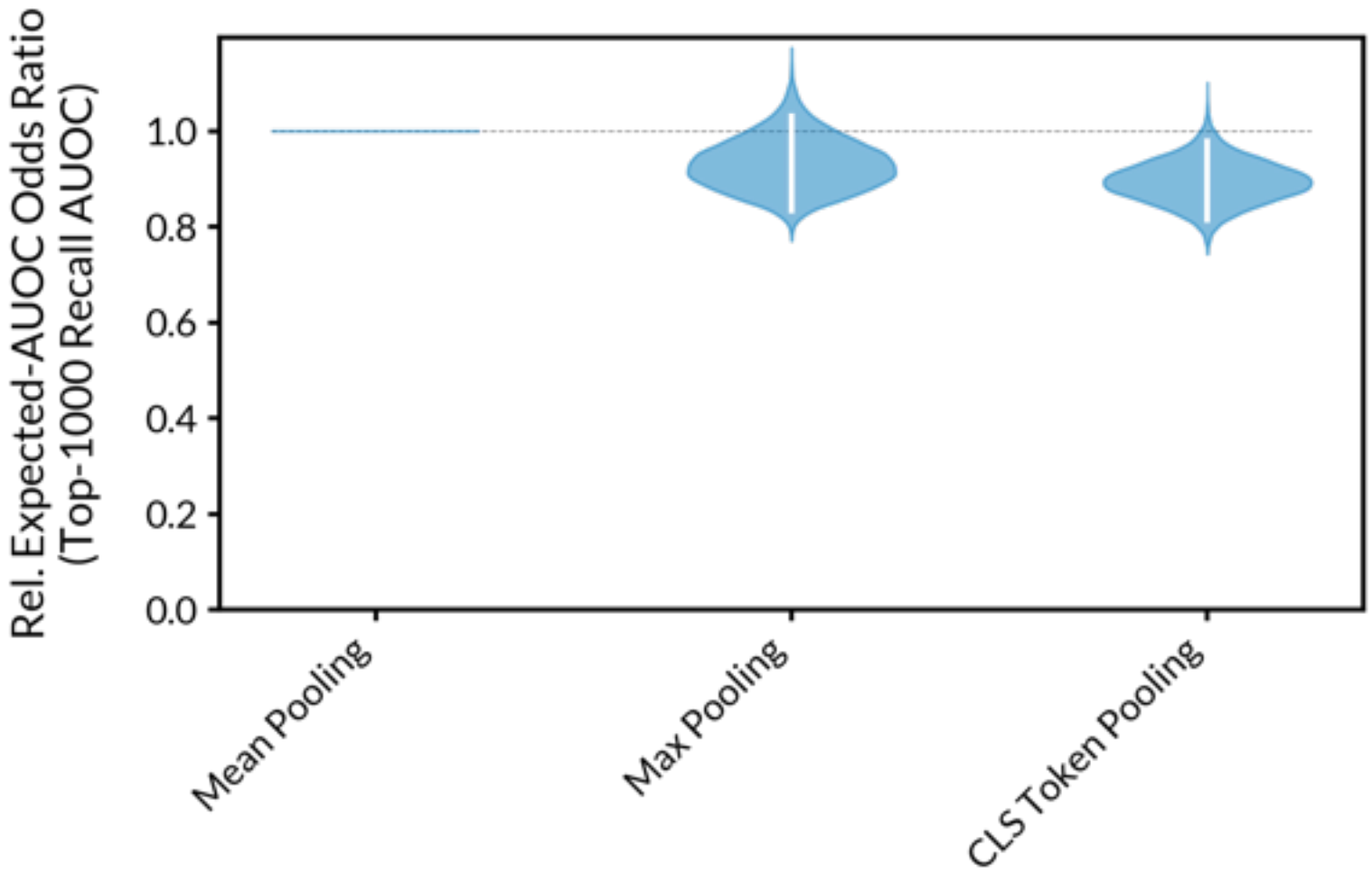

**Figure S 76**: Effect ratios of the transformer pooling strategy relative to mean pooling, averaged over all libraries. Experiments were performed using a Laplace Neural network model on MolFormer embeddings in experimental discovery setting (1,000 experiments, batch size 50). Bars within the violins indicate the 95% highest-density intervals.

## 4.3 Molecular Transformer Fine-Tuning

### *ChemBERTa*

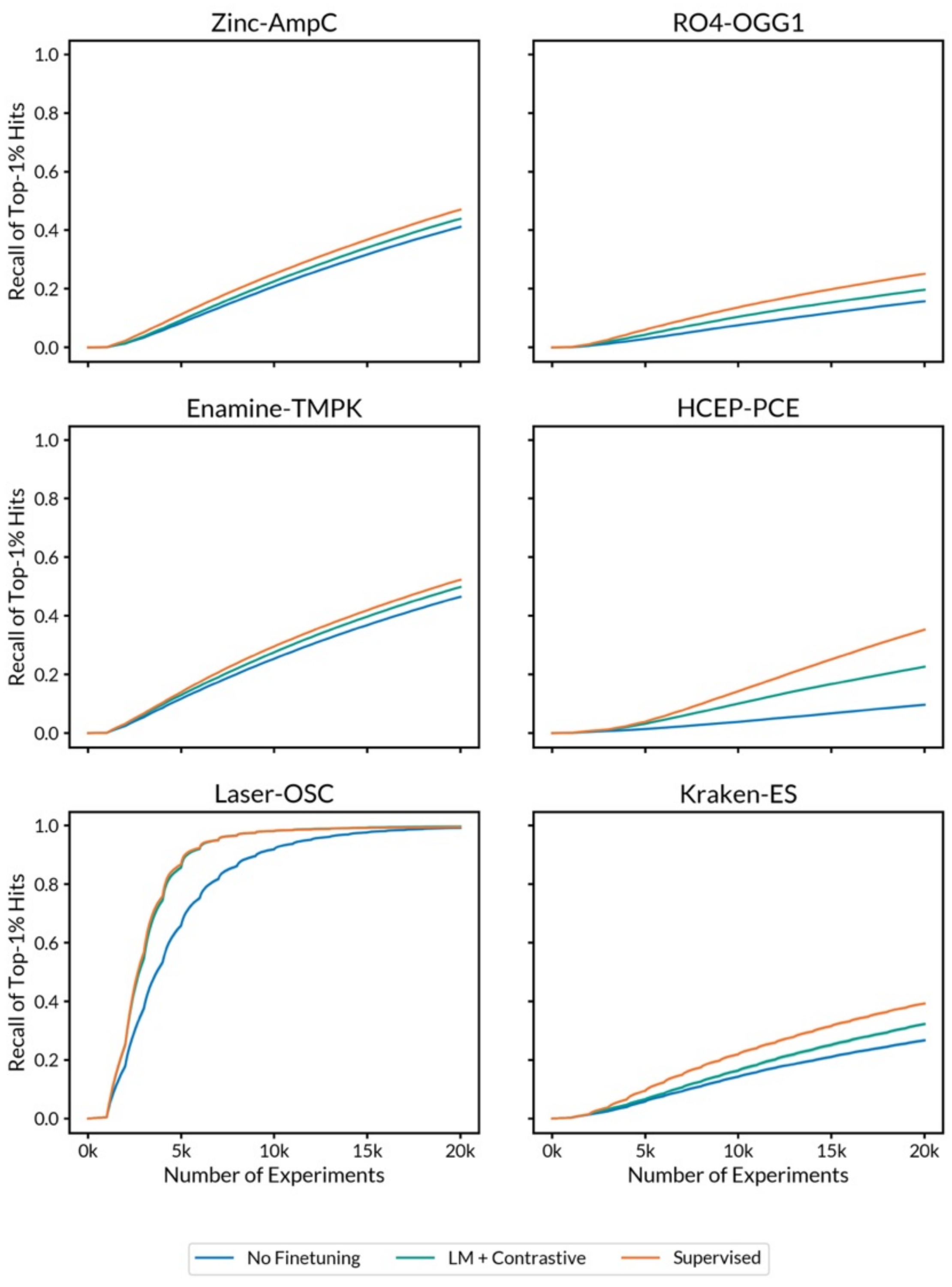

**Figure S 77**: Optimization trajectories after using different strategies to fine-tune ChemBERTa embeddings, evaluated over a budget of 20,000 experiments (batch size of 1,000 experiments). Trajectories are shown as the mean over 20 independent campaigns from different random seed populations. The shaded areas indicate the standard error of the mean.

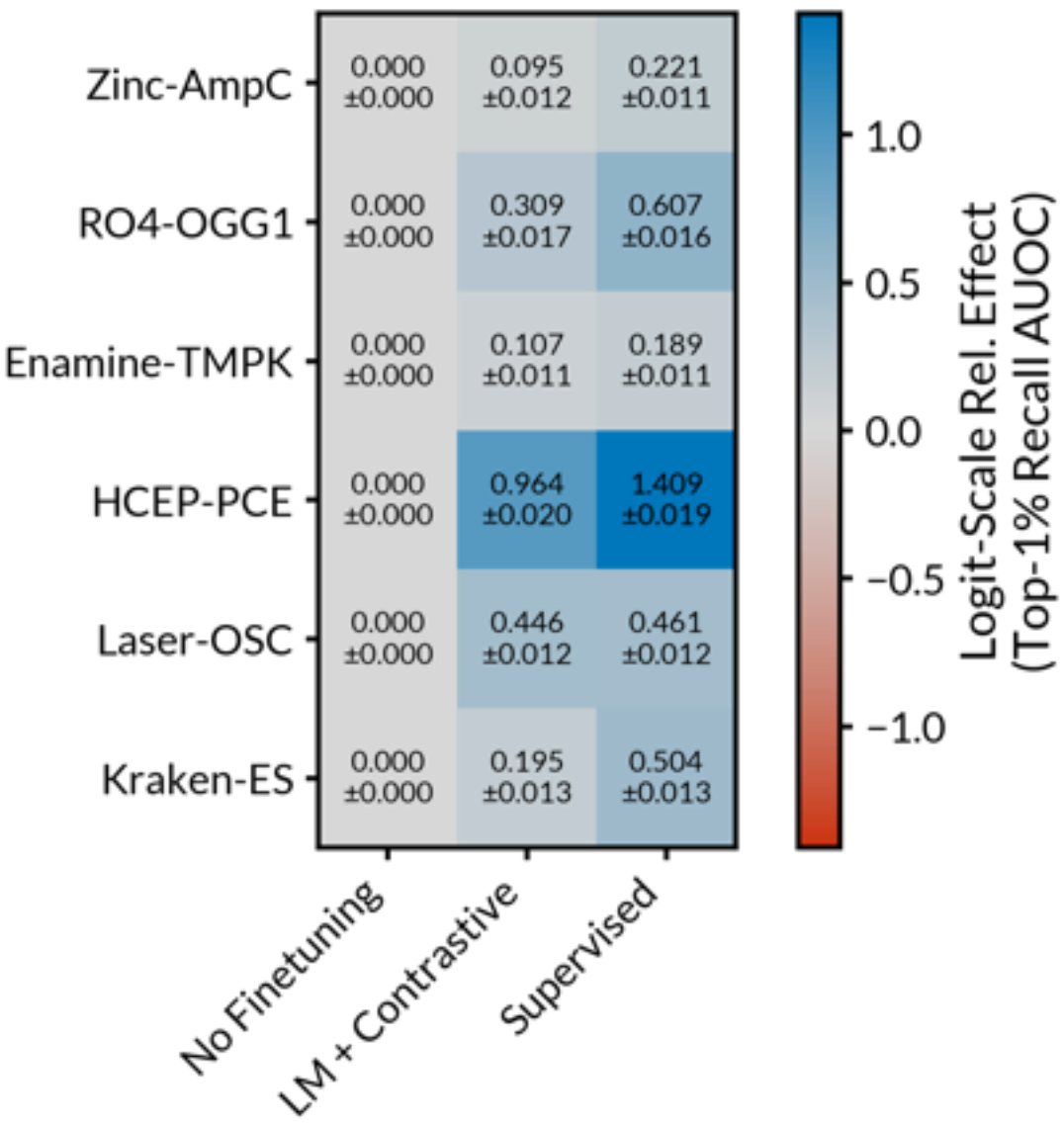

**Figure S 78**: Logit-scale effect of the fine-tuning strategy on ChemBERTa embeddings, differentiated by molecular libraries. Experiments were performed using a Laplace Neural network model in the virtual screening setting (20,000 experiments, batch size 1,000). Means and standard deviations were obtained from 8,000 posterior samples.

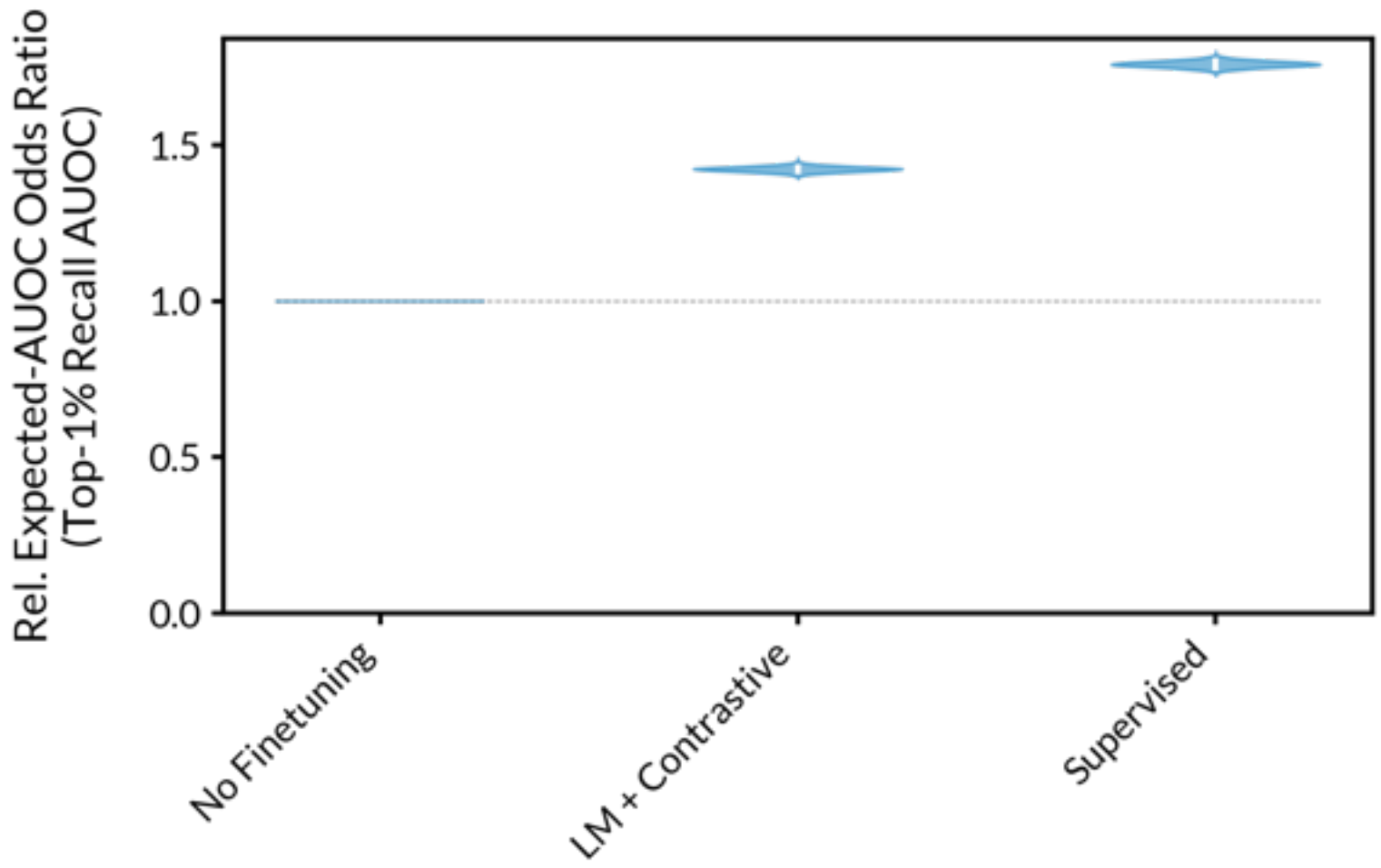

**Figure S 79**: Effect ratios of the fine-tuning strategy on ChemBERTa embeddings, averaged over all libraries. Experiments were performed using a Laplace Neural network model in the virtual screening setting (20,000 experiments, batch size 1,000). Bars within the violins indicate the 95% highest-density intervals.

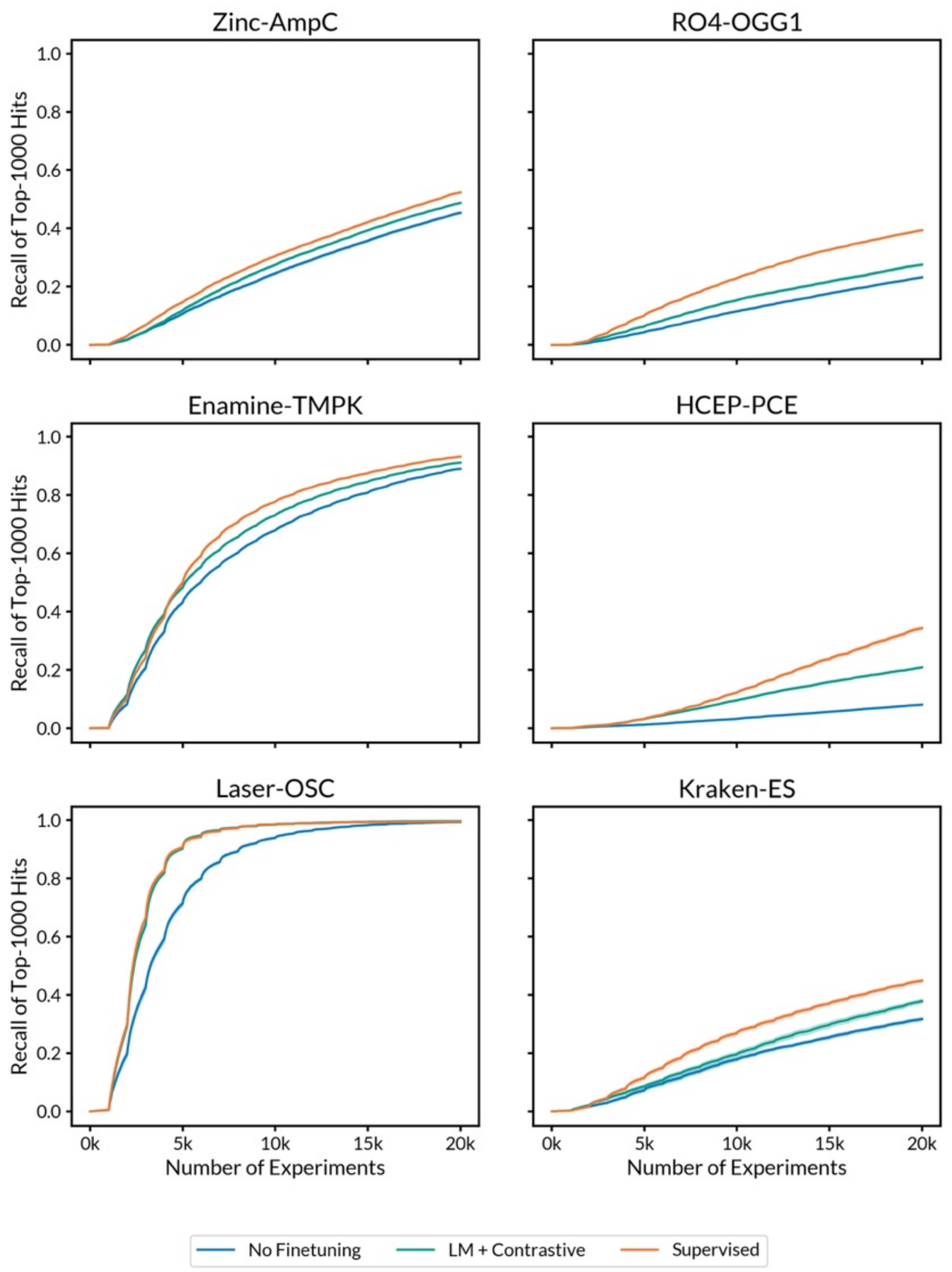

**Figure S 80**: Optimization trajectories after using different strategies to fine-tune ChemBERTa embeddings, evaluated over a budget of 20,000 experiments (batch size of 1,000 experiments). Trajectories are shown as the mean over 20 independent campaigns from different random seed populations. The shaded areas indicate the standard error of the mean.

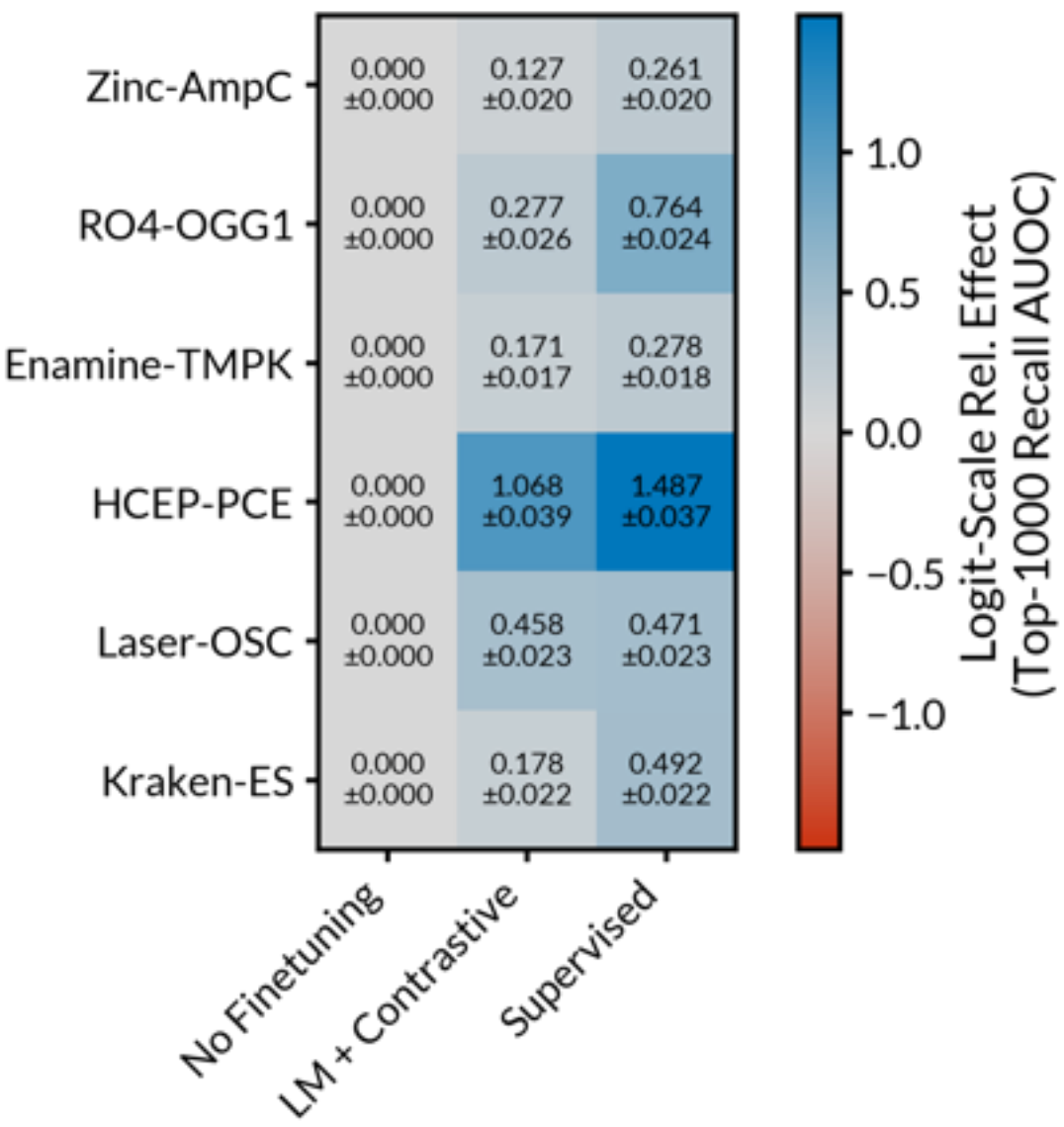

**Figure S 81**: Logit-scale effect of the fine-tuning strategy on ChemBERTa embeddings, differentiated by molecular libraries. Experiments were performed using a Laplace Neural network model in the virtual screening setting (20,000 experiments, batch size 1,000). Means and standard deviations were obtained from 8,000 posterior samples.

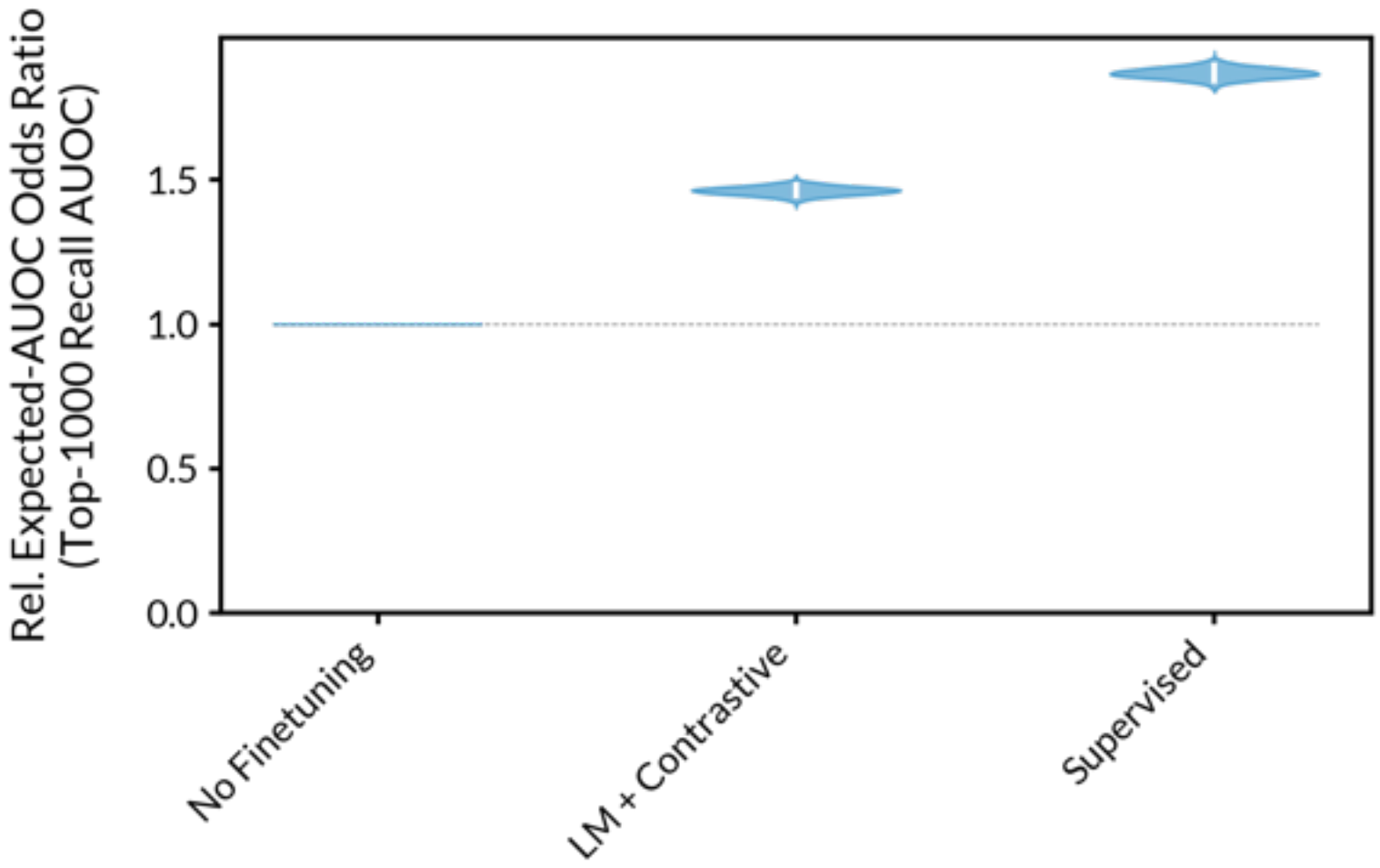

**Figure S 82**: Effect ratios of the fine-tuning strategy on ChemBERTa embeddings, averaged over all libraries. Experiments were performed using a Laplace Neural network model in the virtual screening setting (20,000 experiments, batch size 1,000). Bars within the violins indicate the 95% highest-density intervals.

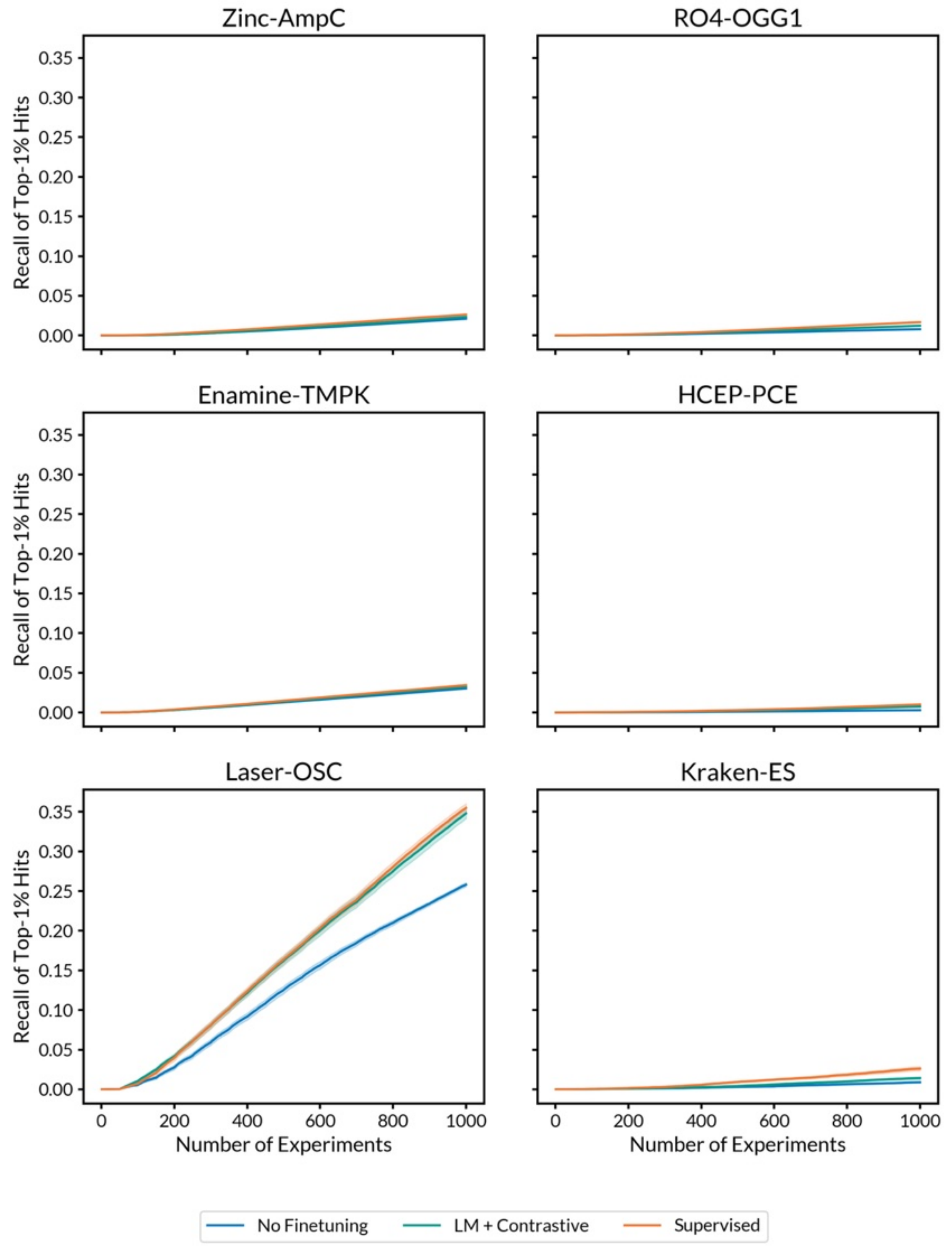

**Figure S 83**: Optimization trajectories after using different strategies to fine-tune ChemBERTa embeddings, evaluated over a budget of 1,000 experiments (batch size of 50 experiments). Trajectories are shown as the mean over 20 independent campaigns from different random seed populations. The shaded areas indicate the standard error of the mean.

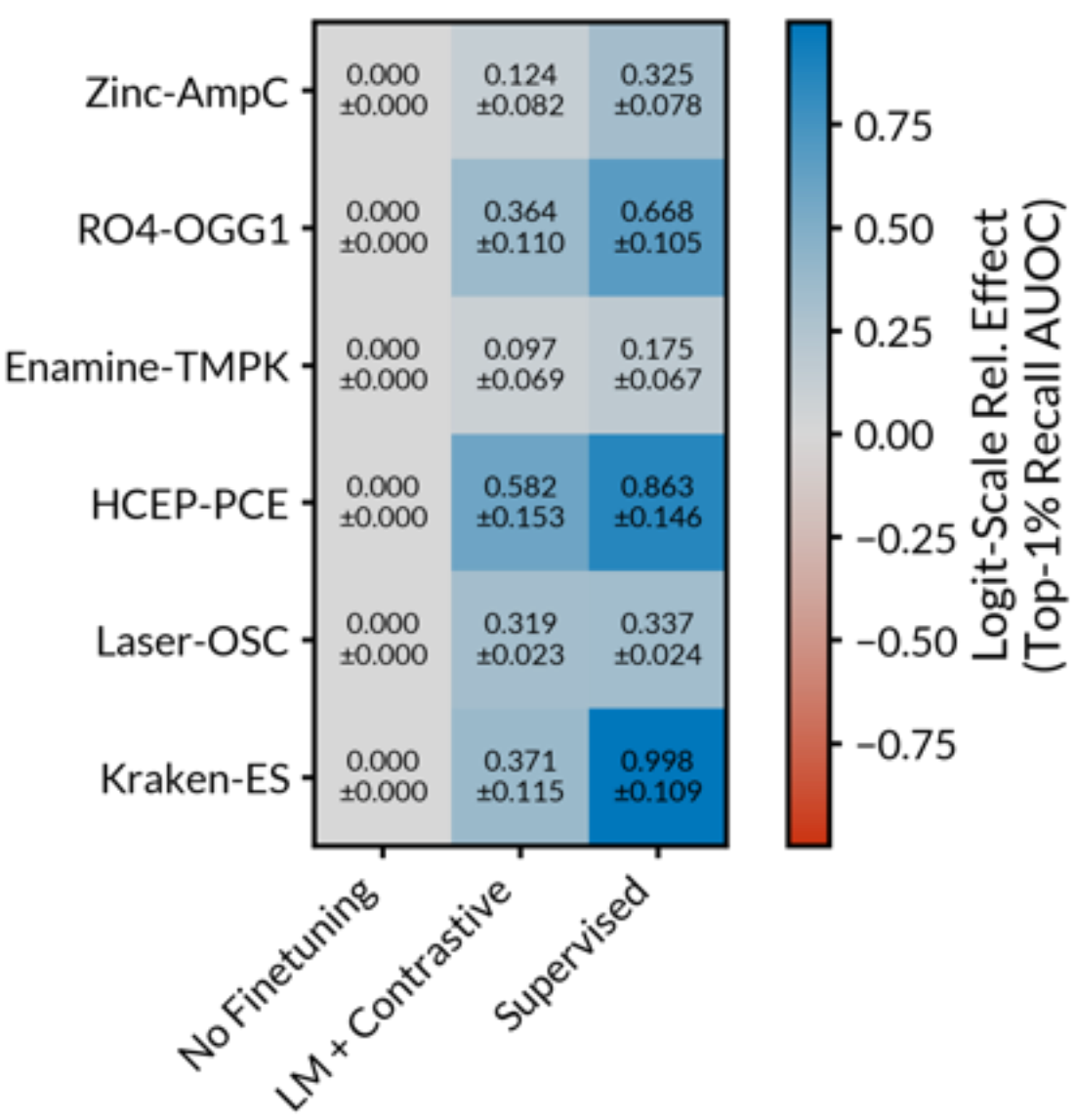

**Figure S 84**: Logit-scale effect of the fine-tuning strategy on ChemBERTa embeddings, differentiated by molecular libraries. Experiments were performed using a Laplace Neural network model in the experimental discovery setting (1,000 experiments, batch size 50). Means and standard deviations were obtained from 8,000 posterior samples.

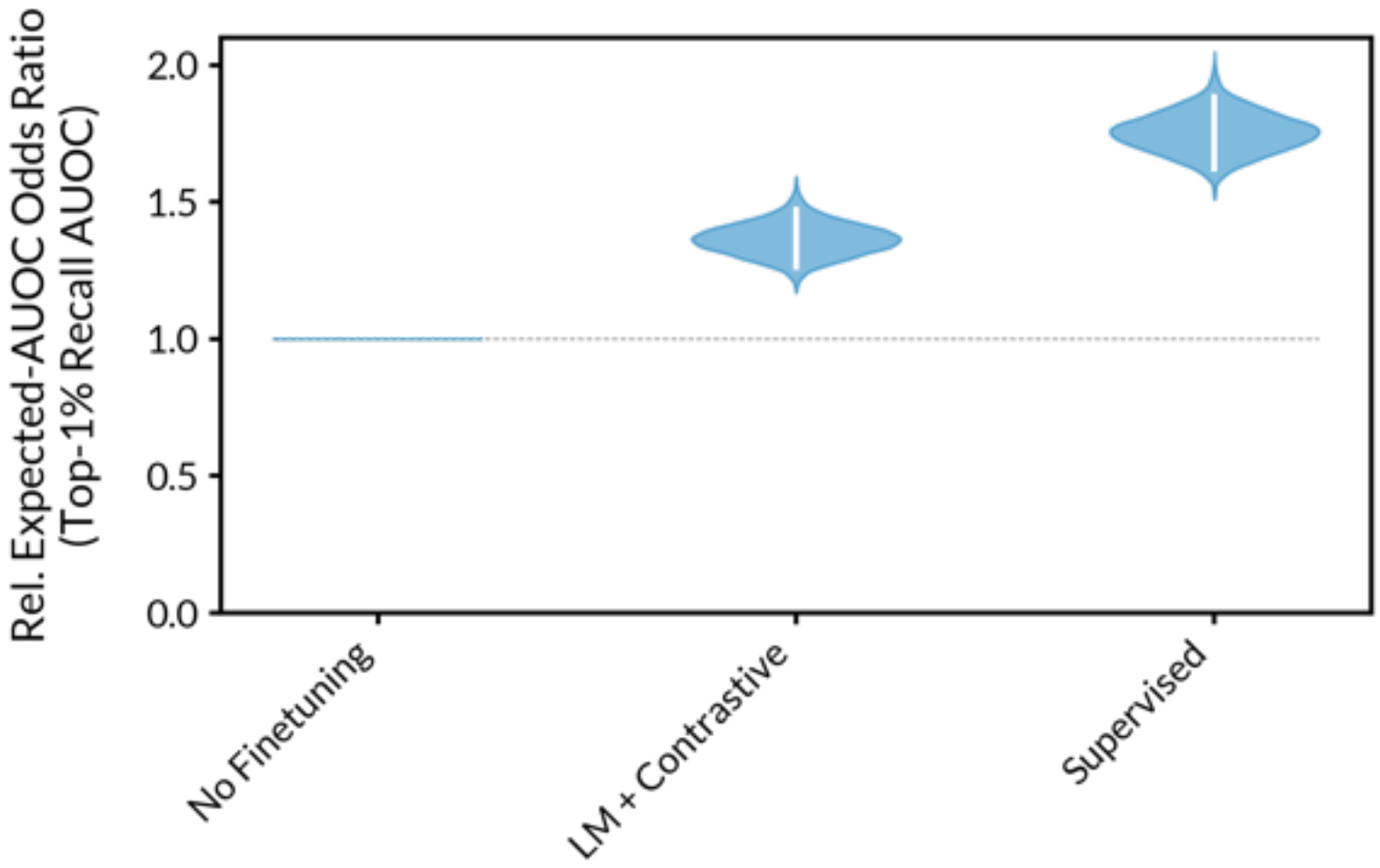

**Figure S 85**: Effect ratios of the fine-tuning strategy on ChemBERTa embeddings, averaged over all libraries. Experiments were performed using a Laplace Neural network model in the experimental discovery setting (1,000 experiments, batch size 50). Bars within the violins indicate the 95% highest-density intervals.

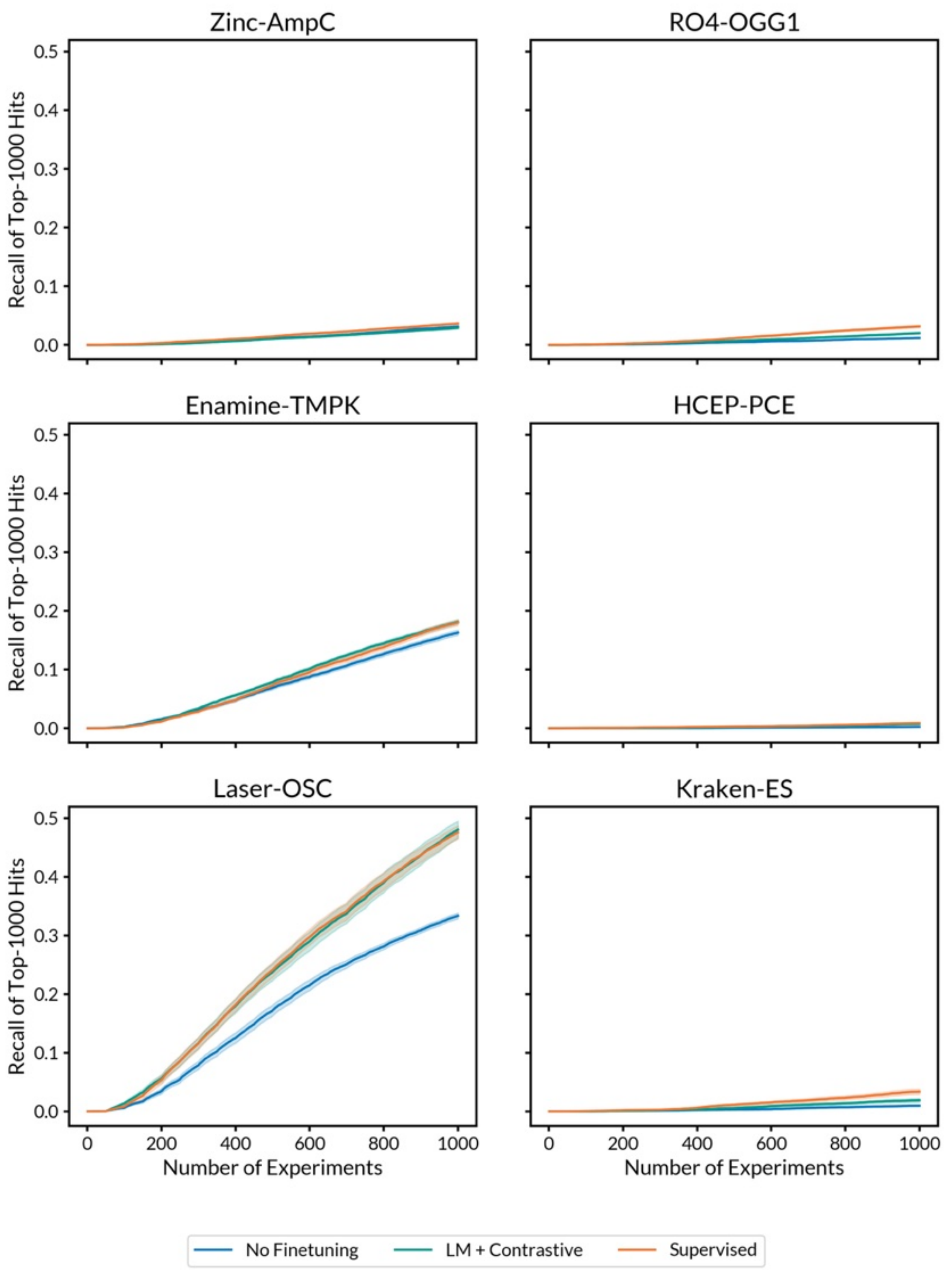

**Figure S 86**: Optimization trajectories after using different strategies to fine-tune ChemBERTa embeddings, evaluated over a budget of 1,000 experiments (batch size of 50 experiments). Trajectories are shown as the mean over 20 independent campaigns from different random seed populations. The shaded areas indicate the standard error of the mean.

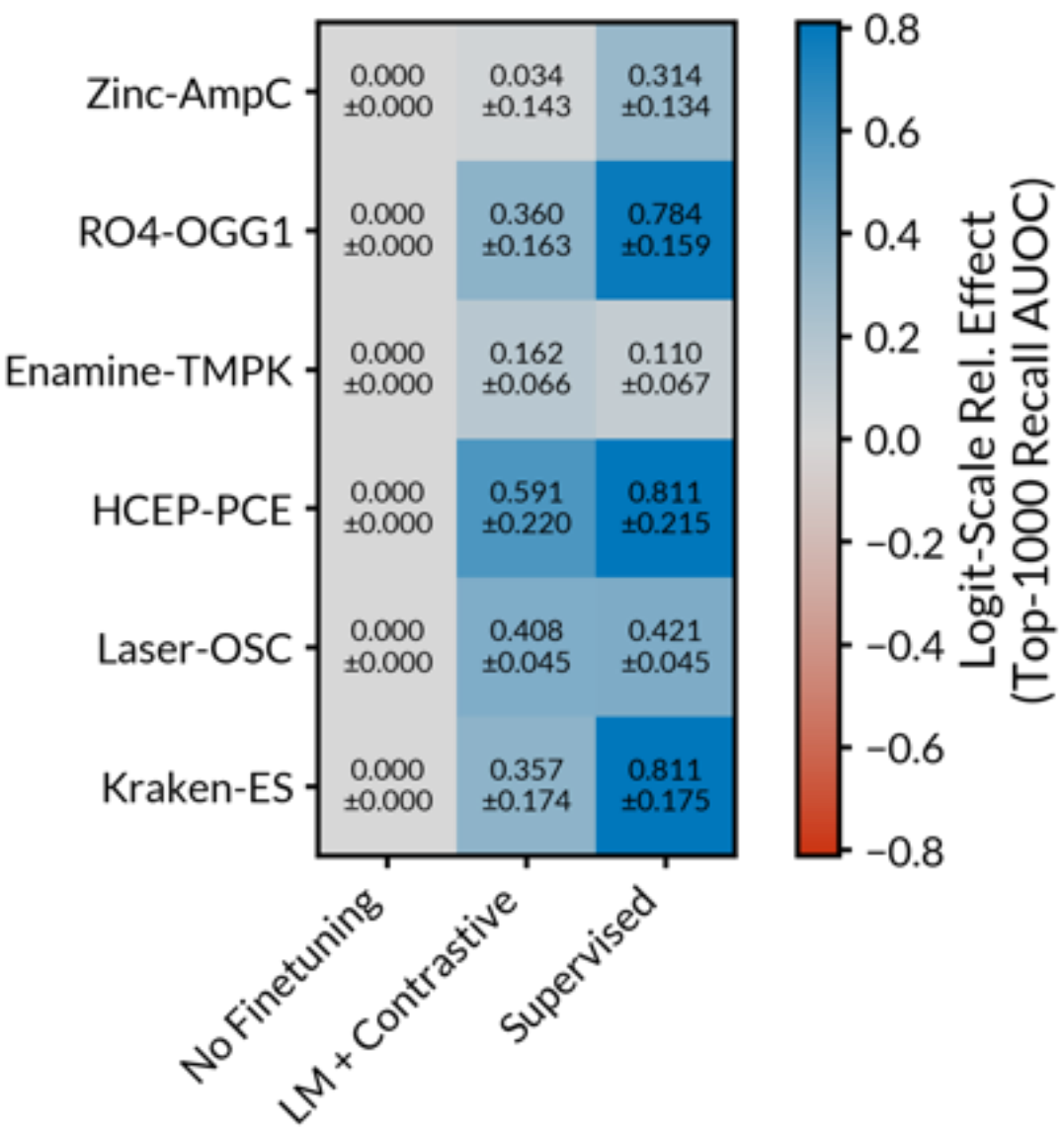

**Figure S 87**: Logit-scale effect of the fine-tuning strategy on ChemBERTa embeddings, differentiated by molecular libraries. Experiments were performed using a Laplace Neural network model in the experimental discovery setting (1,000 experiments, batch size 50). Means and standard deviations were obtained from 8,000 posterior samples.

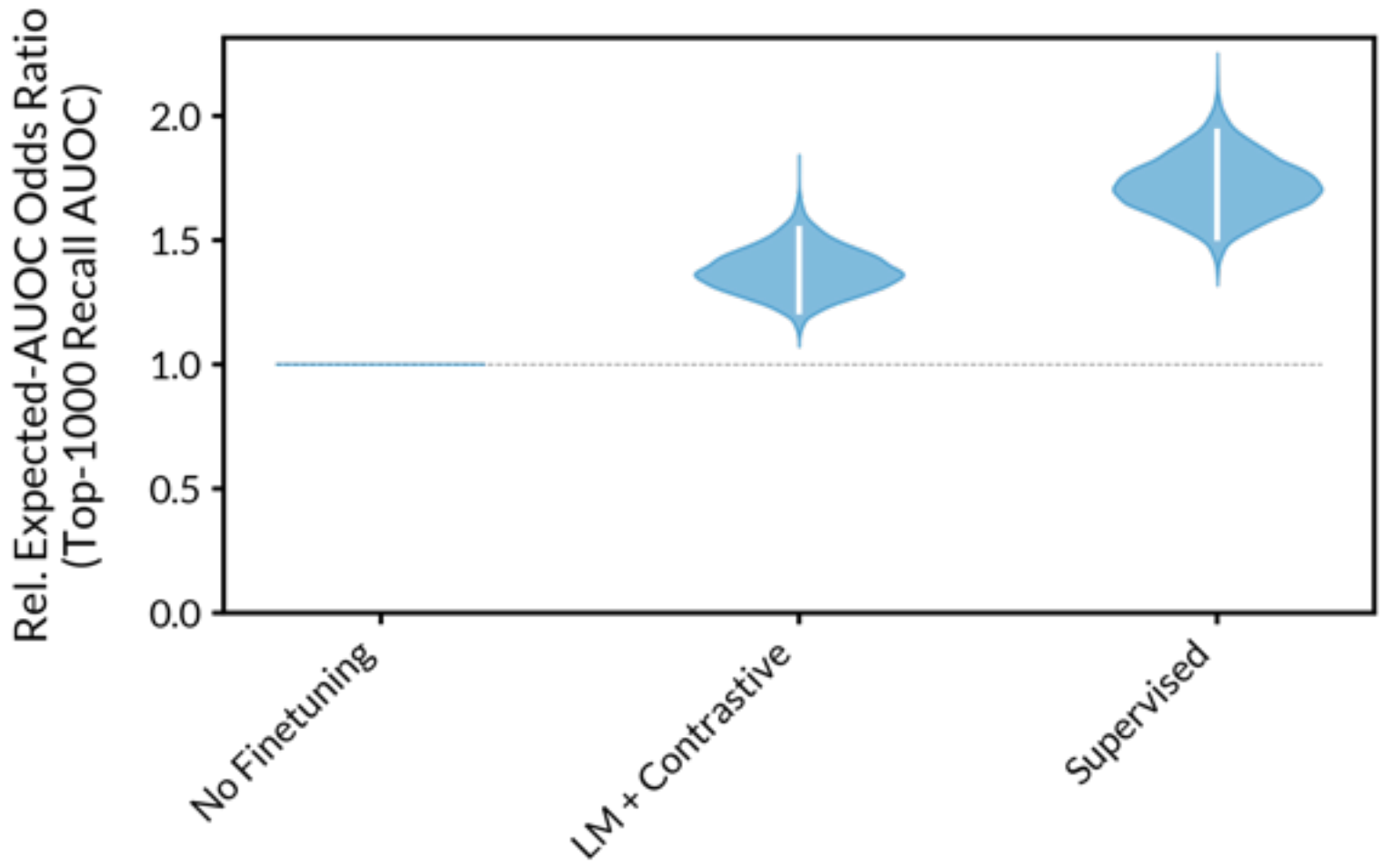

**Figure S 88**: Effect ratios of the fine-tuning strategy on ChemBERTa embeddings, averaged over all libraries. Experiments were performed using a Laplace Neural network model in the experimental discovery setting (1,000 experiments, batch size 50). Bars within the violins indicate the 95% highest-density intervals.

***MolFormer-XL***

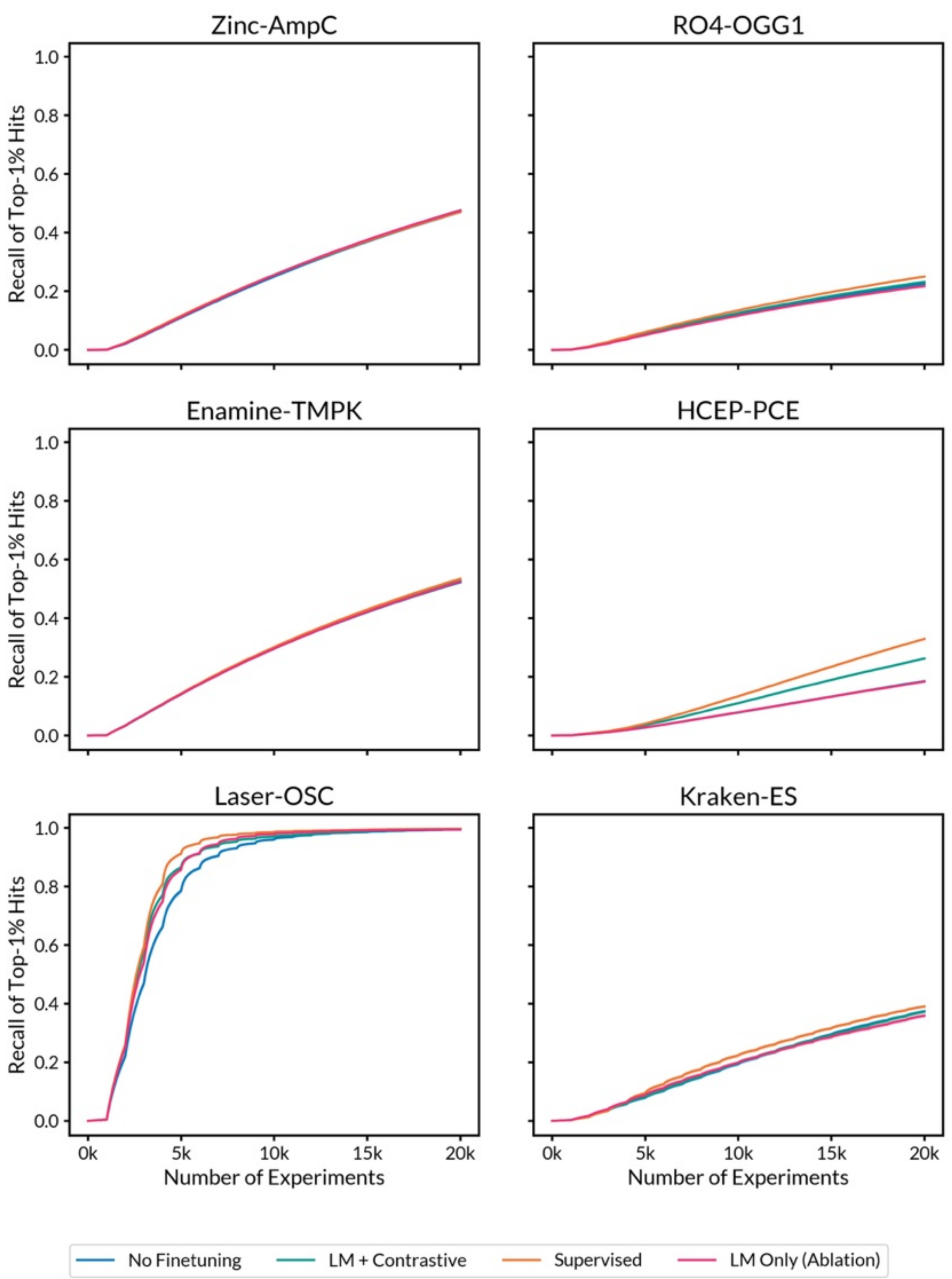

**Figure S 89**: Optimization trajectories after using different strategies to fine-tune MolFormer-XL embeddings, evaluated over a budget of 20,000 experiments (batch size of 1,000 experiments). Trajectories are shown as the mean over 20 independent campaigns from different random seed populations. The shaded areas indicate the standard error of the mean.

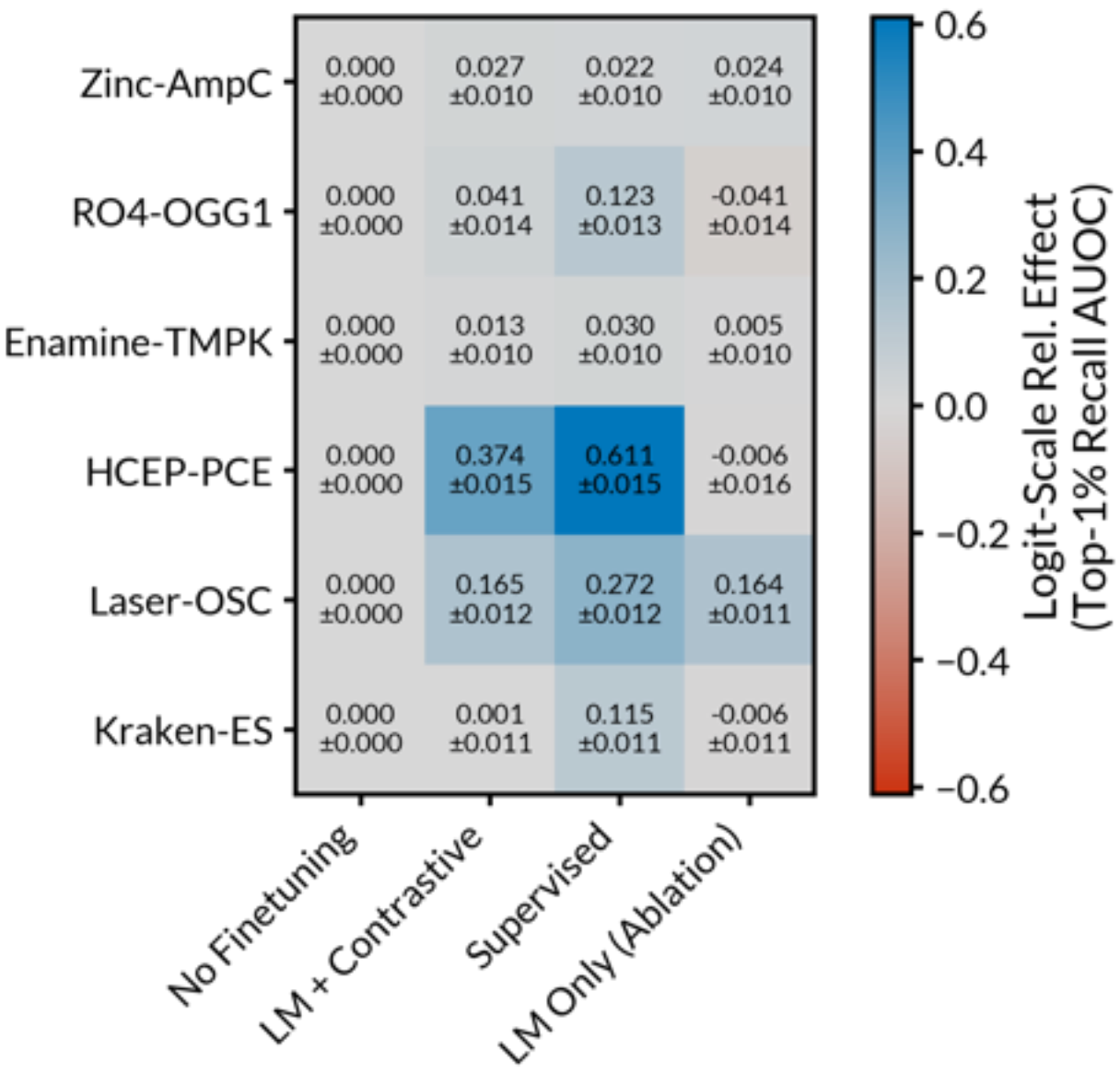

**Figure S 90**: Logit-scale effect of the fine-tuning strategy on MolFormer-XL embeddings, differentiated by molecular libraries. Experiments were performed using a Laplace Neural network model in the virtual screening setting (20,000 experiments, batch size 1,000). Means and standard deviations were obtained from 8,000 posterior samples.

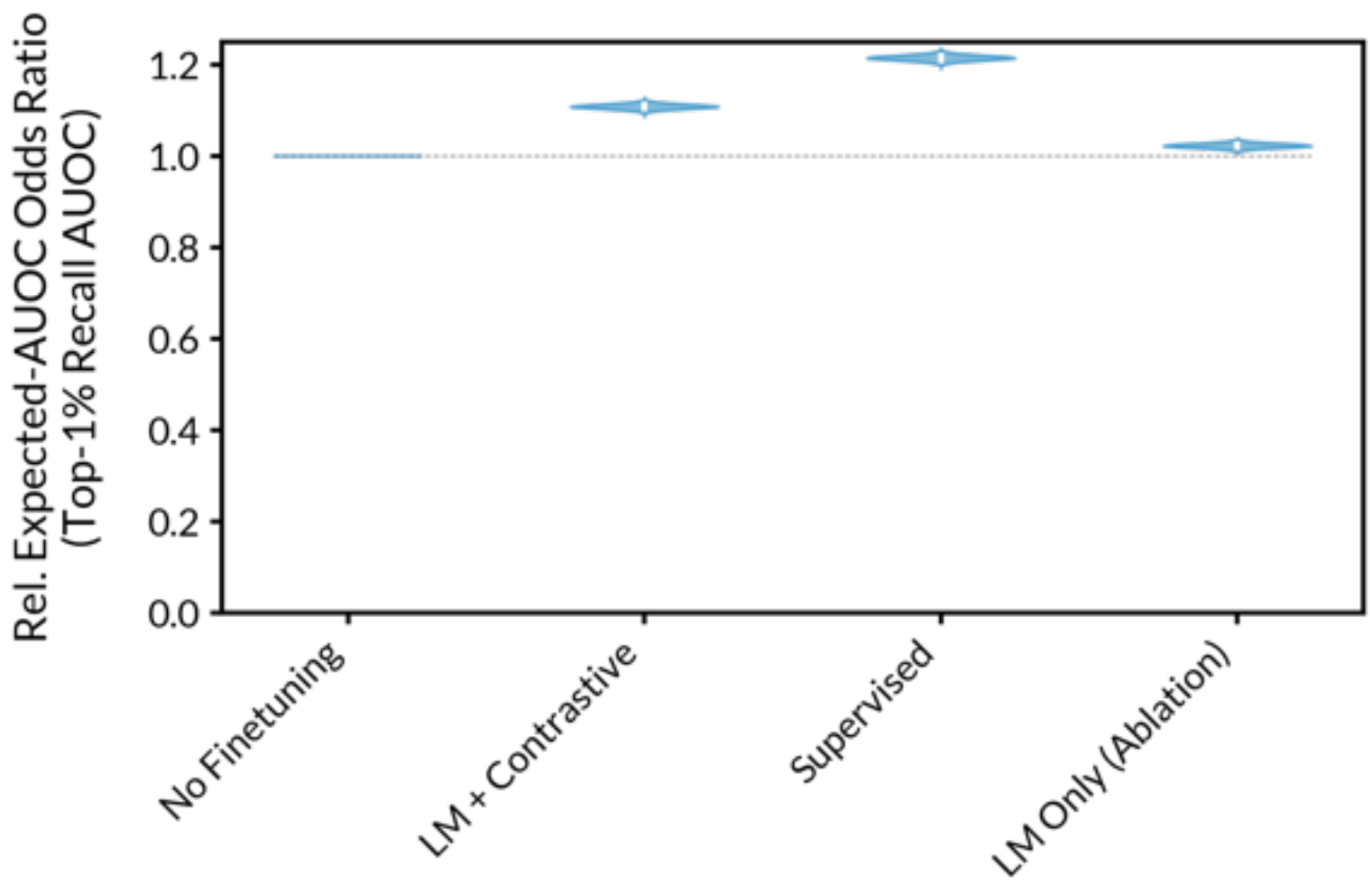

**Figure S 91**: Effect ratios of the fine-tuning strategy on MolFormer-XL embeddings, averaged over all libraries. Experiments were performed using a Laplace Neural network model in the virtual screening setting (20,000 experiments, batch size 1,000). Bars within the violins indicate the 95% highest-density intervals.

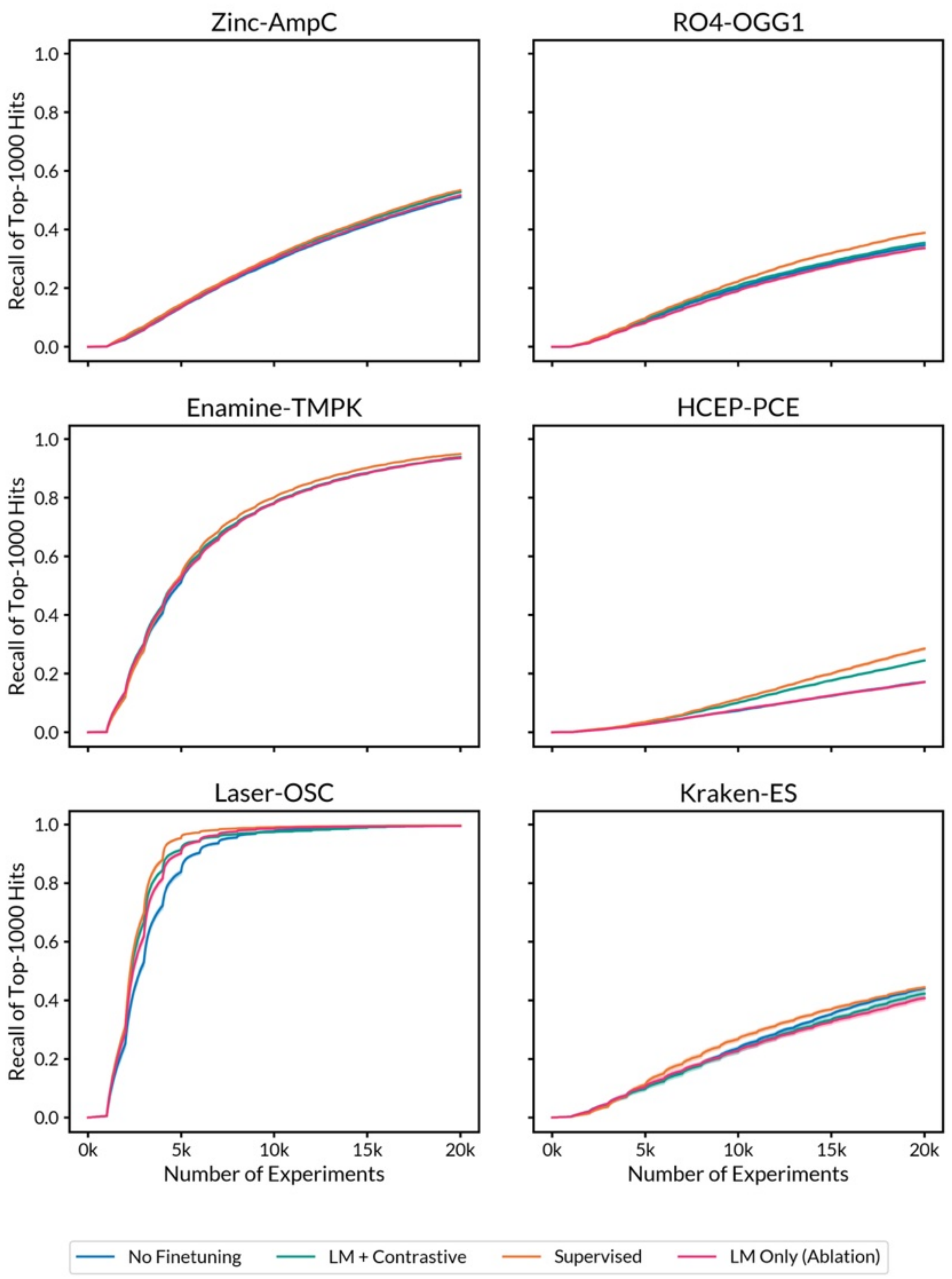

**Figure S 92**: Optimization trajectories after using different strategies to fine-tune MolFormer-XL embeddings, evaluated over a budget of 20,000 experiments (batch size of 1,000 experiments). Trajectories are shown as the mean over 20 independent campaigns from different random seed populations. The shaded areas indicate the standard error of the mean.

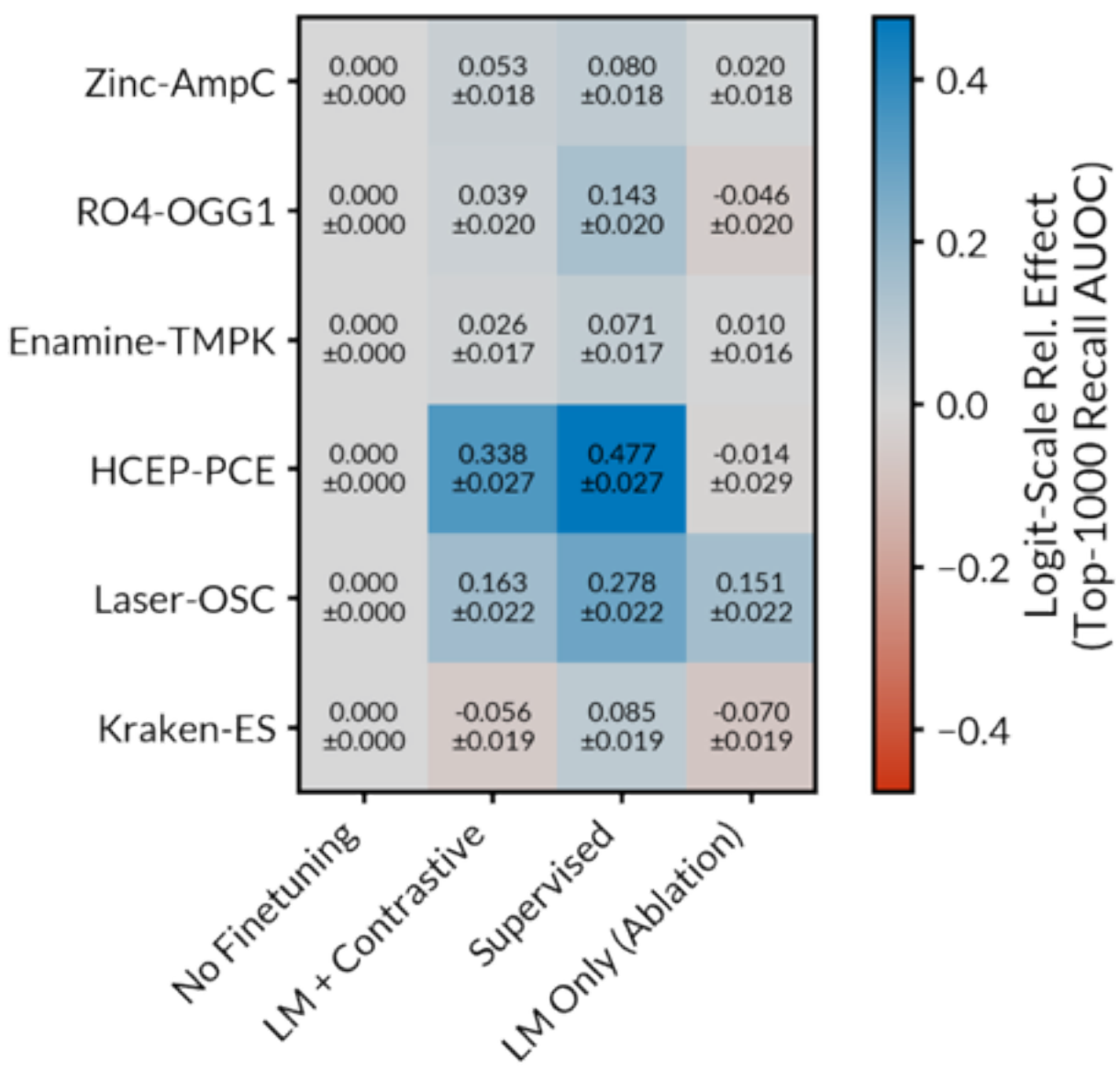

**Figure S 93**: Logit-scale effect of the fine-tuning strategy on MolFormer-XL embeddings, differentiated by molecular libraries. Experiments were performed using a Laplace Neural network model in the virtual screening setting (20,000 experiments, batch size 1,000). Means and standard deviations were obtained from 8,000 posterior samples.

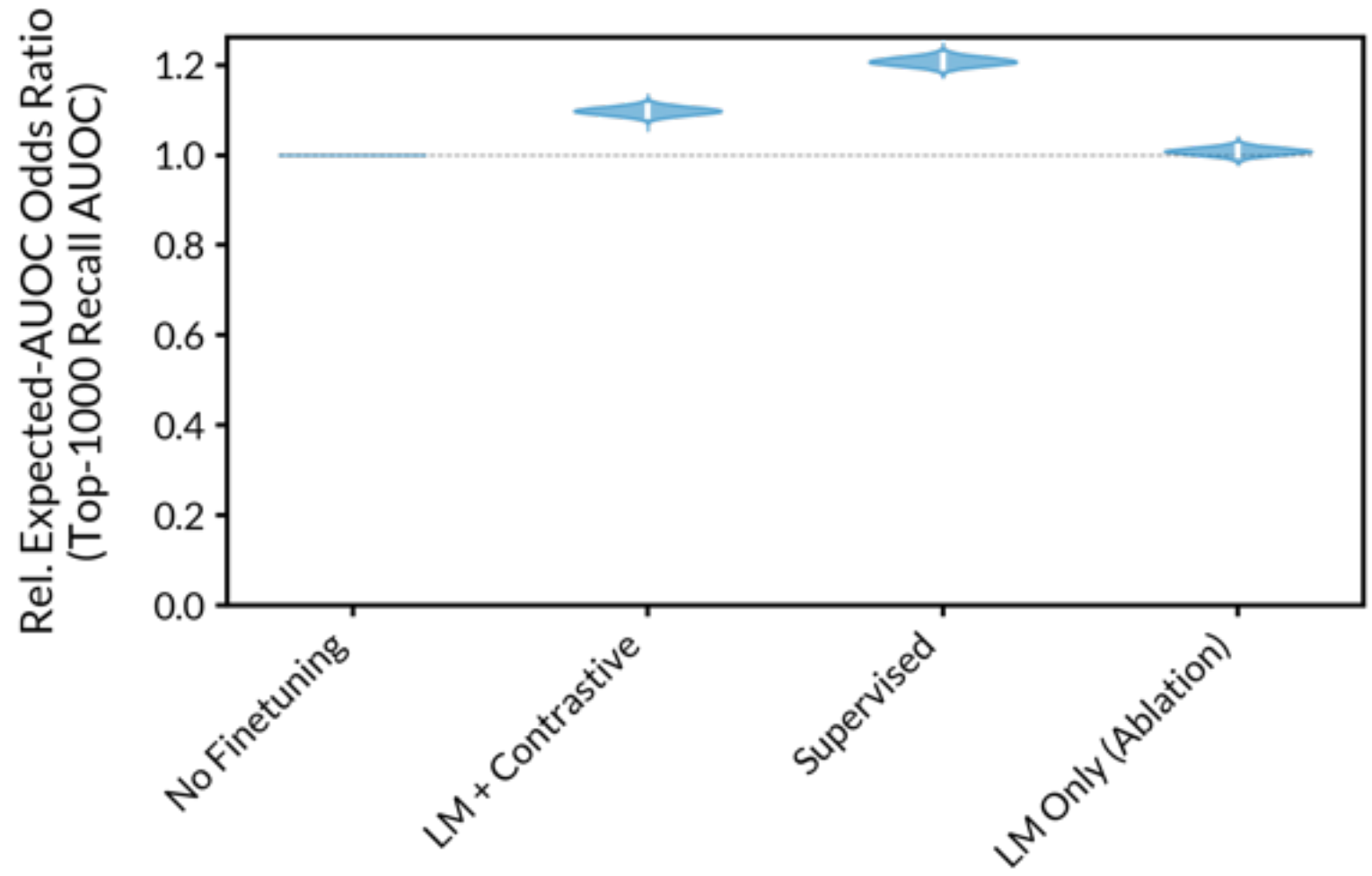

**Figure S 94**: Effect ratios of the fine-tuning strategy on MolFormer-XL embeddings, averaged over all libraries. Experiments were performed using a Laplace Neural network model in the virtual screening setting (20,000 experiments, batch size 1,000). Bars within the violins indicate the 95% highest-density intervals.

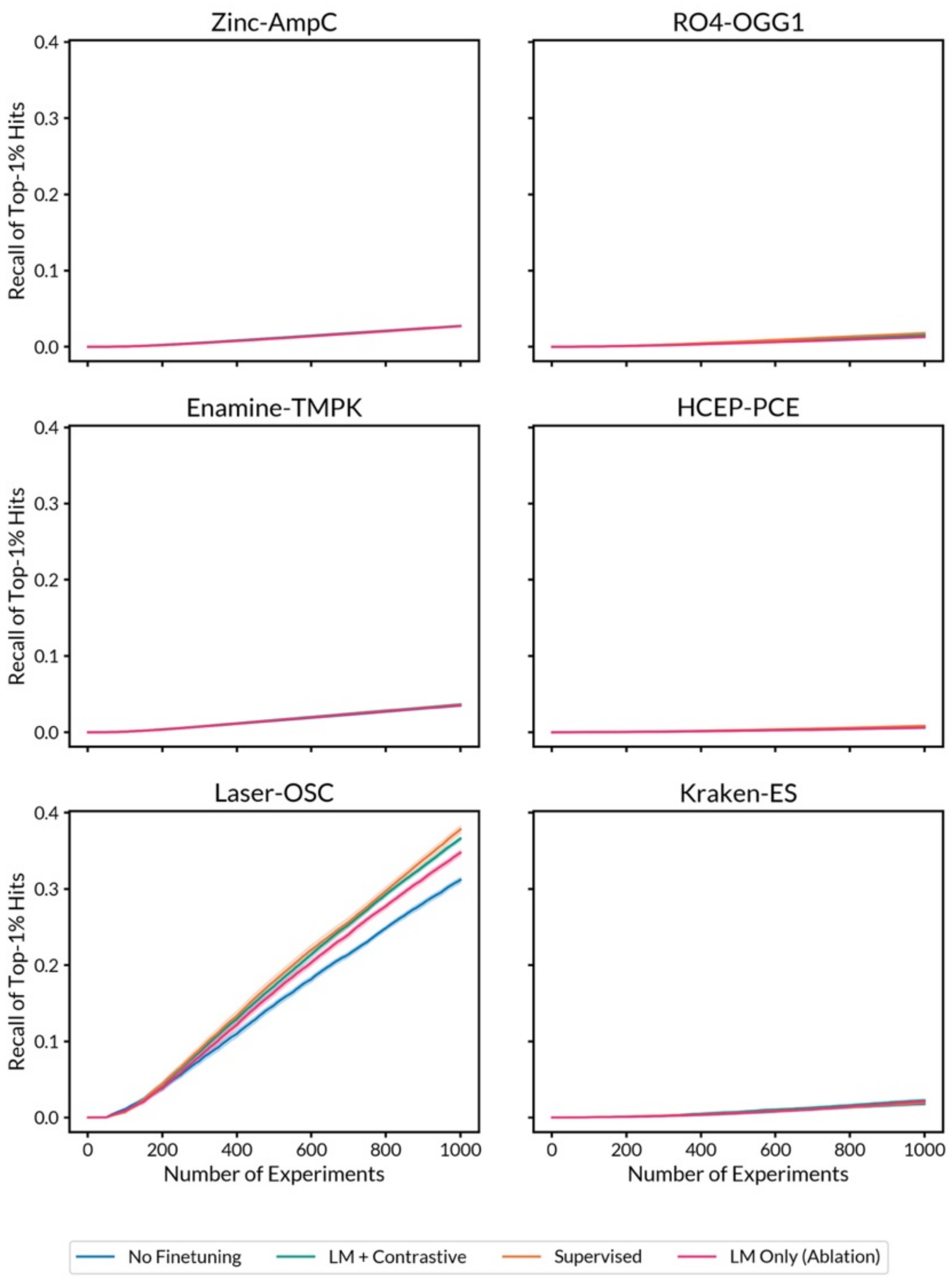

**Figure S 95**: Optimization trajectories after using different strategies to fine-tune MolFormer-XL embeddings, evaluated over a budget of 1,000 experiments (batch size of 50 experiments). Trajectories are shown as the mean over 20 independent campaigns from different random seed populations. The shaded areas indicate the standard error of the mean.

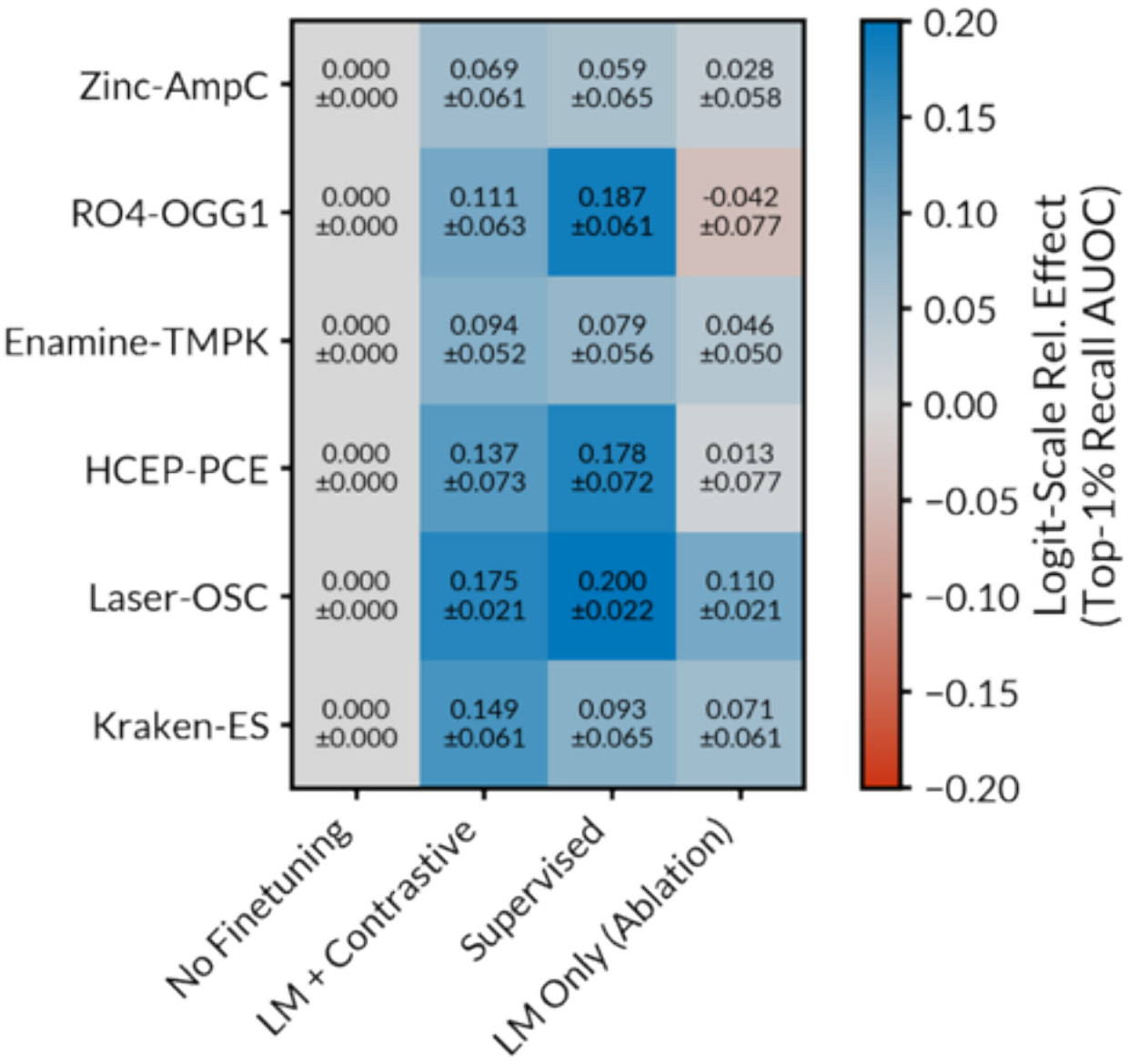

**Figure S 96**: Logit-scale effect of the fine-tuning strategy on MolFormer-XL embeddings, differentiated by molecular libraries. Experiments were performed using a Laplace Neural network model in the experimental discovery setting (1,000 experiments, batch size 50). Means and standard deviations were obtained from 8,000 posterior samples.

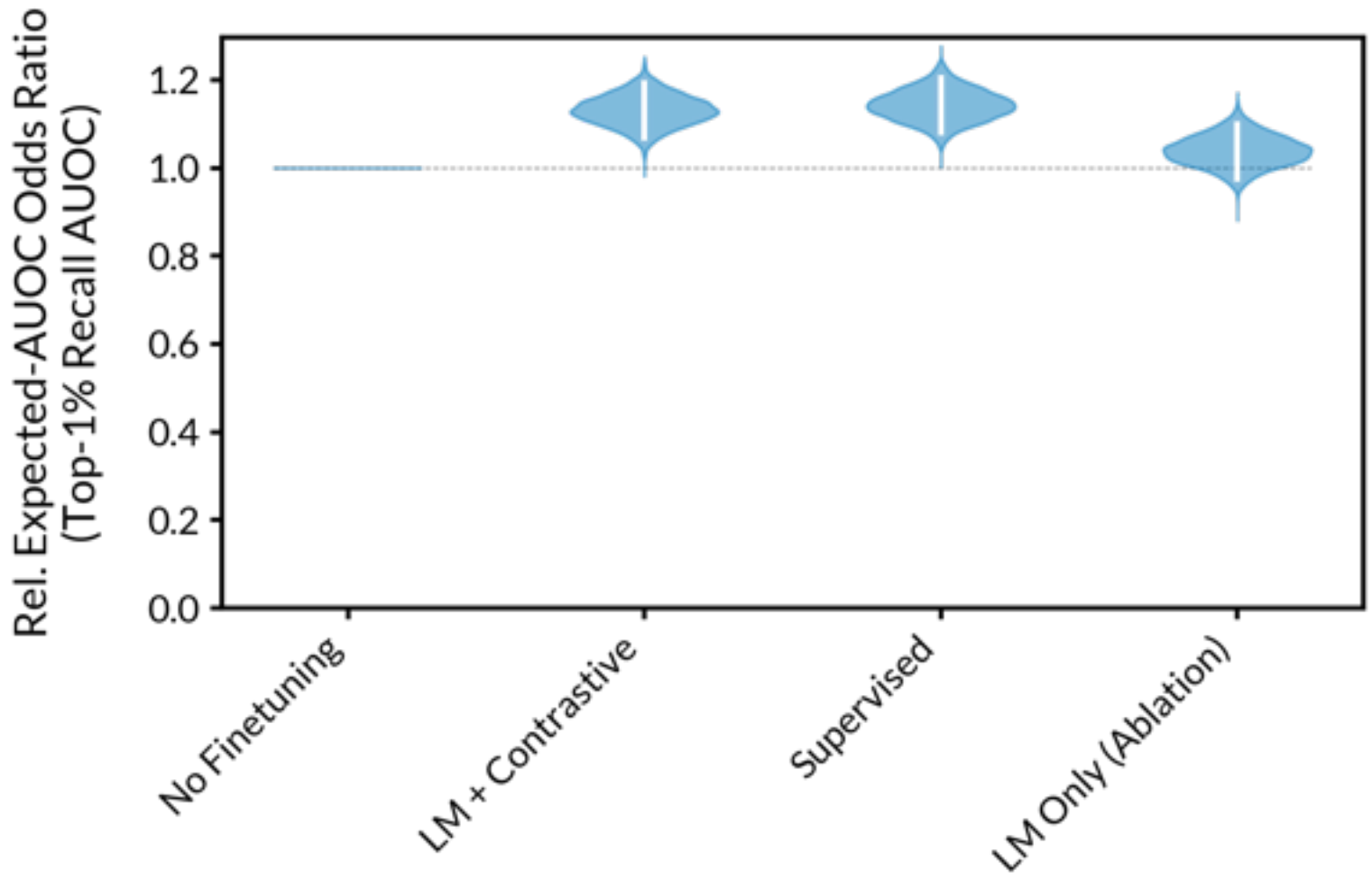

**Figure S 97**: Effect ratios of the fine-tuning strategy on MolFormer-XL embeddings, averaged over all libraries. Experiments were performed using a Laplace Neural network model in the experimental discovery setting (1,000 experiments, batch size 50). Bars within the violins indicate the 95% highest-density intervals.

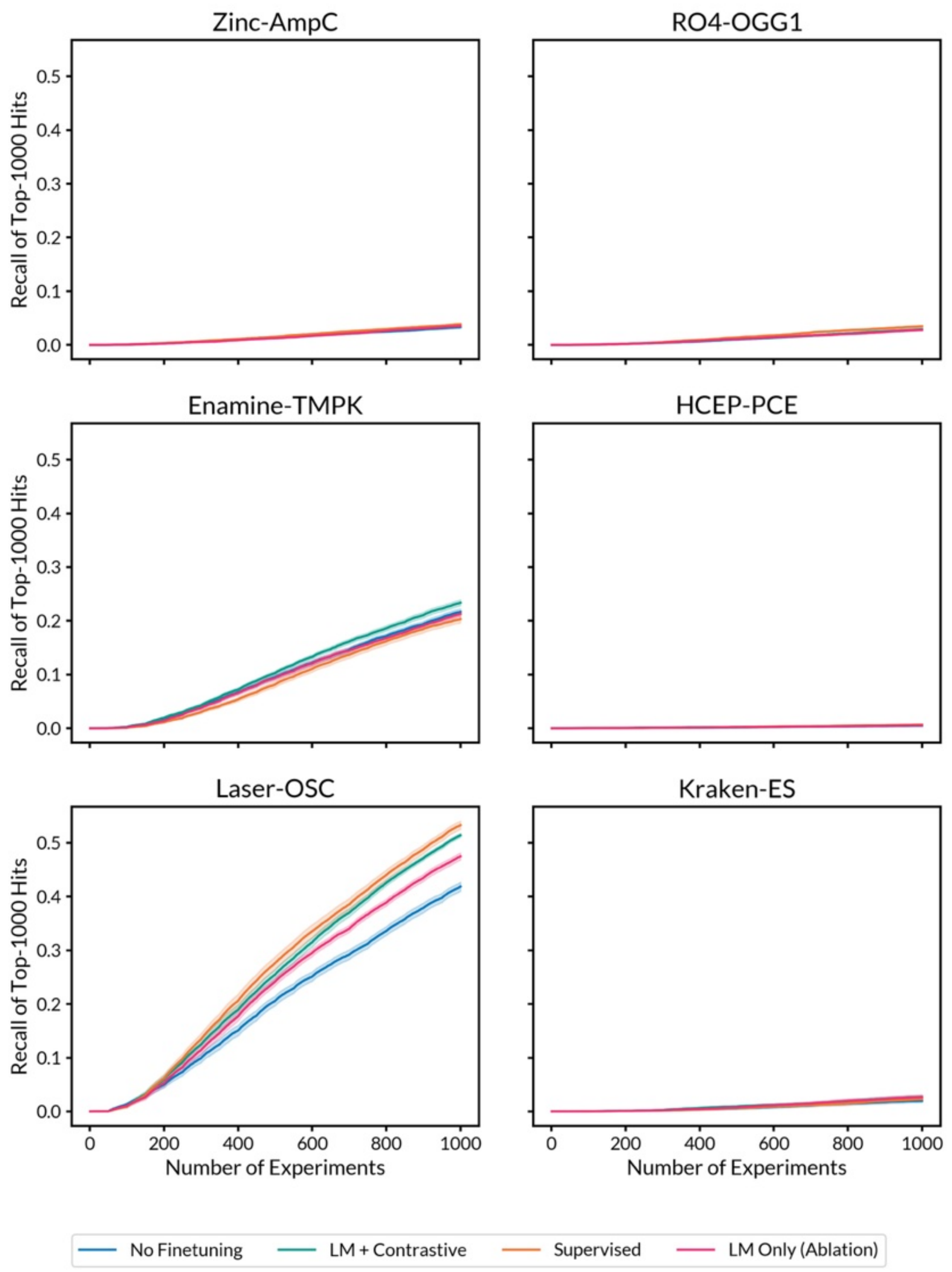

**Figure S 98**: Optimization trajectories after using different strategies to fine-tune MolFormer-XL embeddings, evaluated over a budget of 1,000 experiments (batch size of 50 experiments). Trajectories are shown as the mean over 20 independent campaigns from different random seed populations. The shaded areas indicate the standard error of the mean.

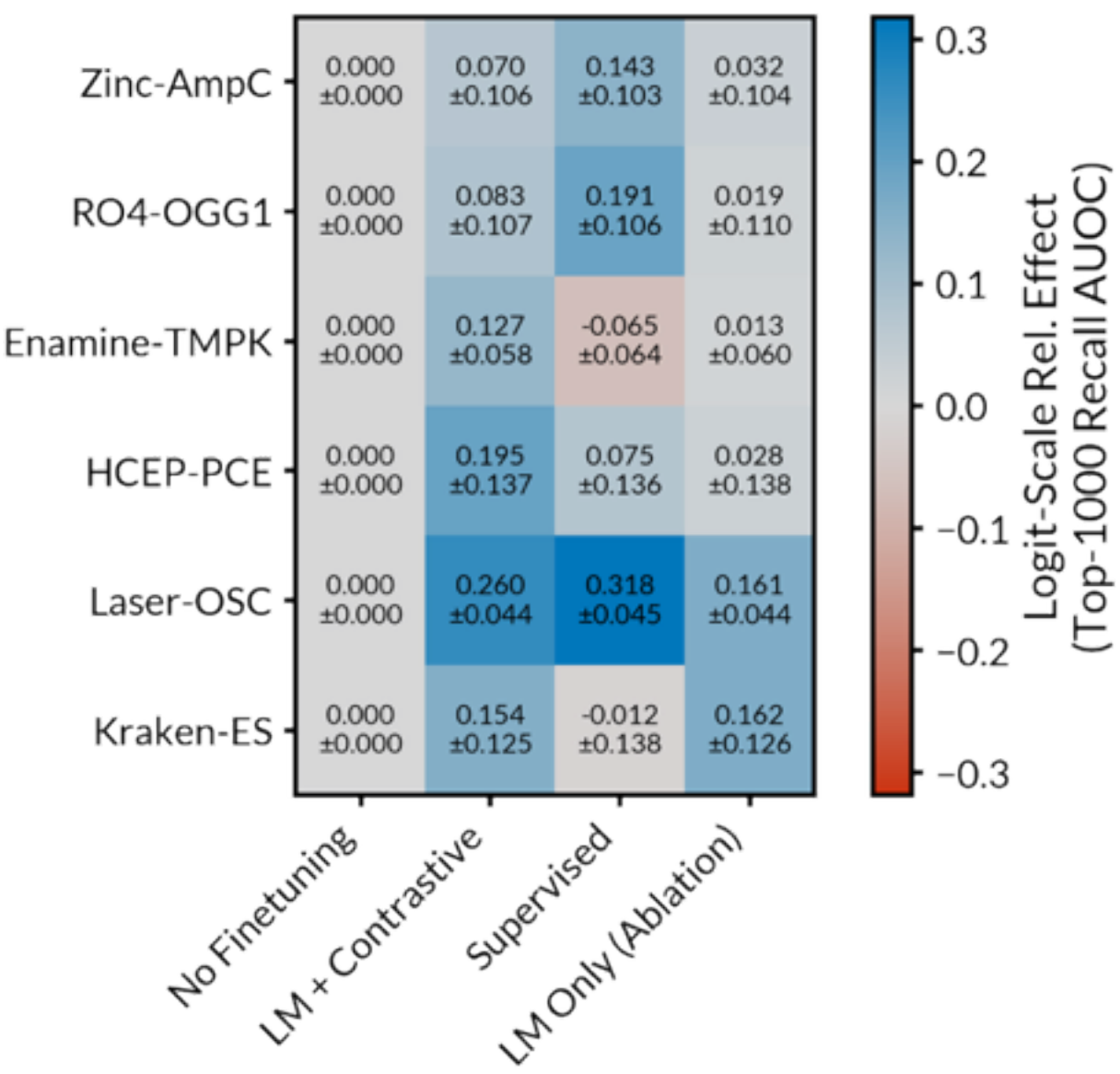

**Figure S 99**: Logit-scale effect of the fine-tuning strategy on MolFormer-XL embeddings, differentiated by molecular libraries. Experiments were performed using a Laplace Neural network model in the experimental discovery setting (1,000 experiments, batch size 50). Means and standard deviations were obtained from 8,000 posterior samples.

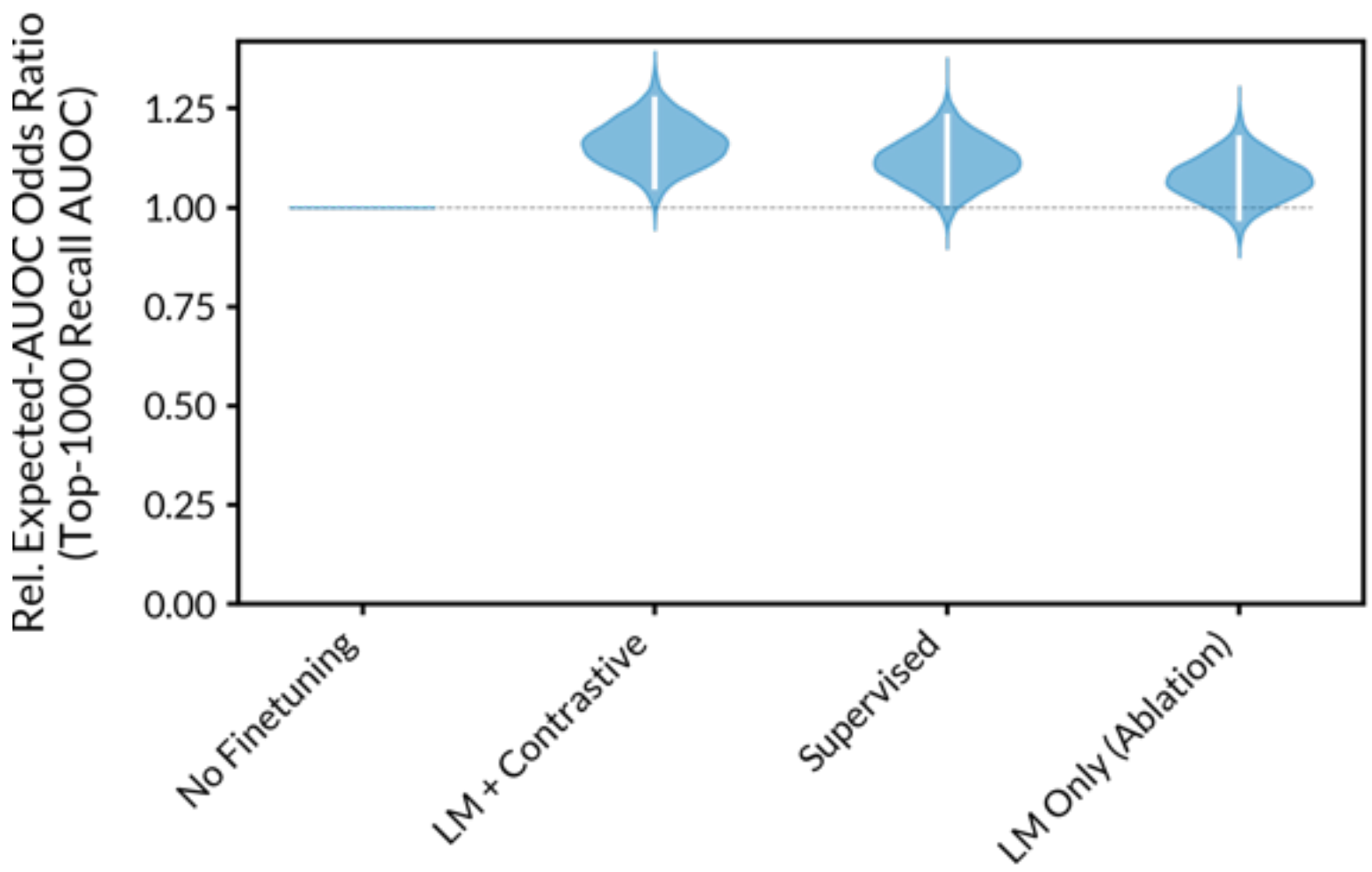

**Figure S 100**: Effect ratios of the fine-tuning strategy on MolFormer-XL embeddings, averaged over all libraries. Experiments were performed using a Laplace Neural network model in the experimental discovery setting (1,000 experiments, batch size 50). Bars within the violins indicate the 95% highest-density intervals.

***SmiTed***

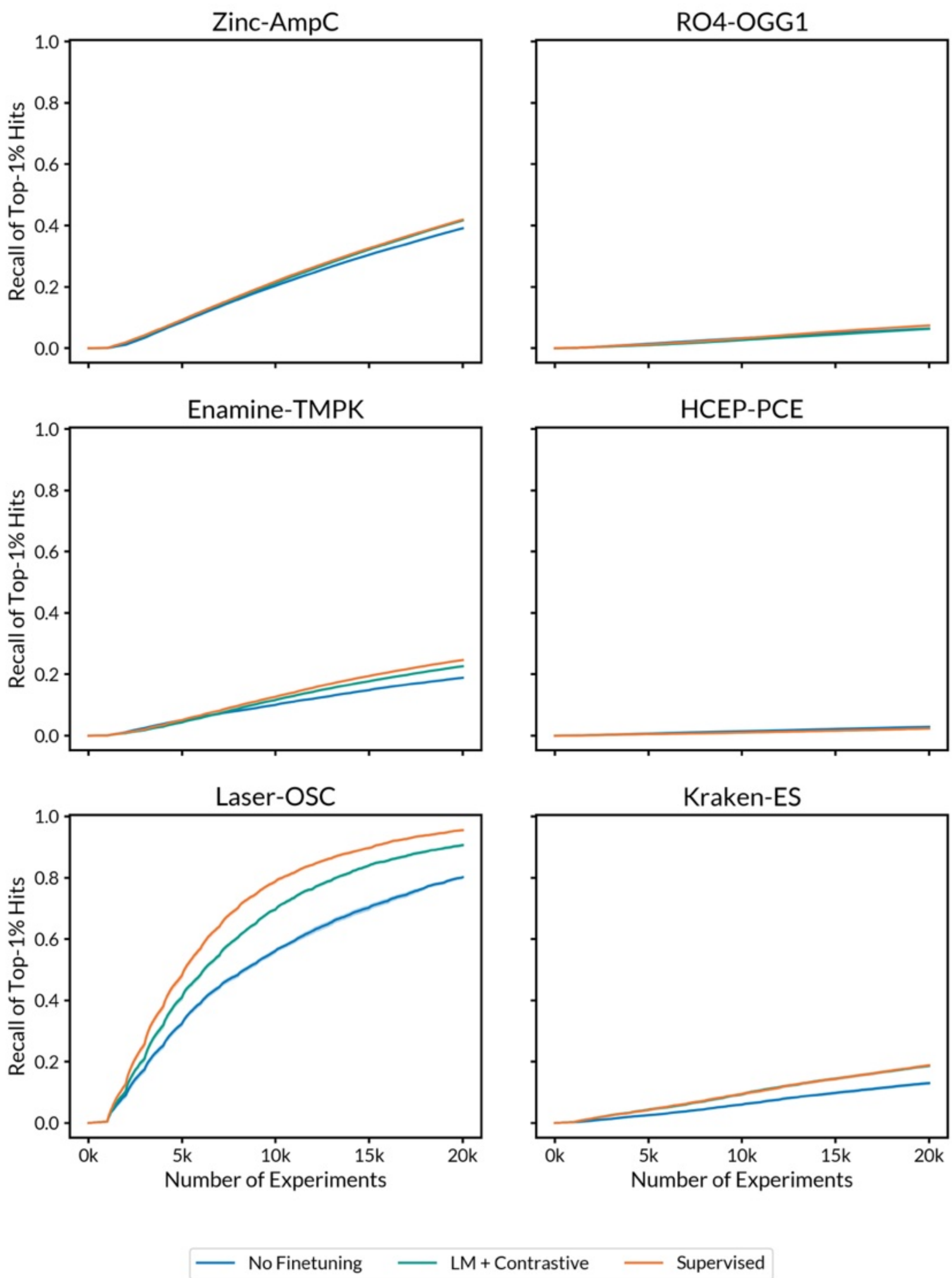

**Figure S 101**: Optimization trajectories after using different strategies to fine-tune SmiTed embeddings, evaluated over a budget of 20,000 experiments (batch size of 1,000 experiments). Trajectories are shown as the mean over 20 independent campaigns from different random seed populations. The shaded areas indicate the standard error of the mean.

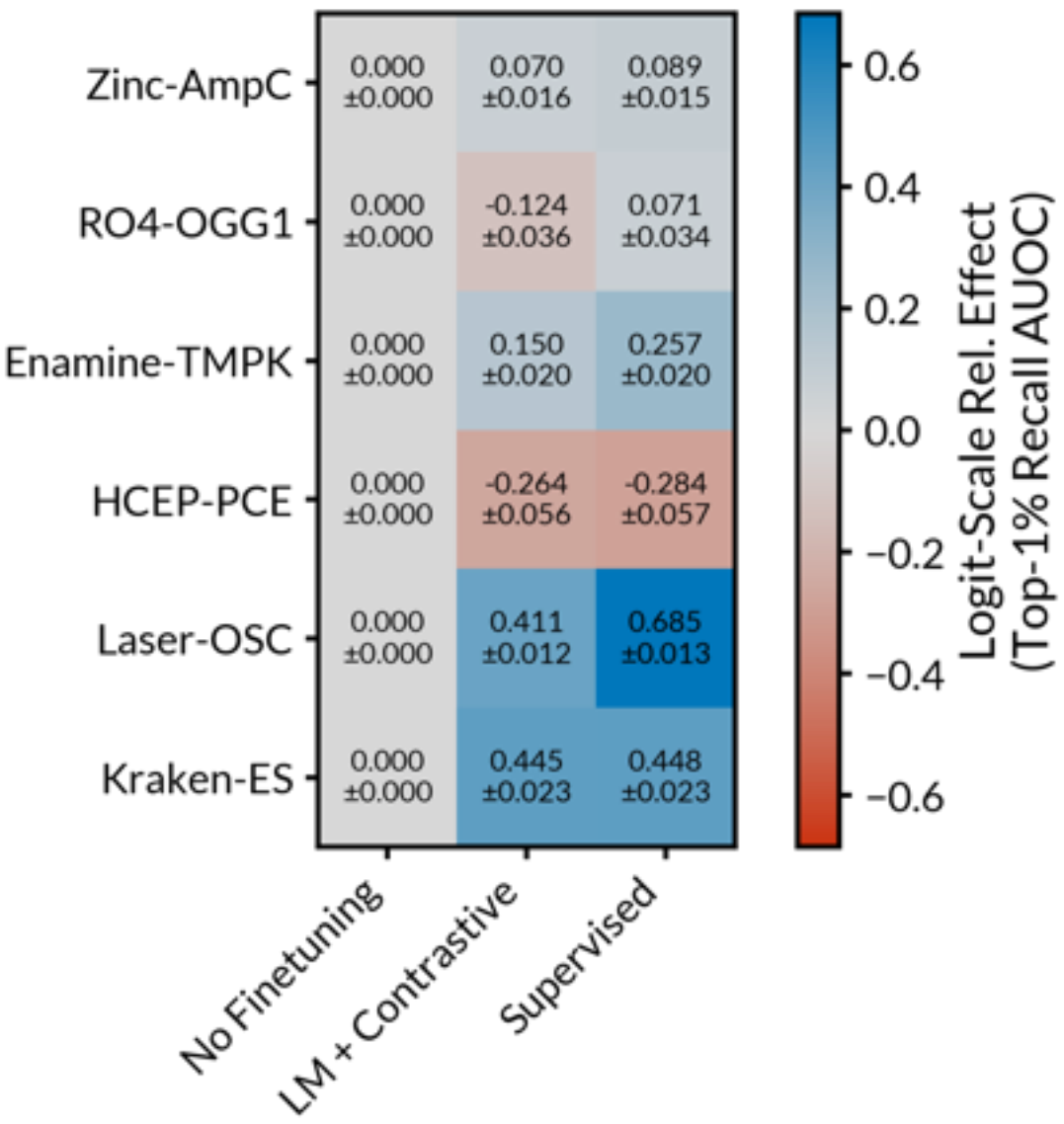

**Figure S 102**: Logit-scale effect of the fine-tuning strategy on SmiTed embeddings, differentiated by molecular libraries. Experiments were performed using a Laplace Neural network model in the virtual screening setting (20,000 experiments, batch size 1,000). Means and standard deviations were obtained from 8,000 posterior samples.

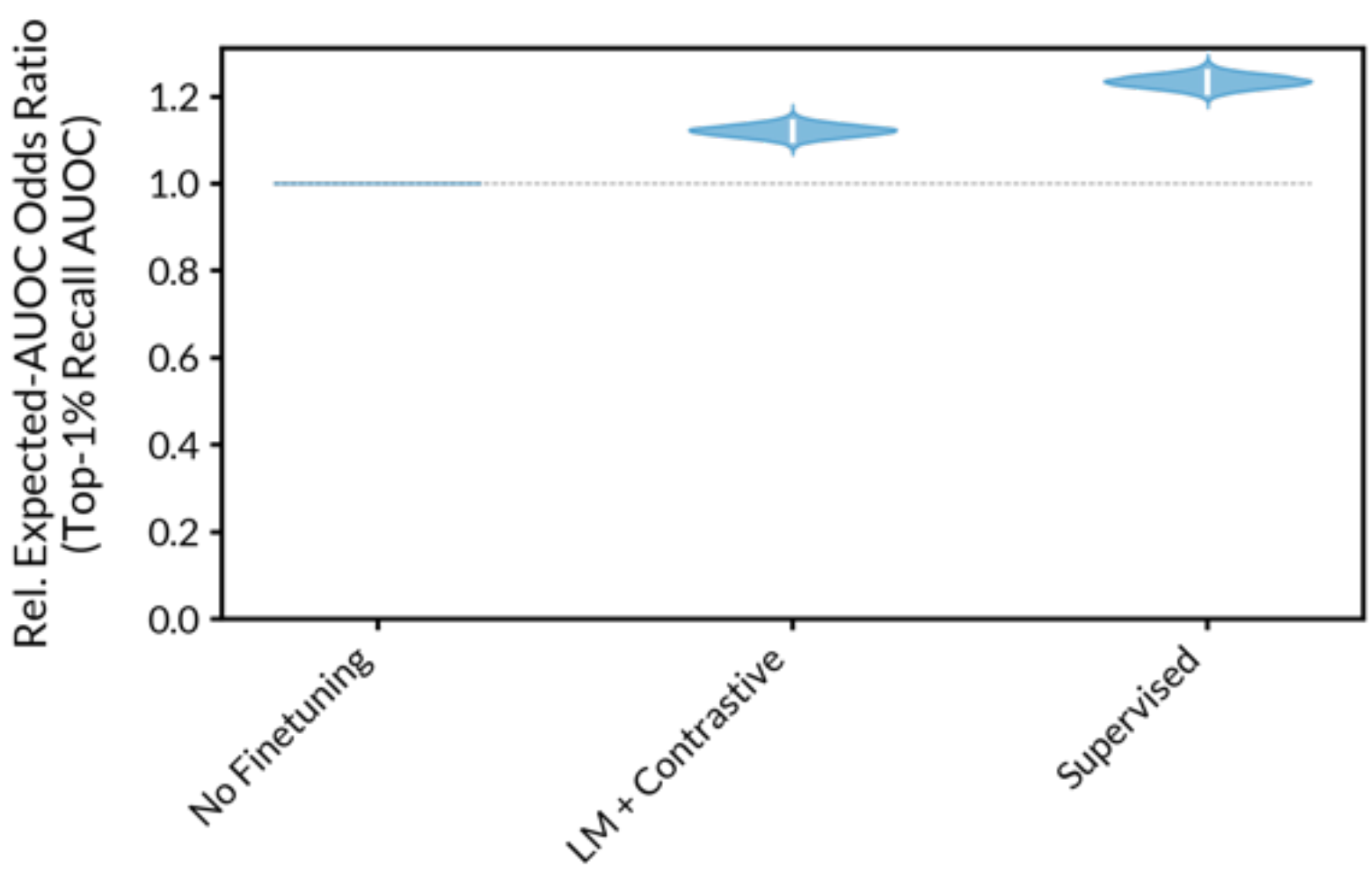

**Figure S 103**: Effect ratios of the fine-tuning strategy on SmiTed embeddings, averaged over all libraries. Experiments were performed using a Laplace Neural network model in the virtual screening setting (20,000 experiments, batch size 1,000). Bars within the violins indicate the 95% highest-density intervals.

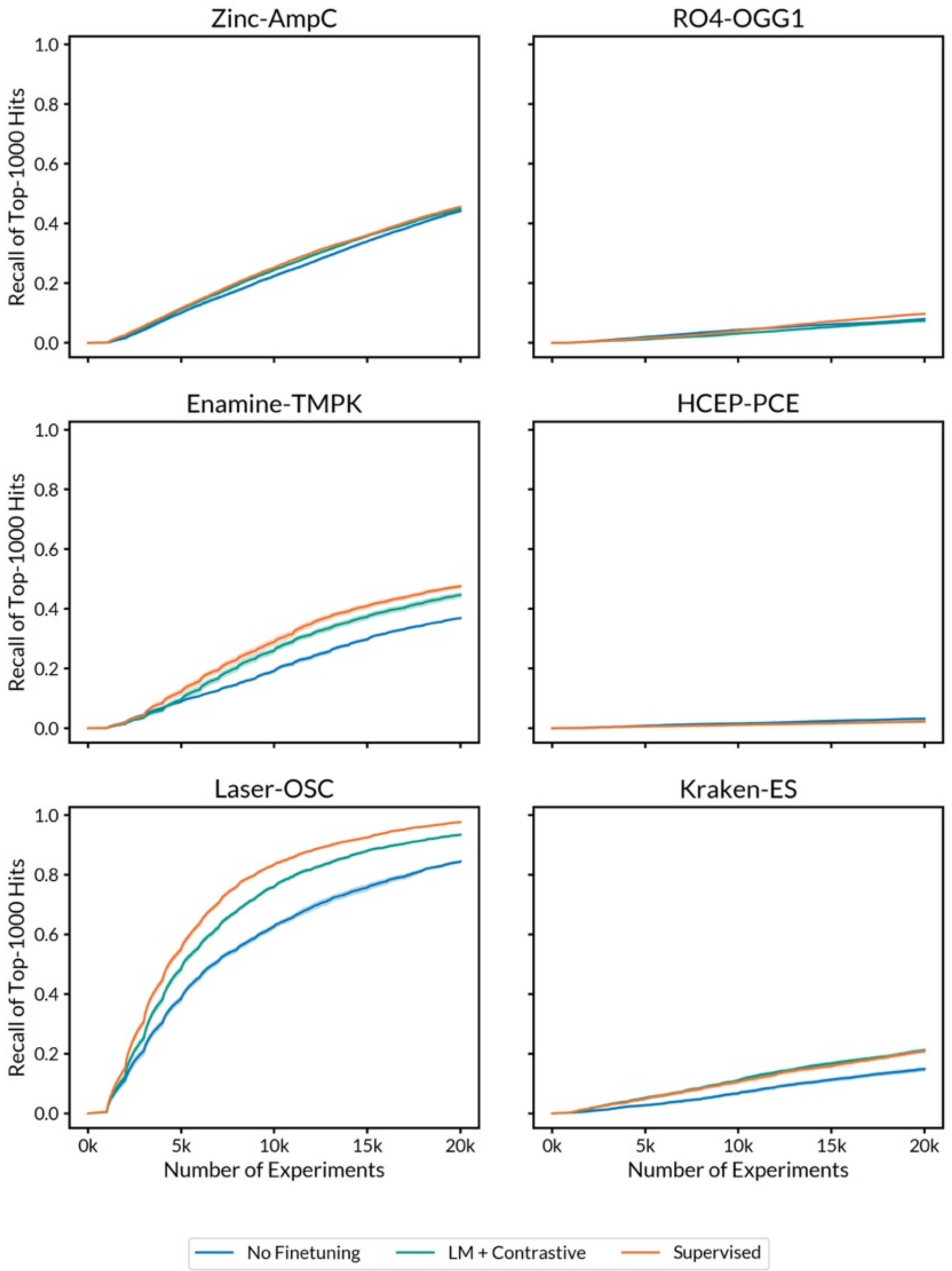

**Figure S 104**: Optimization trajectories after using different strategies to fine-tune SmiTed embeddings, evaluated over a budget of 20,000 experiments (batch size of 1,000 experiments). Trajectories are shown as the mean over 20 independent campaigns from different random seed populations. The shaded areas indicate the standard error of the mean.

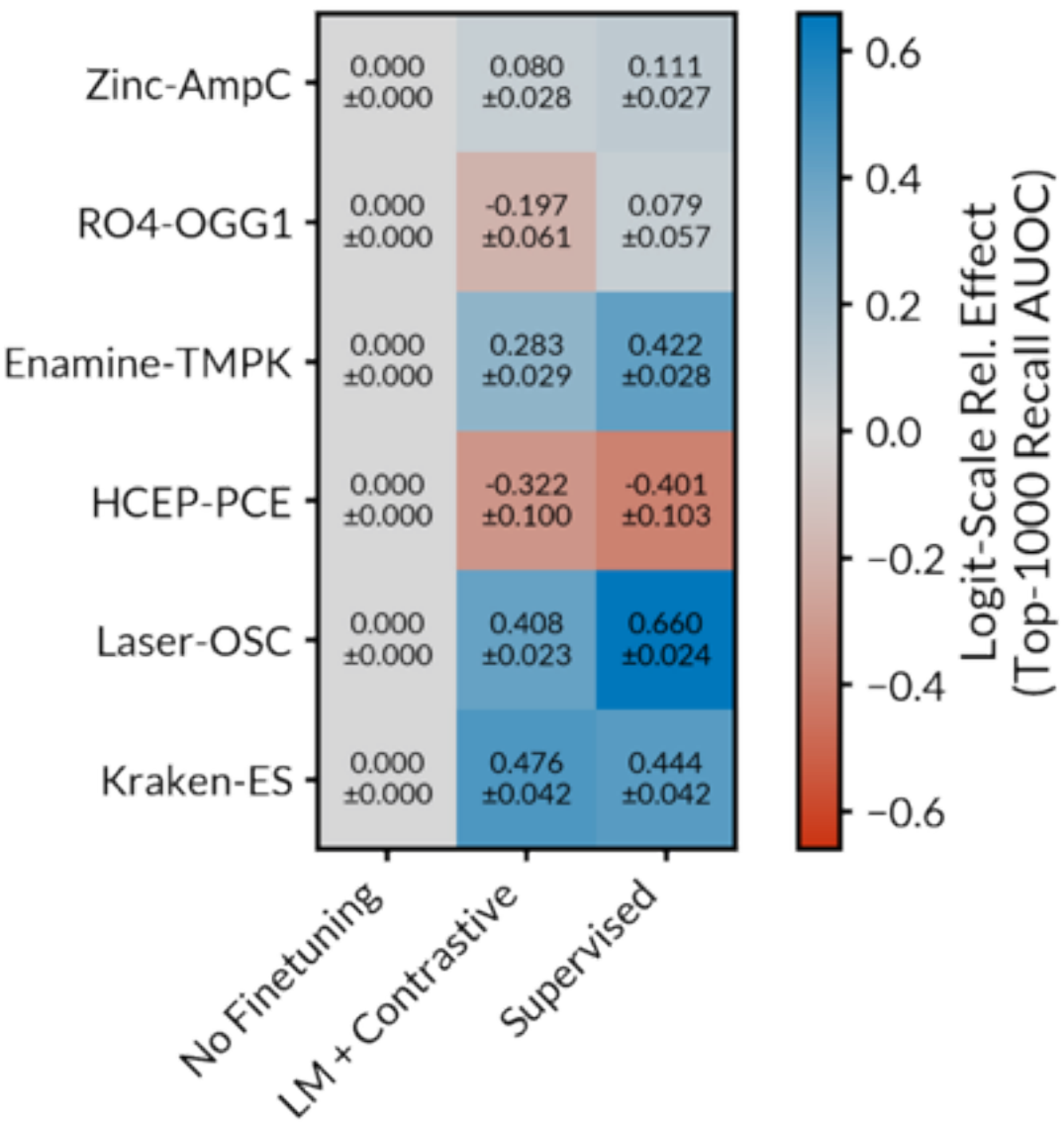

**Figure S 105**: Logit-scale effect of the fine-tuning strategy on SmiTed embeddings, differentiated by molecular libraries. Experiments were performed using a Laplace Neural network model in the virtual screening setting (20,000 experiments, batch size 1,000). Means and standard deviations were obtained from 8,000 posterior samples.

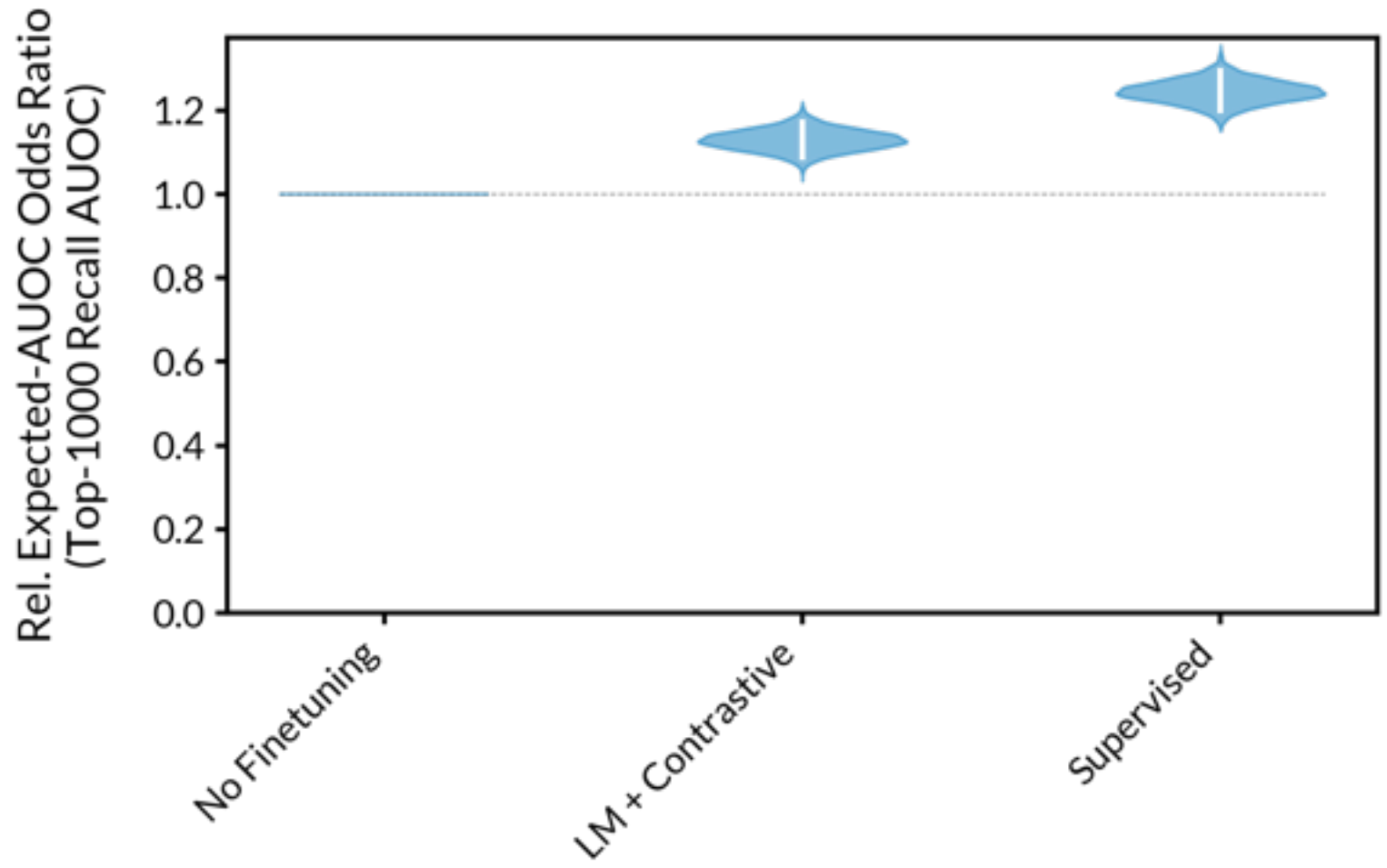

**Figure S 106**: Effect ratios of the fine-tuning strategy on SmiTed embeddings, averaged over all libraries. Experiments were performed using a Laplace Neural network model in the virtual screening setting (20,000 experiments, batch size 1,000). Bars within the violins indicate the 95% highest-density intervals.

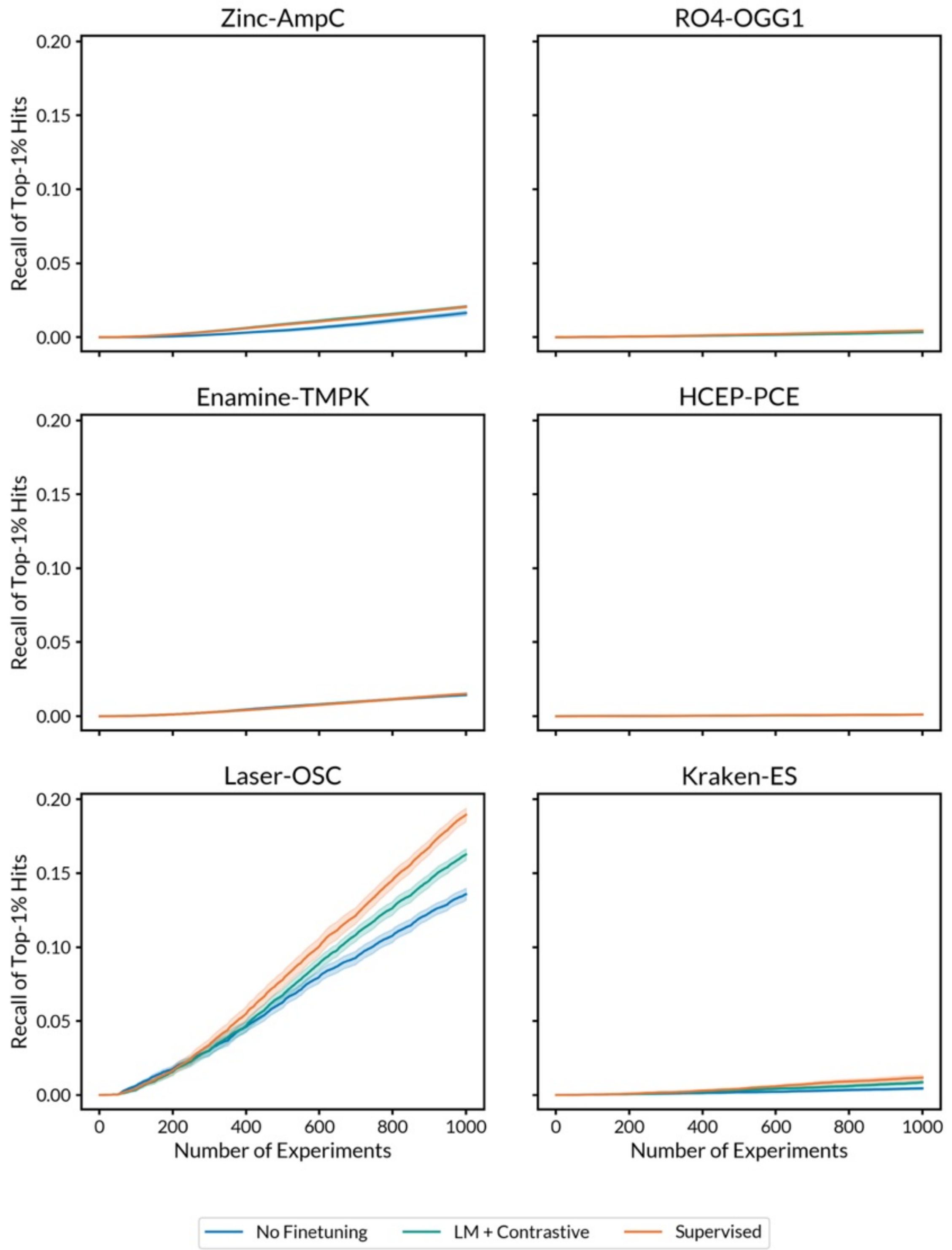

**Figure S 107**: Optimization trajectories after using different strategies to fine-tune SmiTed embeddings, evaluated over a budget of 1,000 experiments (batch size of 50 experiments). Trajectories are shown as the mean over 20 independent campaigns from different random seed populations. The shaded areas indicate the standard error of the mean.

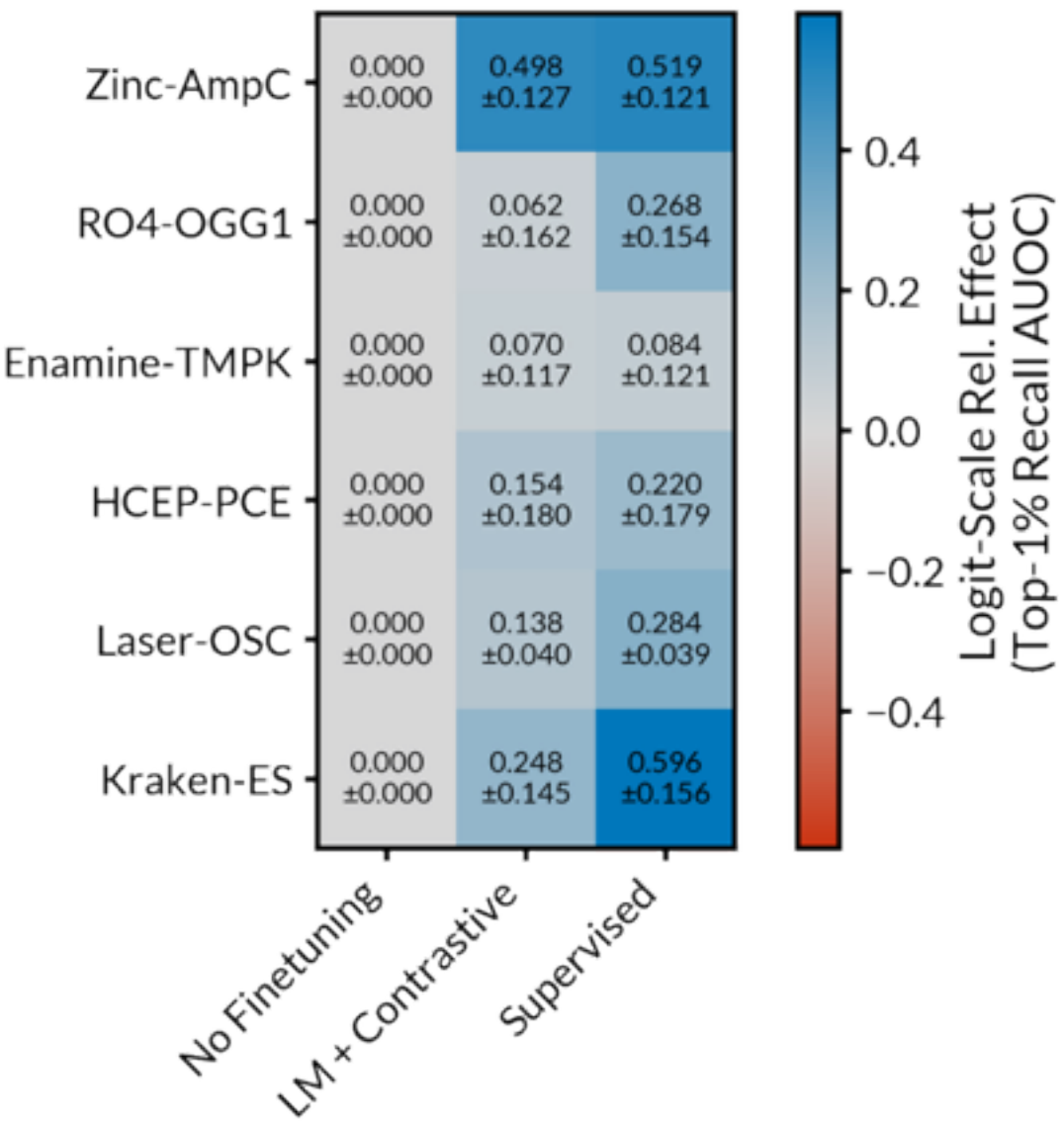

**Figure S 108**: Logit-scale effect of the fine-tuning strategy on SmiTed embeddings, differentiated by molecular libraries. Experiments were performed using a Laplace Neural network model in the experimental discovery setting (1,000 experiments, batch size 50). Means and standard deviations were obtained from 8,000 posterior samples.

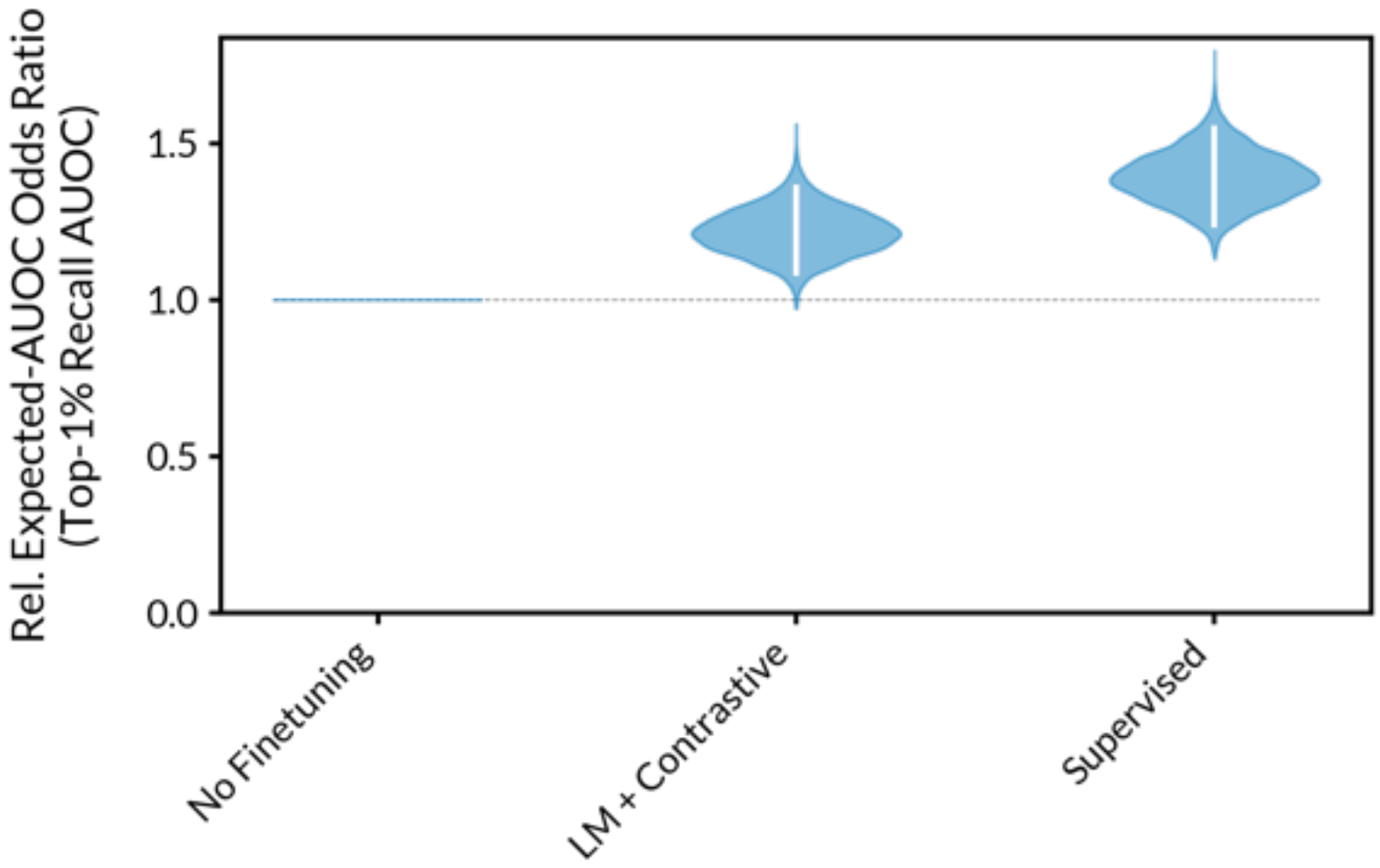

**Figure S 109**: Effect ratios of the fine-tuning strategy on SmiTed embeddings, averaged over all libraries. Experiments were performed using a Laplace Neural network model in the experimental discovery setting (1,000 experiments, batch size 50). Bars within the violins indicate the 95% highest-density intervals.

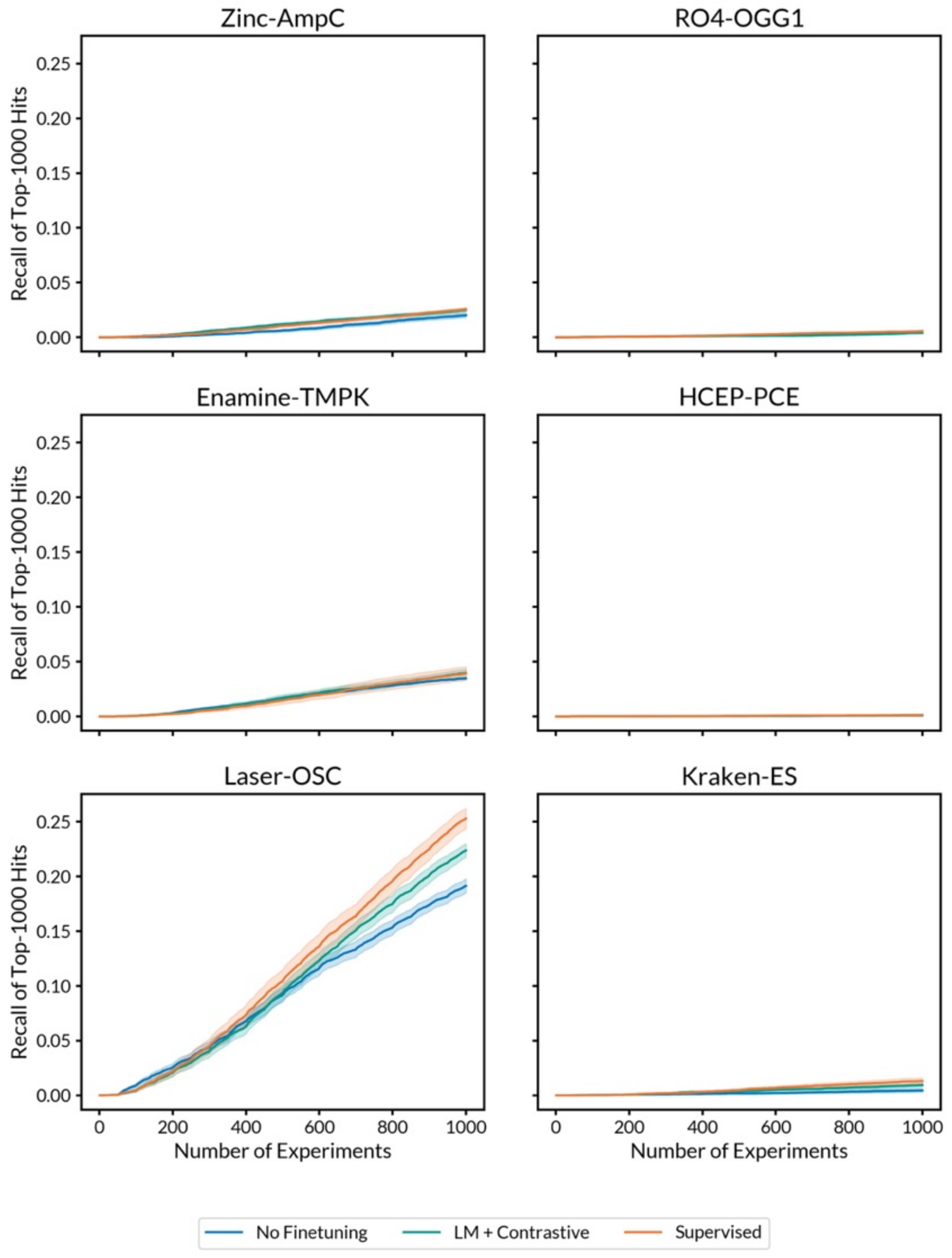

**Figure S 110**: Optimization trajectories after using different strategies to fine-tune SmiTed embeddings, evaluated over a budget of 1,000 experiments (batch size of 50 experiments). Trajectories are shown as the mean over 20 independent campaigns from different random seed populations. The shaded areas indicate the standard error of the mean.

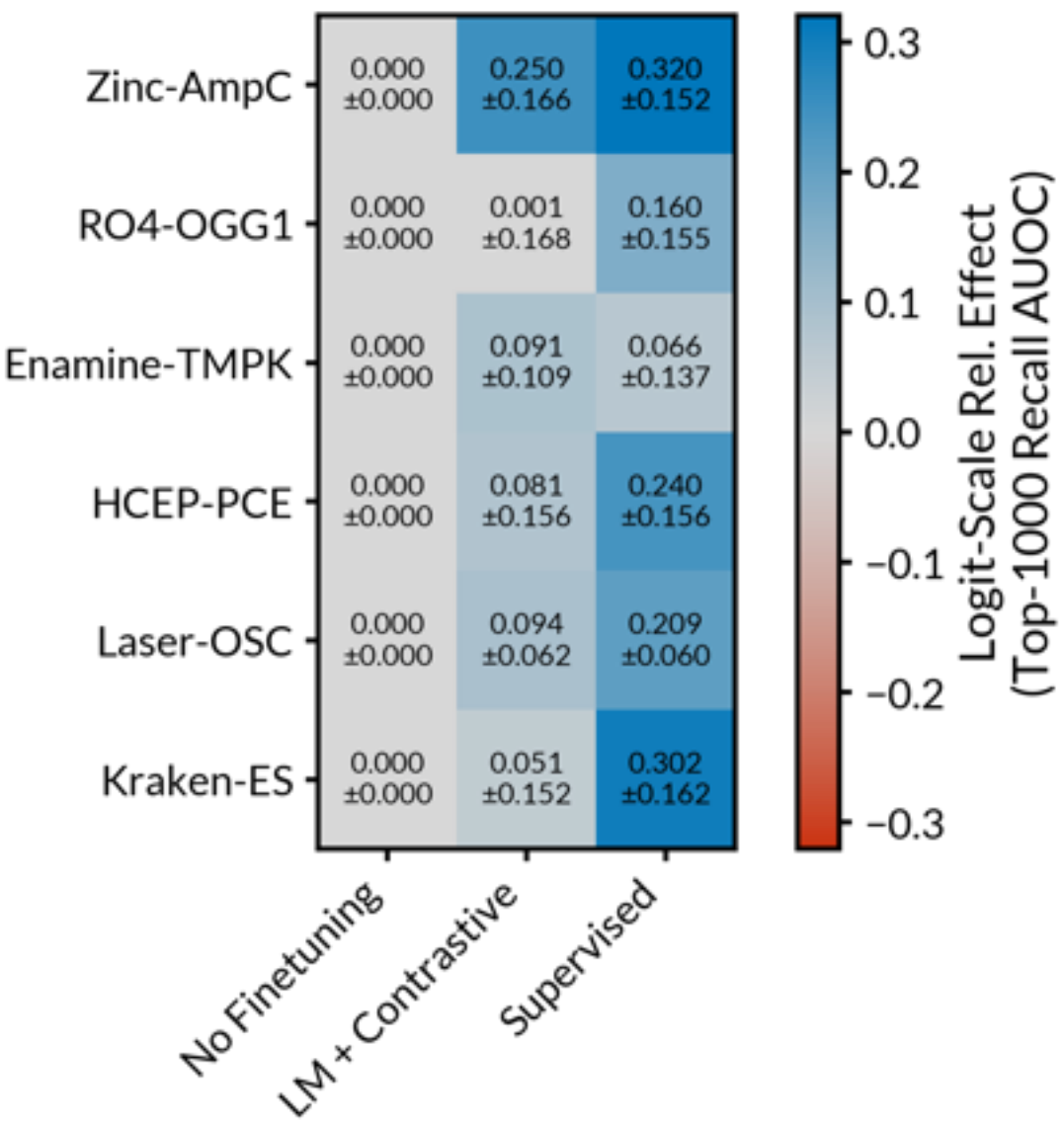

**Figure S 111**: Logit-scale effect of the fine-tuning strategy on SmiTed embeddings, differentiated by molecular libraries. Experiments were performed using a Laplace Neural network model in the experimental discovery setting (1,000 experiments, batch size 50). Means and standard deviations were obtained from 8,000 posterior samples.

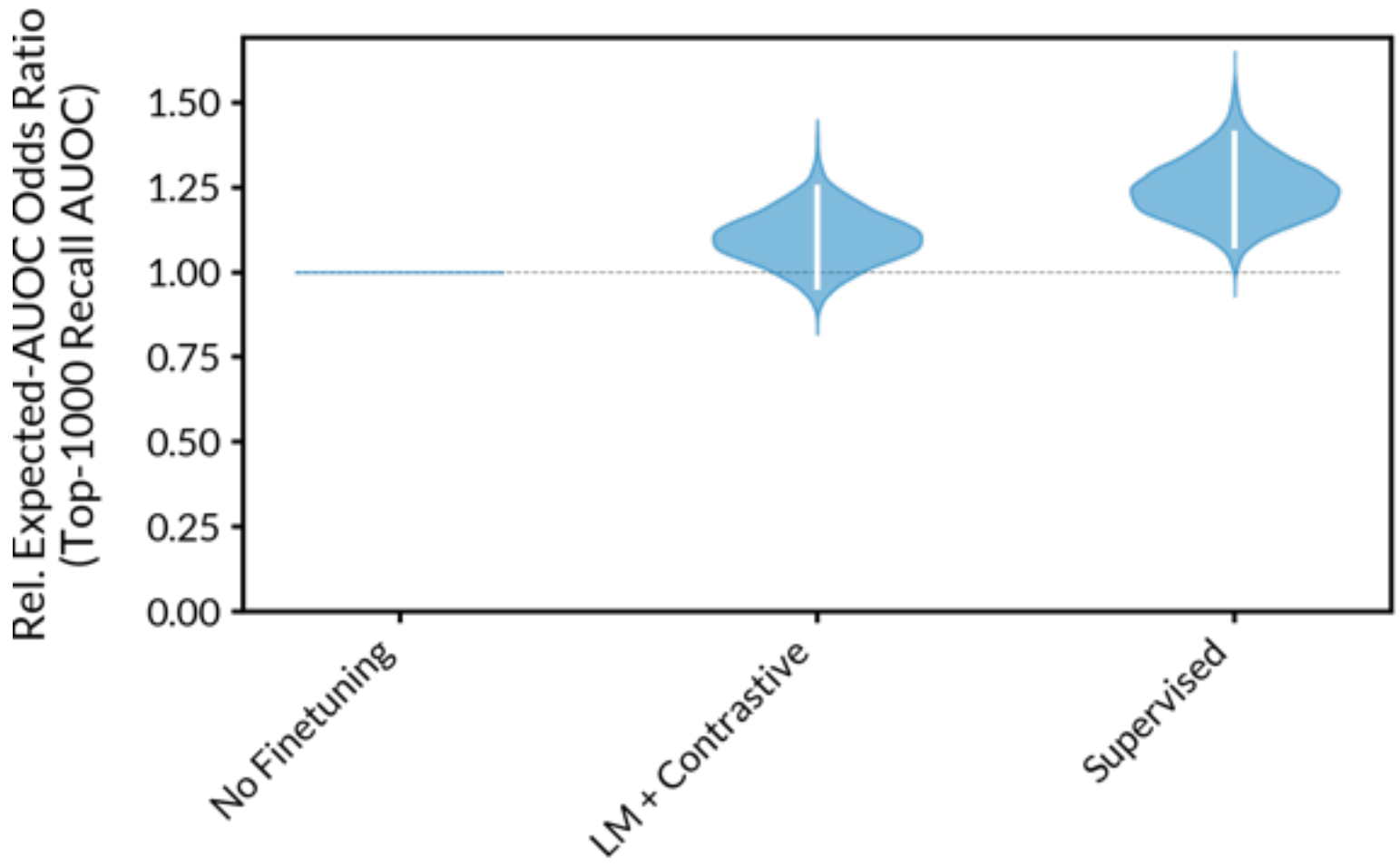

**Figure S 112**: Effect ratios of the fine-tuning strategy on SmiTed embeddings, averaged over all libraries. Experiments were performed using a Laplace Neural network model in the experimental discovery setting (1,000 experiments, batch size 50). Bars within the violins indicate the 95% highest-density intervals.

***T5Chem***

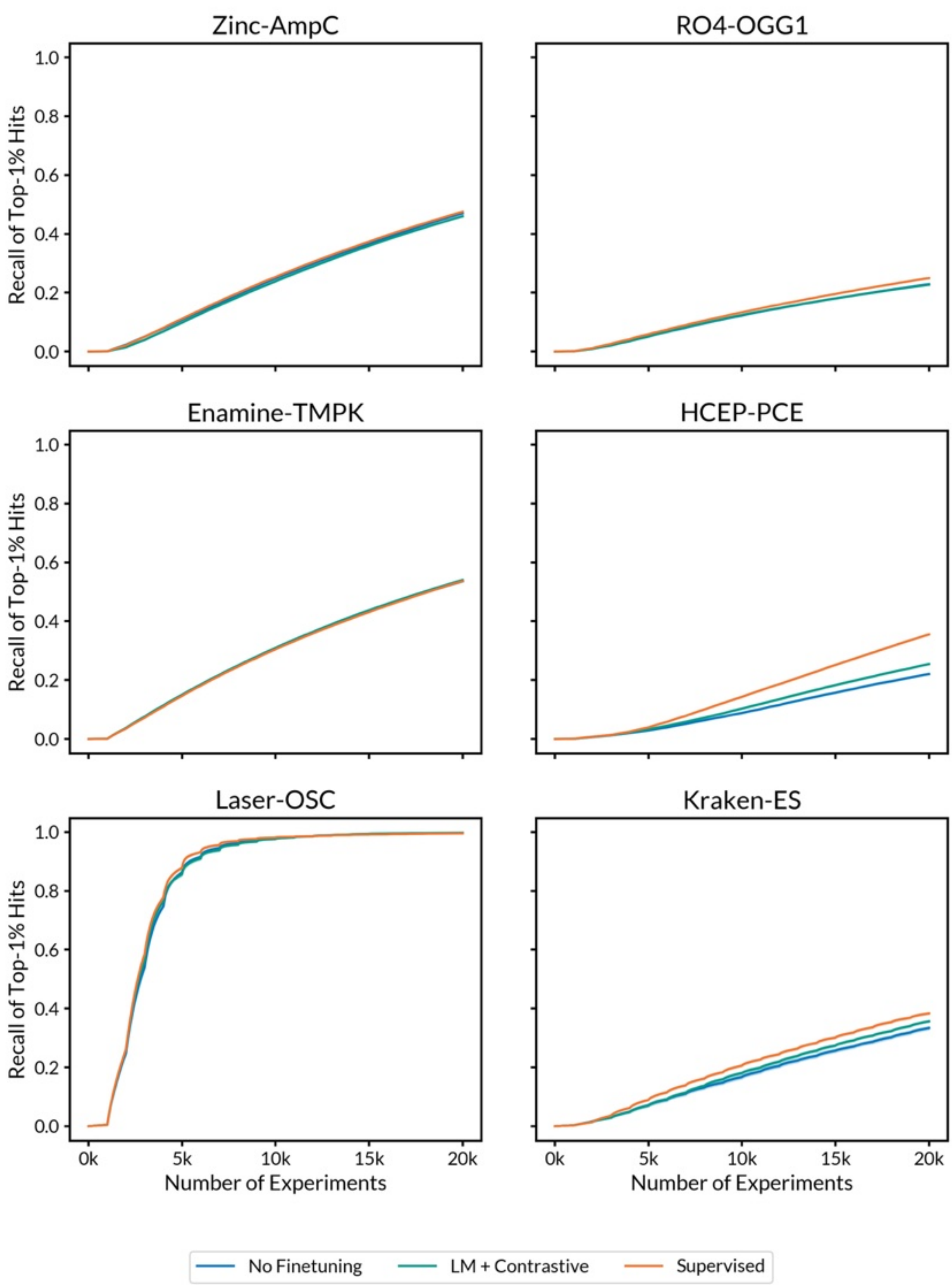

**Figure S 113**: Optimization trajectories after using different strategies to fine-tune T5Chem embeddings, evaluated over a budget of 20,000 experiments (batch size of 1,000 experiments). Trajectories are shown as the mean over 20 independent campaigns from different random seed populations. The shaded areas indicate the standard error of the mean.

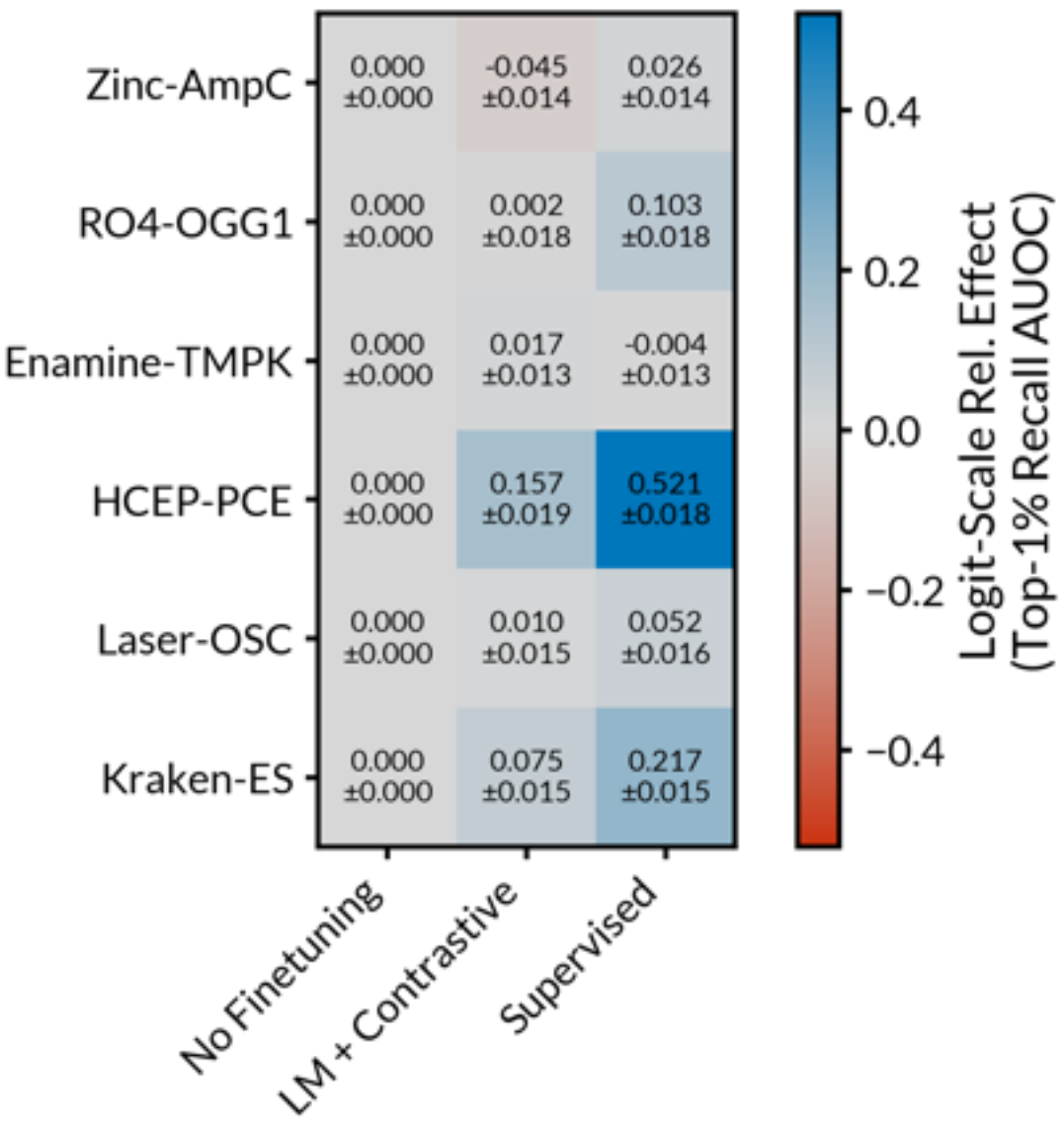

**Figure S 114**: Logit-scale effect of the fine-tuning strategy on T5Chem embeddings, differentiated by molecular libraries. Experiments were performed using a Laplace Neural network model in the virtual screening setting (20,000 experiments, batch size 1,000). Means and standard deviations were obtained from 8,000 posterior samples.

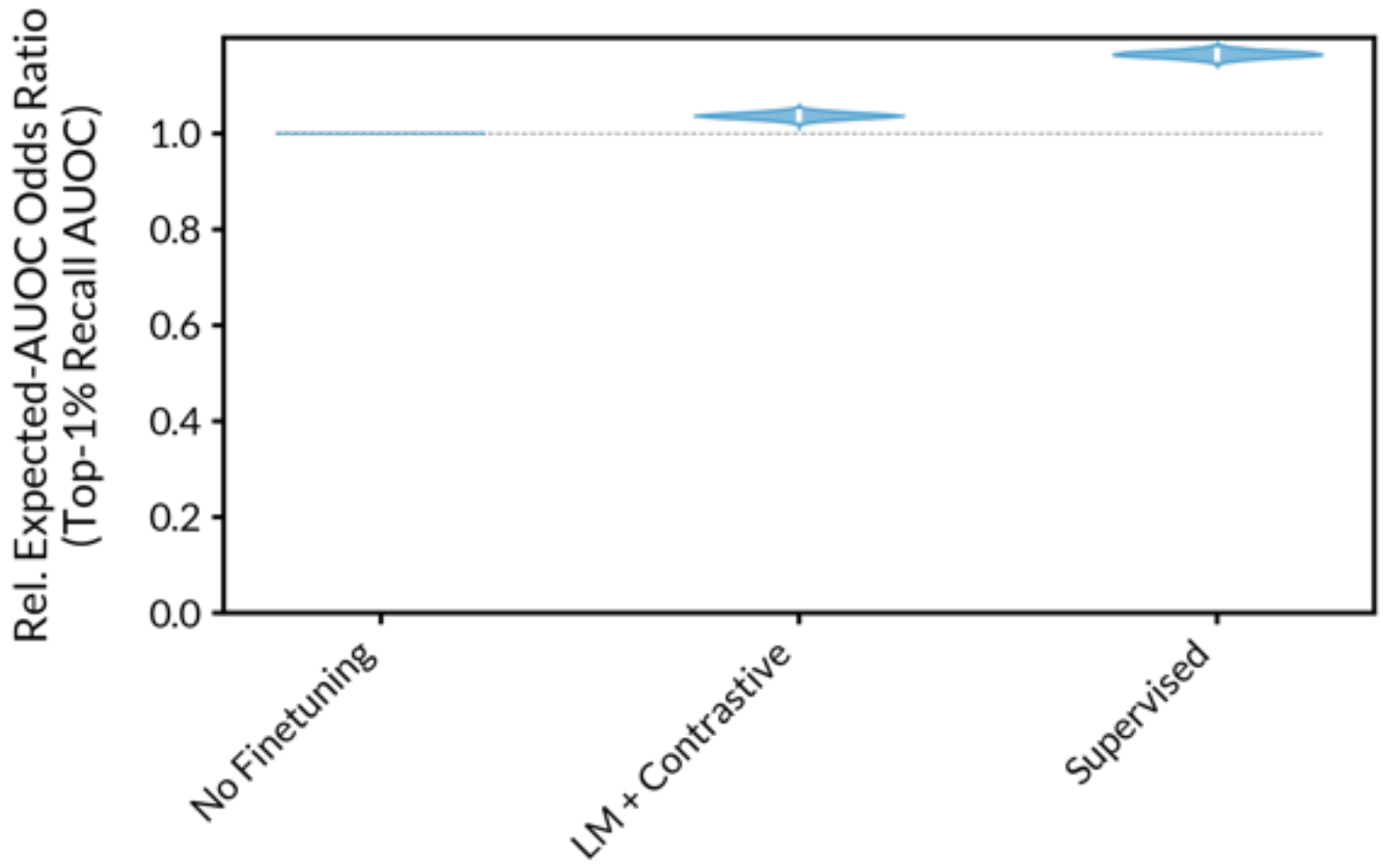

**Figure S 115**: Effect ratios of the fine-tuning strategy on T5Chem embeddings, averaged over all libraries. Experiments were performed using a Laplace Neural network model in the virtual screening setting (20,000 experiments, batch size 1,000). Bars within the violins indicate the 95% highest-density intervals.

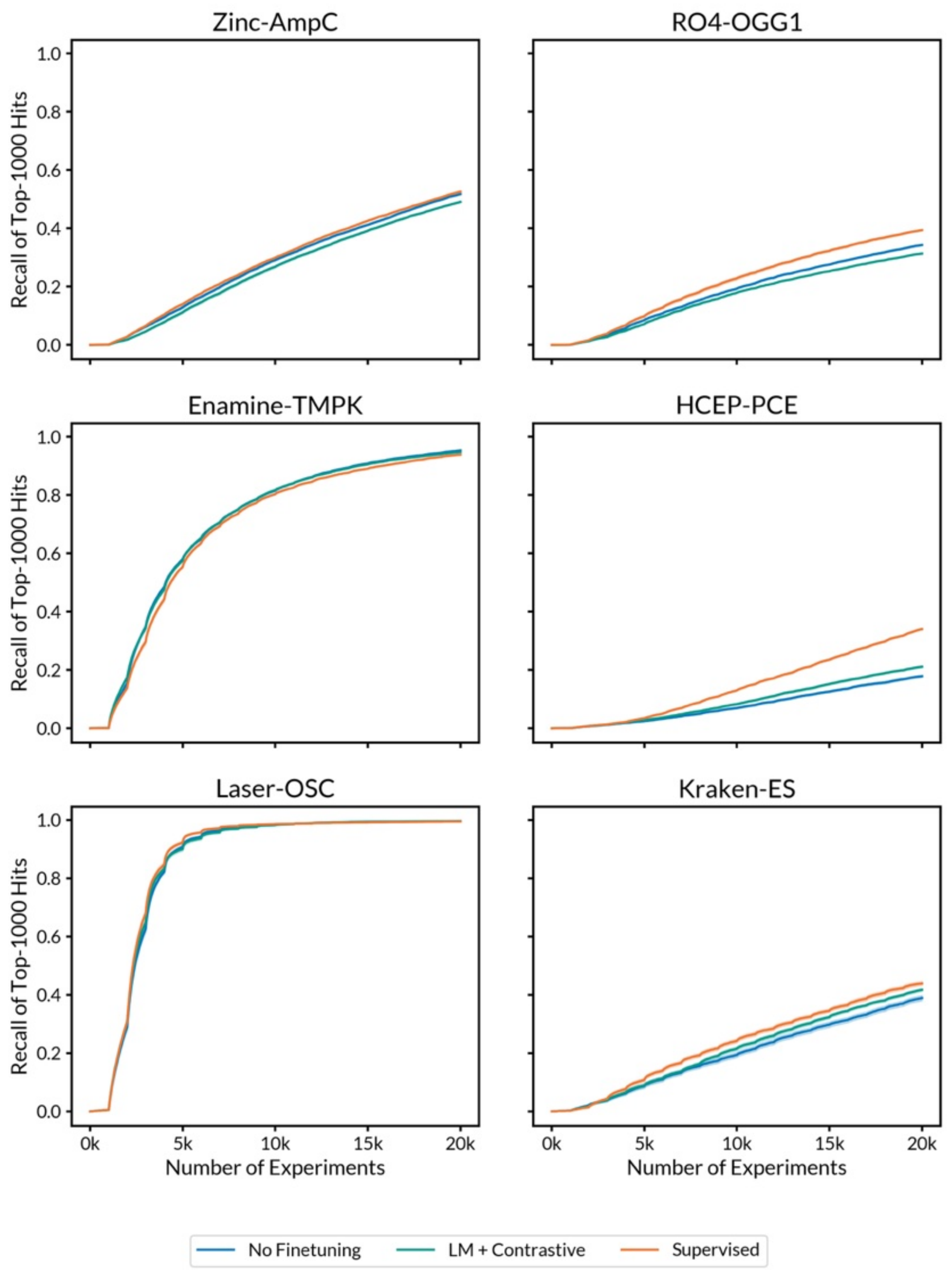

**Figure S 116**: Optimization trajectories after using different strategies to fine-tune T5Chem embeddings, evaluated over a budget of 20,000 experiments (batch size of 1,000 experiments). Trajectories are shown as the mean over 20 independent campaigns from different random seed populations. The shaded areas indicate the standard error of the mean.

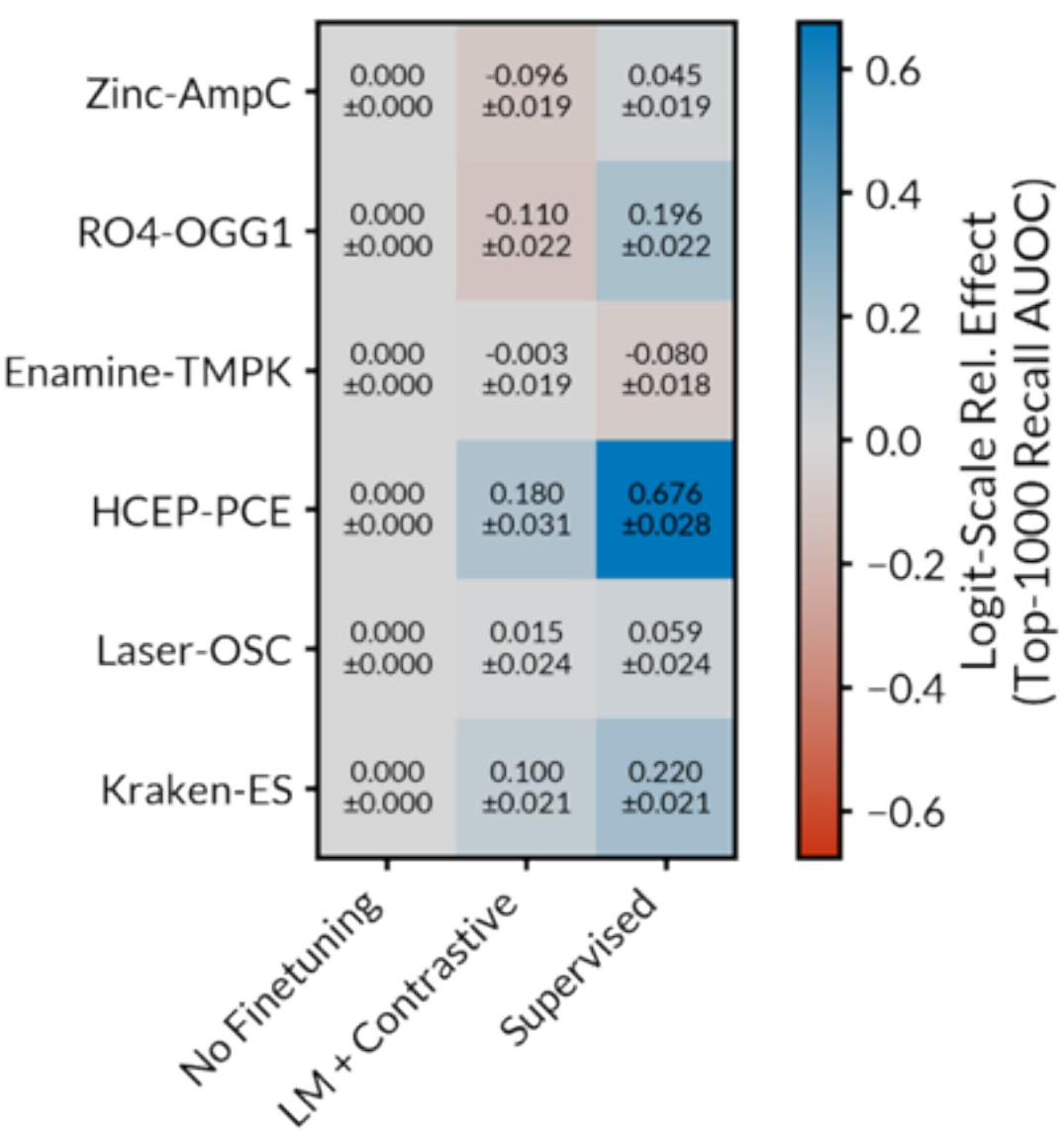

**Figure S 117**: Logit-scale effect of the fine-tuning strategy on T5Chem embeddings, differentiated by molecular libraries. Experiments were performed using a Laplace Neural network model in the virtual screening setting (20,000 experiments, batch size 1,000). Means and standard deviations were obtained from 8,000 posterior samples.

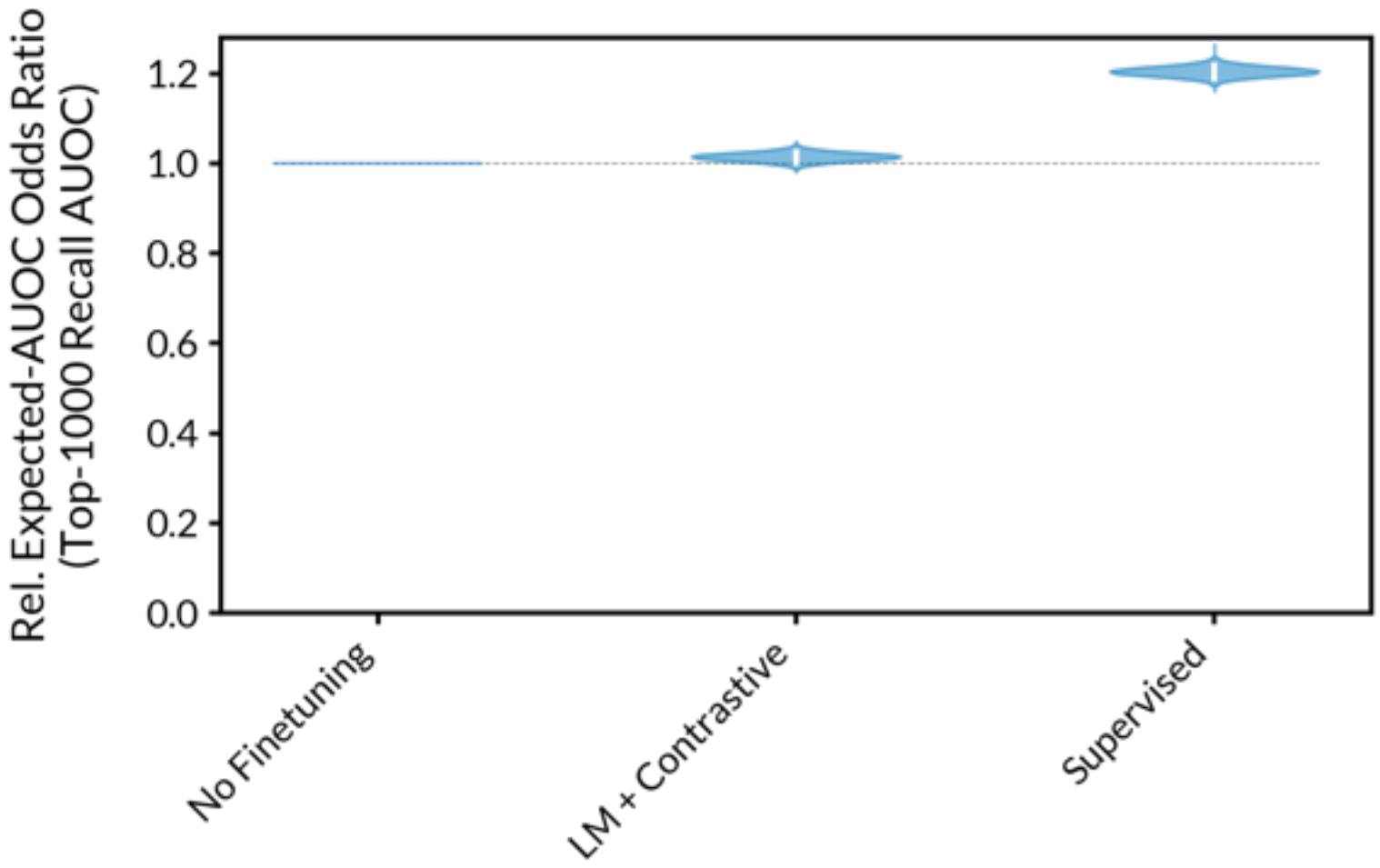

**Figure S 118**: Effect ratios of the fine-tuning strategy on T5Chem embeddings, averaged over all libraries. Experiments were performed using a Laplace Neural network model in the virtual screening setting (20,000 experiments, batch size 1,000). Bars within the violins indicate the 95% highest-density intervals.

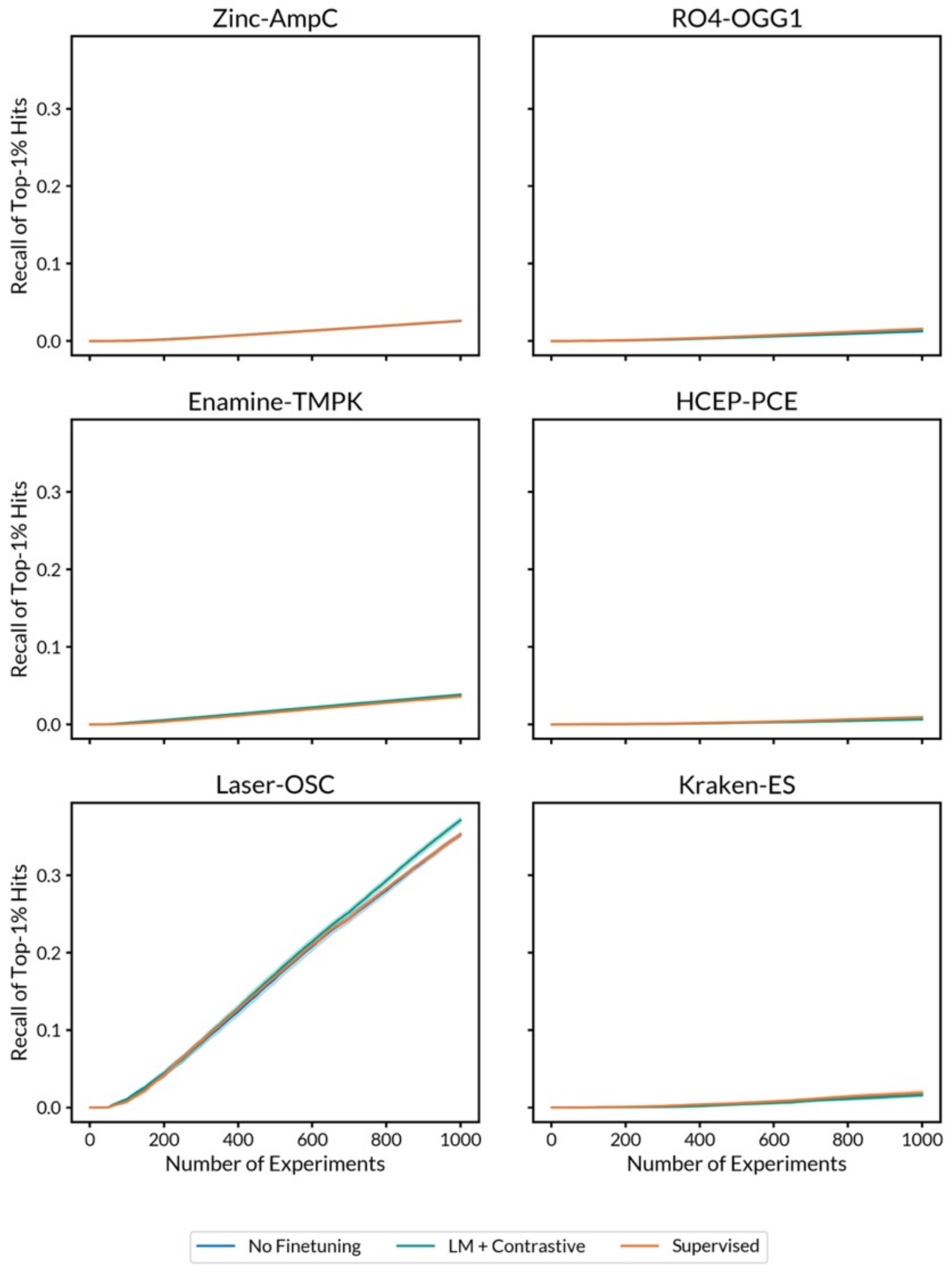

**Figure S 119**: Optimization trajectories after using different strategies to fine-tune T5Chem embeddings, evaluated over a budget of 1,000 experiments (batch size of 50 experiments). Trajectories are shown as the mean over 20 independent campaigns from different random seed populations. The shaded areas indicate the standard error of the mean.

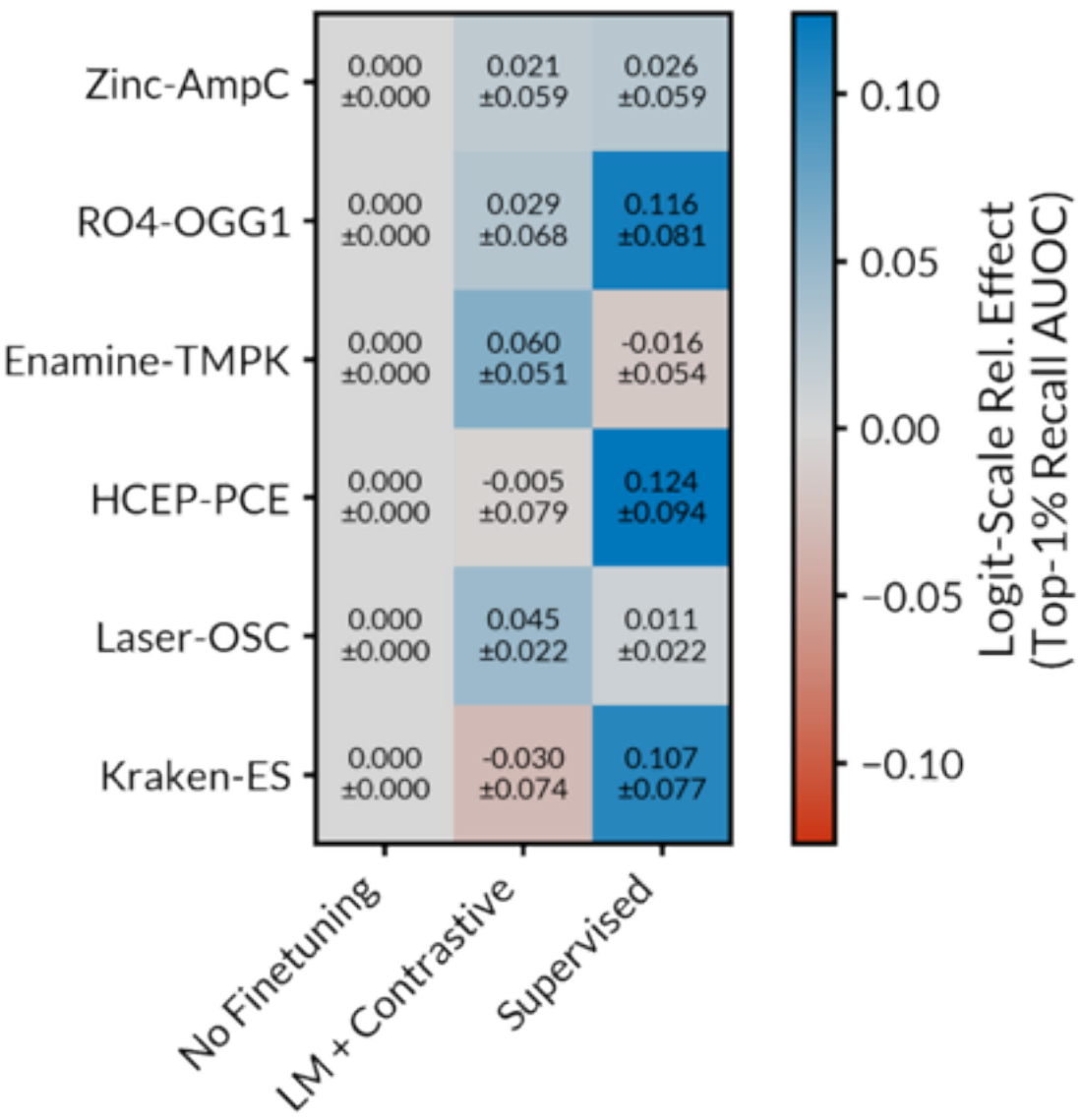

**Figure S 120**: Logit-scale effect of the fine-tuning strategy on T5Chem embeddings, differentiated by molecular libraries. Experiments were performed using a Laplace Neural network model in the experimental discovery setting (1,000 experiments, batch size 50). Means and standard deviations were obtained from 8,000 posterior samples.

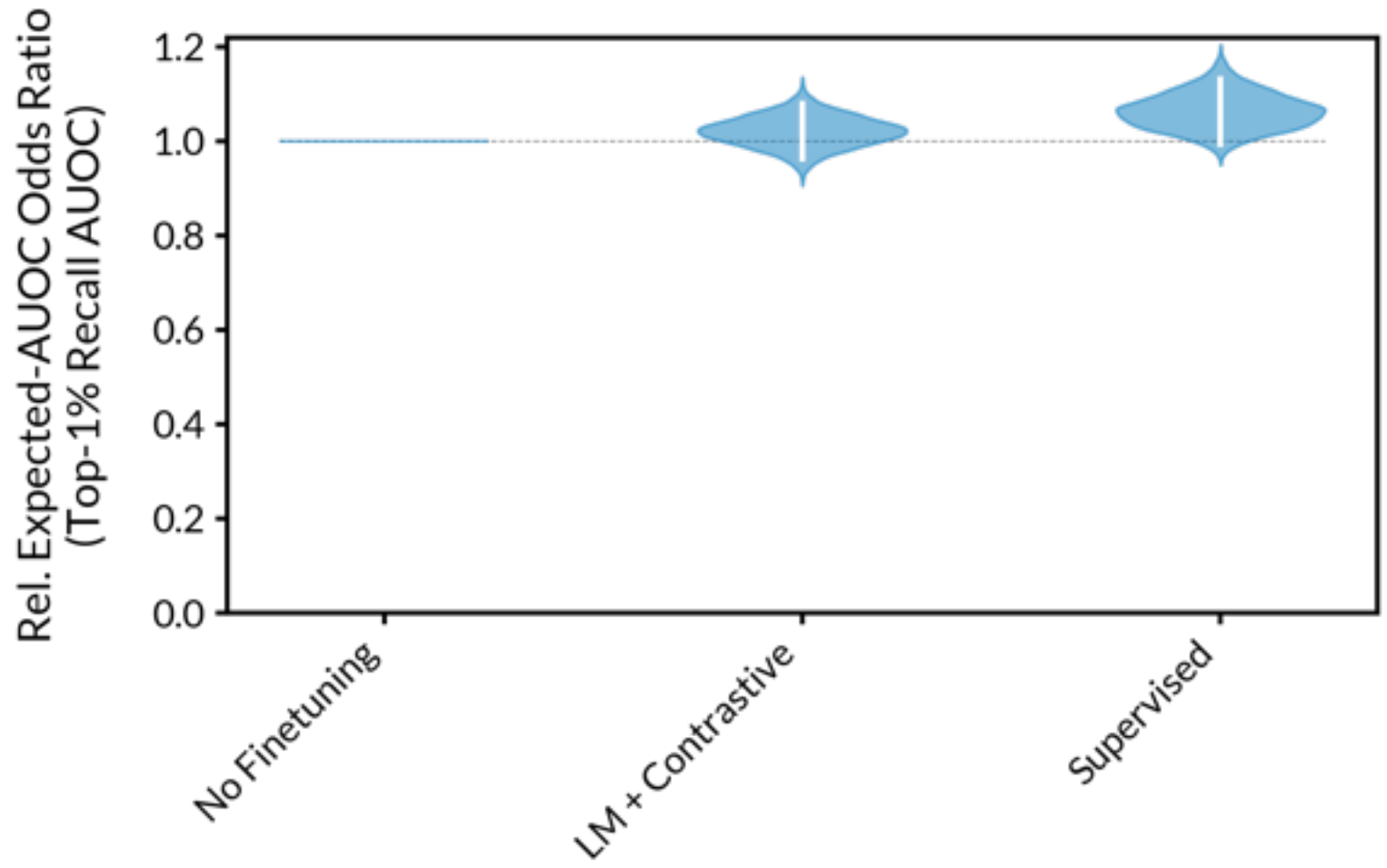

**Figure S 121**: Effect ratios of the fine-tuning strategy on T5Chem embeddings, averaged over all libraries. Experiments were performed using a Laplace Neural network model in the experimental discovery setting (1,000 experiments, batch size 50). Bars within the violins indicate the 95% highest-density intervals.

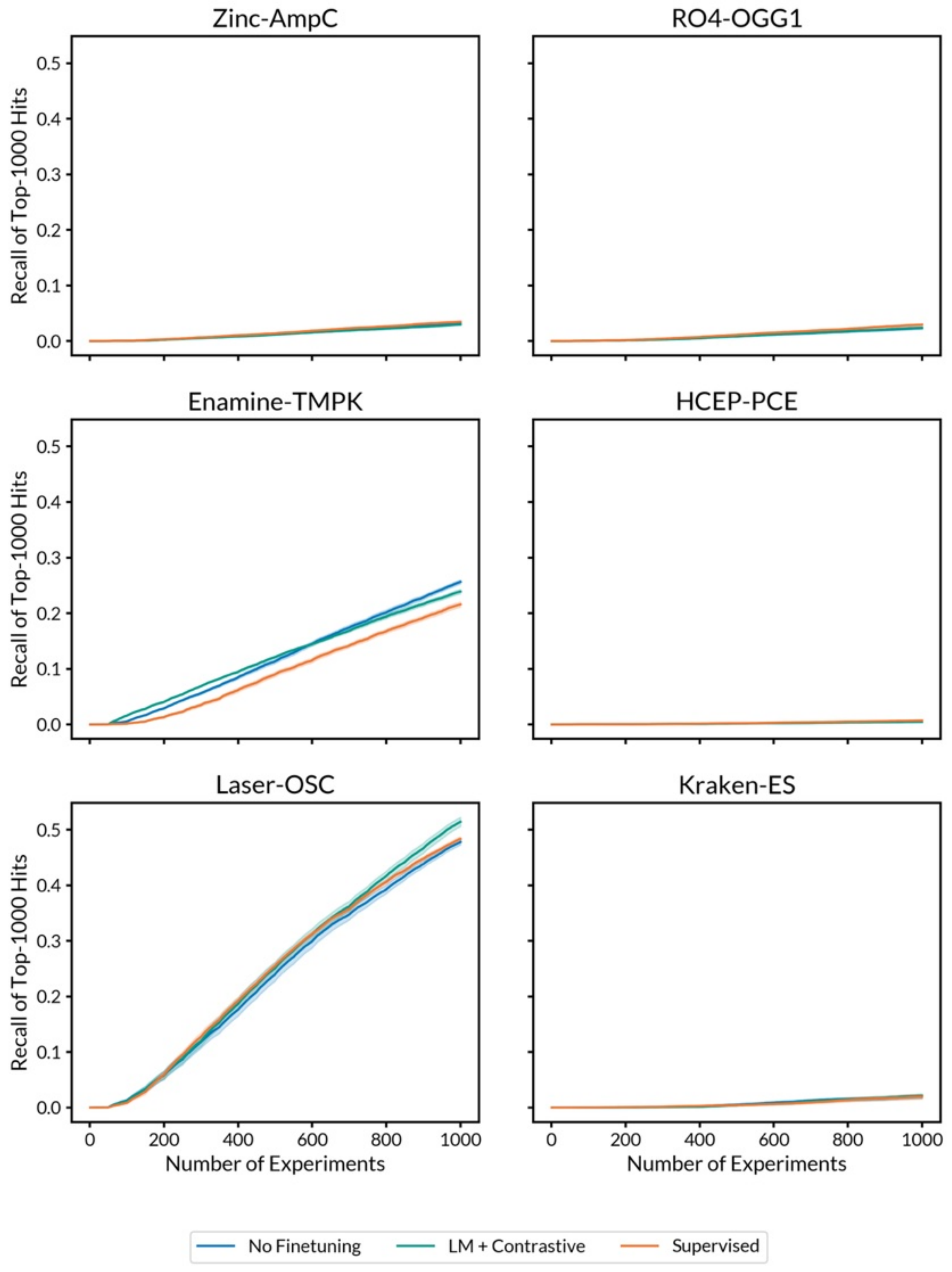

**Figure S 122**: Optimization trajectories after using different strategies to fine-tune T5Chem embeddings, evaluated over a budget of 1,000 experiments (batch size of 50 experiments). Trajectories are shown as the mean over 20 independent campaigns from different random seed populations. The shaded areas indicate the standard error of the mean.

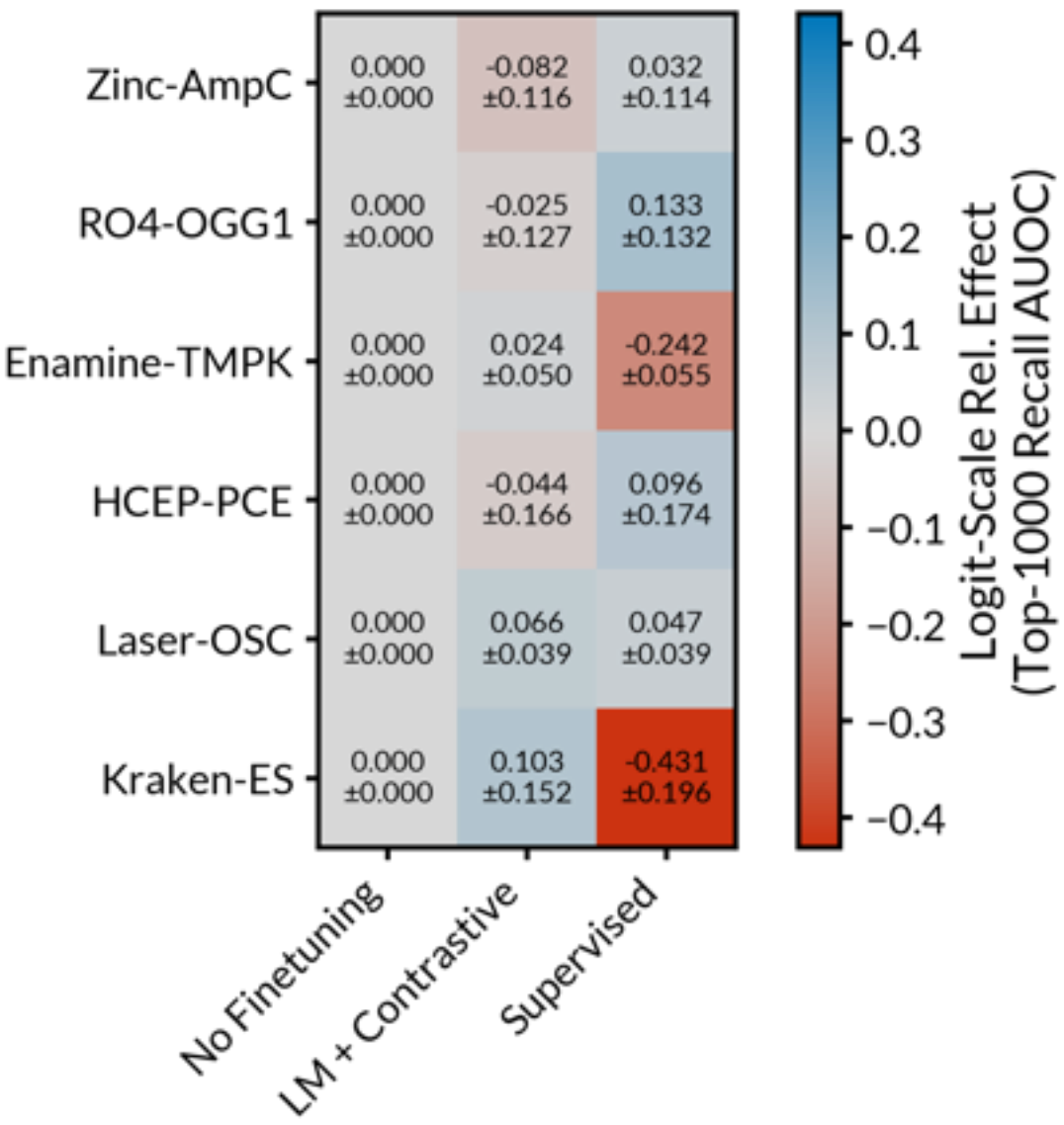

**Figure S 123**: Logit-scale effect of the fine-tuning strategy on T5Chem embeddings, differentiated by molecular libraries. Experiments were performed using a Laplace Neural network model in the experimental discovery setting (1,000 experiments, batch size 50). Means and standard deviations were obtained from 8,000 posterior samples.

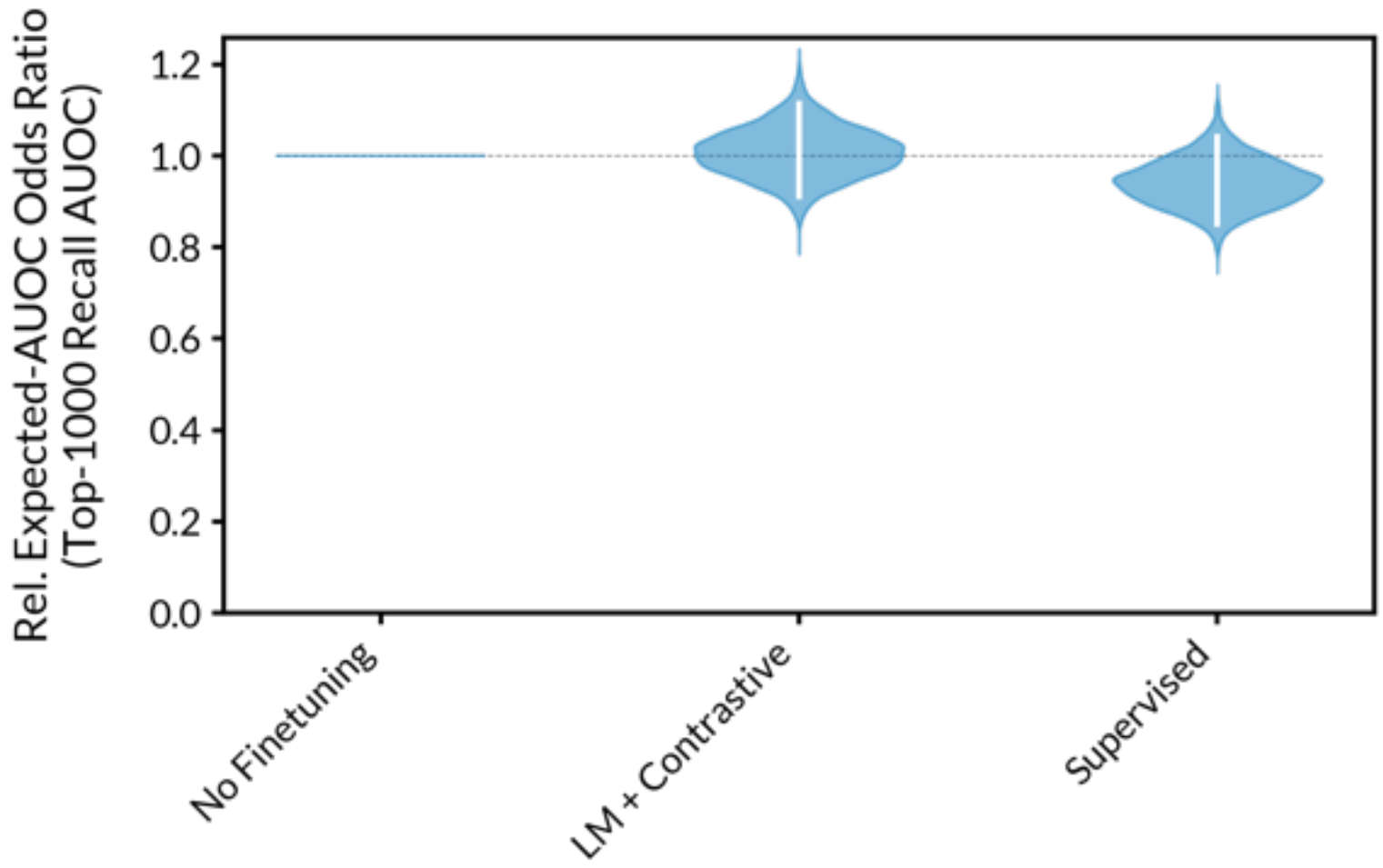

**Figure S 124**: Effect ratios of the fine-tuning strategy on T5Chem embeddings, averaged over all libraries. Experiments were performed using a Laplace Neural network model in the experimental discovery setting (1,000 experiments, batch size 50). Bars within the violins indicate the 95% highest-density intervals.

## 5 Supplementary References